\documentclass[review]{elsarticle}

\journal{Journal of Computational Physics}

\usepackage{amsmath}
\usepackage{amsthm}
\usepackage{dsfont}
\usepackage{amssymb}
\usepackage{graphicx,color}
\usepackage{extarrows}
\usepackage{mathrsfs}
\usepackage{booktabs}
\usepackage{multirow}
\usepackage[justification=raggedright]{caption}
\usepackage{threeparttable}
\usepackage[misc]{ifsym}
\usepackage[linesnumbered,ruled]{algorithm2e}
\usepackage{array}
\usepackage{subfig}
\usepackage{setspace}
\usepackage{float}
\usepackage{xurl}
\usepackage{lineno}
\usepackage{hyperref}

\makeatletter
\@addtoreset{equation}{section}
\makeatother

\graphicspath{ {./} }

\begin{document}
	\begin{frontmatter}
		\title{\bf Inverse Problems for Partial Differential Equations with Jump Discontinuities in Coefficients via Two-Stage Physics-Informed Deep Learning and Statistical Mixture Models}
		
		\author[addr1,addr2]{Zhikun Zhang}
        \ead{zhikunzhang@nwpu.edu.cn}

        \author[addr3,addr6]{Guanyu Pan}
        
        \author[addr2]{Xiangjun Wang}
        
        \author[addr1,addr5]{Yong Xu}

        \author[addr4,addr6]{Guangtao Zhang\corref{mycorrespondingauthor}}
        \cortext[mycorrespondingauthor]{Corresponding author}
        \ead{tao@sgd-ai.com}
		
		\address[addr1]{School of Mathematics and Statistics,\\ Northwestern Polytechnical University, Xi’an 710072, China}
		
		\address[addr2]{School of Mathematics and Statistics,\\ Huazhong University of Science and Technology, Wuhan 430074, China}
		
		\address[addr3]{School of Mathematics and Computational Science,\\ Xiangtan University, Xiangtan 411105, China}
		
		\address[addr4]{Department of Ocean Science and Engineering and Center for Complex Flows and Soft Matter Research,\\ Southern University of Science and Technology, Shenzhen 518055, China}
		
		\address[addr5]{MOE Key Laboratory for Complexity Science in Aerospace,\\ Northwestern Polytechnical University, Xi’an 710072, China}
		
		\address[addr6]{SandGold AI Research, Guangzhou 510642, China}
		
		\begin{abstract}
			Inverse problems involving partial differential equations (PDEs) with discontinuous coefficients arise widely in the modeling of complex spatiotemporal systems with uncertain dynamics and heterogeneous structures. Their solution is challenging because the unknown coefficient fields may exhibit abrupt temporal or spatial transitions, together with additional difficulties arising from nonlinearity, nonsmoothness, and multidimensionality. This work proposes a two-stage physics-informed deep learning framework that combines neural-network-based sampling with statistical inference and constrained parameter refinement. In the first stage, a dual-network physics-informed architecture is used, where a main network approximates the PDE solution and an auxiliary coefficient sub network provides a relaxed continuous surrogate of the true discontinuous coefficient field. A gradient-adaptive weighting strategy is incorporated into the physics residual to improve residual training and enhance sampling reliability near possible discontinuity regions. The sampled coefficient values are then analyzed using Bayesian learning for Gaussian mixture models and birth-death Markov chain model selection, which estimate the number of coefficient regimes and provide heuristic search intervals for coefficient values and candidate transition regions. In the second stage, the inverse problem is reformulated as a constrained physics-informed estimator, in which the coefficient is represented explicitly as a hard piecewise-constant function over the spatiotemporal domain. Numerical experiments on different PDE types with jump-discontinuous coefficients demonstrate that the proposed framework achieves accurate parameter estimation with acceptable computational costs compared to existing methods. This work provides an effective integrated workflow for inverse problems governed by PDEs with discontinuous parameter structures, particularly in nonstationary and heterogeneous systems.
		\end{abstract}
		
		\begin{keyword}
			Partial differential equation, Inverse problem, Parameter identification, Jump-varying coefficient, Physics-informed neural network, Gaussian mixture model, Markov chain Monte Carlo
		\end{keyword}
		
	\end{frontmatter}
	
	{\baselineskip=14pt
		\tableofcontents
		\hrulefill}
	
	\newpage
	
	\section{Introduction}\label{sec1}
	Partial differential equations (PDEs) provide a fundamental framework for modeling spatiotemporal physical processes in science and engineering. In many applications, the coefficients in these equations encode material properties, transport rates, source effects, or medium-dependent parameters, and they often need to be inferred from indirect observations \cite{isakov1988uniqueness, uhlmann2014inverse}. The inverse identification of such coefficients becomes especially difficult when they vary over space and time, and the difficulty is further amplified when they contain abrupt changes or jump discontinuities. Such discontinuities lead to nonsmooth parameter fields and localized transition regions, which may manifest as temporal change points, spatial interfaces, or more general spatiotemporal discontinuity sets in heterogeneous media \cite{chan2003identification, chan2004level, hao2011convergence}.
	
	Such scenarios arise in a wide range of PDE models and application domains. For example, in wave and Helmholtz equations, coefficients determine propagation speeds or medium properties, and discontinuities may correspond to layered media, inclusions, or material interfaces \cite{erlangga2004class, guru2009electromagnetic, metcalfe2012global, ning2014nonlinear, tong2016acoustic}. Heat and diffusion equations may involve discontinuous thermal conductivities or diffusivities \cite{hammond2011analytical, costin2012borel}, while fluid and transport models may contain spatially or temporally varying viscosity, diffusivity, or source terms \cite{bednarova2013spatial, bader2022hybrid}. Similar issues arise in geophysical, biological, environmental, and material systems, where effective parameters may vary across spatial subdomains or time intervals \cite{ferguson2009multivariate, sato2012seismic, farkas2023solving, han2025climate}. In these settings, the solution field may remain relatively smooth even when the underlying coefficient field is nonsmooth, making coefficient recovery from state data ill-conditioned. Therefore, accurately identifying both stable coefficient values and discontinuity locations from limited observations is a central task in PDE inverse problems with heterogeneous or regime-dependent parameters. This motivates the present focus on PDE coefficients that are piecewise constant or nearly piecewise constant over spatiotemporal domains.
	
	Classical parameter-identification methods for PDEs include regularized optimization, reduced-order modeling, and Bayesian inference. Output least-squares formulations combined with Tikhonov regularization have been used to recover conductivity or diffusion parameters \cite{engl2000new}. Reduced-basis and model-reduction techniques can accelerate repeated forward solves and Gauss-Newton iterations in inverse PDE problems \cite{druskin2007combining, jin2008fast}. Bayesian and hierarchical inference methods provide a probabilistic framework for parameter estimation and can be combined with Markov chain Monte Carlo (MCMC), stochastic collocation, or sparse Bayesian learning techniques \cite{xun2013parameter, frasso2016parameter, yuan2017sparse}. These methods have achieved substantial progress, but their implementation may become expensive or sensitive to discretization, prior modeling, and regularization choices when the unknown coefficient field is discontinuous, multidimensional or only indirectly observed. These considerations motivate data-assisted approaches that combine observational information with physical constraints.
	
	Recently, physics-informed neural networks (PINNs) have emerged as a mesh-free, data-assisted framework for PDE forward and inverse problems by penalizing the residuals of the governing equations together with data and boundary losses \cite{raissi2019physics}. PINNs have been widely used for solution approximation and inverse parameter estimation from sparse data \cite{karniadakis2021physics}. Subsequent developments have improved their stability and expressiveness through adaptive activation functions \cite{jagtap2020adaptive}, gradient-enhanced formulations \cite{yu2022gradient}, coupled architectures \cite{zhang2023cpinns}, and residual or self-adaptive weighting strategies \cite{wang2021understanding, mcclenny2023self, zhang2023constrained, anagnostopoulos2024residual}. For inverse problems with variable coefficients, additional network components or branch-network structures have also been introduced to approximate coefficient functions and to address the mismatch between the dimensionality of the unknown coefficient and that of the solution field \cite{mattey2022novel, miao2023vc, dong2024cp, zhang2025data}. These methods provide useful tools for learning smoothly varying coefficients from data, but their performance may deteriorate when the target coefficient contains sharp jump discontinuities.
	
	Despite these advances, discontinuous coefficient identification remains difficult for standard PINNs and existing variable-coefficient extensions. Conversely, a purely discrete parameterization requires prior knowledge of the number of regimes and the approximate locations of change points or interfaces, which are usually unavailable. As a result, the plateau values of piecewise-constant coefficients may be biased, and the corresponding change points or spatial interfaces may be inaccurately localized. Adaptive weighting can improve training stability, but it does not by itself convert a smooth coefficient approximation into a discontinuous representation. These observations motivate a framework that combines continuous neural approximation, statistical regime extraction, and constrained discontinuous-parameter refinement.
	
	This work develops a two-stage physics-informed learning framework for PDE inverse problems with jump-discontinuous coefficients. In the first stage, a gradient-adaptive weighted sub-main PINNs (GWS-PINNs) architecture is employed as a neural-network sampler. The main network approximates the PDE solution, while the auxiliary coefficient sub network learns a smoothed continuous surrogate of the underlying piecewise-constant coefficient field over the spatiotemporal interior. To improve sampling stability near possible discontinuity interfaces, a gradient-adaptive weighting strategy is incorporated into the physics residual, which suppresses the influence of high-gradient transition regions and enhances the reliability of coefficient samples in approximately constant regimes. The sampled coefficient values are then processed by a statistical mixture learner. Specifically, a Gaussian mixture model (GMM) within a Bayesian inference framework is used to extract discrete coefficient levels, while a birth-death Markov chain (BDMC) is adopted as a model-selection mechanism to infer the number of coefficient states. For multidimensional spatial or spatiotemporal coefficients, a dimension-by-dimension or flattened sampling procedure maps the coefficient samples to one-dimensional sequences for statistical inference and then maps the inferred transition information back to the original domain. This statistical step provides heuristic search intervals for the coefficient values and localized candidate regions containing possible temporal change points or spatial discontinuity interfaces.
	
	In the second stage, the inverse problem is reformulated as a constrained change-point detection PINNs (CCD-PINNs) estimator. Instead of continuing to represent the coefficient by a smooth neural function, the coefficient field is replaced by an explicit finite-dimensional piecewise-constant parameterization. The discrete coefficient values are optimized within the heuristic search intervals supplied by the mixture learner, and the corresponding change points or discontinuity boundaries are refined inside the candidate change regions. The solution network is initialized by transfer learning from the first-stage main network, and residual collocation points are enriched near the candidate transition regions to improve sensitivity to discontinuities. A classification-based refinement mechanism is further used to sharpen the region-to-state assignment and to construct a hard piecewise-constant estimator of the coefficient field. If no parameter variation is detected, the proposed estimator naturally reduces to a standard PINNs-based inverse formulation with interval-constrained constant parameters. Otherwise, it identifies both the plateau values and the associated temporal or spatial transition structures. In this way, this framework transforms a preliminary smooth coefficient surrogate into an explicit, constrained, and interpretable piecewise-constant reconstruction while simultaneously producing a high-fidelity surrogate solution.
	
	The main contributions of this work are summarized as follows.

    \begin{itemize}
    \item A two-stage physics-informed workflow is developed for inverse problems involving PDEs with piecewise-constant, jump-discontinuous coefficients. The workflow connects a continuous neural approximation in the first stage with an explicit discontinuous coefficient representation in the second stage, thereby avoiding the need to prescribe the coefficient states and transition locations in advance.

    \item In the first stage, a dual-network physics-informed architecture is used to jointly approximate the PDE solution and construct a relaxed continuous surrogate of the unknown coefficient field. A gradient-adaptive residual weighting strategy reduces the influence of high-gradient transition regions, allowing the coefficient sub network to provide more stable samples of approximately constant coefficient regimes.

    \item A Bayesian Gaussian mixture model combined with birth-death model selection serves as the statistical bridge between the two learning stages. It estimates the number of coefficient states and supplies candidate intervals for the coefficient values together with localized regions containing possible temporal change points or spatial discontinuity interfaces.

    \item In the second stage, the smooth coefficient surrogate is replaced by a constrained piecewise-constant parameterization. Transfer learning, locally enriched residual sampling, and classification-based region assignment are combined to refine the coefficient levels and discontinuity locations within the statistically identified candidate ranges.

    \end{itemize}
    
    Overall, the proposed jump-varying coefficients PINNs (JVC-PINNs) framework integrates physics-informed coefficient sampling, statistical regime identification, and constrained discontinuity refinement into a unified computational workflow. The numerical experiments in this study evaluate this workflow on representative synthetic PDE inverse problems with temporal and spatial coefficient jumps.

	The remainder of this study is organized as follows. 
	Section~\ref{sec2} presents the mathematical formulation of spatiotemporal systems with discontinuously varying coefficients and introduces the proposed two-stage framework, including the neural-network-based sampler, the statistical model learner, and the constrained parameter refinement estimator. 
	Section~\ref{sec3} reports numerical experiments on representative PDEs with jump discontinuities in their coefficients. 
	Section~\ref{sec4} conducts an ablation study to evaluate the contributions of different loss components. 
	Section~\ref{sec5} compares the proposed method with existing methods in terms of accuracy and computational efficiency.
	Section~\ref{sec6} presents applications of the proposed framework to physical-field reconstruction.
	Section~\ref{sec7} investigates the robustness to different levels of observational noise and presents a diagnostic of predictive variability in reconstructed solution and parameter estimation.
	Section~\ref{sec8} summarizes some limitations of this framework.
	Finally, Section~\ref{sec9} concludes the paper and discusses future directions.
	
	\section{Methodology}\label{sec2}
	This section presents a two-stage framework, termed JVC-PINNs, for identifying discontinuously varying coefficients in PDEs. The two PINNs-based stages are connected through Bayesian statistical mixture modeling. After formulating the inverse problem for PDEs with finite piecewise-constant coefficients in the spatiotemporal interior, the first stage employs the GWS-PINNs sampler, in which a main network approximates the PDE solution while an auxiliary coefficient network, regularized by a gradient-adaptive weighting strategy, produces a smoothed continuous surrogate of the coefficient field and serves as an interior sampler of the underlying piecewise-constant states. These samples are then passed to a statistical mixture method, where GMMs capture the discrete parameter levels and a BDMC adaptively determines the number of states. Meanwhile, a dimension-by-dimension flattening procedure extends this one-dimensional analysis to multidimensional spatiotemporal coefficients and yields heuristic search intervals together with candidate parameter change regions.
	
\begin{figure}[t]
		\centering
		\includegraphics[width=1\linewidth]{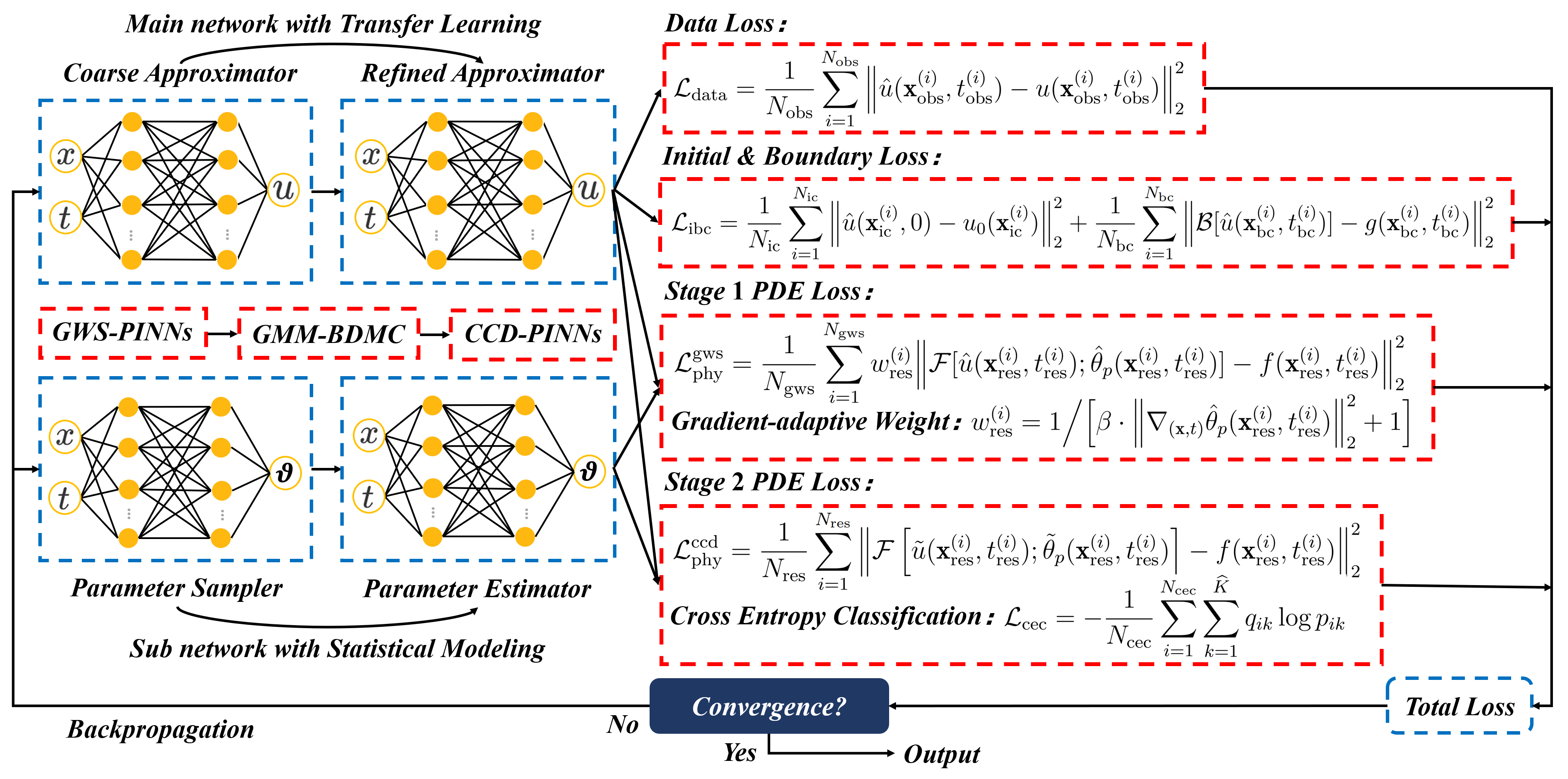}
		\caption{Schematic diagram of JVC-PINNs framework, including GWS-PINNs and CCD-PINNs stages, which are connected by GMM-BDMC statistical procedure. \label{fig0}}
	\end{figure}

	In the second stage, the inverse problem is recast as the CCD-PINNs estimator. Instead of continuing to approximate the coefficient field by a smooth neural function, CCD-PINNs performs constrained learning of the physical parameters and their discontinuity locations. Specifically, the coefficient is reparameterized using a low-dimensional set of piecewise-constant values and change-region variables, all of which are confined to the search intervals or regions supplied by the statistical learning results. Moreover, the solution network is fine-tuned through transfer learning, and residual sampling is densified inside the candidate change regions to improve sensitivity to discontinuities. In this way, the proposed framework transforms the preliminary smooth coefficient surrogate into an explicitly constrained piecewise-constant representation, while simultaneously delivering a high-fidelity surrogate solution and an interpretable reconstruction of the discontinuous coefficient field in heterogeneous spatiotemporal PDE systems. An overview of the JVC-PINNs framework is shown in Figure~\ref{fig0}.
	
	\subsection{Mathematical Setup}
	The PDEs with variable coefficients have been extensively studied in cases involving continuously varying parameters, with numerous numerical and theoretical frameworks proposed for such systems. However, scenarios where coefficients exhibit discontinuous and abrupt changes across spatiotemporal domains, often driven by underlying switching mechanisms in complex systems, remain underexplored and pose unique challenges for inverse problems. This study focuses precisely on PDEs with discontinuously varying coefficients, seeking to accurately identify and infer these parameters from observed system outputs while maintaining computational tractability and interpretability.
	
	To begin with, consider a general initial-boundary value problem with a solution function $u(\mathbf{x}, t)$, where $\mathbf{x} = [x_1, x_2, \ldots, x_d]$ denotes the $d$-dimensional spatial variable and $t$ denotes time. Let $\Omega \subset \mathbb{R}^d$ be a bounded spatial domain with boundary $\partial\Omega$ and interior $\Omega^{\circ}$, and let $[t_0,T]\subset \mathbb{R}$ be the temporal interval. The spatiotemporal domain is denoted by
	\begin{equation}
		\mathcal{Q} = \Omega \times [t_0,T], \quad \mathcal{Q}^{\circ} = \Omega^{\circ} \times (t_0,T),
	\end{equation}
	where $\mathcal{Q}^{\circ}$ denotes the interior of the spatiotemporal domain. Without loss of generality, set $t_0=0$ in the formulation below. The governing system is formulated as
	\begin{equation}
		\begin{cases}
			\mathcal{F}\big[u(\mathbf{x},t);\,\theta_p(\mathbf{x},t)\big] = f(\mathbf{x},t), 
			& (\mathbf{x},t)\in \Omega^{\circ}\times (0,T], \\[2pt]
			\mathcal{B}[u(\mathbf{x},t)] = g(\mathbf{x},t),
			& (\mathbf{x},t)\in \partial\Omega \times (0,T], \\[2pt]
			u(\mathbf{x},0) = u_0(\mathbf{x}), 
			& \mathbf{x}\in \Omega,
		\end{cases}
		\label{eq:pde_ibc}
	\end{equation}
	where $\mathcal{F}[\cdot]$ denotes a differential operator acting on $u$ and its spatiotemporal derivatives, $\mathcal{B}[\cdot]$ is a boundary operator, e.g., Dirichlet, Neumann, or Robin type, $g(\mathbf{x},t)$ specifies the boundary data, $f(\mathbf{x},t)$ is a source term, $u_0(\mathbf{x})$ provides the initial condition, and $\theta_p(\mathbf{x},t)$ denotes the spatially and temporally varying physical parameter field to be estimated from observations of the physical solution field $u(\mathbf{x}, t)$, which can be a single parameter or $m$ parameters in this PDE system.
	
	When the PDE coefficients vary over the spatiotemporal domain $\mathcal{Q}$, the parameter field $\theta_p(\mathbf{x},t)$ may exhibit deterministic or random spatial and temporal variations. This work specifically focuses on cases of finitely many piecewise-constant coefficients, a common scenario in which PDE parameters exhibit jump discontinuities through state transitions in complex spatiotemporal systems. Given a finite partition $\{\mathcal{Q}_r\}_{r=1}^{R}$ of $\mathcal{Q}^{\circ}$ satisfying $\bigcup_{r=1}^{R}\mathcal{Q}_r=\mathcal{Q}^{\circ}$ and $\mathcal{Q}_r\cap\mathcal{Q}_{r'}=\emptyset$ for $r\ne r'$, the coefficient field is assumed to take finitely many constant states. Let $\{\theta_k\}_{k=1}^{K}$ denote the distinct coefficient states and let $ z:\{1,\dots,R\}\to\{1,\dots,K\} $ be the region-to-state map, where $R$ denotes the number of subdomains, and $K$ denotes the number of different coefficient states. This yields
	\begin{equation}
		\theta_p(\mathbf{x},t) = \sum_{r=1}^{R} \theta_{z(r)} \mathbf{1}_{\mathcal{Q}_r} (\mathbf{x},t),
		\label{eq:piecewise_param}
	\end{equation}
	where $K\le R$ in general, since the same coefficient state may appear in multiple disconnected subdomains.
	
	Further assume that the discontinuity interfaces of $\theta_p(\mathbf{x},t)$ lie strictly within the interior $\mathcal{Q}^{\circ}$, that is, the set of jump discontinuities in coefficients satisfies
	\begin{equation}
		\mathcal{J}_{\theta} := \bigcup_{\substack{r\ne r'\\ z(r)\ne z(r')}} \left( \overline{\mathcal{Q}_r} \cap \overline{\mathcal{Q}_{r'}} \right) \subset \mathcal{Q}^{\circ}.
		\label{jump_discontinuities}
	\end{equation}
	This assumption ensures that parameter variations do not occur at the initial time $t_0=0$ or on the spatial boundary $\partial\Omega\times(0,T]$. This assumption is consistent with many physical scenarios in which switching behavior is generated by interior dynamics rather than by imposed boundary or initial data.
	
	With the above setup, the inverse problem addressed in this paper is stated as follows. Given the known differential operator $\mathcal{F}$, the boundary operator $\mathcal{B}$, the source term $f$, the initial condition $u_0$, a set of possibly noisy discrete observations $\{u(\mathbf{x}_{\text{obs}}^{(i)}, t_{\text{obs}}^{(i)})\}_{i=1}^{N_{\text{obs}}}$, the boundary data $\{g(\mathbf{x}_{\mathrm{bc}}^{(i)},t_{\mathrm{bc}}^{(i)})\}_{i=1}^{N_{\mathrm{bc}}}$ and the initial data $\{u_0(\mathbf{x}_{\mathrm{ic}}^{(i)})\}_{i=1}^{N_{\mathrm{ic}}}$ generated by \eqref{eq:pde_ibc}, the goal is to identify the discontinuously varying coefficient field $\theta_p(\mathbf{x},t)$ of the piecewise-constant form \eqref{eq:piecewise_param}, including the distinct coefficient states $\{\theta_k\}_{k=1}^{K}$, the partition $\{\mathcal{Q}_r\}_{r=1}^{R}$, and the region-to-state map $z(r)$. Equivalently, the task is to recover both the coefficient levels and the interior jump interfaces $\mathcal{J}_{\theta}$. In the purely one-dimensional temporal case, these interfaces reduce to a finite set of change points $\{\tau_l\}_{l=1}^{L}$, where $L$ is the number of candidate change regions, typically $L=R-1$ for an ordered temporal partition. This part provides the modeling foundation for the inversion and identification tasks discussed in subsequent sections.
	
	\subsection{Neural Network Sampler}
	In previous works, neural networks have been demonstrated to be a powerful universal function approximator. To address PDEs with spatiotemporal switching coefficients, one necessary modification is to extend the PINNs architecture by incorporating an additional sub network for learning the unknown spatially and temporally varying parameters. This subsection introduces the first stage of this two-stage framework, referred to as GWS-PINNs, in which a physics-informed deep learning model is used to construct a coefficient sampler over the spatiotemporal domain $\mathcal{Q}$.
	
	In a general parameter inversion problem, given a set of observations $\{u(\mathbf{x}_{\text{obs}}^{(i)}, t_{\text{obs}}^{(i)}) \}_{i=1}^{N_{\text{obs}}}$ at the training point set $\{(\mathbf{x}_{\text{obs}}^{(i)}, t_{\text{obs}}^{(i)}) \}_{i=1}^{N_{\text{obs}}}$ from PDE solutions, together with the boundary data $\{g(\mathbf{x}_{\text{bc}}^{(i)}, t_{\text{bc}}^{(i)})\}_{i=1}^{N_{\text{bc}}}$ on $\partial\Omega\times(0,T]$ and the initial data $\{u_0(\mathbf{x}_{\text{ic}}^{(i)})\}_{i=1}^{N_{\text{ic}}}$ on $\Omega\times\{0\}$, the aim is to infer the unknown coefficient field $\theta_p(\mathbf{x},t)$ in the PDE system \eqref{eq:pde_ibc}. In the standard PINNs framework \cite{raissi2019physics}, the solution $u(\mathbf{x}, t)$ is approximated by a neural network $\hat{u}(\mathbf{x}, t)$, and the parameters to be inferred are treated as learnable variables $\hat{\theta}_p$ optimized via stochastic gradient descent. The total loss is decomposed into a data fitting term, a PDE physics residual term, a boundary condition term, and an initial condition term, given by
	\begin{equation}\label{Loss_sample}
		\begin{aligned}
			\mathcal{L}_{\text{data}} &= \frac{1}{N_{\text{obs}}}\sum_{i=1}^{N_{\text{obs}}} \Big\| \hat{u}(\mathbf{x}_{\text{obs}}^{(i)}, t_{\text{obs}}^{(i)}) - u(\mathbf{x}_{\text{obs}}^{(i)}, t_{\text{obs}}^{(i)}) \Big\|_2^2,\\
			\mathcal{L}^{\text{std}}_{\text{phy}} &= \frac{1}{N_{\text{res}}}\sum_{i=1}^{N_{\text{res}}} \Big\| \mathcal{F}[\hat{u}(\mathbf{x}_{\text{res}}^{(i)}, t_{\text{res}}^{(i)}); \hat{\theta}_p] - f(\mathbf{x}_{\text{res}}^{(i)}, t_{\text{res}}^{(i)}) \Big\|_2^2,\\
			\mathcal{L}_{\text{bc}} &= \frac{1}{N_{\text{bc}}}\sum_{i=1}^{N_{\text{bc}}} \Big\| \mathcal{B}[\hat{u}(\mathbf{x}_{\text{bc}}^{(i)}, t_{\text{bc}}^{(i)})] - g(\mathbf{x}_{\text{bc}}^{(i)}, t_{\text{bc}}^{(i)}) \Big\|_2^2,\\
			\mathcal{L}_{\text{ic}} &= \frac{1}{N_{\text{ic}}}\sum_{i=1}^{N_{\text{ic}}} \Big\| \hat{u}(\mathbf{x}_{\text{ic}}^{(i)}, 0) - u_0(\mathbf{x}_{\text{ic}}^{(i)}) \Big\|_2^2,\\
			\mathcal{L}_{\text{total}} &= \gamma_0 \mathcal{L}_{\text{data}} + \gamma_1 \mathcal{L}^{\text{std}}_{\text{phy}} + \gamma_2 \mathcal{L}_{\text{bc}} + \gamma_3 \mathcal{L}_{\text{ic}},
		\end{aligned}
	\end{equation}
	where $N_{\text{obs}}$, $N_{\text{res}}$, $N_{\text{bc}}$, and $N_{\text{ic}}$ denote the numbers of observation points, residual points, boundary points, and initial points, respectively. Generally set $\gamma_0=1$, and the other nonnegative weights $\gamma_1$, $\gamma_2$ and $\gamma_3$ regulate the relative influence of the physics residual and the initial-boundary terms in the total loss. Their values need to be set appropriately based on factors such as the problem scale, the magnitude of observation noise, and other considerations.
	
	Since $\hat{\theta}_p$ is merely a learnable parameter updated directly through gradient descent, its convergence depends on the choice of initial values. Inappropriate initialization may lead to prolonged training or convergence to inaccurate parameter values. Additionally, standard PINNs are naturally formulated for PDEs with constant or smoothly parameterized coefficients and are not directly suited to PDEs with discontinuously varying coefficients of the form \eqref{eq:piecewise_param}. To address the parameter inversion problem for PDEs with spatiotemporal jump-varying coefficients, modifications to the standard PINNs network architecture are required.
	
	Given that neural networks can approximate arbitrary continuous functions, recent research has explored incorporating an additional sub network into PINNs to fit the unknown varying coefficients in PDEs \cite{miao2023vc}. To address the inversion and identification task of spatiotemporal varying coefficients, this work constructs a network architecture consisting of a main network $\hat{u}(\mathbf{x},t)$ that approximates the PDE solution $u(\mathbf{x},t)$, and a sub network $\hat{\theta}_p(\mathbf{x},t)$ that samples the varying coefficient field over the spatiotemporal domain. Since the parameter jumps are assumed to occur strictly in the interior $\mathcal{Q}^{\circ}$, all residual points used for the sub network are drawn from $\mathcal{Q}^{\circ}$, i.e., the boundary $\partial\Omega\times(0,T]$ and the initial slice $\Omega\times\{0\}$ are excluded from parameter sampling. Replacing the scalar parameter $\hat{\theta}_p$ by the sub network output $\hat{\theta}_p(\mathbf{x},t)$, the data term $\mathcal{L}_{\text{data}}$, the boundary term $\mathcal{L}_{\text{bc}}$, and the initial term $\mathcal{L}_{\text{ic}}$ remain unchanged, while the PDE physics residual term is redefined as
	\begin{equation}
		\mathcal{L}_{\text{phy}}^{\text{vc}} = \frac{1}{N_{\text{res}}} \sum_{i=1}^{N_{\text{res}}} \Big\| \mathcal{F}[\hat{u}(\mathbf{x}_{\text{res}}^{(i)}, t_{\text{res}}^{(i)});\; \hat{\theta}_p(\mathbf{x}_{\text{res}}^{(i)}, t_{\text{res}}^{(i)})] - f(\mathbf{x}_{\text{res}}^{(i)}, t_{\text{res}}^{(i)}) \Big\|_2^2, \quad (\mathbf{x}_{\text{res}}^{(i)}, t_{\text{res}}^{(i)}) \in \mathcal{Q}^{\circ}.
	\end{equation}
	
	On the other hand, it is necessary to further optimize the physics residual loss of standard PINNs. For parameter fields with strong discontinuities or jump regions, relying solely on sub networks is insufficient to effectively suppress abnormal oscillation phenomena in parameter sampling results near discontinuities. To address this issue, exploring weighting strategies during PINNs training has become a mainstream approach. Residual-based adaptive sampling has also been used to allocate more collocation points to difficult regions \cite{wu2023comprehensive}, which to some extent mitigates underfitting or overfitting of the solution approximation in discontinuous or nonsmooth regions during training. Even within a neural network framework that already incorporates a sub network, it is still possible to apply weighting to the PDE residual term. Specifically, weights that vary with gradients can be introduced into the physics residual, so that $\mathcal{L}_{\text{phy}}$ is further refined into a gradient-adaptive form as
	\begin{equation}
		\mathcal{L}_{\text{phy}}^{\text{gws}} = \frac{1}{N_{\text{res}}} \sum_{i=1}^{N_{\text{res}}} w_{\text{res}}^{(i)} \Big\| \mathcal{F}[\hat{u}(\mathbf{x}_{\text{res}}^{(i)}, t_{\text{res}}^{(i)});\; \hat{\theta}_p(\mathbf{x}_{\text{res}}^{(i)}, t_{\text{res}}^{(i)})] - f(\mathbf{x}_{\text{res}}^{(i)}, t_{\text{res}}^{(i)}) \Big\|_2^2, \quad (\mathbf{x}_{\text{res}}^{(i)}, t_{\text{res}}^{(i)}) \in \mathcal{Q}^{\circ},
	\end{equation}
	where
	\begin{equation}
		w_{\text{res}}^{(i)} = 1\Big/\Big[\beta \cdot \left\|\nabla_{(\mathbf{x},t)} \hat{\theta}_p(\mathbf{x}_{\text{res}}^{(i)},t_{\text{res}}^{(i)}) \right\|_2^2 + 1\Big].
	\end{equation}
	Here, $w_{\text{res}}^{(i)} > 0$ is the gradient-adaptive weight controlling the contribution of each residual point, $\nabla_{(\mathbf{x},t)}$ denotes the spatiotemporal gradient operator acting on the sub-network output, and $\beta$ is the gradient-adaptive weight hyperparameter. This mechanism down-weights residual points in regions where the sampled coefficient field exhibits large gradients near discontinuity interfaces $\mathcal{J}_{\theta}$, while assigning relatively larger weights to smoother regions. This weighting scheme for $\hat{\theta}_p(\mathbf{x},t)$ reduces the contribution of residual points located in high-gradient regions and thereby emphasizes samples from approximately constant regions.
	
	Combining the unchanged data, boundary, and initial losses with the redefined gradient-adaptive physics residual, the total loss of the GWS-PINNs sampler is given by
	\begin{equation}
		\mathcal{L}_{\text{sam}} = \gamma_0\mathcal{L}_{\text{data}} + \gamma_1 \mathcal{L}_{\text{phy}}^{\text{gws}} + \gamma_2 \mathcal{L}_{\text{bc}} + \gamma_3 \mathcal{L}_{\text{ic}}.
	\end{equation}
	
	In contrast to time-domain partitioning approaches such as Bayesian sparsity \cite{li2023diffusion} and neural networks \cite{mattey2022novel}, the sub network can approximate a broad range of coefficient fields, including piecewise-constant coefficients, stochastically time-varying parameters, spatially heterogeneous fields, and more complex distributions. In addition, GWS-PINNs flexibly directs learning within the PDE domain using adaptive weighting to suppress instability in regions with large gradients caused by discontinuous parameter variations, which helps mitigate numerical oscillations and overfitting near these regions, and thus improves the physical consistency between the neural networks and real systems.
	
	Given observed data from PDE solutions, the main network and sub network respectively output $\hat{u}(\mathbf{x},t)$ and $\hat{\theta}_p(\mathbf{x},t)$. Physics residuals are constructed using automatic differentiation and refined by adaptive weights to form the final loss for backpropagation. Once trained, $\hat{u}(\mathbf{x},t)$ provides a continuous approximation of the PDE solution, while $\hat{\theta}_p(\mathbf{x},t)$ yields a continuous surrogate of the underlying piecewise-constant coefficient field. Although the sub-network output remains continuous and thus can only approximate the true jump-varying parameter \eqref{eq:piecewise_param} in a smoothed manner, the gradient-adaptive weighting confines the transition regions to be sufficiently narrow, so that $\hat{\theta}_p(\mathbf{x},t)$ evaluated at a finite collection of interior points in $\mathcal{Q}^{\circ}$ constitutes a reliable empirical sample of the discontinuous coefficient field. These samples will then be passed to the statistical learner introduced in the next subsection, in which a GMM combined with a BDMC model is employed to extract the discrete parameter states and to locate the corresponding jump regions.
	
	\subsection{Statistical Mixture Modeling}
	After the first-stage training of GWS-PINNs, evaluating the sub network $\hat{\theta}_p(\mathbf{x},t)$ at a finite set of interior collocation points in $\mathcal{Q}^{\circ}$ yields a collection of samples of the underlying piecewise-constant coefficient field. Flattening these samples into a one-dimensional sequence yields a sequence
	\begin{equation}
		\mathbf{y} = \{ y_n \}_{n=1}^{N},
	\end{equation}
	where each $y_n$ is the sub-network output at the $n$-th interior point under the prescribed rule.
    
    For a one-dimensional temporal coefficient $\theta(t)$, the natural time ordering is used directly. For a $d$+1-dimensional spatiotemporal coefficient $\theta_p(\mathbf{x},t)$, the flattening can be performed in different ways, such as traversing the interior points along coordinate axes or randomly mixing all interior samples, as long as the multidimensional interior data in $\mathcal{Q}^{\circ}$ are transformed into a one-dimensional sequence.
    
    Because the GMM-BDMC learner depends only on the empirical distribution of the sampled coefficient values, different bijective flattening strategies lead to identical statistical targets up to permutation of sample indices. Consequently, flattening serves only as a computational representation and does not affect the subsequent recovery of discontinuity regions in the original spatiotemporal domain. After statistical inference, the identified indices can be mapped back to the original spatiotemporal domain through the inverse of the flattening operation. In this way, the inference of discontinuously varying coefficients across space and time is reduced to a unified one-dimensional statistical learning task.
	
	Since the sub-network sampling inevitably carries bias from observation noise and network approximation error, a Gaussian assumption is adopted around each latent constant state. Combined with the finite piecewise-constant structure \eqref{eq:piecewise_param}, the marginal distribution of $y_n$ is modeled as a GMM
	\begin{equation}
		\Pr(y_n \mid \boldsymbol{\vartheta})=\sum_{k=1}^{K} \eta_k\, f_{\mathcal{N}}(y_n;\, \mu_k, \sigma_k^2),\quad \sum_{k=1}^{K} \eta_k=1,
	\end{equation}
	with parameters $\boldsymbol{\vartheta}=\{\mu_k,\,\sigma_k,\,\eta_k\}_{k=1}^{K}$ and latent allocations $\mathbf{S}=\{S_n\}_{n=1}^N$, $S_n\in\{1,\dots,K\}$. Here $f_{\mathcal{N}}$ denotes the univariate Gaussian density. The parameters are inferred via a Bayesian MCMC sampler with conditionally conjugate priors, summarized as
	\begin{equation}
		\sigma_k^2\sim \mathcal{IG}(c_0,C_0),\quad \mu_k\mid \sigma_k^2\sim \mathcal{N}(b_0,\sigma_k^2),\quad \boldsymbol{\eta}\sim \mathcal{D}(1,\dots,1),
	\end{equation}
	where $\mathcal{N}$, $\mathcal{IG}$ and $\mathcal{D}$ denote the Gaussian, inverse-Gamma and Dirichlet distributions, respectively. Conjugacy yields closed-form Gibbs sampler updates
	\begin{equation}
		\sigma_k^2 \,|\, \mathbf{y},\mathbf{S}\sim \mathcal{IG}(c_k(\mathbf{S}),C_k(\mathbf{S})),\quad
		\mu_k \,|\, \sigma_k^2,\mathbf{y},\mathbf{S}\sim \mathcal{N}(b_k(\mathbf{S}),B_k(\mathbf{S})),\quad
		\boldsymbol{\eta} \,|\, \mathbf{S}\sim \mathcal{D}(e_1(\mathbf{S}),\dots,e_{K}(\mathbf{S})),
	\end{equation}
	where
	\begin{equation}
		\begin{aligned}
			c_k(\mathbf{S})&=c_0+\dfrac{1}{2}N_k(\mathbf{S}),\quad 
			C_k(\mathbf{S})=C_0+\dfrac{1}{2}\!\left(N_k(\mathbf{S})\,s_{y,k}^2(\mathbf{S})+\dfrac{N_k(\mathbf{S})}{N_k(\mathbf{S})+1}(\bar{y}_k(\mathbf{S})-b_0)^2\right),\\
			b_k(\mathbf{S})&=\dfrac{b_0+N_k(\mathbf{S})\bar{y}_k(\mathbf{S})}{N_k(\mathbf{S})+1},\quad
			B_k(\mathbf{S})=\dfrac{\sigma_k^2}{N_k(\mathbf{S})+1},\quad
			e_k(\mathbf{S})=N_k(\mathbf{S})+1,
		\end{aligned}
	\end{equation}
	with $N_k(\mathbf{S})$, $\bar{y}_k(\mathbf{S})$, and $s_{y,k}^2(\mathbf{S})$ denoting the count, mean, and variance of the samples allocated to component $k$.
	
	A key difficulty is that the number of GMM components $K$ is unknown a priori. Information-criterion-based approaches require fitting all candidate models for comparison, which is computationally expensive. Instead, using a BDMC mechanism that treats $K$ itself as a random variable evolving in continuous time. For a $K$-component GMM $\mathcal{M}_{K}$, a birth occurs at a constant rate $b(s)\equiv\lambda_b$, which appends a new component with weight $\eta_{K+1}\sim\mathcal{Be}(1,K)$ sampled from a Beta distribution and rescaling $\eta_j^{\text{new}}=\eta_j(1-\eta_{K+1})$ for $j\le K$, while each existing component $k$ dies at rate
	\begin{equation}
		d_k=\frac{\lambda_b\cdot \Pr(\mathbf{y}\mid \mathcal{M}_{K-1},\boldsymbol{\vartheta}^{-k})\,\Pr(\mathcal{M}_{K-1})}{K\cdot \Pr(\mathbf{y}\mid \mathcal{M}_{K},\boldsymbol{\vartheta})\,\Pr(\mathcal{M}_{K})},\quad k=1,\dots,K,
	\end{equation}
	where $\boldsymbol{\vartheta}^{-k}$ denotes the parameter set obtained by removing the parameters associated with the $k$-th component from the original GMM parameter set as
	\begin{equation}
		\boldsymbol{\vartheta}^{-k} = \{\mu_1,\cdots,\mu_{k-1},\mu_{k+1},\cdots,\mu_{K}, \sigma_1,\cdots,\sigma_{k-1},\sigma_{k+1},\cdots,\sigma_{K}, \eta_1,\cdots,\eta_{k-1},\eta_{k+1},\cdots,\eta_{K} \},
	\end{equation}
	with total death rate $d(s)=\sum_{k=1}^{K} d_k$, inter-event time $\Delta s\sim\mathcal{E}(b(s)+d(s))$ from an exponential distribution and rescales the remaining weights as $\eta_j^{\mathrm{new}}=\eta_j/(1-\eta_k)$ for $j<k$ and $\eta_j^{\mathrm{new}}=\eta_{j+1}/(1-\eta_k)$ for $j=k,\ldots,K-1$. A birth is chosen with probability $b(s)/(b(s)+d(s))$ and the death of component $k$ with probability $d_k/(b(s)+d(s))$. Between jump events, Gibbs sampling is performed at the current $K$ to update $(\boldsymbol{\vartheta},\mathbf{S})$. Letting $T_{K}$ denote the total time that the chain spends in state $\mathcal{M}_{K}$ over the evolution horizon $[0,s_{\max}]$, the optimal component number is estimated as
	\begin{equation}
		\widehat{K}=\arg\max_{K} \frac{T_{K}}{s_{\max}}.
	\end{equation}
	With $\widehat{K}$ fixed, discarding the first $M_0$ burn-in samples and averaging the remaining $M$ posterior draws yields the component-wise estimates
	\begin{equation}
		\widehat{\mu}_k=\frac{1}{M}\sum_{j=1}^{M}\mu_{k,j},\quad 
		\widehat{\sigma}_k=\frac{1}{M}\sum_{j=1}^{M}\sigma_{k,j},\quad k=1,\dots,\widehat{K}.
	\end{equation}
	To avoid label switching, the mixture components are relabeled after each posterior draw such that $\mu_1<\mu_2<\cdots<\mu_{\widehat{K}}$. Thus the framework infers the discrete coefficient states and candidate transition regions from the first-stage PINNs samples obtained using discrete PDE solution observations. For more details on statistical mixture model learning, refer to \cite{fruhwirth2006finite}.
	
	Based on the statistical mixture modeling results, heuristic search intervals are constructed to constrain the second-stage PINNs inverse problem. For each identified state $k$, the candidate interval for the coefficient value is defined as
    \begin{equation}
        \mathcal{I}_k = \big[\,\widehat{\mu}_k-\widehat{\sigma}_k,\,
        \widehat{\mu}_k+\widehat{\sigma}_k\,\big], \quad k=1,\dots,\widehat{K},
        \label{eq:param_interval}
    \end{equation}
    where $\widehat{\sigma}_k$ describes the estimated within-component dispersion of the Stage~1 coefficient samples and $\mathcal{I}_k$ is used as a practical search range for the coefficient value in the subsequent constrained refinement. These intervals are heuristic and are not Bayesian credible intervals for the coefficient states or posterior uncertainty intervals for the component means.
    
    As a localization rule, sample indices in the sequence $\mathbf{y}$ whose coefficient values fall outside every interval $\mathcal{I}_k$ are treated as candidate transition samples. This rule is motivated by the observation that the Stage~1 continuous surrogate generally exhibits small within-state variation on coefficient plateaus and larger deviations across smoothed transition layers. It is not intended to provide an exact statistical classification, which means plateau samples may occasionally fall outside their associated intervals, while transition samples may remain inside one of the intervals. The resulting set is therefore used only as a coarse candidate region, whose transition structure is subsequently refined by CCD-PINNs. Mapping these indices back to the original spatiotemporal domain through the inverse of the flattening operation yields the candidate discontinuity set
	\begin{equation}
		\mathcal{X}=\Big\{(\mathbf{x},t)\in\mathcal{Q}^{\circ}:\, \hat{\theta}_p(\mathbf{x},t)\notin \bigcup_{k=1}^{\widehat{K}} \mathcal{I}_k\Big\},
		\label{eq:jump_interval}
	\end{equation}
	which is used as a candidate search region for the subsequent refinement of the discontinuity locations. If $\widehat{K} = 1$, indicating that no variation is detected, set $\mathcal{X}=\varnothing$ and skip the change-point or interface refinement while retaining the coefficient refinement in Stage~2.
    
	This candidate transition set is further divided according to geometric connectivity in the original spatiotemporal domain. Specifically, connected points are grouped into the same candidate change region, while disconnected parts are treated as different regions with the number of candidate change regions $\widehat{L}$ as
	\begin{equation}
		\mathcal{X} = \bigcup_{l=1}^{\widehat{L}}
		\mathcal{X}_{l}, \quad \mathcal{X}_{l}\cap \mathcal{X}_{l'}=\varnothing,\quad l \neq l',
		\label{eq:jump_region_decomposition}
	\end{equation}
	where each $\mathcal{X}_{l}$ is a geometrically connected component. For purely temporal coefficients, \eqref{eq:jump_region_decomposition} reduces to a finite union of candidate change-point time intervals. The pair $\big(\{\mathcal{I}_k\}_{k=1}^{\widehat{K}},\{\mathcal{X}_{l}\}_{l=1}^{\widehat{L}}\big)$ provides both admissible parameter ranges and localized jump regions, which will serve in the next subsection as interval constraints on the learnable parameter vector and as a refinement indicator for residual point sampling in the second-stage PINNs inverse estimator. 
	
	The above analysis is formulated for a single jump-varying physical parameter in the spatiotemporal domain. When the governing equation contains $m$ discontinuously varying coefficients, the GWS-PINNs sampling and GMM-BDMC inference procedures are applied to each parameter separately, producing parameter-specific search intervals and candidate change regions. These statistical results are then passed individually to the subsequent CCD-PINNs refinement stage, where each coefficient is constrained and refined according to its own candidate intervals and change regions. In this way, the final jump-varying parameters are learned separately while preserving the distinct discontinuity structures of different physical coefficients.
	
	\subsection{Constrained Parameter Refinement}
	Based on the statistical mixture learner, the number of discrete parameter states $\widehat{K}$, the candidate parameter intervals $\{\mathcal{I}_k\}_{k=1}^{\widehat{K}}$, and the candidate jump regions $\{\mathcal{X}_{l}\}_{l=1}^{\widehat{L}}$ have been obtained. The second stage of the proposed framework is a constrained parameter refinement, denoted by CCD-PINNs. Its role is not to relearn a smooth coefficient field, but to refine the piecewise-constant coefficient field \eqref{eq:piecewise_param} under the constraints inferred from the statistical learning step. In contrast to time-domain partitioning approaches such as network-based change-point detection for PDEs \cite{dong2024cp}, the framework can more effectively solve inverse problems involving PDEs with spatiotemporal jump coefficients guided by prior information from statistical learning.
	
	In the refinement stage, the main network $\tilde{u}(\mathbf{x},t)$ for the PDE solution is initialized with the trained main network of the GWS-PINNs approximator in the previous stage. The coefficient values are treated as trainable variables and are constrained by
	\begin{equation}
		\hat{\theta}_k\in\mathcal{I}_k, \quad
		k=1,\ldots,\widehat{K}.
		\label{eq:CCD_theta_constraint}
	\end{equation}
	Then the refinement of jump locations is performed inside the candidate regions $\{\mathcal{X}_{l}\}_{l=1}^{\widehat{L}}$. Instead of assigning an independent binary classifier to each candidate region, introducing a multiclass classifier as
	\begin{equation}
		g(\mathbf{x},t) = \left( g^{(1)} (\mathbf{x},t), \ldots, g^{(\widehat{K})}(\mathbf{x},t) \right) \in \mathbb{R}^{\widehat{K}},
	\end{equation}
	where $g^{(k)}(\mathbf{x},t)$ denotes the logit associated with the $k$-th coefficient state, $\widehat{K}$ is the number of coefficient states identified by the GMM-BDMC step. For a refinement point $(\mathbf{x}_i,t_i)$, the probability that it belongs to the $k$-th coefficient state is defined by the softmax function
	\begin{equation}
		p_{ik} =  \exp\left(g^{(k)}(\mathbf{x}_i,t_i)\right) \Big/
		\sum_{j=1}^{\widehat{K}} \exp\left(g^{(j)}(\mathbf{x}_i,t_i)\right) , \quad k=1,\ldots,\widehat{K}.
		\label{eq:CCD_softmax}
	\end{equation}
	A provisional label for $(\mathbf{x}_i,t_i)$ is assigned according to the nearest GMM state mean as
	\begin{equation}
		s_i = \arg\min_{1\le k\le \widehat{K}}
		\left| \hat{\theta}_p(\mathbf{x}_i,t_i) - \widehat{\mu}_k
		\right|.
		\label{eq:CCD_hidden_label}
	\end{equation}
	Equivalently, its one-hot representation is denoted by
	\begin{equation}
		q_{ik} = \begin{cases}
			1, & k=s_i,\\[2pt] 0, & k\ne s_i.
		\end{cases}
		\label{eq:CCD_one_hot}
	\end{equation}
	The corresponding cross entropy classification loss is defined as
	\begin{equation}
		\mathcal{L}_{\mathrm{cec}} = - \frac{1}{N_{\mathrm{cec}}}
		\sum_{i=1}^{N_{\mathrm{cec}}} \sum_{k=1}^{\widehat{K}} q_{ik}\log p_{ik},
		\label{eq:CCD_cec_loss}
	\end{equation}
	where $N_{\mathrm{cec}}$ is the number of refinement points used to train the classifier. This loss encourages each refinement point to be assigned to one of the $\widehat{K}$ coefficient states and refines the region-to-state map $\hat{z}(r)$.
	
	For one-dimensional temporal switching, the classification boundary degenerates into a set of change points. Therefore, training the classifier inside $\{\mathcal{X}^{\tau}_{l}\}_{l=1}^{\widehat{L}}$ gives the refined temporal jump locations
	\begin{equation}
		\{\widehat{\tau}_l\}_{l=1}^{\widehat{L}}, \quad
		\widehat{\tau}_l\in\mathcal{X}^{\tau}_{l},
		\label{eq:CCD_tau_estimate}
	\end{equation}
	where $ 0<\widehat{\tau}_1<\widehat{\tau}_2<\cdots<\widehat{\tau}_{\widehat{L}}<T$. For multidimensional spatial or spatiotemporal discontinuities, the same classification procedure yields an estimate of the discontinuity boundary associated with \eqref{jump_discontinuities} as
	\begin{equation}
		\widehat{\mathcal{J}}_{\theta} = \bigcup_{l=1}^{\widehat{L}}
		\widehat{\mathcal{J}}_{l}, \quad \widehat{\mathcal{J}}_{l} \subset \mathcal{X}_{l}.
		\label{eq:CCD_boundary_estimate}
	\end{equation}
	Thus, the candidate jump regions supplied by the statistical learner are further refined into either pointwise change locations or multidimensional jump interfaces.
	
	Let $\tilde{\theta}_p(\mathbf{x},t)$ denote the current hard piecewise-constant coefficient estimator during the CCD-PINNs refinement stage. With the refined coefficient values and the refined partition, the discontinuous coefficient is represented as a hard piecewise-constant function
	\begin{equation}
		\tilde{\theta}_p(\mathbf{x},t) = \sum_{r=1}^{\widehat{R}} \hat{\theta}_{\hat{z}(r)} \mathbf{1}_{\widehat{\mathcal Q}_r}(\mathbf{x},t), \quad \hat{z}(r) \in\{1,\dots,\widehat{K}\},
		\label{eq:CCD_hard_theta}
	\end{equation}
	where the subdomains $\{\widehat{\mathcal{Q}}_r\}_{r=1}^{\widehat{R}}$ are determined by $\{\widehat{\tau}_l\}_{l=1}^{\widehat{L}}$ in one-dimensional temporal cases and by $\widehat{\mathcal{J}}_{\theta}$ in multidimensional cases. Additionally, $\widehat{R}$ is the number of refined piecewise-constant subdomains, and $\hat{z}(r)$ assigns the $r$-th estimated subdomain to one of the $\widehat{K}$ coefficient states.
	
	Then the physics residual in Stage~2 CCD-PINNs is obtained by replacing the smooth coefficient sub network with the hard piecewise-constant coefficient as
	\begin{equation}
		\mathcal{L}_{\mathrm{phy}}^{\mathrm{ccd}} = \frac{1}{N_{\mathrm{ccd}}}
		\sum_{i=1}^{N_{\mathrm{ccd}}} \left\| \mathcal{F} \left[ \tilde{u}(\mathbf{x}_{\mathrm{res}}^{(i)},t_{\mathrm{res}}^{(i)});
		\tilde{\theta}_{p}(\mathbf{x}_{\mathrm{res}}^{(i)},t_{\mathrm{res}}^{(i)}) \right] - f(\mathbf{x}_{\mathrm{res}}^{(i)},t_{\mathrm{res}}^{(i)}) \right\|_2^2, \quad (\mathbf{x}_{\text{res}}^{(i)}, t_{\text{res}}^{(i)}) \in \mathcal{Q}^{\circ}.
		\label{eq:CCD_phy_loss}
	\end{equation}
	The residual set contains uniformly sampled interior points and additional points sampled from the candidate jump regions, which improves the resolution of the physics loss near discontinuities and provides additional points for training the local classifier.
	
	Combining the data, physics, boundary, initial, and classification terms, the total loss of CCD-PINNs is defined as
	\begin{equation}
		\mathcal{L}_{\mathrm{est}} =
		\gamma_0 \mathcal{L}_{\mathrm{data}} + \gamma_1 \mathcal{L}_{\mathrm{phy}}^{\mathrm{ccd}} +
		\gamma_2 \mathcal{L}_{\mathrm{bc}} + \gamma_3 \mathcal{L}_{\mathrm{ic}} + \gamma_4\mathcal{L}_{\mathrm{cec}},
		\label{eq:CCD_total_loss}
	\end{equation}
	where $\gamma_4$ is a new nonnegative weight. The final output of the constrained refinement stage is the hard piecewise-constant coefficient estimator associated with \eqref{eq:piecewise_param}. The refined solution $\tilde{u}(\mathbf{x},t)$ produced by the main network is also returned.
	
	For the PDE cases with multiple discontinuously varying parameters, suppose that there are $m$ coefficient fields to be estimated, denoted by $\{\theta_p^{(j)}(\mathbf{x},t)\}_{j=1}^{m}$. The constrained refinement stage outputs a hard piecewise-constant estimator for each parameter separately, given by
	\begin{equation}
		\left( \tilde{\theta}_p^{(1)}(\mathbf{x},t), \cdots,
		\tilde{\theta}_p^{(m)}(\mathbf{x},t) \right) = \left(
		\sum_{r=1}^{\widehat{R}^{(1)}} 
		\hat{\theta}^{(1)}_{\hat{z}^{(1)}(r)}
		\mathbf{1}_{\widehat{\mathcal{Q}}^{(1)}_r}(\mathbf{x},t),\: \cdots,\:
		\sum_{r=1}^{\widehat{R}^{(m)}} 
		\hat{\theta}^{(m)}_{\hat{z}^{(m)}(r)}
		\mathbf{1}_{\widehat{\mathcal{Q}}^{(m)}_r}(\mathbf{x},t)
		\right).
		\label{eq:CCD_multi_param_estimator}
	\end{equation}
	Here, $\widehat{R}^{(j)}$, $\widehat{\mathcal{Q}}^{(j)}_r$, and $\hat{z}^{(j)}(r)$ denote the number of refined subdomains, the $r$-th estimated subdomain, and the corresponding region-to-state map for the $j$-th parameter, respectively. Each parameter is refined according to its own heuristic search intervals and candidate change regions obtained from the statistical mixture modeling step.
	
	In summary, this framework provides a two-stage methodology for identifying discontinuously varying coefficients in PDEs. In the first stage, GWS-PINNs is employed, in which a main network approximates the PDE solution and a sub network generates a smooth surrogate of the underlying piecewise-constant coefficient field. The resulting coefficient samples are then processed using a Bayesian Gaussian mixture model combined with a BDMC method to identify the discrete coefficient levels and localize candidate transition regions. In the second-stage CCD-PINNs, the continuous surrogate is replaced with a finite-dimensional, piecewise-constant representation. The coefficient values are refined within the heuristic search intervals $\{\mathcal{I}_k\}_{k=1}^{\widehat{K}}$, while the change points or discontinuity boundaries are refined within the candidate regions $\{\mathcal{X}_{l}\}_{l=1}^{\widehat{L}}$ through constrained local multiclass classification. The main novelty of the proposed framework lies in integrating adaptive residual weighting, sub-network-based coefficient sampling, Bayesian mixture modeling, and classification-based constrained refinement to reconstruct discontinuous coefficients accurately while reducing the search space and improving stability near discontinuities.
	
	\section{Numerical Example}\label{sec3}
	To validate the effectiveness of the proposed two-stage JVC-PINNs framework for inverse problems involving PDEs with discontinuously varying coefficients, this section presents a series of numerical experiments on representative spatiotemporal dynamical systems. The proposed method first employs GWS-PINNs to generate a smooth surrogate of the unknown coefficient field. The resulting samples are then processed by the GMM-BDMC statistical learner to infer the number of discrete coefficient states, heuristic search intervals, and candidate transition regions. Finally, CCD-PINNs refines the piecewise-constant coefficient values and discontinuity locations under these statistical constraints. The numerical experiments therefore evaluate both the coefficient-sampling capability of the first-stage network and the constrained parameter-estimation accuracy of the second-stage refinement model. The numerical cases considered in this study are adapted from existing studies, and the reference solutions are generated using a high-precision finite-element method.
	
	For PDEs with discontinuously time-varying coefficients, this section considers wave, heat, Burgers, and Navier-Stokes equations. For each equation, multiple coefficient-jump scenarios are designed to represent abrupt parameter changes, multi-state switching behavior, and different jump magnitudes that may arise in complex physical systems. The numerical results are used to assess the accuracy, stability, and robustness of the proposed framework in recovering both the discrete coefficient levels and the corresponding temporal change points. In addition, the refined solution network produced by CCD-PINNs is examined to determine whether the recovered discontinuous coefficient structure preserves the physical consistency of the reconstructed PDE solution.
	
	To further assess the applicability of the proposed method to discontinuously varying spatial coefficients, particularly in multidimensional settings, this section further considers the wave and Helmholtz equations as representative examples. The coefficients are assumed to be piecewise constant over both regular and irregular spatial subdomains. These examples test whether the proposed pipeline can infer discrete spatial coefficient states and refine the corresponding discontinuity interfaces. The results demonstrate the stability of the proposed JVC-PINNs framework for identifying and estimating jump-varying coefficients across different PDE types, jump patterns, and spatial geometries.
	
	Stage~1 used GWS-PINNs with a main solution network $\hat{u}$ and an auxiliary coefficient sub network $\hat{\theta}_p$. Stage~2 used CCD-PINNs, initialized the solution network using the trained Stage~1 network, and replaced the smooth coefficient sub network with a constrained piecewise-constant function sub network $\tilde{\theta}_p$. The observation data, physics residual, boundary condition, initial condition, and classification terms were weighted by $\gamma_0$, $\gamma_1$, $\gamma_2$, $\gamma_3$, and $\gamma_4$, respectively. The observation weight was fixed at $\gamma_0=1$, while $\gamma_1$, $\gamma_2$ and $\gamma_3$ were selected separately for each case without $\gamma_4$ used, as reported in each PDE case. Stage~2 generally used the same weight settings as Stage~1, with the additional classification weight fixed at $\gamma_4=50$ when the classification loss was activated. When used, the classification term was a cross-entropy loss, and the classification sub network used ReLU activations. All other networks used tanh activations. Unless stated otherwise, the gradient-adaptive weighting hyperparameter was set to $\beta=1$, the Adam optimizer was used, and random seeds were fixed for NumPy, PyTorch, and CUDA.
    
    The GMM-BDMC learner used $K_{\max}=5$, an initial value of $K=3$, birth rate $\lambda_b=1$, stopping time $200$, $25$ inner Gibbs updates, burn-in length $M_0=2000$, posterior sample size $M=4000$, $c_0=2.5$, $C_0=0.5s_y^2$, and $b_0=\bar{y}$. Here, $\bar{y}$ and $s_y^2$ denote the sample mean and variance of the coefficient samples $\{y_n\}_{n=1}^{N}$ generated by GWS-PINNs. A symmetric Dirichlet prior $(1,\ldots,1)$ was used for the mixture weights. Details of the neural-network training are provided after the introduction of each PDE type. All numerical experiments were conducted on a computer equipped with an NVIDIA GeForce RTX 5060 Ti GPU using PyTorch 2.11.0 and cu130. The code used in this study is available at \url{https://github.com/Zhikun0416/Inverse-Problem-for-Partial-Differential-Equations-with-Jump-Discontinuities-in-Coefficients}.
	
	\subsection{Wave Equation}
	First, consider a 1+1D nonlinear wave equation, which means define in one-dimensional space and one-dimensional time, namely the sine-Gordon equation, with a time-varying physical coefficient $\alpha(t)>0$, which denotes the squared wave speed. This example is adapted from \cite{yuan2023machine} and the governing equation is given by
	\begin{equation}
		\begin{cases}
			u_{tt} = \alpha(t) u_{xx} + \sin u, & \quad (x,t)\in U \times (0,T],\\[2pt]
			u(x,t) = 0, & \quad (x,t)\in \partial U \times (0,T],\\[2pt]
			u_t(x,0) = 0, & \quad x \in U,\\[2pt]
			u(x,0) = g(x), & \quad x \in U.
		\end{cases}
	\end{equation}
	Let $U=[0,1]$, $T=10$, and the initial function be
	\begin{equation}
		g(x)=\sin \pi x.
	\end{equation}
	Two different scenarios of time-varying coefficients are considered.
	
	\textbf{Case~1.1:}
	\begin{equation}
		\alpha(t)=1, \quad t\in[0,10].
	\end{equation}
	
	\textbf{Case~1.2:}
	\begin{equation}
		\alpha(t)=
		\begin{cases}
			0.5, & t\in[0,5), \\[2pt]
			1, & t\in[5,10].
		\end{cases}
	\end{equation}
	
\begin{figure}[p]
		\centering
		\subfloat[Spatiotemporal solution]{
			\begin{minipage}[t]{0.44\linewidth}
				\includegraphics[width=1\linewidth]{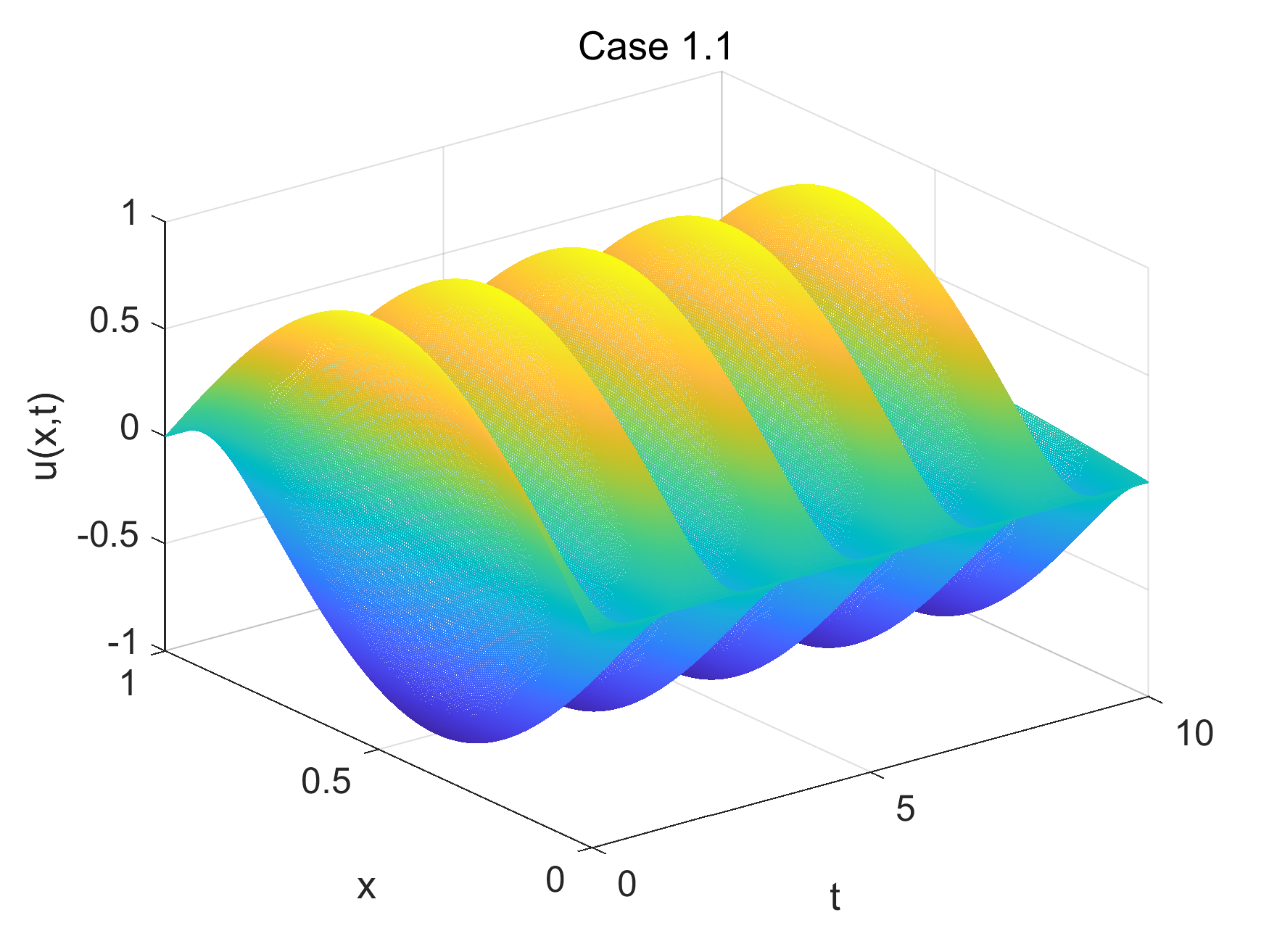}
			\end{minipage}
		}
		\subfloat[Parameter inverse result]{
			\begin{minipage}[t]{0.44\linewidth}
				\includegraphics[width=1\linewidth]{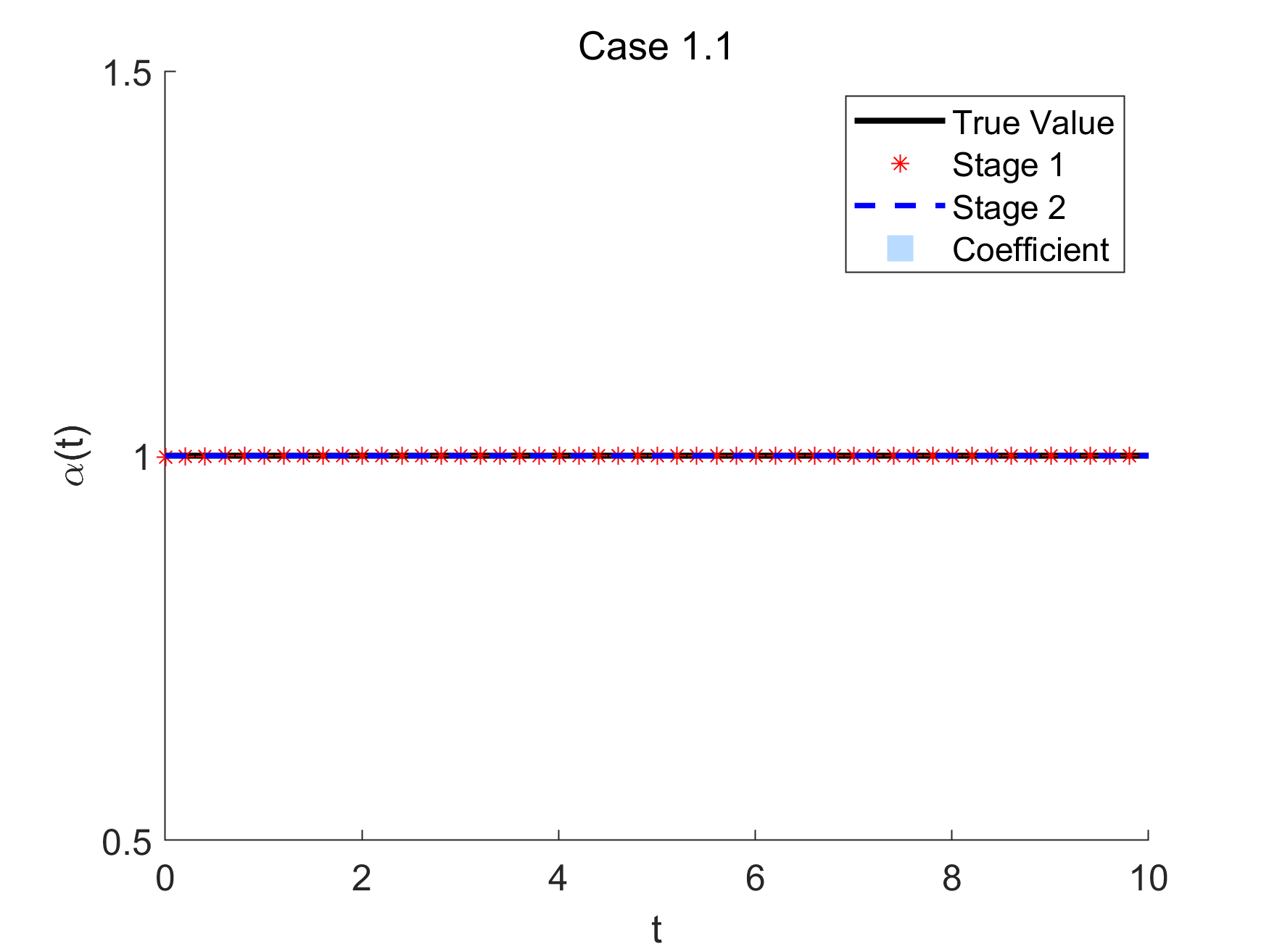}
			\end{minipage}
		}\\
		\subfloat[Reference solution]{
			\begin{minipage}[t]{0.315\linewidth}
				\includegraphics[width=1\linewidth]{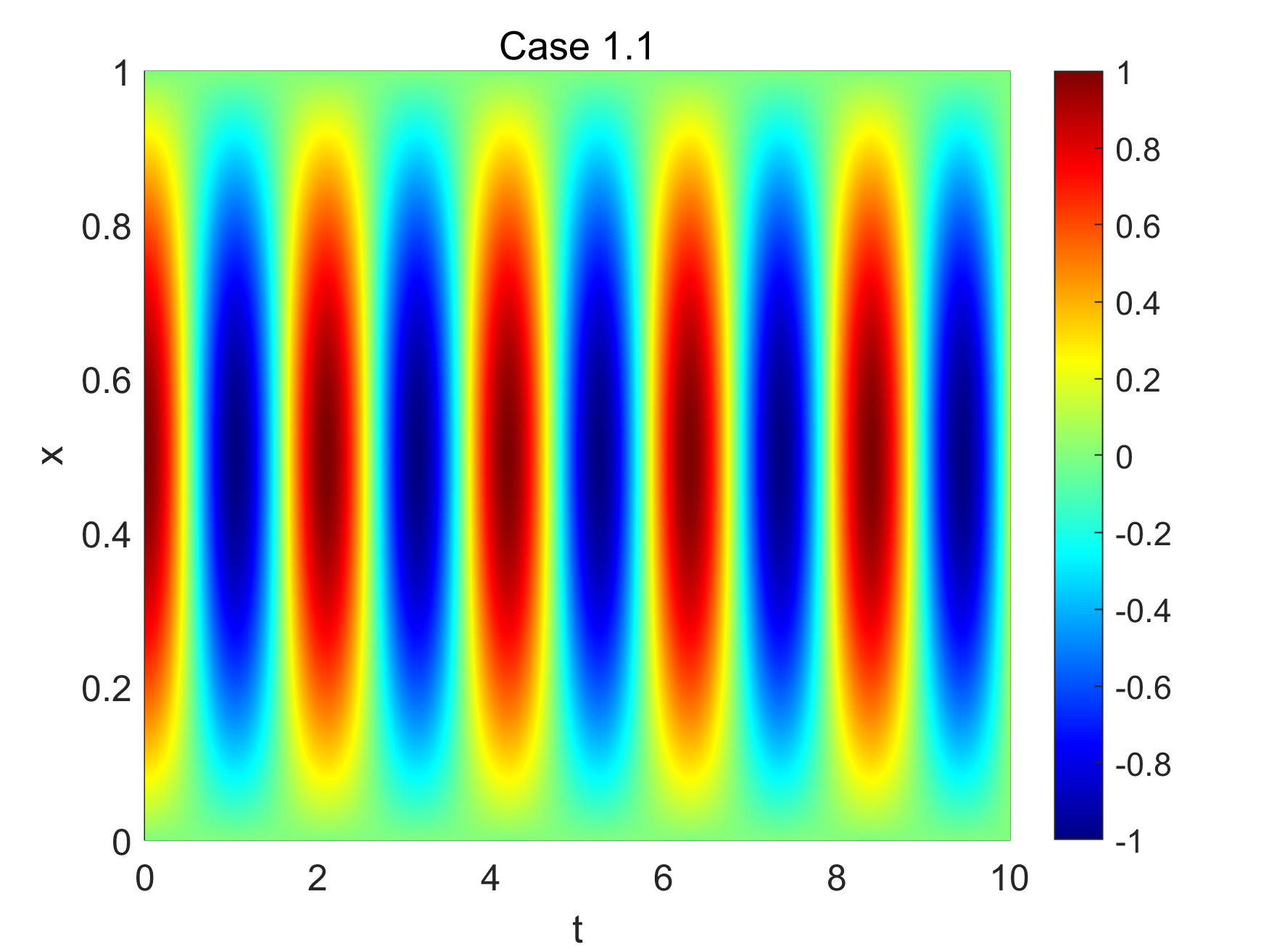}
			\end{minipage}
		}
		\subfloat[Predicted solution]{
			\begin{minipage}[t]{0.315\linewidth}
				\includegraphics[width=1\linewidth]{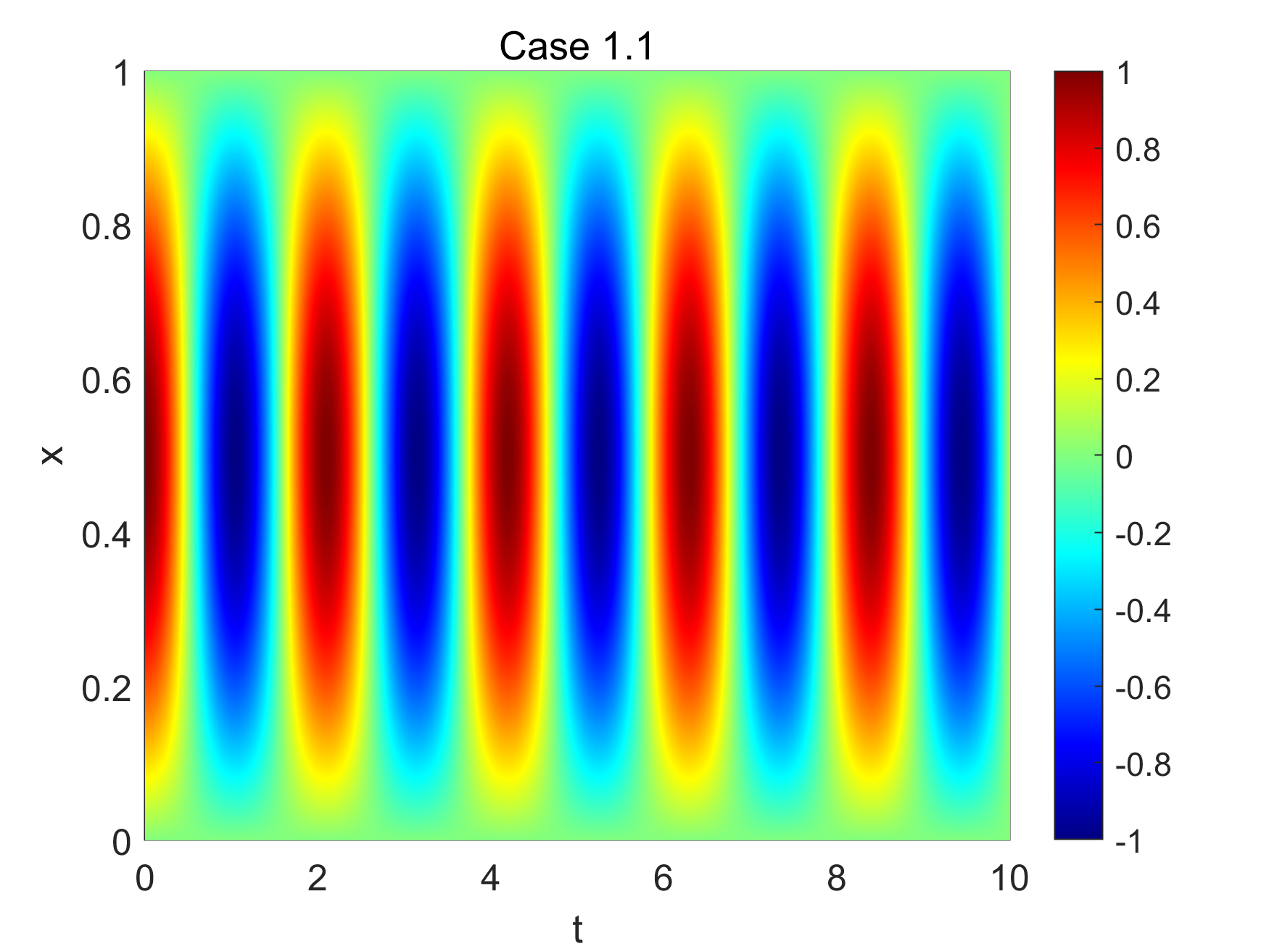}
			\end{minipage}
		}
		\subfloat[Absolute error]{
			\begin{minipage}[t]{0.315\linewidth}
				\includegraphics[width=1\linewidth]{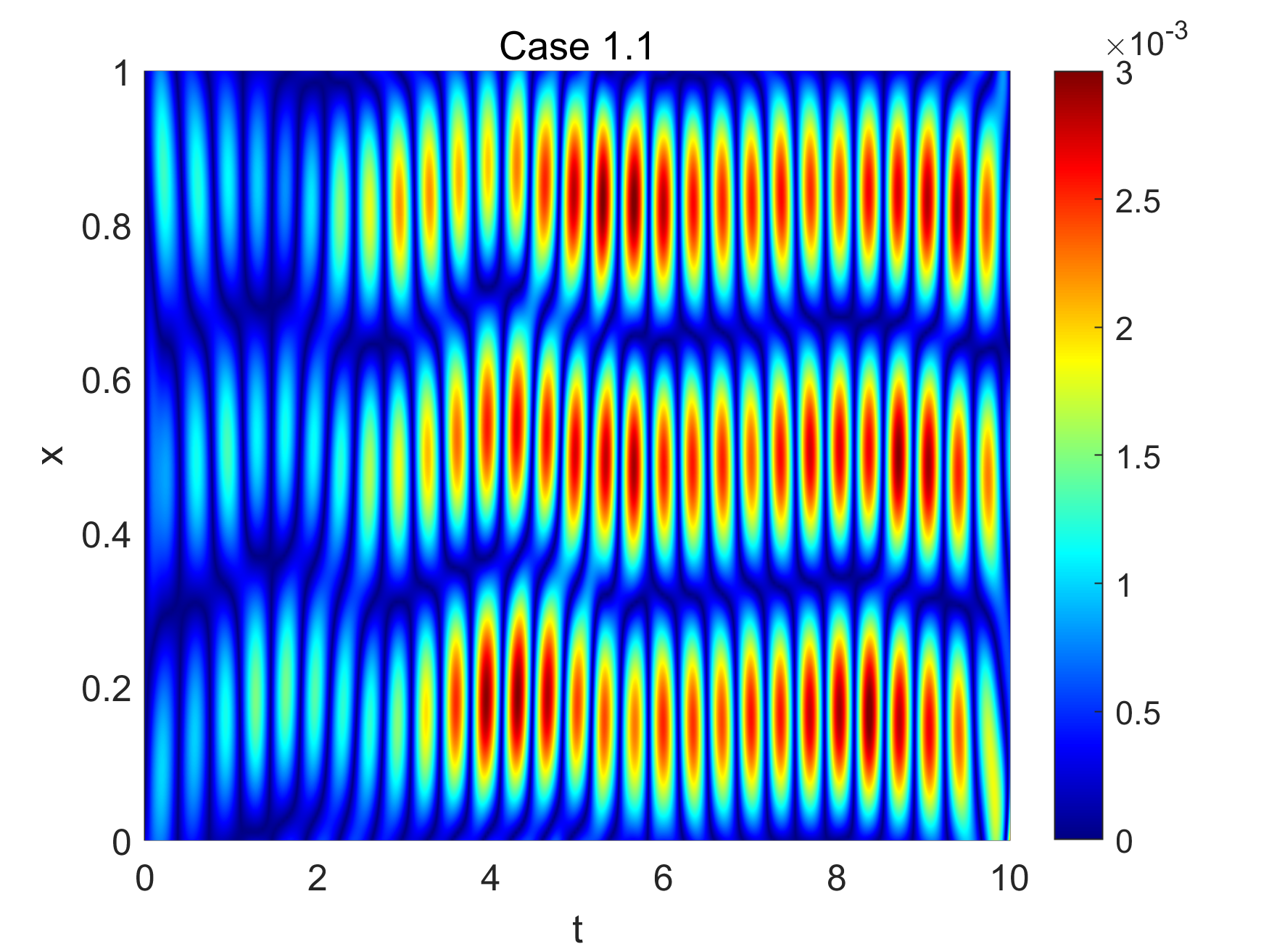}
			\end{minipage}
		}\\
		\subfloat[Spatiotemporal solution]{
			\begin{minipage}[t]{0.44\linewidth}
				\includegraphics[width=1\linewidth]{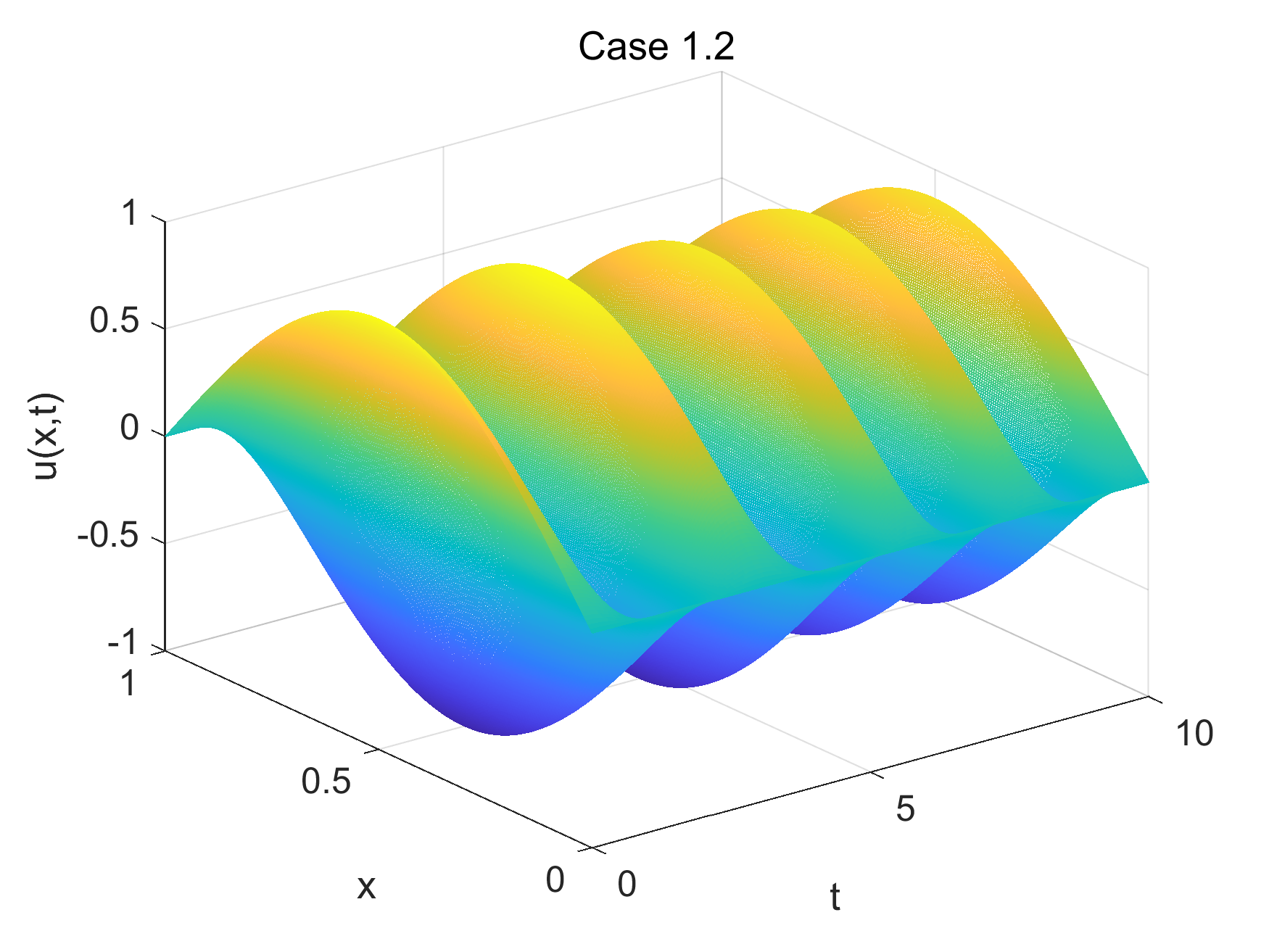}
			\end{minipage}
		}
		\subfloat[Parameter inverse result]{
			\begin{minipage}[t]{0.44\linewidth}
				\includegraphics[width=1\linewidth]{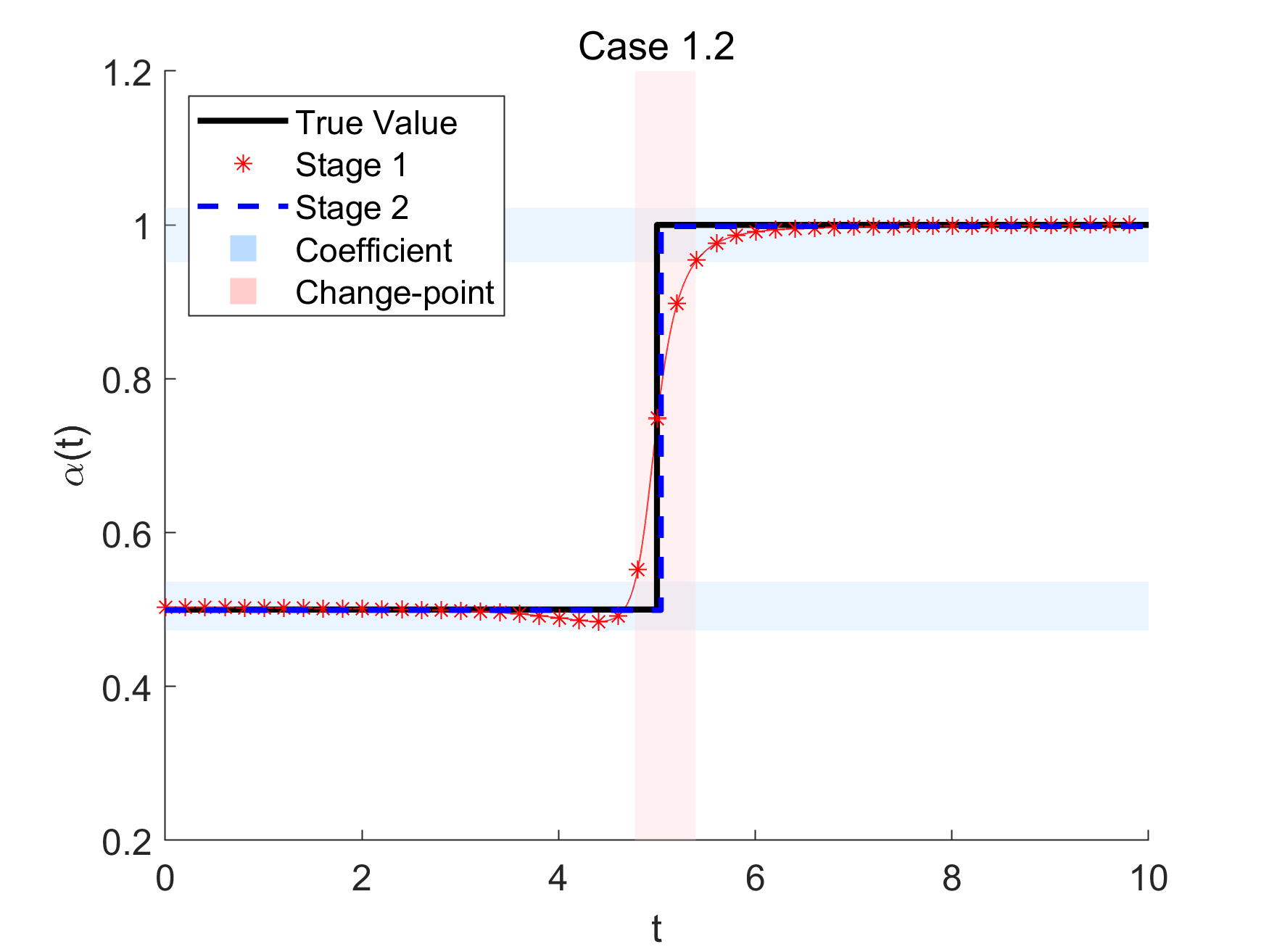}
			\end{minipage}
		}\\
		\subfloat[Reference solution]{
			\begin{minipage}[t]{0.315\linewidth}
				\includegraphics[width=1\linewidth]{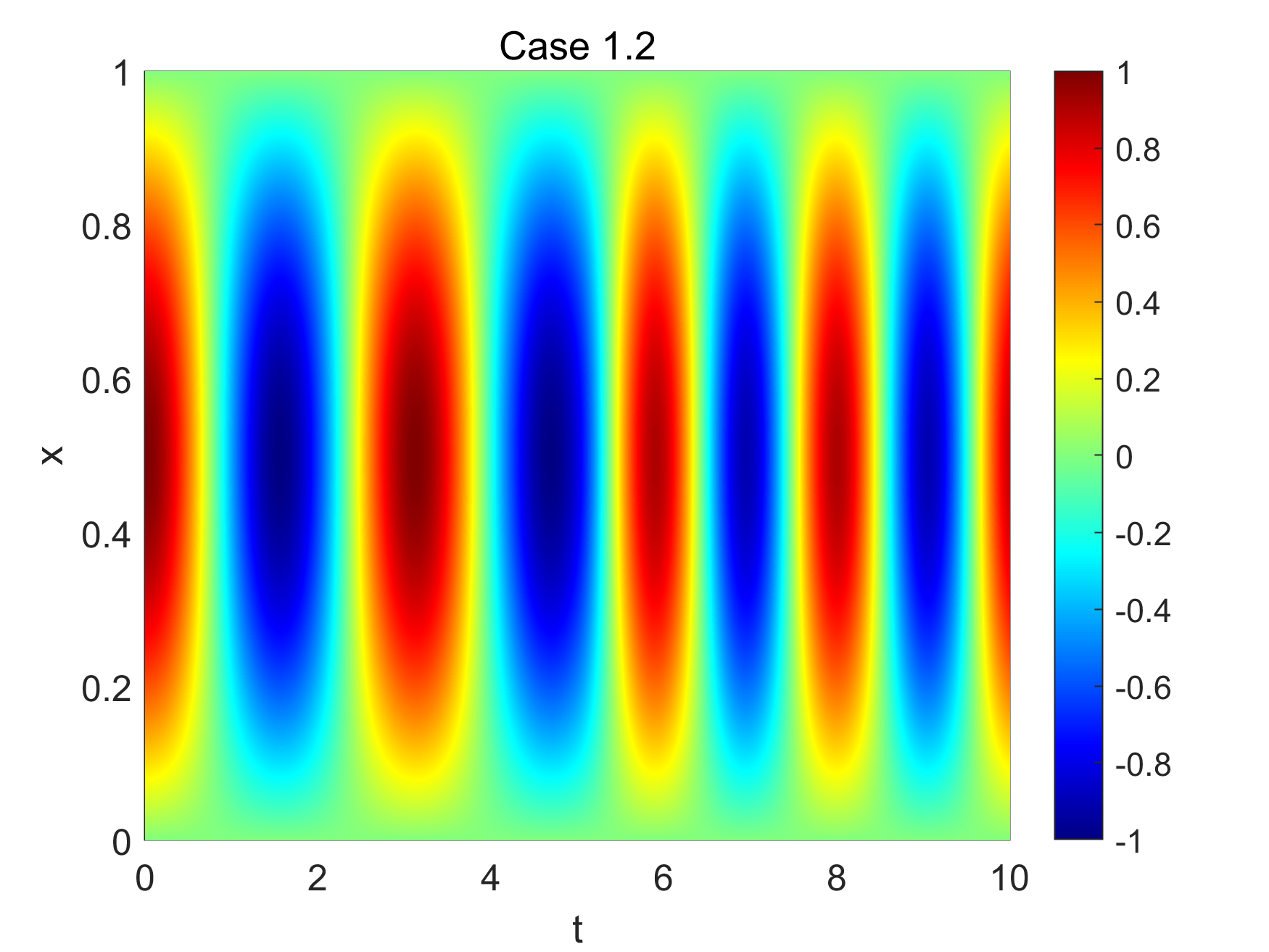}
			\end{minipage}
		}
		\subfloat[Predicted solution]{
			\begin{minipage}[t]{0.315\linewidth}
				\includegraphics[width=1\linewidth]{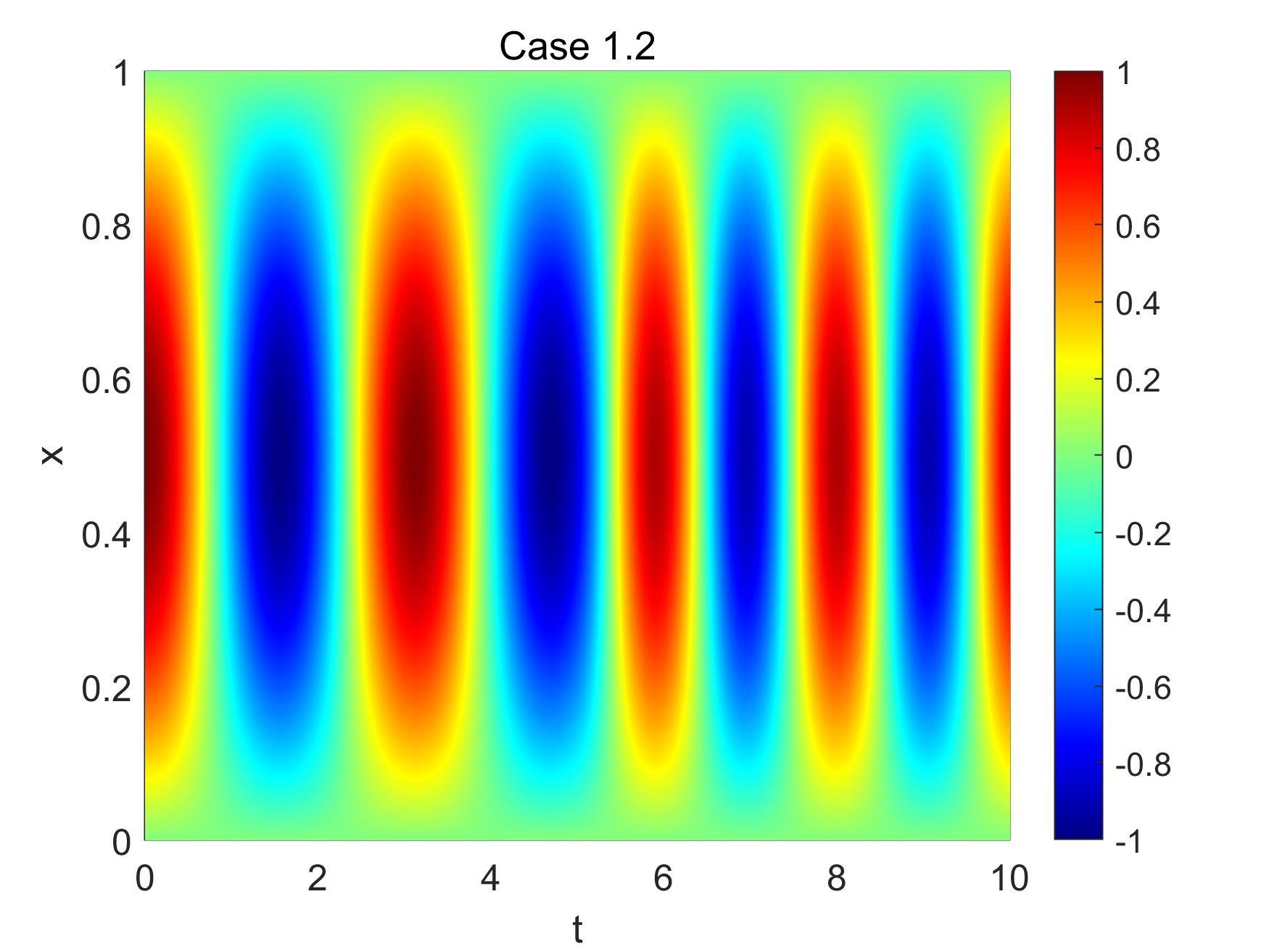}
			\end{minipage}
		}
		\subfloat[Absolute error]{
			\begin{minipage}[t]{0.315\linewidth}
				\includegraphics[width=1\linewidth]{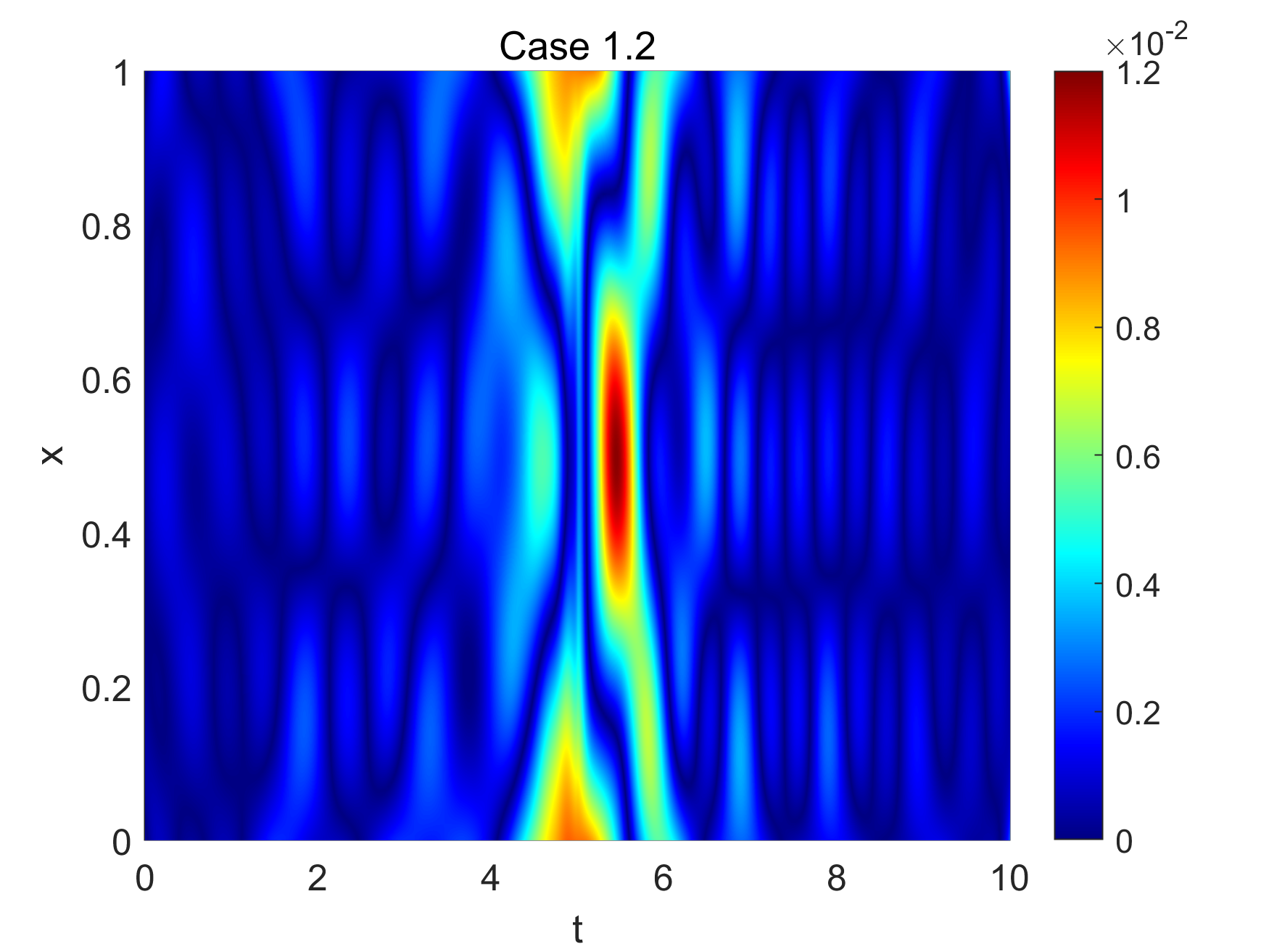}
			\end{minipage}
		}
		\caption{Numerical results for wave equations with discontinuously time-varying coefficient $\alpha(t)$. \label{fig1}}
	\end{figure}

	The spatiotemporal solution fields of these two wave equations and the corresponding two-stage parameter inversion results for the coefficient $\alpha(t)$ are shown in Figure~\ref{fig1}. In the parameter inversion plots, the Stage~1 result represents the soft continuous approximation generated by the GWS-PINNs coefficient sub network. Since the sub network is a continuous neural function, the Stage~1 approximation may smooth the true discontinuity and form a transition layer near the jump location. The Stage~2 result represents the hard piecewise-constant estimator refined by CCD-PINNs under the statistical constraints supplied by GMM-BDMC. The solid black line represents the true coefficient values. The red markers and blue dashed lines represent the coefficient estimates obtained in Stage~1 and Stage~2, respectively. The blue shaded bands denote the heuristic coefficient search intervals inferred from the posterior GMM components, while the red shaded bands denote the candidate temporal change-point intervals used in the constrained refinement stage.
	
	The mean squared error (MSE) between the reference solution and the predicted solution is defined as
	\begin{equation}
		MSE_{\hat{u}} = \frac{1}{N_{\text{obs}}} \sum_{i=1}^{N_{\text{obs}}} \Big[ \hat{u}(\mathbf{x}_{\text{obs}}^{(i)},t_{\text{obs}}^{(i)}) - u(\mathbf{x}_{\text{obs}}^{(i)},t_{\text{obs}}^{(i)}) \Big]^2,
	\end{equation}
	where $\hat{u}$ denotes the solution network obtained from the first-stage GWS-PINNs, which can be replaced with $\tilde{u}$ from the second-stage CCD-PINNs. The reported MSEs, the estimated parameter values and the percentage of signed relative errors are rounded to four decimal places where appropriate, where errors are computed using the unrounded parameter estimates. The same precision setting is adopted for the remaining numerical examples.
	
	Case~1.1 and Case~1.2 used random seed $101$ with $\theta_p=\alpha(t)$ on $(x,t)\in[0,1]\times[0,10]$. The Stage~1 main network was a Fourier-feature network with layers $[2,200,200,1]$, and the coefficient network had layers $[1,50,50,50,50,1]$. Adam ran for $50000$ iterations. The learning rate for $\hat{u}$ was $10^{-3}$, and the learning rate for $\hat\theta_p$ was $5\times10^{-4}$ in Case~1.1 and $10^{-3}$ in Case~1.2. The training sets used $50\times80=4000$ residual grid points, $6000$ strict interior observation points, $4000$ random initial-condition points on $t=0$, and $4000$ random boundary points split across the two spatial boundaries. The remaining loss weights were $\gamma_1=1$, $\gamma_2=1$, $\gamma_3=0.5$. In Stage~2, $\alpha(t)$ was represented by a hard step function initialized using the GMM-BDMC intervals and the stable plateau means obtained in Stage~1. The same Adam learning rate $10^{-4}$ was used for the main network and sub network.
	
	For Case~1.1, the GMM-BDMC statistical learner identifies $\widehat{K}=1$, indicating that the coefficient contains only one constant state. Therefore, no temporal change point is introduced in the second-stage refinement. In this situation, CCD-PINNs automatically degenerates into a constrained constant-coefficient inverse estimator, and the estimated coefficient remains consistent with the reference value. This result confirms that the proposed framework does not artificially introduce discontinuities when the underlying PDE coefficient is constant.
	
	For Case~1.2, the GMM-BDMC learner identifies two coefficient states and provides search intervals for the two parameter values, together with a candidate temporal change-point interval. The Stage~1 coefficient approximation in GWS-PINNs is continuous and thus produces a smooth transition near the true jump time. However, this relaxed approximation still contains sufficient statistical information for the GMM-BDMC model to separate the underlying coefficient states. The subsequent CCD-PINNs refinement further converts the continuous surrogate into a sharp step-function representation and yields a refined change point within the candidate interval. Therefore, the Stage~2 estimator provides an explicit and interpretable reconstruction of the discontinuously time-varying coefficient.
	
	The reconstructed solution based on the second-stage main network $\tilde{u}$ and the comparison of its absolute error with the reference solution are shown in the corresponding rows of Figure~\ref{fig1}. The results indicate that the reconstructed solution is close to the reference solution and that the absolute error is well controlled. These observations demonstrate that the hard piecewise-constant coefficient refined by CCD-PINNs can also preserve the physical consistency of the solution approximation. The effectiveness of the proposed algorithm will be further validated in subsequent multidimensional spatial examples.
	
	\begin{figure}[p]
		\centering
		\subfloat[Reference solution t=0]{
			\begin{minipage}[t]{0.315\linewidth}
				\includegraphics[width=1\linewidth]{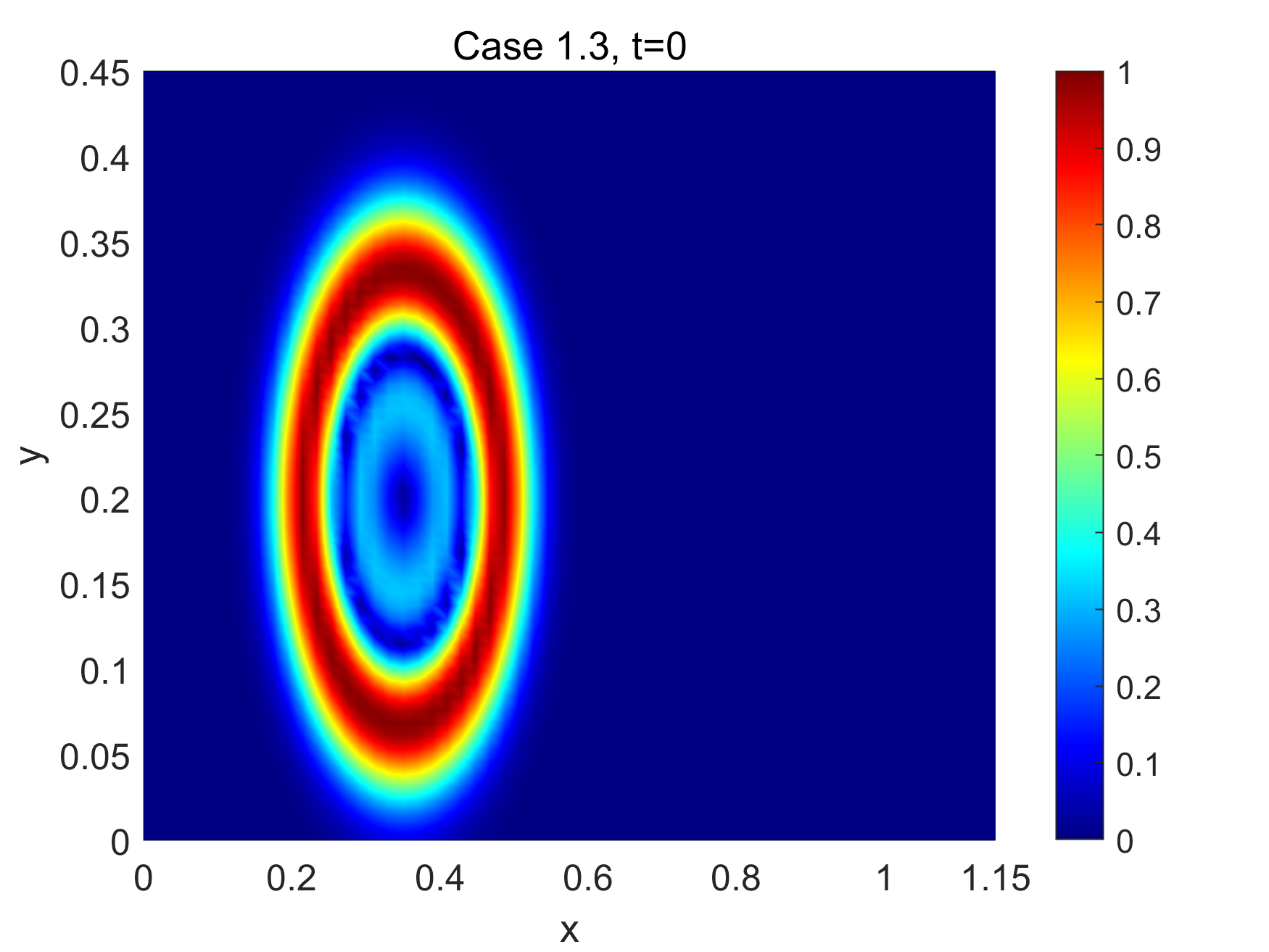}
			\end{minipage}
		}
		\subfloat[Predicted solution t=0]{
			\begin{minipage}[t]{0.315\linewidth}
				\includegraphics[width=1\linewidth]{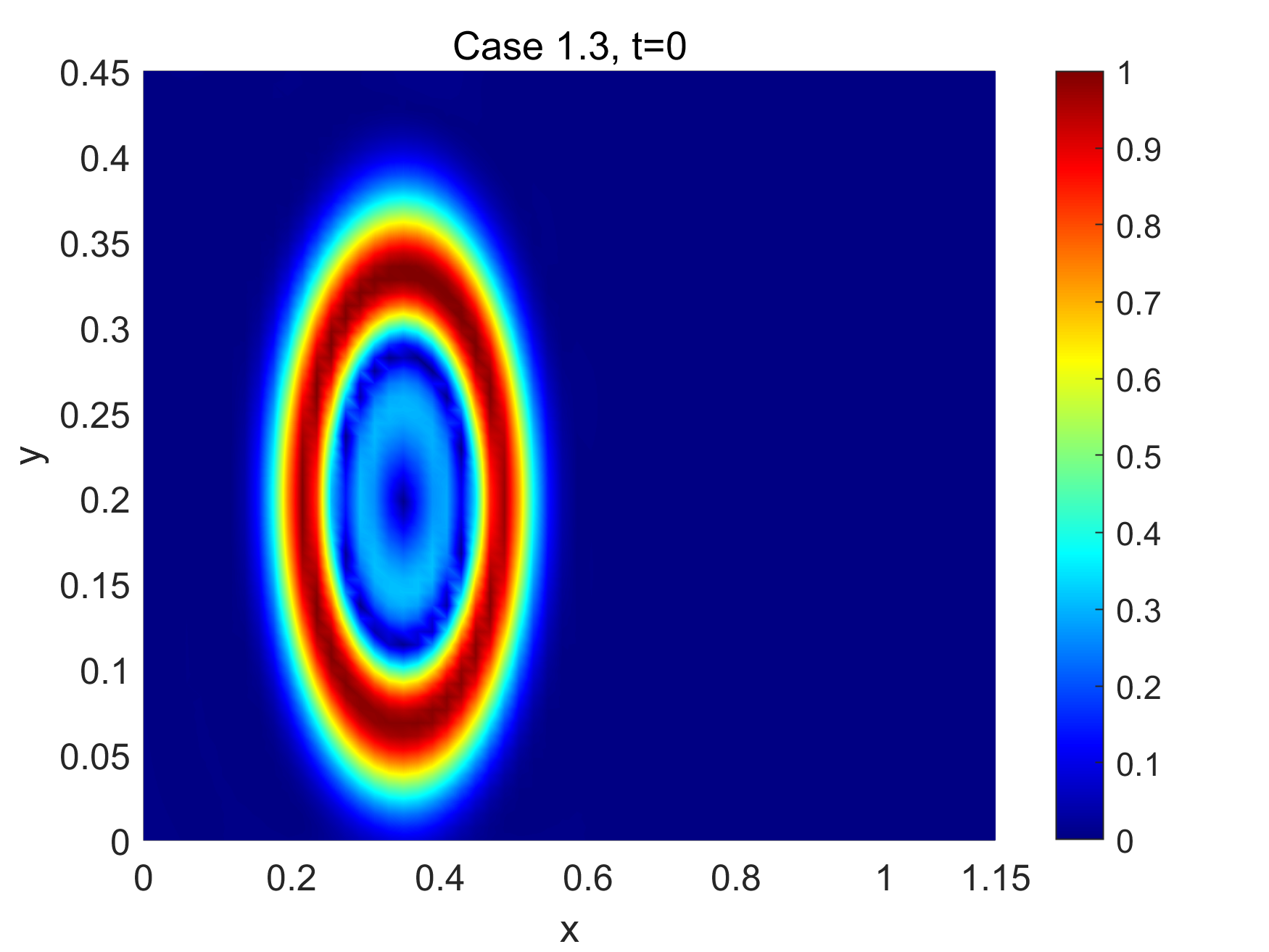}
			\end{minipage}
		}
		\subfloat[Absolute error t=0]{
			\begin{minipage}[t]{0.315\linewidth}
				\includegraphics[width=1\linewidth]{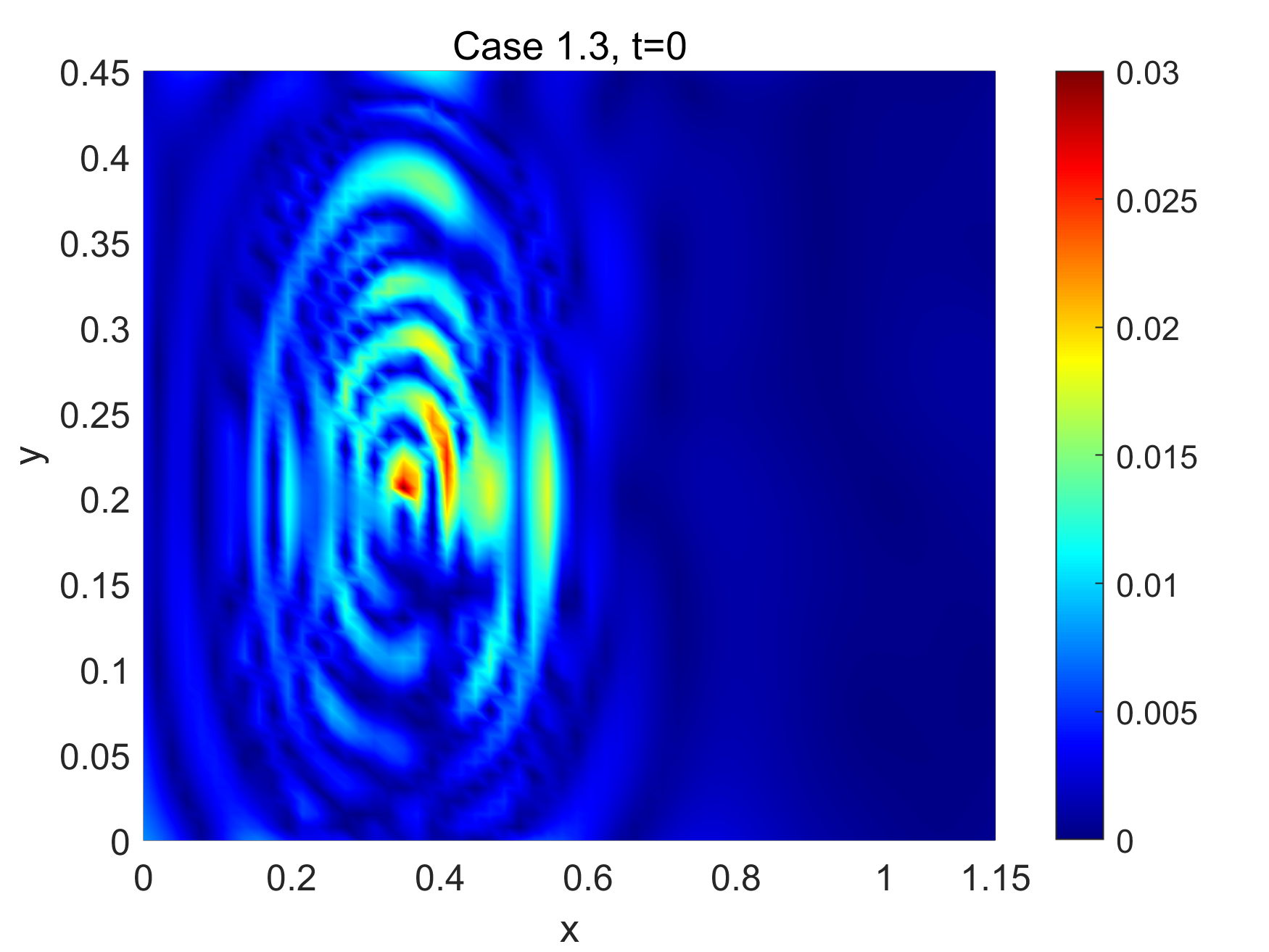}
			\end{minipage}
		}\\
		\subfloat[Reference solution t=0.15]{
			\begin{minipage}[t]{0.315\linewidth}
				\includegraphics[width=1\linewidth]{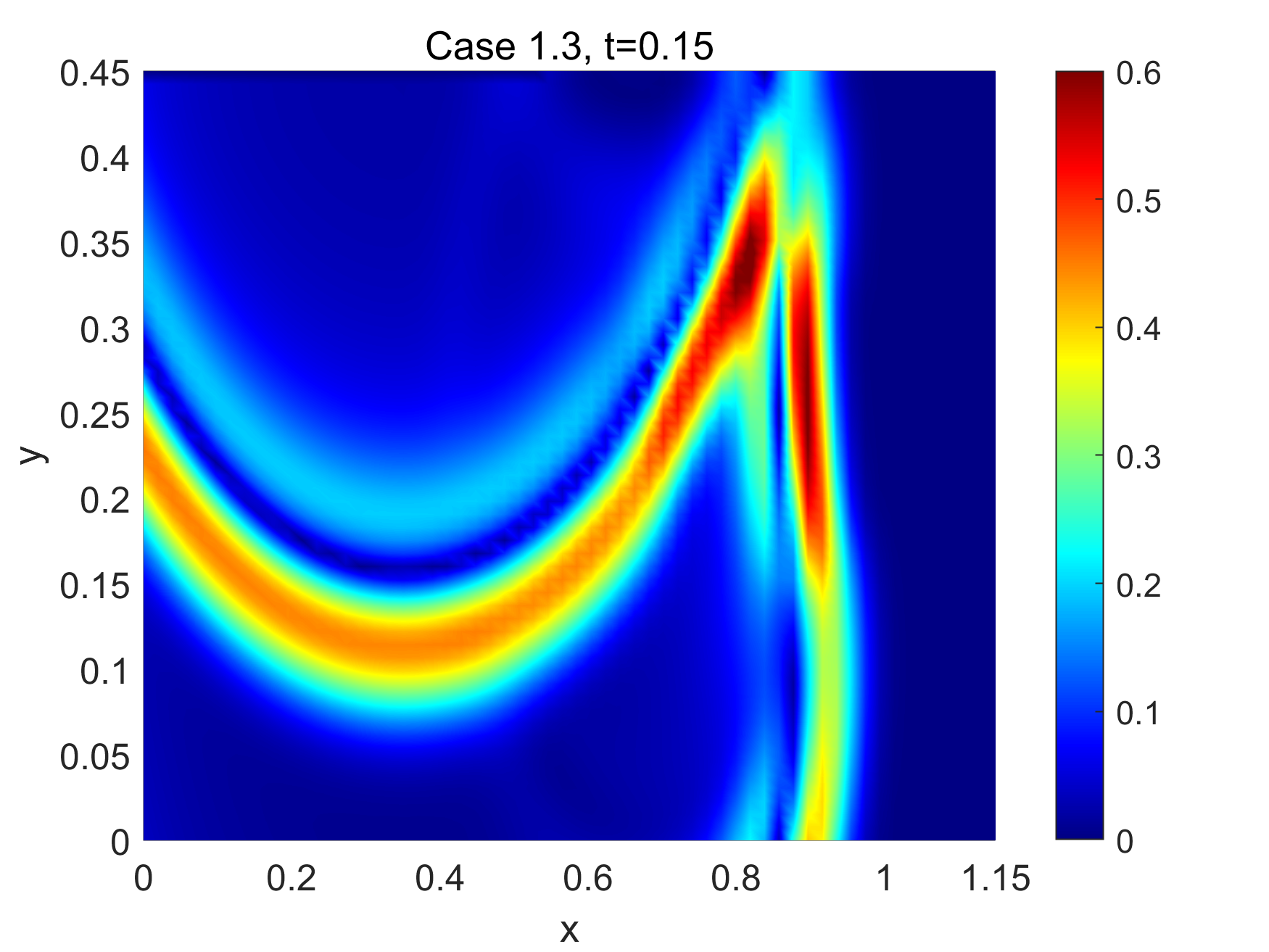}
			\end{minipage}
		}
		\subfloat[Predicted solution t=0.15]{
			\begin{minipage}[t]{0.315\linewidth}
				\includegraphics[width=1\linewidth]{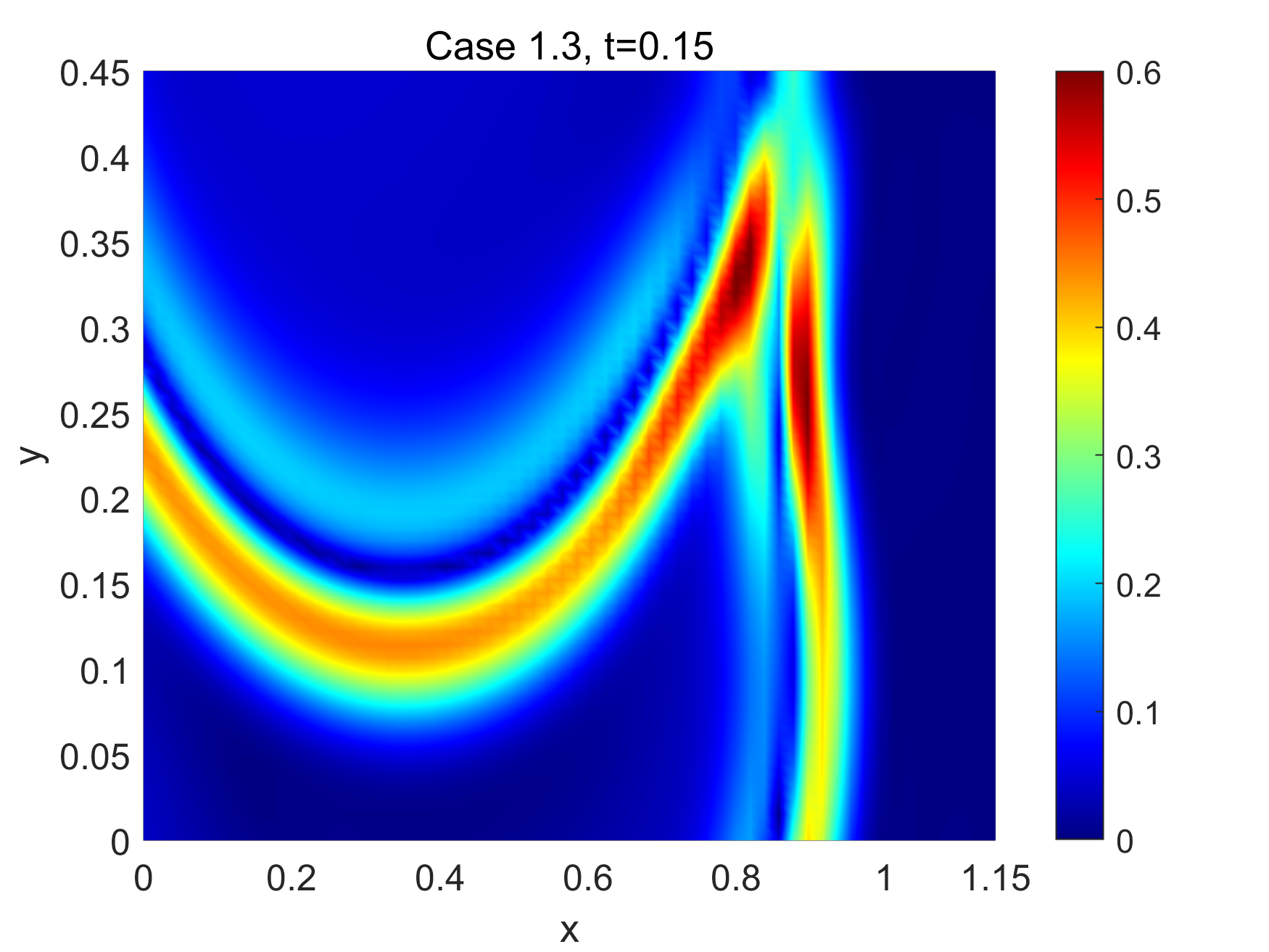}
			\end{minipage}
		}
		\subfloat[Absolute error t=0.15]{
			\begin{minipage}[t]{0.315\linewidth}
				\includegraphics[width=1\linewidth]{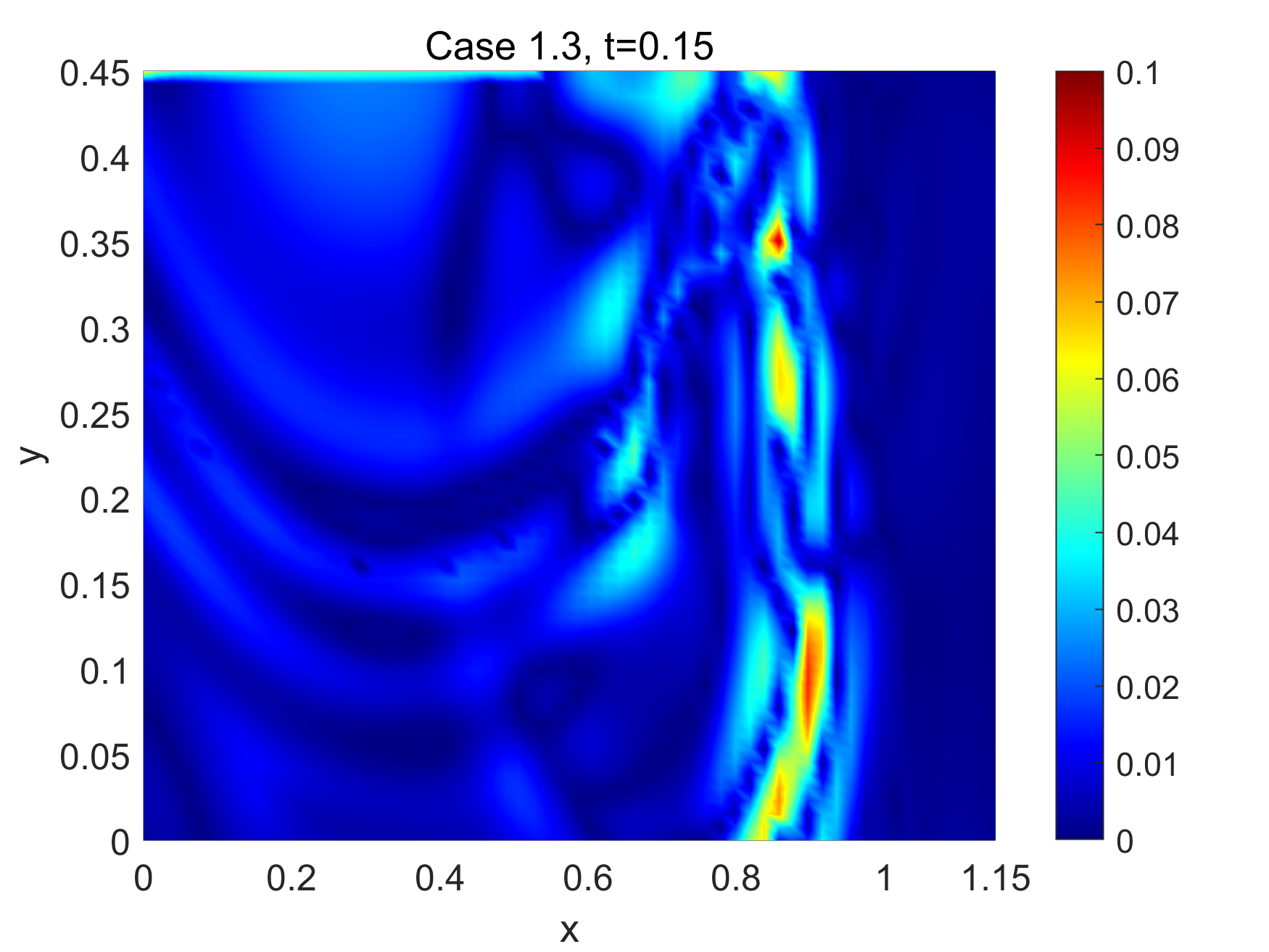}
			\end{minipage}
		}\\
		\subfloat[Reference value $\alpha(x,y)$]{
			\begin{minipage}[t]{0.315\linewidth}
				\includegraphics[width=1\linewidth]{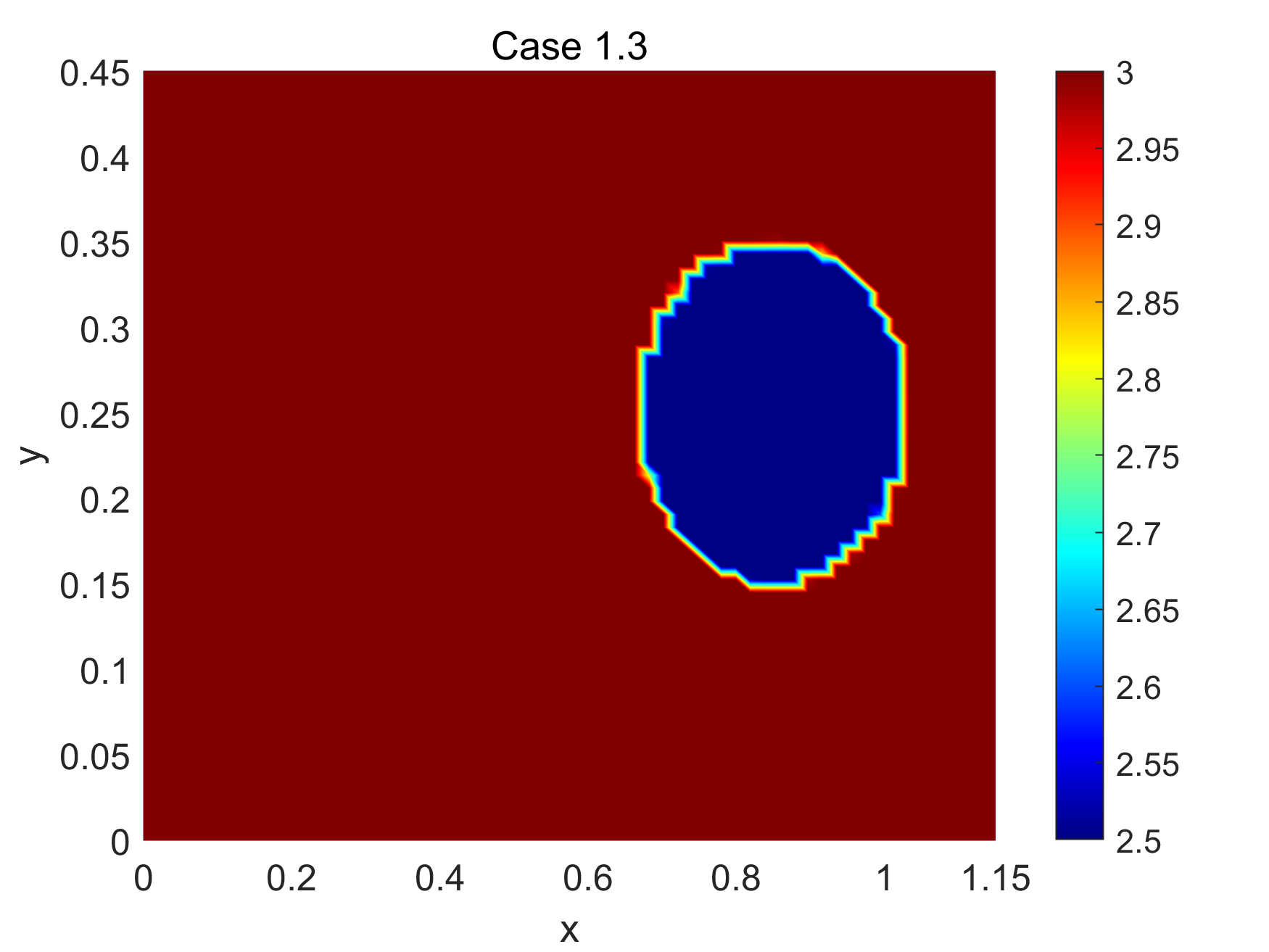}
			\end{minipage}
		}
		\subfloat[Parameter inverse result]{
			\begin{minipage}[t]{0.315\linewidth}
				\includegraphics[width=1\linewidth]{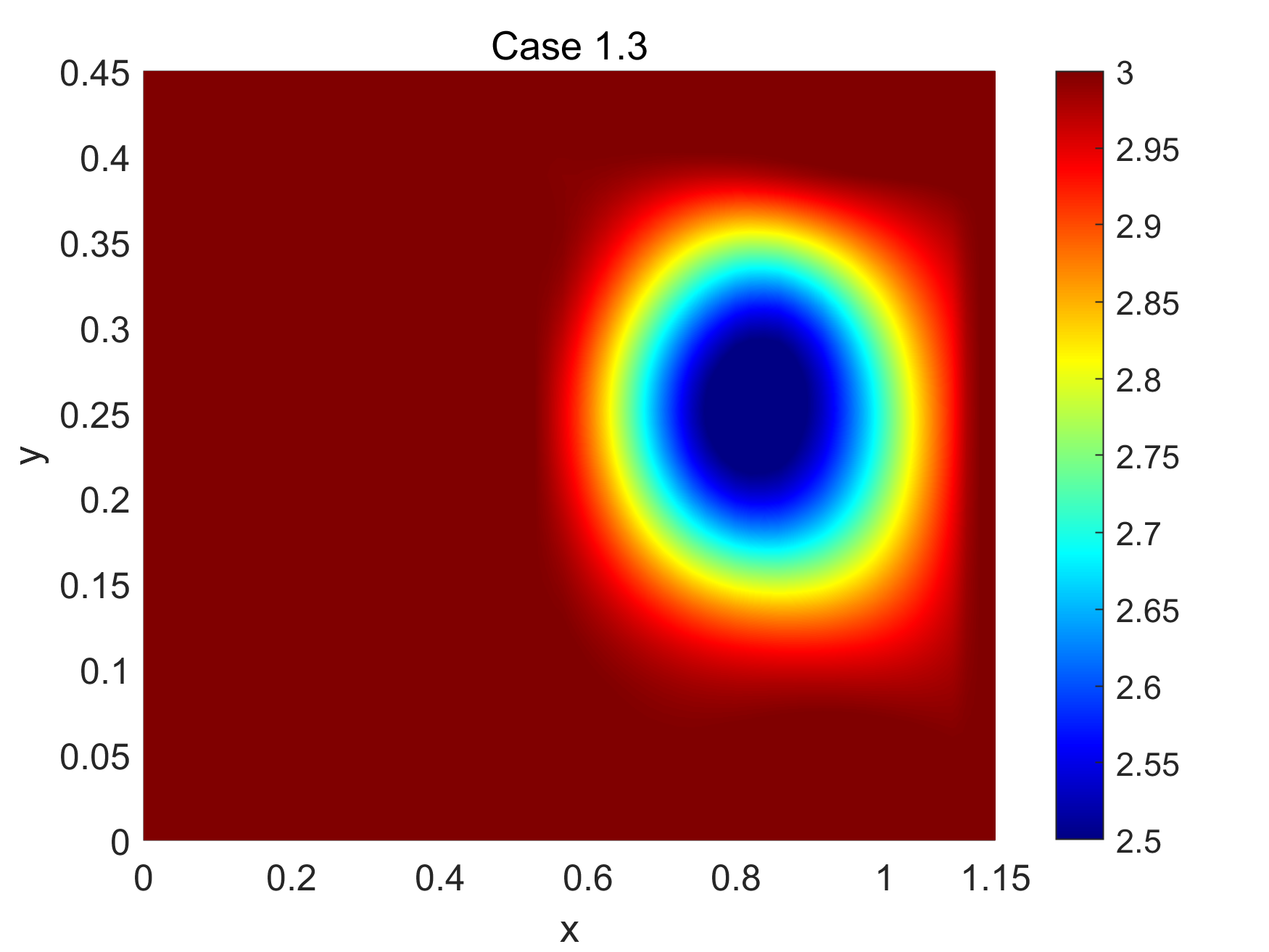}
			\end{minipage}
		}
		\subfloat[Stage~1 absolute error]{
			\begin{minipage}[t]{0.315\linewidth}
				\includegraphics[width=1\linewidth]{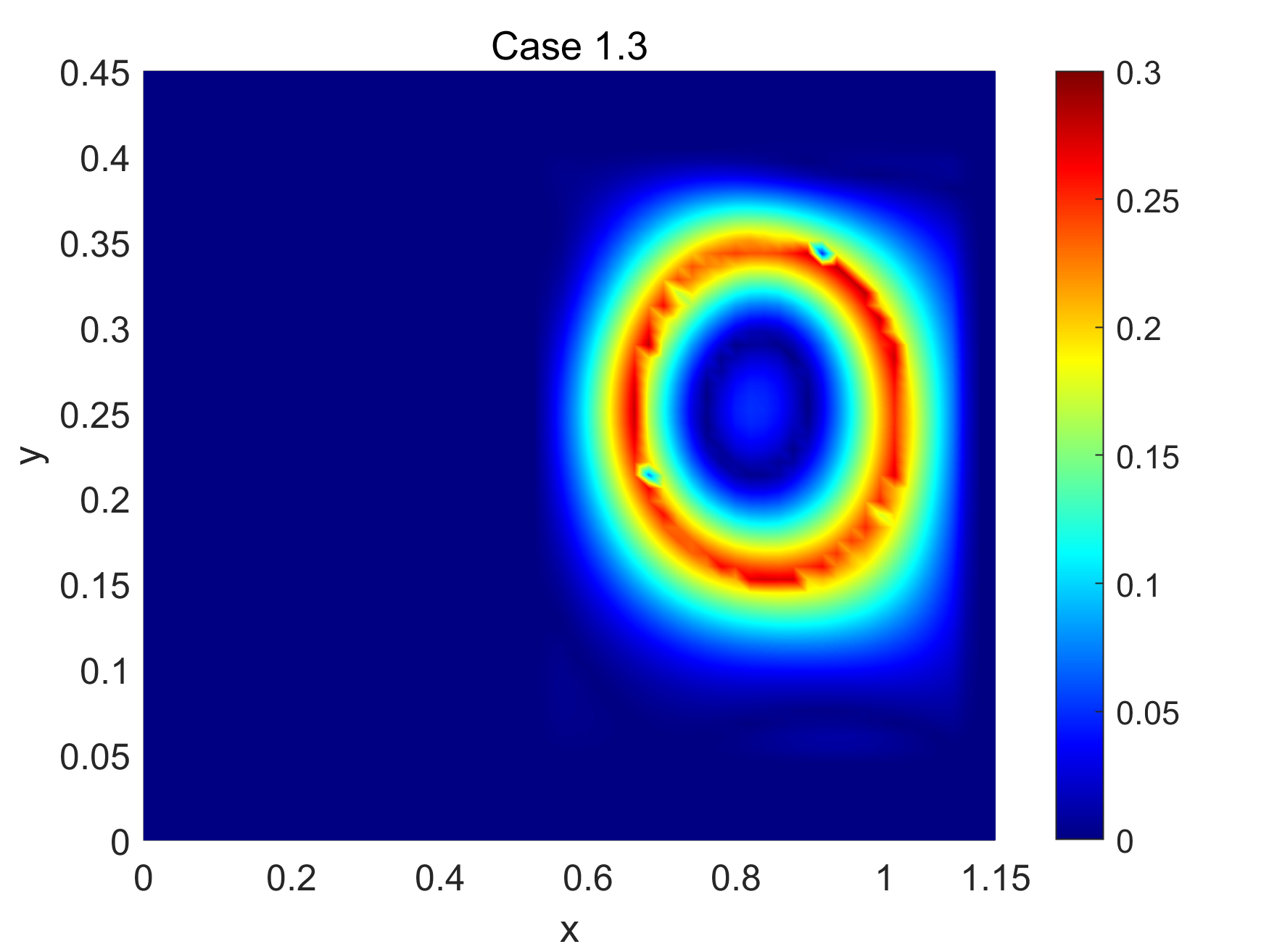}
			\end{minipage}
		}\\
		\subfloat[Parameter estimate result]{
			\begin{minipage}[t]{0.315\linewidth}
				\includegraphics[width=1\linewidth]{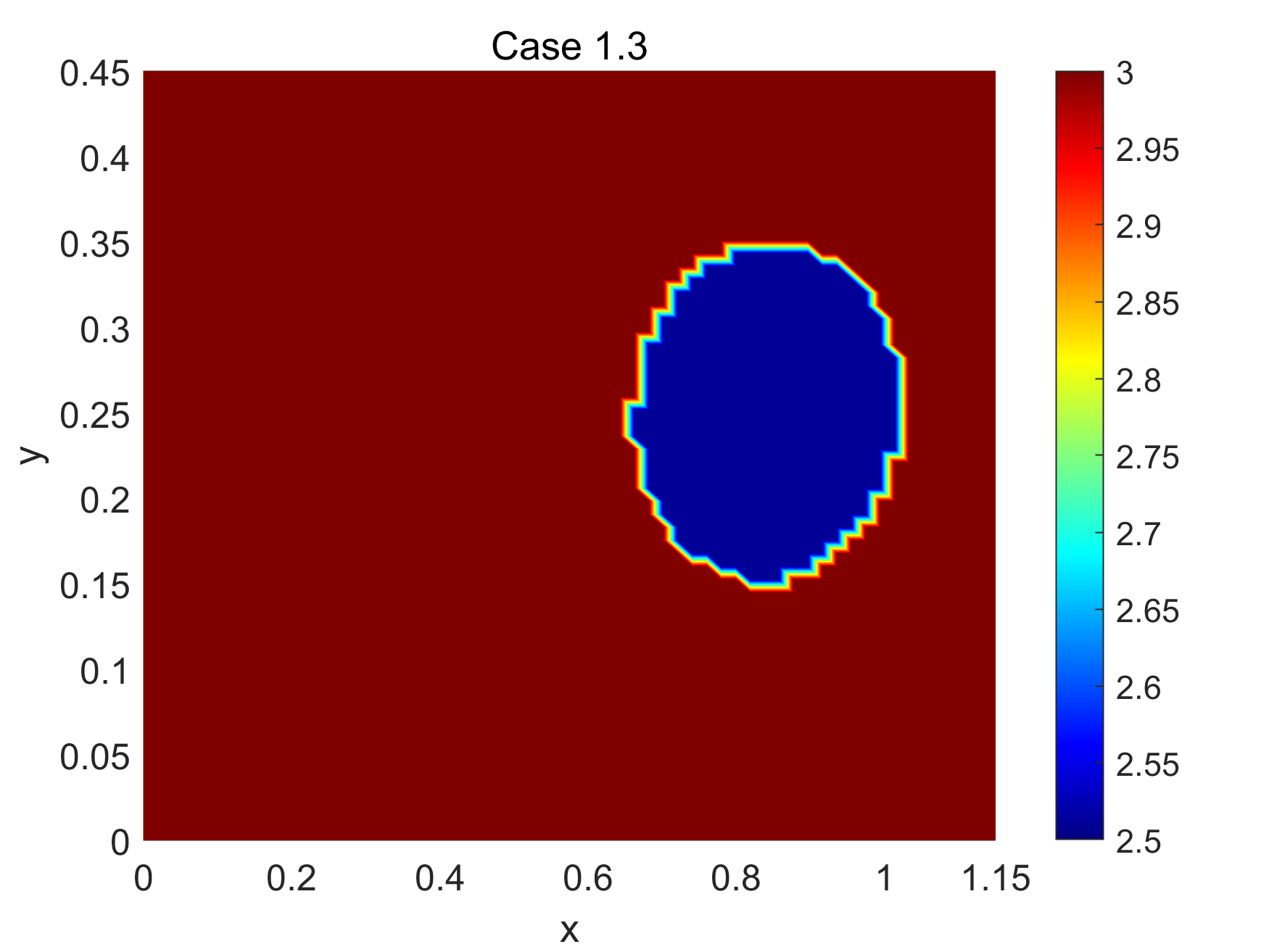}
			\end{minipage}
		}
		\subfloat[Stage~2 absolute error]{
			\begin{minipage}[t]{0.315\linewidth}
				\includegraphics[width=1\linewidth]{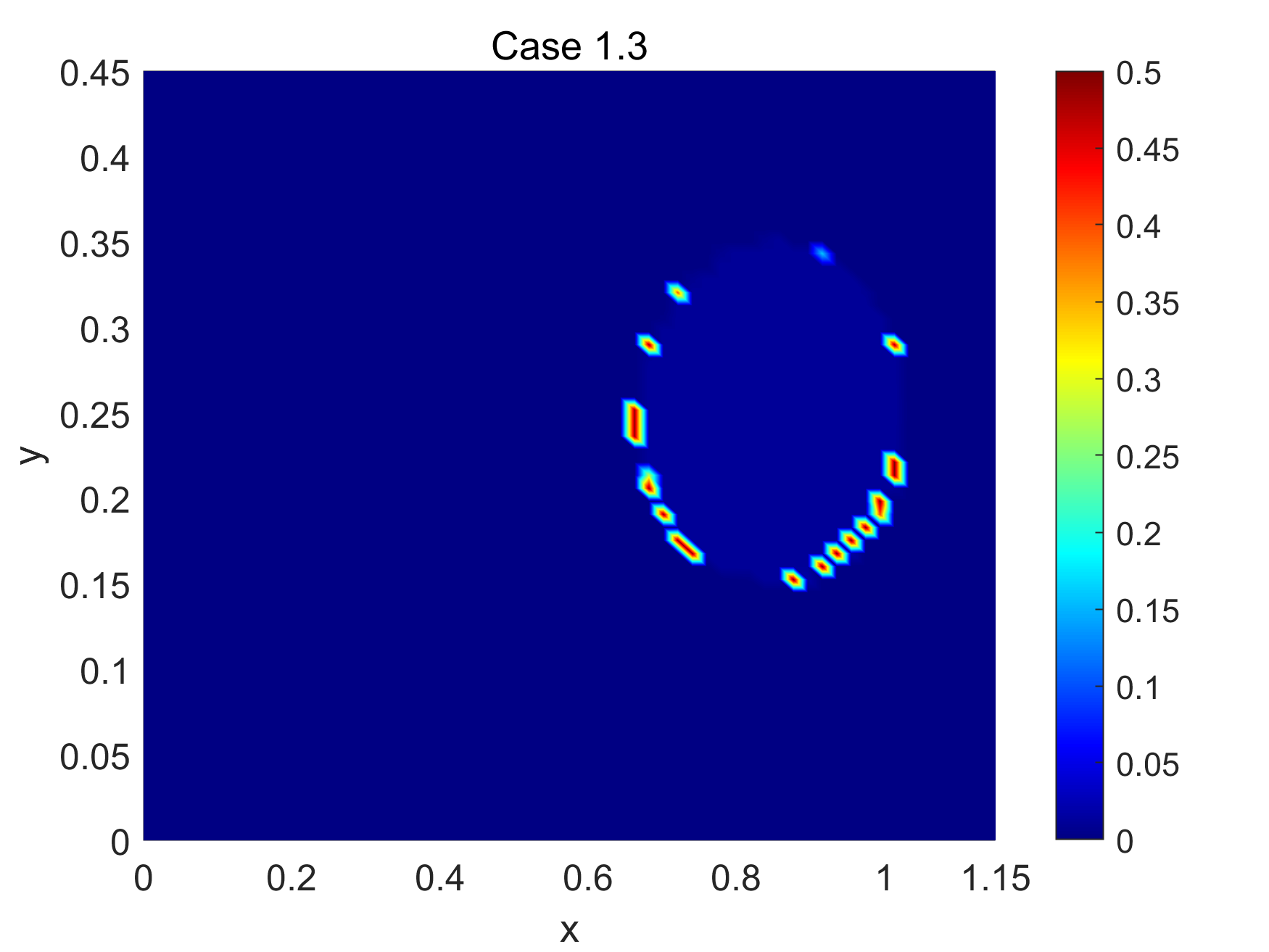}
			\end{minipage}
		}
		\subfloat[Change boundary detection]{
			\begin{minipage}[t]{0.315\linewidth}
				\includegraphics[width=1\linewidth]{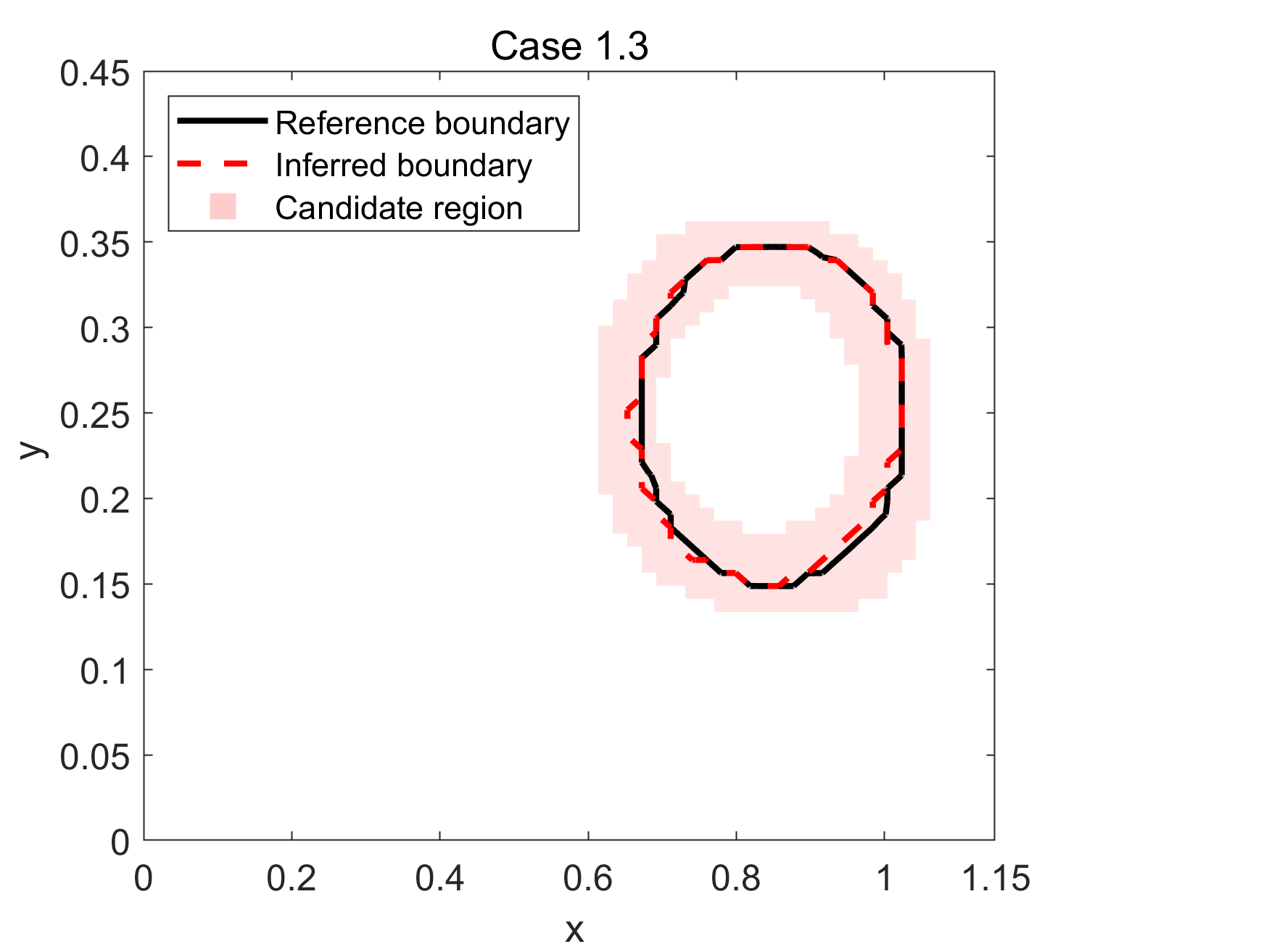}
			\end{minipage}
		}
		\caption{Numerical results for the wave equation with discontinuously 2D space-varying coefficient $\alpha(x,y)$. \label{fig2}}
	\end{figure}
	
	Next, a 2+1D linear wave equation with a 2D space-varying coefficient $\alpha(x,y)$ is considered, following \cite{rasht2022physics}. The governing equation is
	\begin{equation}
		u_{tt}=\alpha(x,y)\Delta u, \quad (x,y,t)\in U\times(0,T],
	\end{equation}
	where $U=[0,1.15]\times[0,0.45]\subset\mathbb{R}^2$ and $T=0.2$. The wave formation and propagation process in this spatial domain is illustrated in Figure~\ref{fig2}. In this numerical example, an approximately circular irregular subdomain $U_2\subset U$ is located on the right side of the spatial domain, and the coefficient $\alpha(x,y)$ denotes the squared wave speed and takes different values inside and outside this region. The specific setting is as follows.
	
	\textbf{Case~1.3:}
	\begin{equation}
		\alpha(x,y)=
		\begin{cases}
			2.5, & (x,y)\in U_2,\\[2pt]
			3, & (x,y)\in U\setminus U_2.
		\end{cases}
	\end{equation}

    For Case~1.3, the Stage~1 solution network used tanh layers $[3,100,\ldots,100,1]$ with eight hidden layers, while the coefficient sub network used the architecture $[2,20,20,20,20,20,1]$. Both networks were trained with Adam for $200000$ epochs using a learning rate of $10^{-4}$. The PDE residual batches contained $40000$ Sobol points and were refreshed every $200$ epochs. In addition, $3600$ snapshot points per frame and $5000$ free-surface points were used. The random seed was set to $101$. The remaining loss weights were $\gamma_1=\gamma_2=\gamma_3=0.1$. The Case~1.3 detector used a $60\times60$ patch grid over the spatial domain, and the subsequent Stage~2 refinement was performed for $1000$ iterations with a learning rate of $5\times10^{-4}$. The training data consisted only of reference solutions at the three time slices $t=0$, $t=0.015$, and $t=0.15$.
	
	The quantitative results of the proposed JVC-PINNs framework for three cases are summarized in Table~\ref{tab1}-\ref{tab4}. Specifically, Table~\ref{tab1} reports the two-stage coefficient inversion results, where the Stage~1 column gives the coefficient search intervals inferred from the GWS-PINNs samples and GMM-BDMC statistical learner, and the Stage~2 column gives the refined hard piecewise-constant estimates obtained by CCD-PINNs. Table~\ref{tab2} presents the temporal change-point detection results for one-dimensional time-varying coefficients, including the candidate change-point intervals from Stage~1 and the refined change-point estimates from Stage~2. For spatially varying coefficients, treating the adjacent boundaries of two platforms with different coefficients as closed curve functions in a 2D plane, and Table~\ref{tab3} reports the MSE of the detected jump boundaries, measuring the discrepancy between the predicted discontinuity interfaces and the reference interfaces. In addition, Table~\ref{tab4} compares the PDE solution approximation errors of the main network in the two stages. The results of the subsequent examples are listed in these four tables.
	
	For Case~1.3, the comparison between the reconstructed solution from the second-stage main network $\tilde{u}$ and the reference solution, together with the corresponding absolute error, is shown in the first two rows of Figure~\ref{fig2}, where two representative time instants $t=0$ and $t=0.15$ are selected. The spatial coefficient inversion results are presented in the third and fourth rows of Figure~\ref{fig2}, including the Stage~1 relaxed continuous approximation, the Stage~2 hard piecewise-constant reconstruction, the heuristic coefficient search intervals and candidate discontinuity region inferred by GMM-BDMC, and the comparison between the predicted jump boundary and the reference discontinuity boundary. The Stage~1 result, produced by the GWS-PINNs coefficient sub network, captures the global spatial distribution of the coefficient but smooths the discontinuity near the interface between $U_2$ and $U\setminus U_2$.
    
    Based on the heuristic coefficient search intervals and candidate discontinuity region supplied by GMM-BDMC, the Stage~2 CCD-PINNs refinement constrains the possible values of the two spatial coefficient states and sharpens the spatial jump interface into a hard piecewise-constant estimator. The reference boundary is determined by the true spatial partition between $U_2$ and $U\setminus U_2$, while the predicted boundary is obtained from the refined Stage~2 hard classification or piecewise-constant coefficient reconstruction. The comparison shows that the proposed framework can recover not only the two discrete coefficient levels but also the spatial interface separating different parameter states. Therefore, the two-stage structure is essential. Stage~1 provides a smooth coefficient sampler suitable for statistical mixture modeling, whereas Stage~2 sharpens the transition and produces an explicit discontinuous coefficient reconstruction.
	
	These wave equation examples demonstrate that the proposed JVC-PINNs framework can automatically distinguish constant coefficients from jump-varying coefficients through the inferred number of GMM components $\widehat{K}$. When $\widehat{K}=1$, no change point or interface is introduced, and the framework reduces to a constant-coefficient inverse problem. When $\widehat{K}>1$, the statistically inferred spatiotemporal coefficient and change-point candidate intervals or change-interface regions are passed to CCD-PINNs for constrained refinement. As a result, the relaxed continuous approximation from Stage~1 is converted into an interpretable hard piecewise-constant estimator in Stage~2. The successful extension from temporal jumps to spatial discontinuity interfaces further confirms the scalability of the proposed framework for multidimensional PDE inverse problems. Additional PDE types are considered in the following subsections to further examine the effectiveness of the method.
	
	\subsection{Heat Equation}
	The reaction-diffusion model, also known as the Fisher equation, is considered as a 1+1D heat equation with a time-varying diffusion coefficient. In the following model, $c(t)\geq 0$ denotes the diffusion parameter, which changes over time. This example is taken from \cite{li2023diffusion}, and the governing equation is given by
	\begin{equation}
		\begin{cases}
			u_t = c(t) u_{xx} + u - u^2, & \quad (x,t)\in U \times (0,T],\\[2pt]
			u(x,t) = 0, & \quad (x,t)\in \partial U \times (0,T],\\[2pt]
			u(x,0) = g(x), & \quad x \in U.
		\end{cases}
	\end{equation}
	Let $U=[-6,6]$, $T=10$, and the initial function be
	\begin{equation}
		g(x) =
		\begin{cases}
			\exp\big(10(x+1)\big), & x < -1, \\[2pt]
			1, & -1 \leq x \leq 1, \\[2pt]
			\exp\big(-10(x-1)\big), & x > 1.
		\end{cases}
	\end{equation}
	This setting has a nonsmooth initial condition, and two different scenarios of time-varying coefficients are considered.
	
	\textbf{Case~2.1:}
	\begin{equation}
		c(t)=
		\begin{cases}
			0.2, & t\in [0,5),\\[2pt]
			0.1, & t\in [5,10].
		\end{cases}
	\end{equation}
	
	\textbf{Case~2.2:}
	\begin{equation}
		c(t)=
		\begin{cases}
			0.05, & t\in [0,2) \cup [8,10],\\[2pt]
			0.1, & t\in [2,4) \cup [6,8),\\[2pt]
			0.2, & t\in [4,6).
		\end{cases}
	\end{equation}
	
	For Case~2.1 and Case~2.2, $\theta_p=c(t)$ on $(t,x)\in[0,10]\times[-6,6]$. Case~2.1 used main-network layers $[2,50,50,50,50,1]$ and sub-network layers $[1,2,2,2,2,2,1]$. Case~2.2 used main-network layers $[2,20,\ldots,20,1]$ with $20$ hidden layers and sub-network layers $[1,10,\ldots,10,1]$ with $10$ hidden layers. Both cases used seed $314$ and Adam optimizer, with Case~2.1 trained for $50000$ iterations and Case~2.2 trained for $200000$ iterations, learning rates $10^{-3}$ for $\hat{u}$ and $10^{-4}$ for $\hat{\theta}_p$, and exponential decay factors $0.9$ and $0.8$. The residual set had $50\times80=4000$ points, and the observation set had $4000$ fixed randomly sampled points. Case~2.1 and Case~2.2 used $\gamma_1=\gamma_2=1$ and $\gamma_3=0$ because of the nonsmooth initial condition. Stage~2 used a hard-step $c(t)$ initialized from GMM-BDMC, with Adam learning rates $5\times10^{-4}$ for $\tilde u$ and $10^{-3}$ for change points and heights. The configured Stage~2 budget was $2500$ iterations.
	
	\begin{figure}[p]
		\centering
		\subfloat[Spatiotemporal solution]{
			\begin{minipage}[t]{0.44\linewidth}
				\includegraphics[width=1\linewidth]{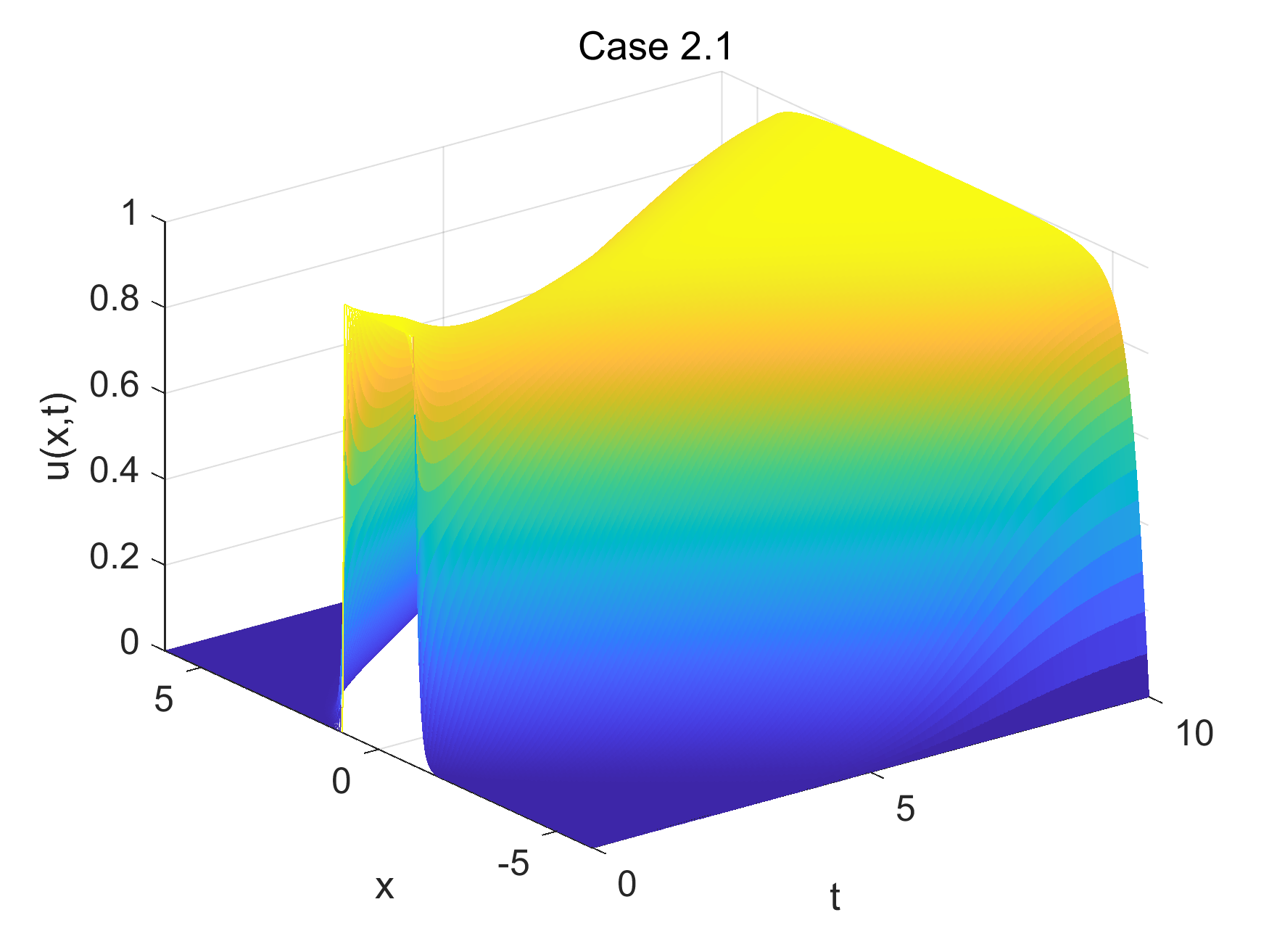}
			\end{minipage}
		}
		\subfloat[Parameter inverse result]{
			\begin{minipage}[t]{0.44\linewidth}
				\includegraphics[width=1\linewidth]{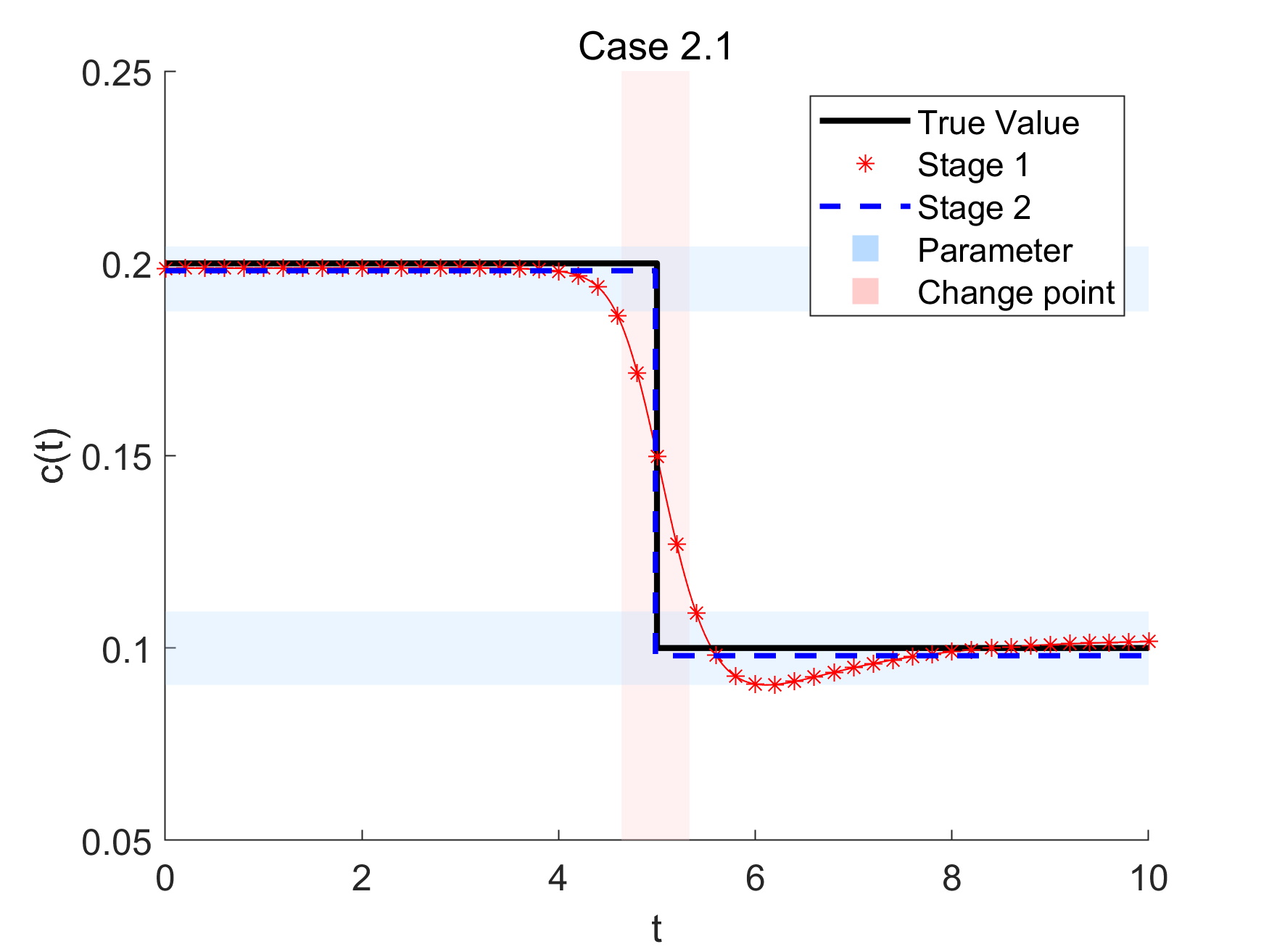}
			\end{minipage}
		}\\
		\subfloat[Reference solution]{
			\begin{minipage}[t]{0.315\linewidth}
				\includegraphics[width=1\linewidth]{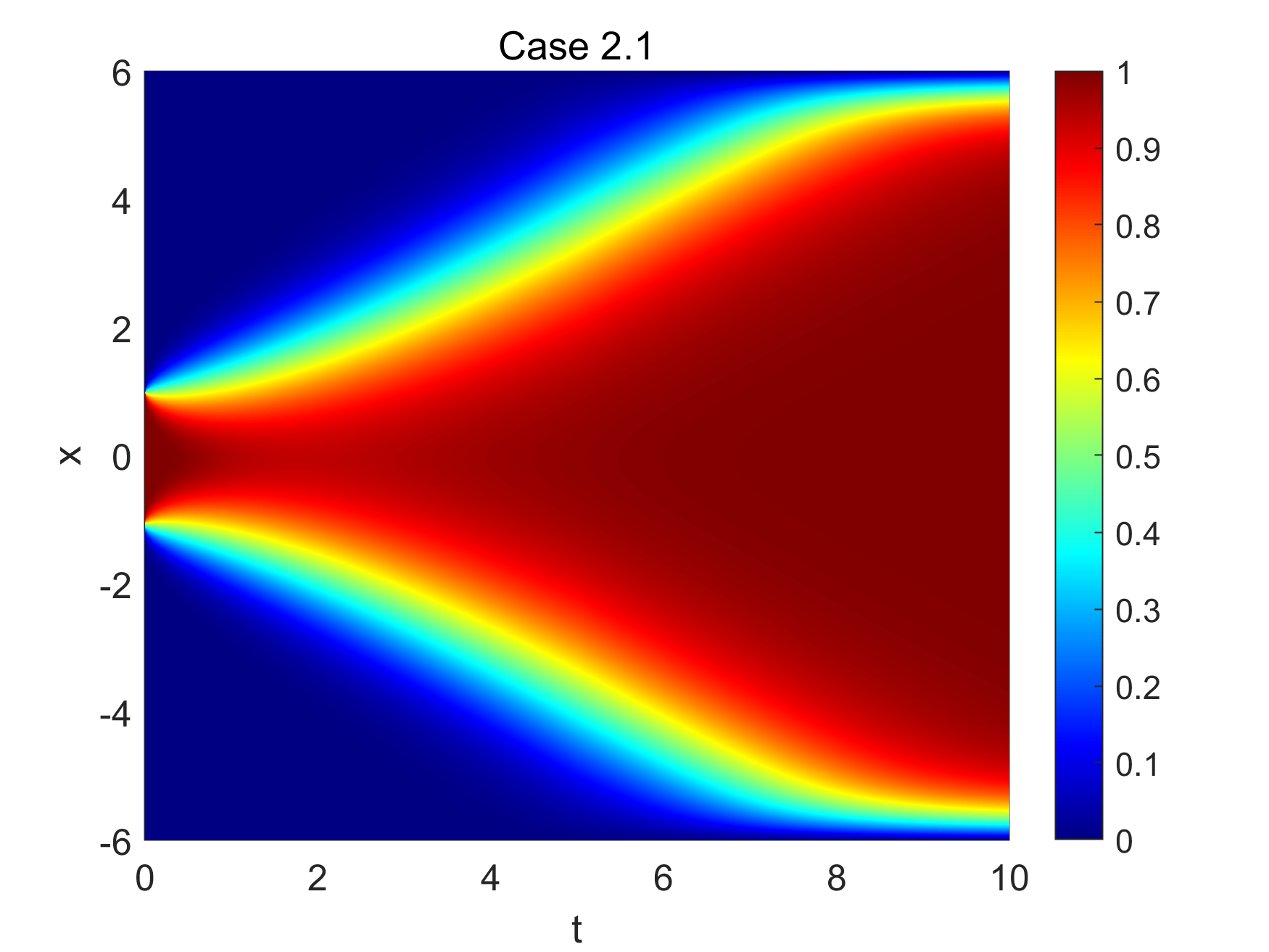}
			\end{minipage}
		}
		\subfloat[Predicted solution]{
			\begin{minipage}[t]{0.315\linewidth}
				\includegraphics[width=1\linewidth]{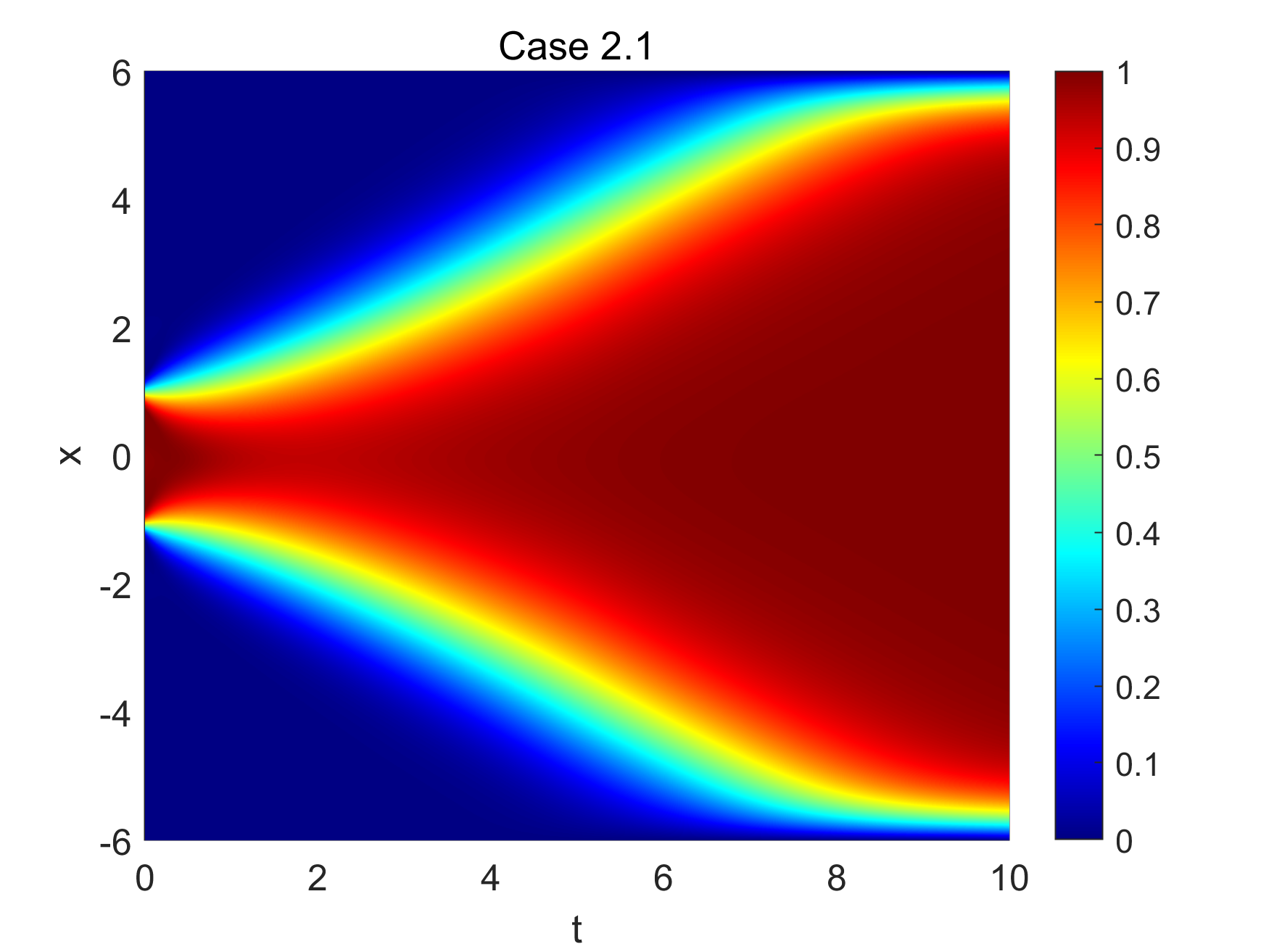}
			\end{minipage}
		}
		\subfloat[Absolute error]{
			\begin{minipage}[t]{0.315\linewidth}
				\includegraphics[width=1\linewidth]{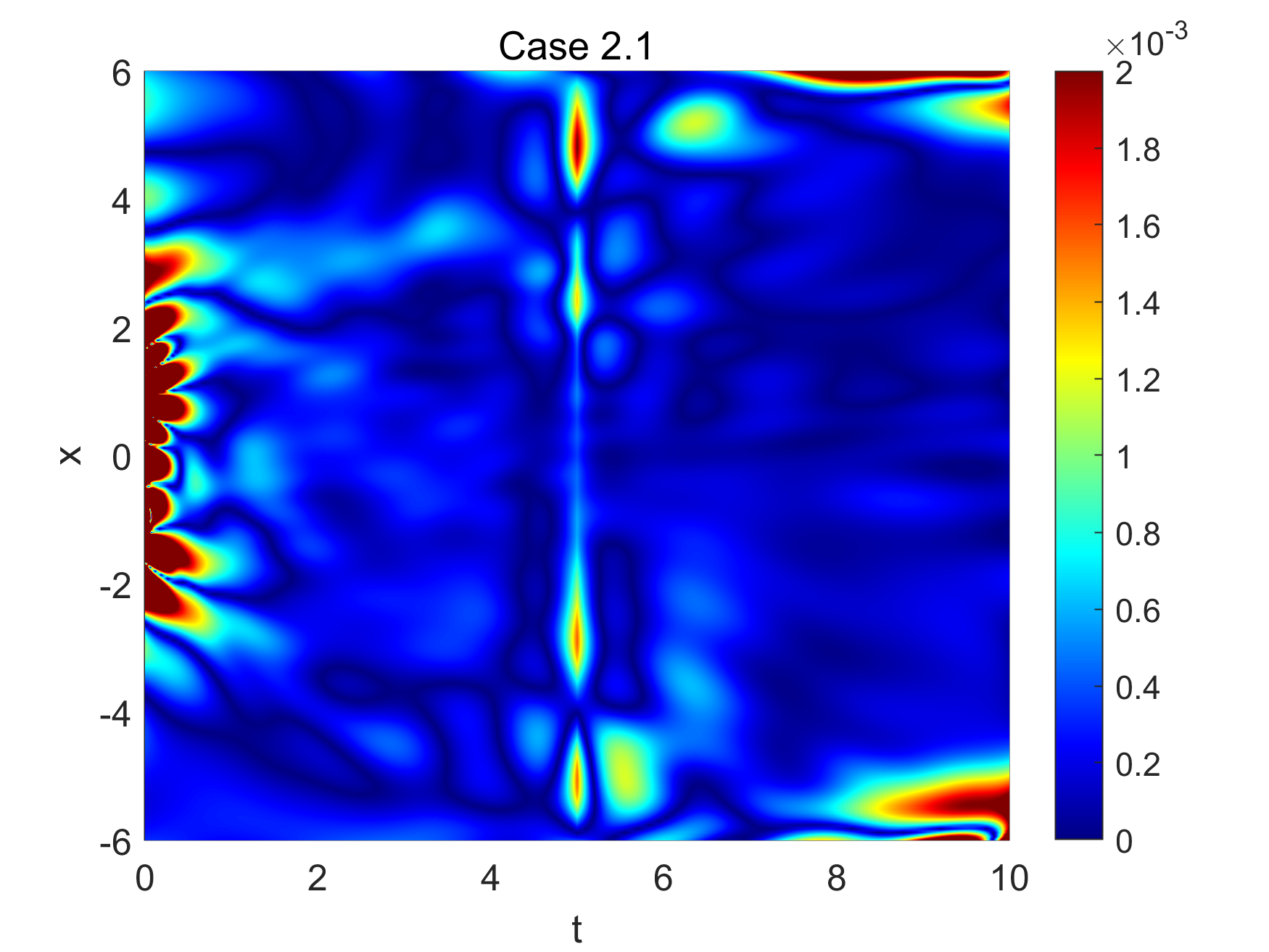}
			\end{minipage}
		}\\
		\subfloat[Spatiotemporal solution]{
			\begin{minipage}[t]{0.44\linewidth}
				\includegraphics[width=1\linewidth]{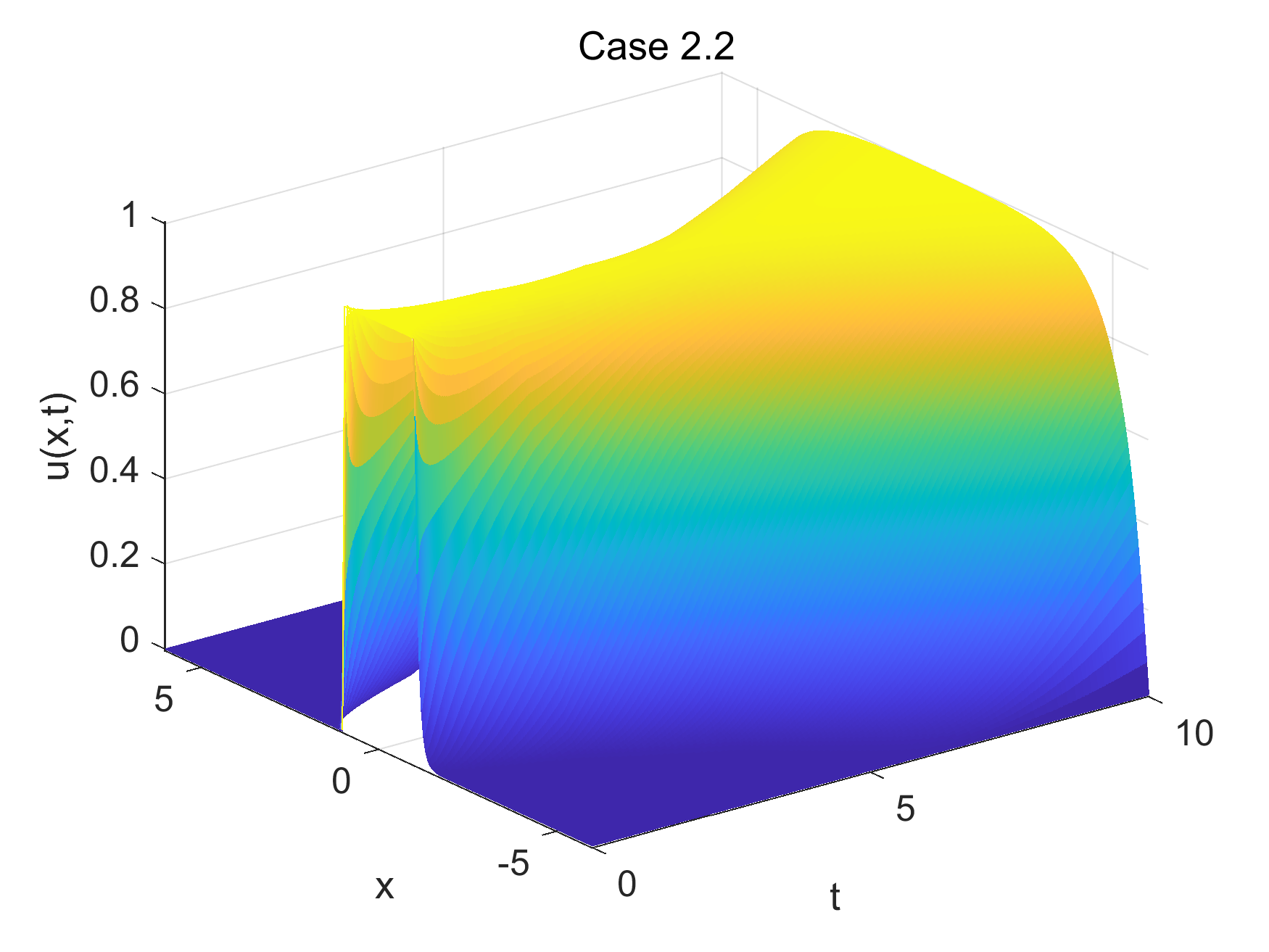}
			\end{minipage}
		}
		\subfloat[Parameter inverse result]{
			\begin{minipage}[t]{0.44\linewidth}
				\includegraphics[width=1\linewidth]{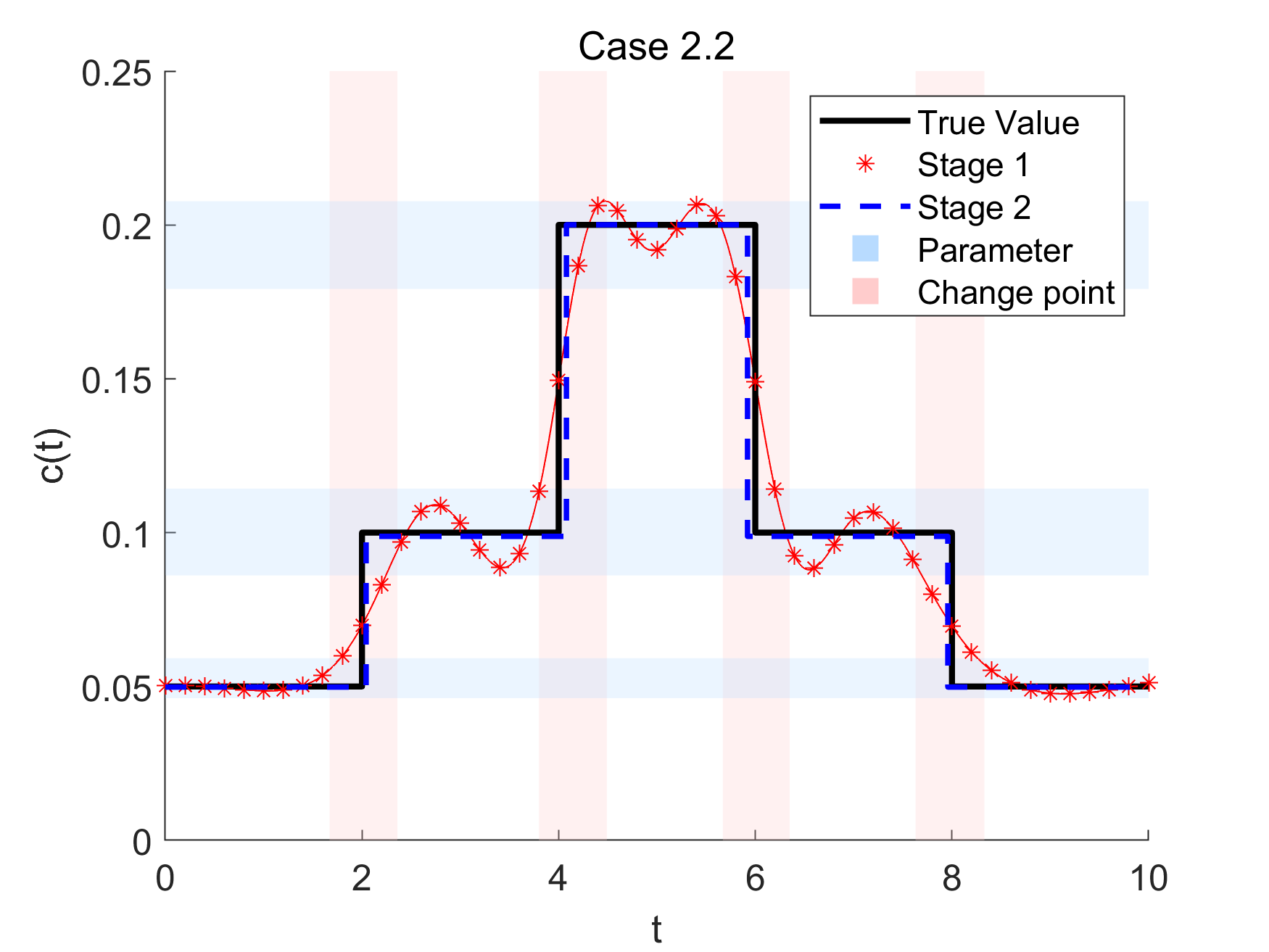}
			\end{minipage}
		}\\
		\subfloat[Reference solution]{
			\begin{minipage}[t]{0.315\linewidth}
				\includegraphics[width=1\linewidth]{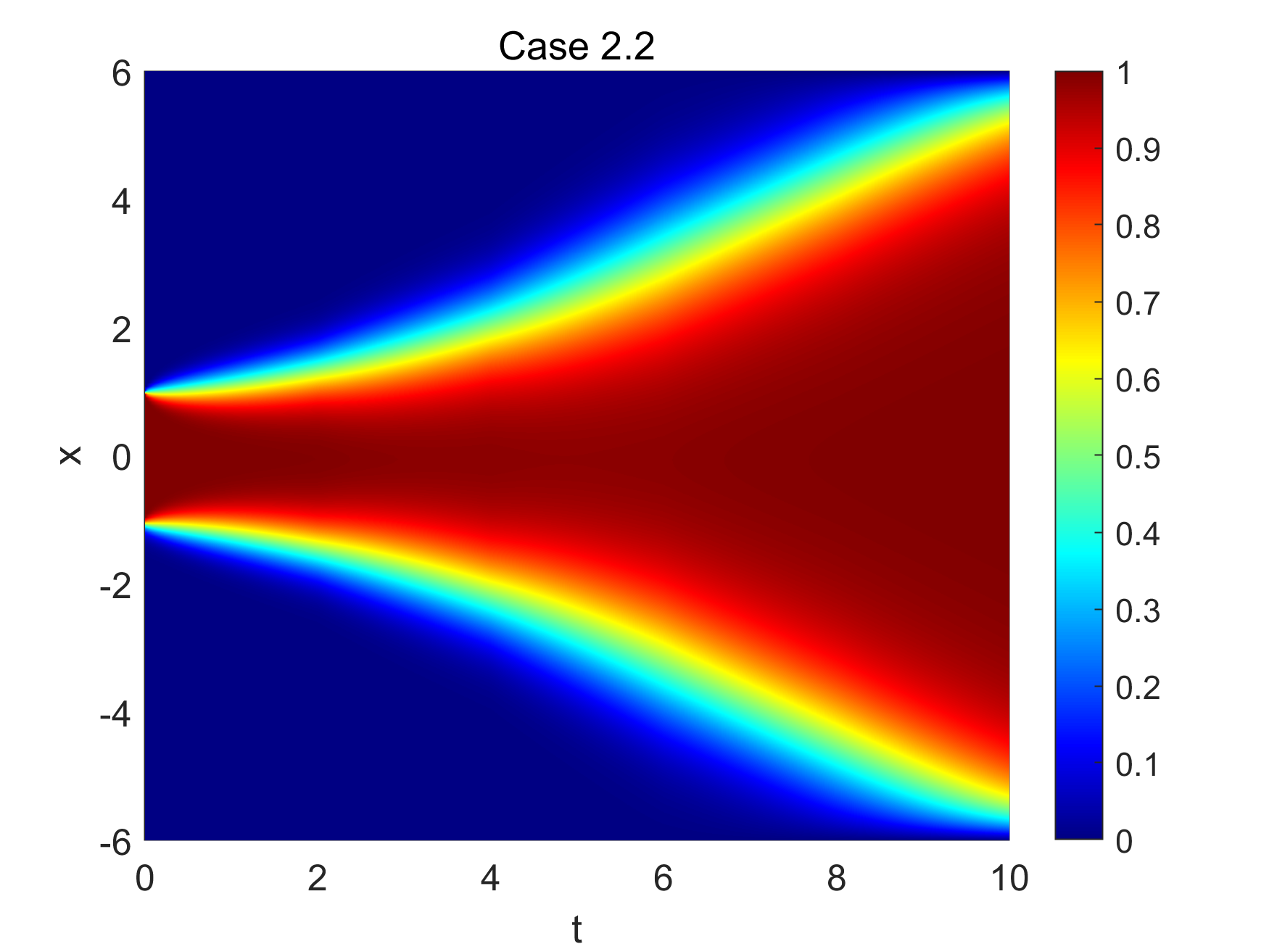}
			\end{minipage}
		}
		\subfloat[Predicted solution]{
			\begin{minipage}[t]{0.315\linewidth}
				\includegraphics[width=1\linewidth]{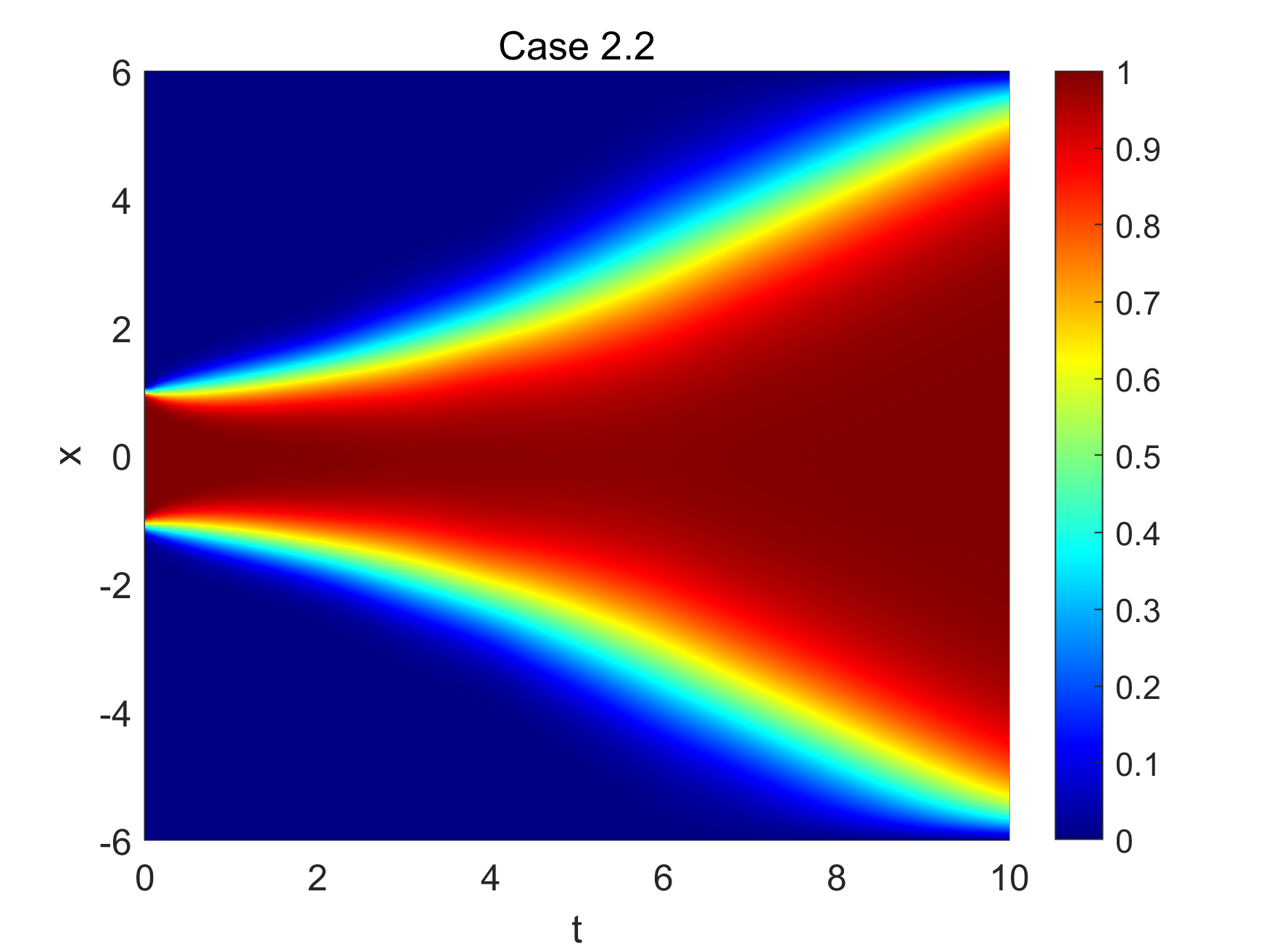}
			\end{minipage}
		}
		\subfloat[Absolute error]{
			\begin{minipage}[t]{0.315\linewidth}
				\includegraphics[width=1\linewidth]{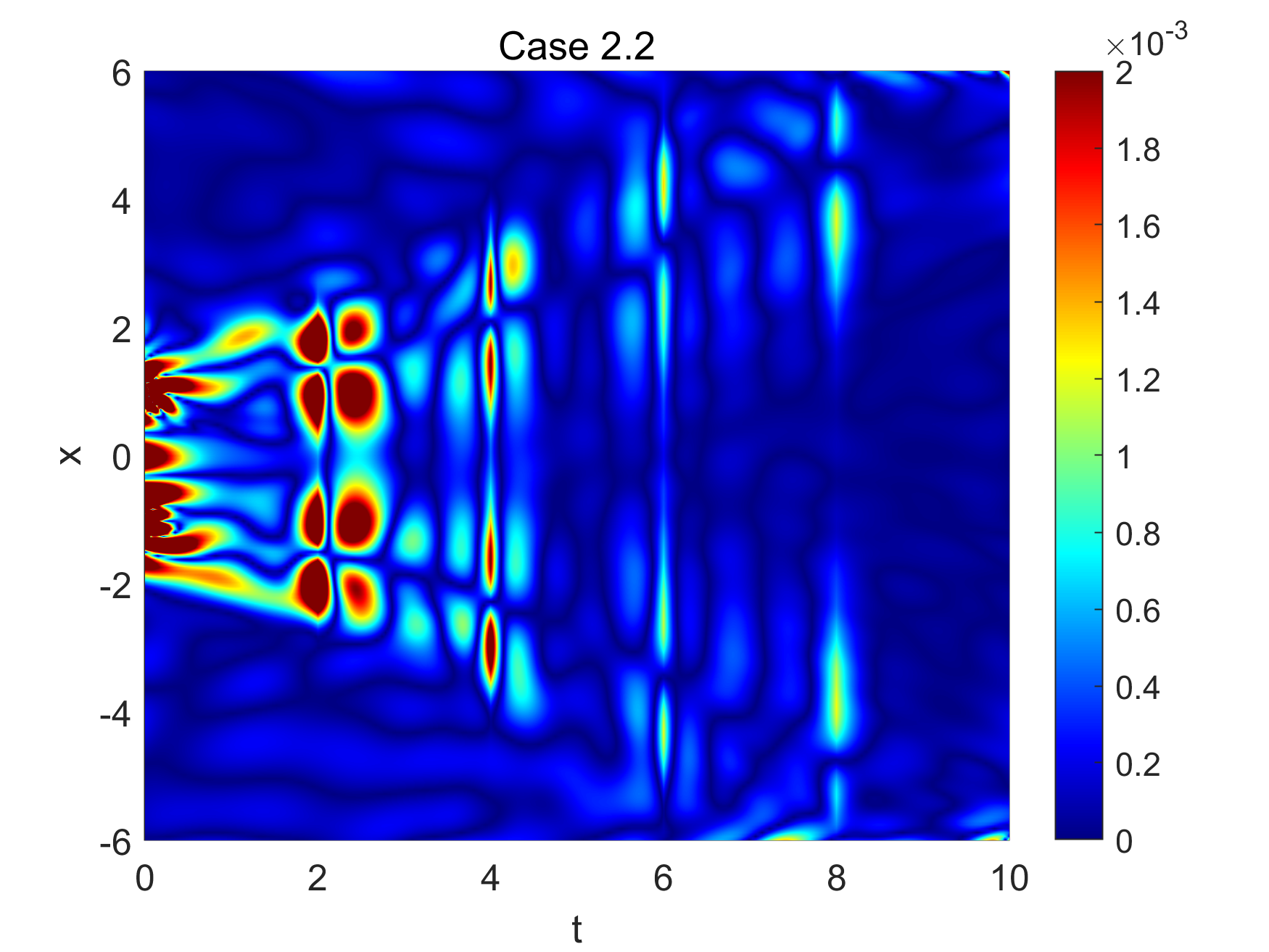}
			\end{minipage}
		}
		\caption{Numerical results for heat equations with discontinuously time-varying coefficient $c(t)$. \label{fig3}}
	\end{figure}
	
	The solution plots for these two heat equation cases and the corresponding two-stage parameter inversion results for the diffusion coefficient $c(t)$ are shown in Figure~\ref{fig3}. In the parameter inversion plots, the Stage~1 result represents the relaxed continuous coefficient approximation generated by the GWS-PINNs sub network, while the Stage~2 result represents the hard piecewise-constant estimator refined by CCD-PINNs. The coefficient search intervals and candidate change-point intervals are inferred by the GMM-BDMC statistical learner and are used as constraints in the second-stage refinement. The reconstructed solution from the second-stage main network $\tilde{u}$ and the corresponding absolute error with respect to the reference solution are also displayed in Figure~\ref{fig3}. Table~\ref{tab1} reports the two-stage coefficient inversion results for $c(t)$, where the Stage~1 column gives the heuristic intervals obtained from GMM-BDMC and the Stage~2 column gives the refined coefficient estimates from CCD-PINNs. Table~\ref{tab2} reports the temporal change-point detection results, including the candidate change-point intervals from Stage~1 and the refined change-point estimates from Stage~2. Table~\ref{tab4} presents the solution approximation errors of the main network in both stages.
	
	For Case~2.1, the GMM-BDMC learner identifies two coefficient states and one candidate change-point interval. Although the Stage~1 coefficient approximation smooths the discontinuity near the true jump time $t=5$, it provides sufficient information for the statistical learner to infer the two diffusion states and localize the candidate transition region. The Stage~2 CCD-PINNs refinement then converts the relaxed continuous approximation into a sharp step-function estimator and produces a refined change point within the candidate interval.
	
	For Case~2.2, the diffusion coefficient switches among three discrete states over several temporal subintervals. The GMM-BDMC learner successfully identifies the three coefficient levels and multiple candidate change-point intervals. The Stage~1 curve captures the overall switching pattern but remains continuous near each jump. In contrast, the Stage~2 estimator recovers a hard piecewise-constant structure with refined temporal change points. These results indicate that the proposed JVC-PINNs framework is effective not only for a single jump but also for multi-state and multi-jump coefficient switching.
	
	The results for the heat equation are consistent with those obtained for the wave equation. The first-stage GWS-PINNs sampler provides a continuous but informative approximation of the jump-varying coefficient, the GMM-BDMC statistical learner extracts the discrete coefficient states and candidate transition intervals, and the second-stage CCD-PINNs refinement yields an explicit piecewise-constant coefficient reconstruction. The reconstructed solution $\tilde{u}$ remains close to the reference solution, showing that the refined coefficient representation preserves the physical consistency of the PDE surrogate.
	
	It is also worth emphasizing the role of the gradient-adaptive weighting strategy in GWS-PINNs. The adaptive weight $w_{\mathrm{res}}^{(i)}$ is not designed to increase the residual contribution near the coefficient jumps, which are usually hard-to-learn regions due to discontinuous parameter variation and possible nonsmooth solution behavior. Instead, the weight becomes larger in smoother regions where the coefficient field varies slowly or remains constant. This design encourages the neural network to learn the physically consistent solution and coefficient behavior in easy-to-learn regions, while avoiding excessive fitting of the artificial transition layer near jumps. As a result, the main network $\hat{u}(\mathbf{x},t)$ and the coefficient sub network $\hat{\theta}_p(\mathbf{x},t)$ in Stage~1 can produce stable solution and coefficient surrogates. These surrogates then provide reliable samples for GMM-BDMC, enabling accurate inference of the coefficient states and candidate change-point intervals. Therefore, the gradient-adaptive weighting strategy plays an important role in improving the stability of the first-stage sampling and the accuracy of the subsequent Stage~2 constrained refinement.
	
	\begin{figure}[p]
		\centering
		\subfloat[Spatiotemporal solution]{
			\begin{minipage}[t]{0.315\linewidth}
				\includegraphics[width=1\linewidth]{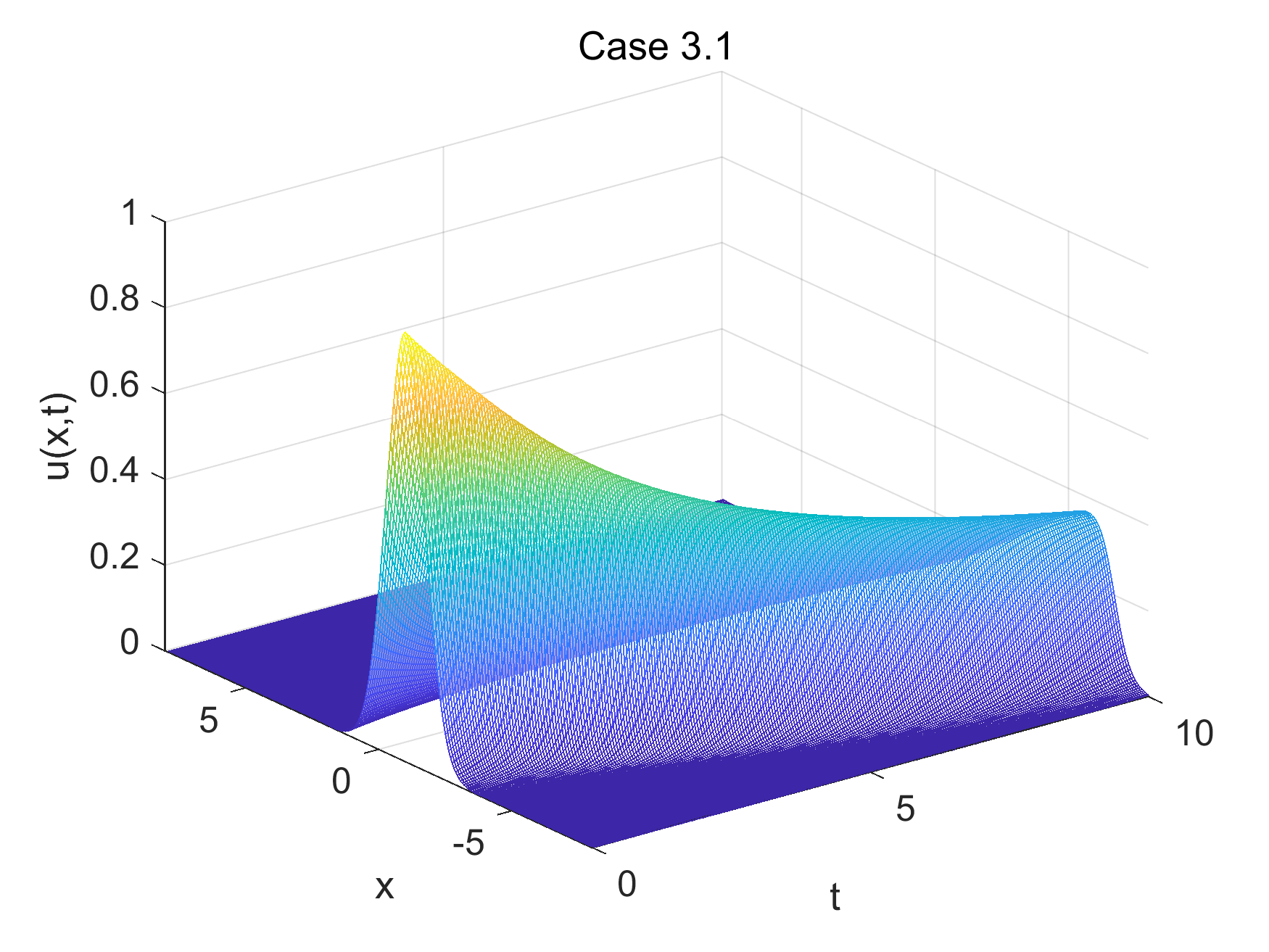}
			\end{minipage}
		}
		\subfloat[Parameter inverse result]{
			\begin{minipage}[t]{0.315\linewidth}
				\includegraphics[width=1\linewidth]{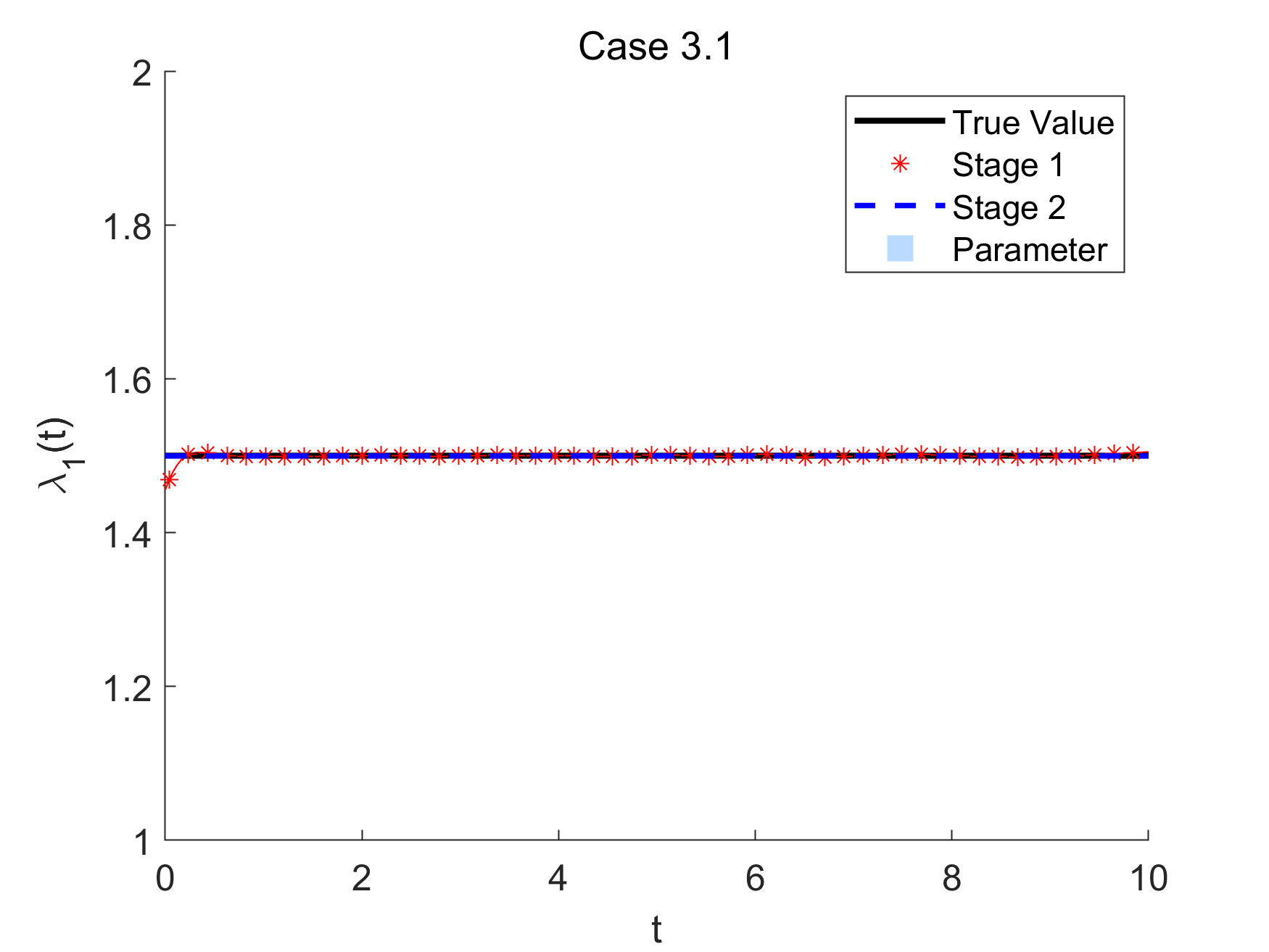}
			\end{minipage}
		}
		\subfloat[Parameter inverse result]{
			\begin{minipage}[t]{0.315\linewidth}
				\includegraphics[width=1\linewidth]{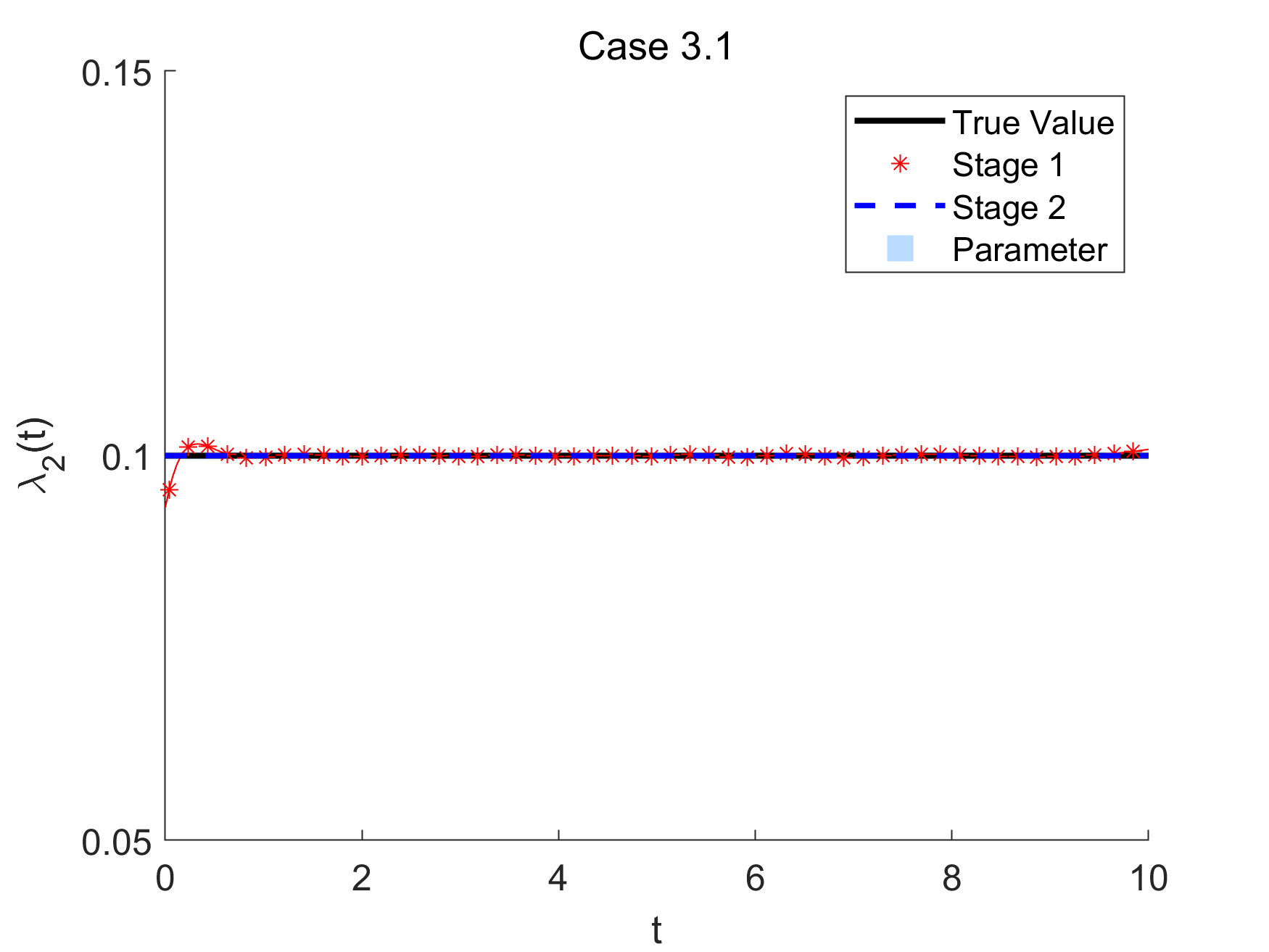}
			\end{minipage}
		}
		\\
		\subfloat[Reference solution]{
			\begin{minipage}[t]{0.315\linewidth}
				\includegraphics[width=1\linewidth]{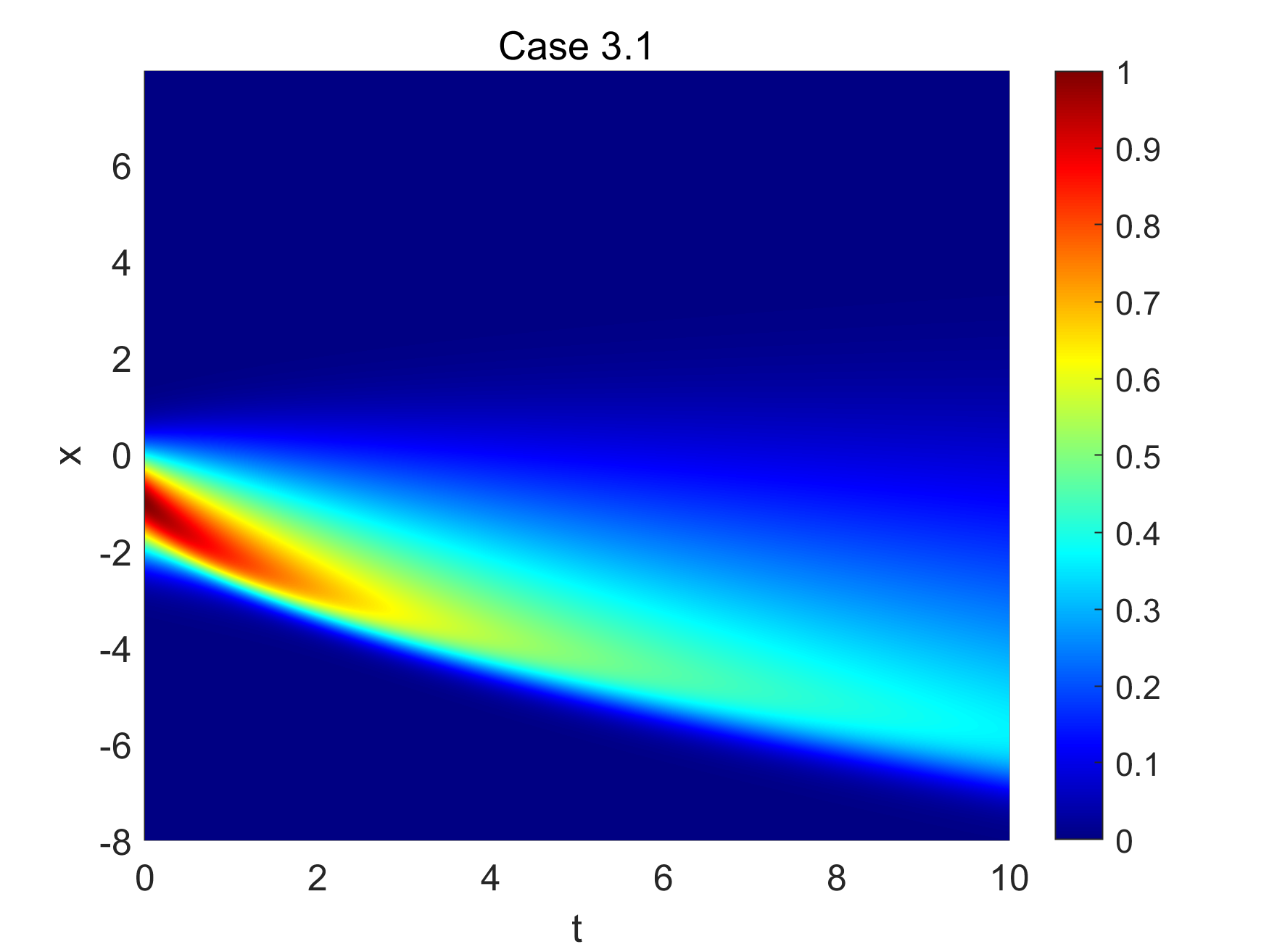}
			\end{minipage}
		}
		\subfloat[Predicted solution]{
			\begin{minipage}[t]{0.315\linewidth}
				\includegraphics[width=1\linewidth]{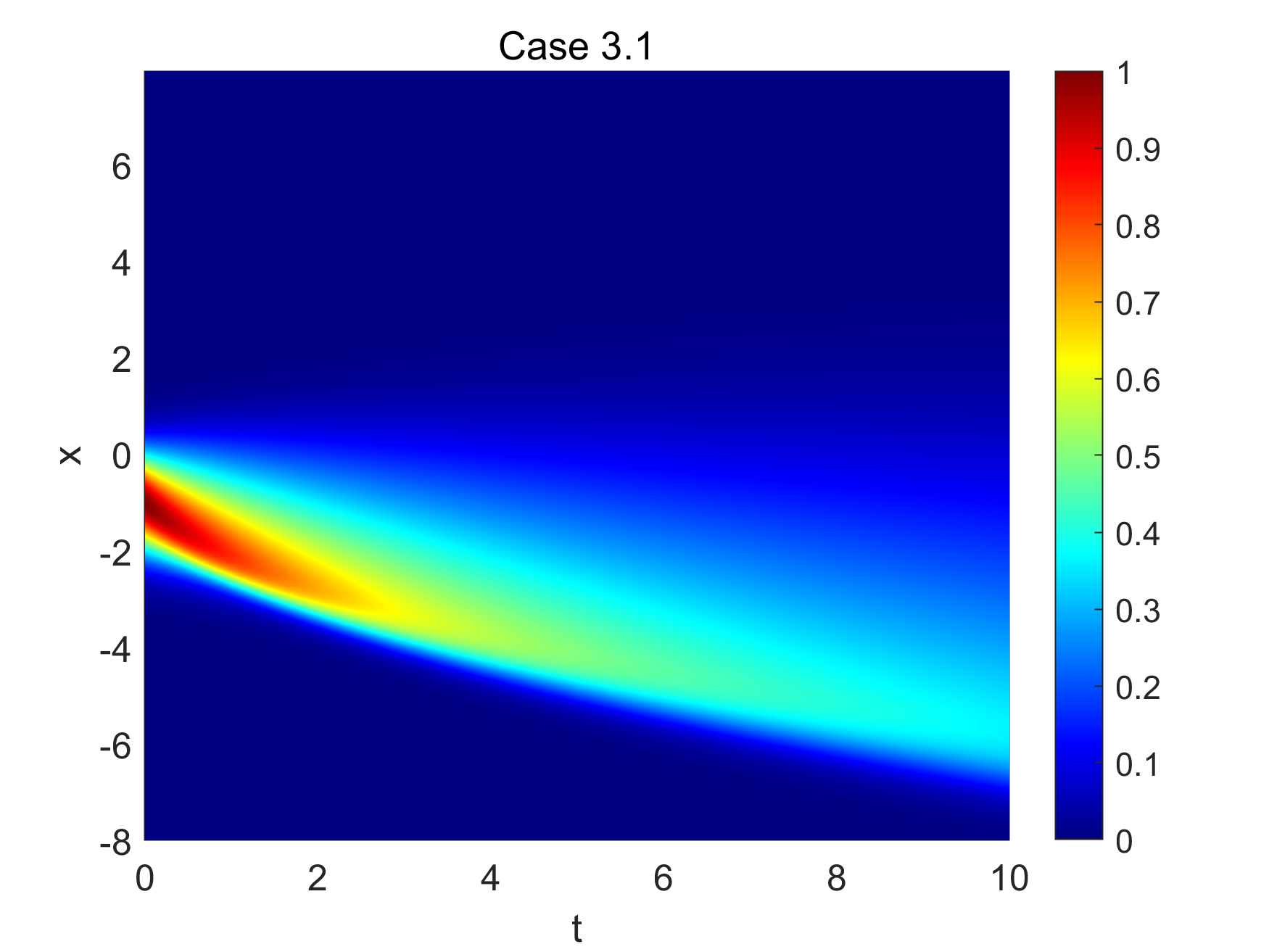}
			\end{minipage}
		}
		\subfloat[Absolute error]{
			\begin{minipage}[t]{0.315\linewidth}
				\includegraphics[width=1\linewidth]{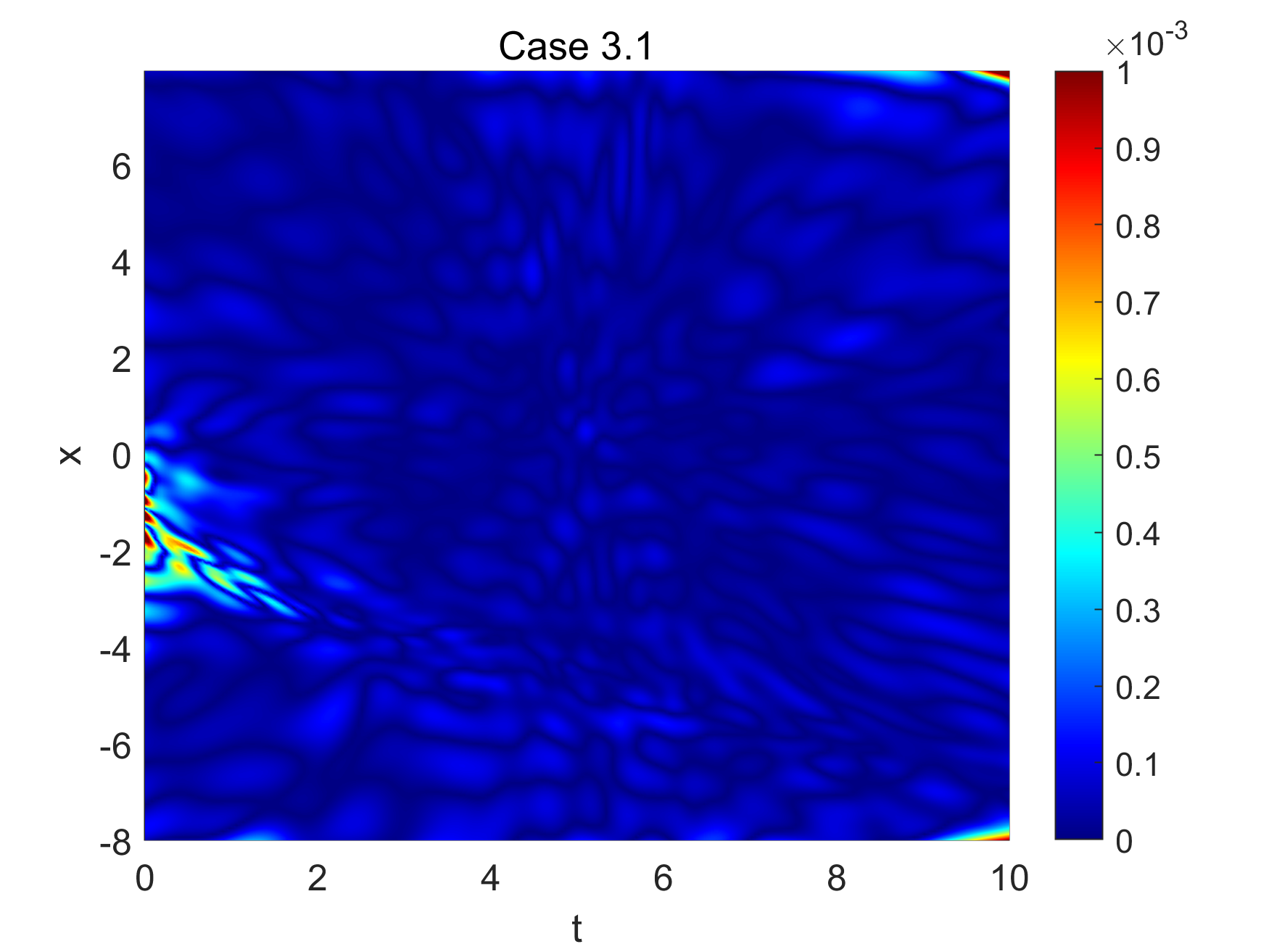}
			\end{minipage}
		}\\
		\subfloat[Spatiotemporal solution]{
			\begin{minipage}[t]{0.315\linewidth}
				\includegraphics[width=1\linewidth]{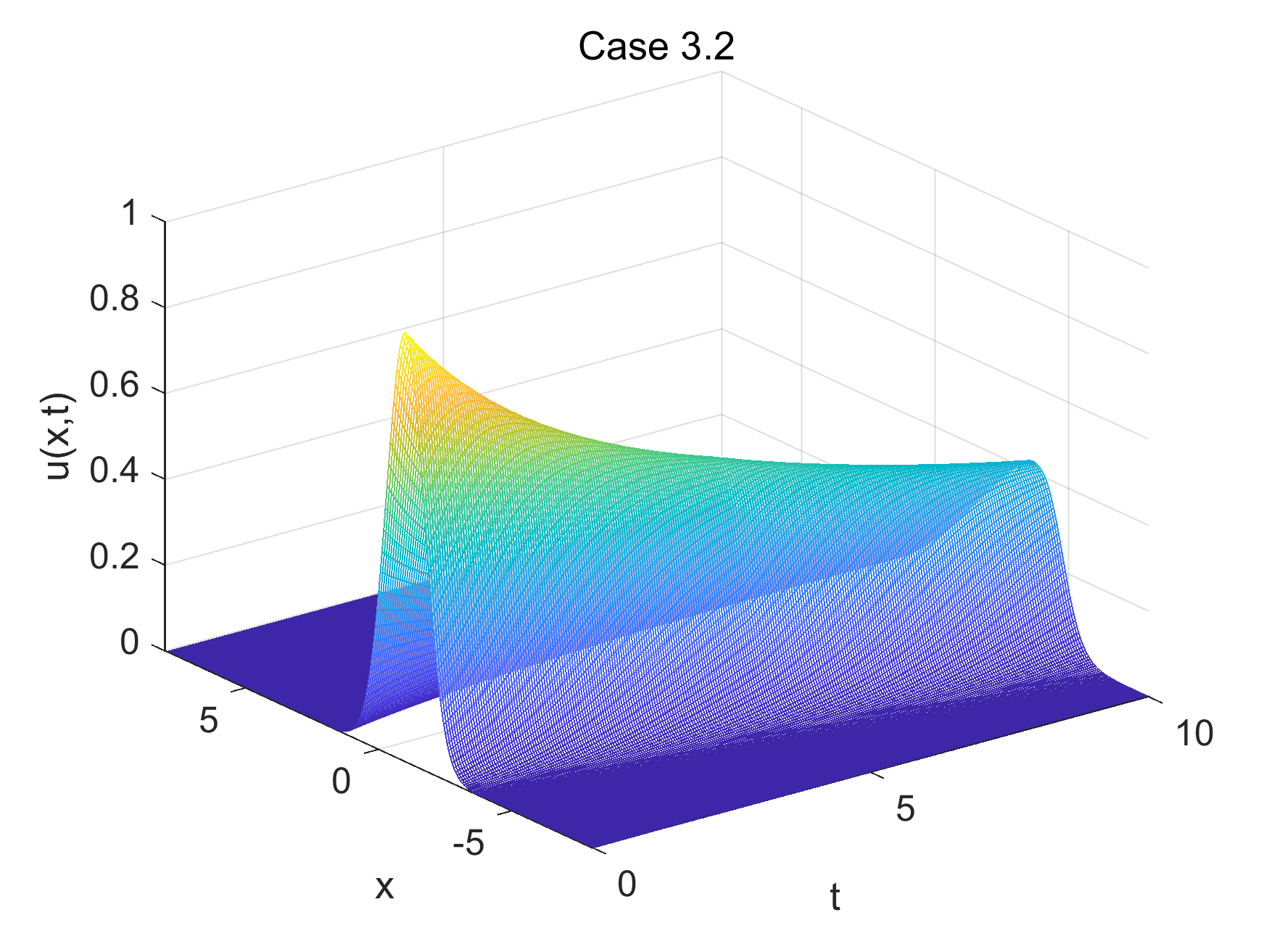}
			\end{minipage}
		}
		\subfloat[Parameter inverse result]{
			\begin{minipage}[t]{0.315\linewidth}
				\includegraphics[width=1\linewidth]{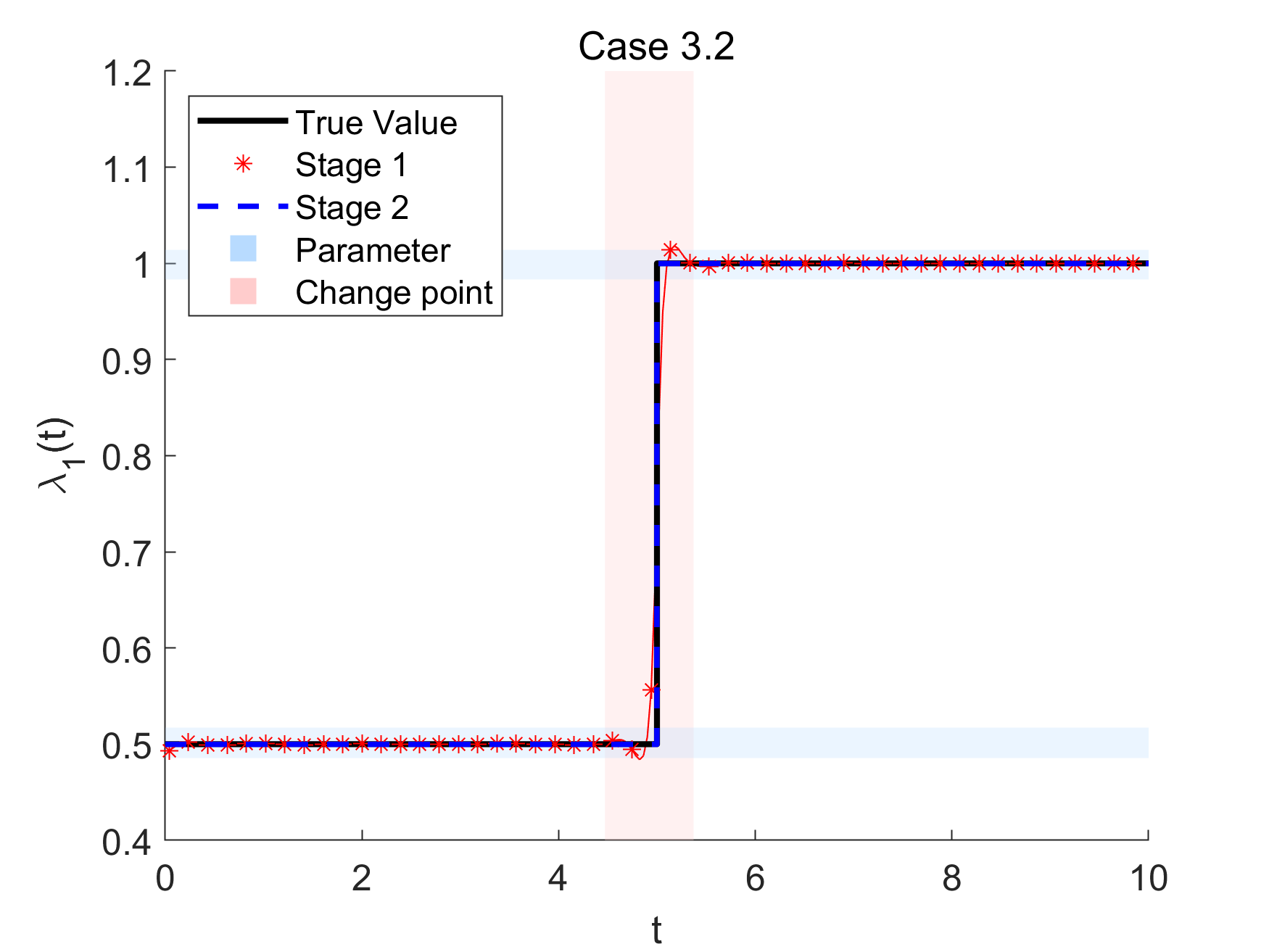}
			\end{minipage}
		}
		\subfloat[Parameter inverse result]{
			\begin{minipage}[t]{0.315\linewidth}
				\includegraphics[width=1\linewidth]{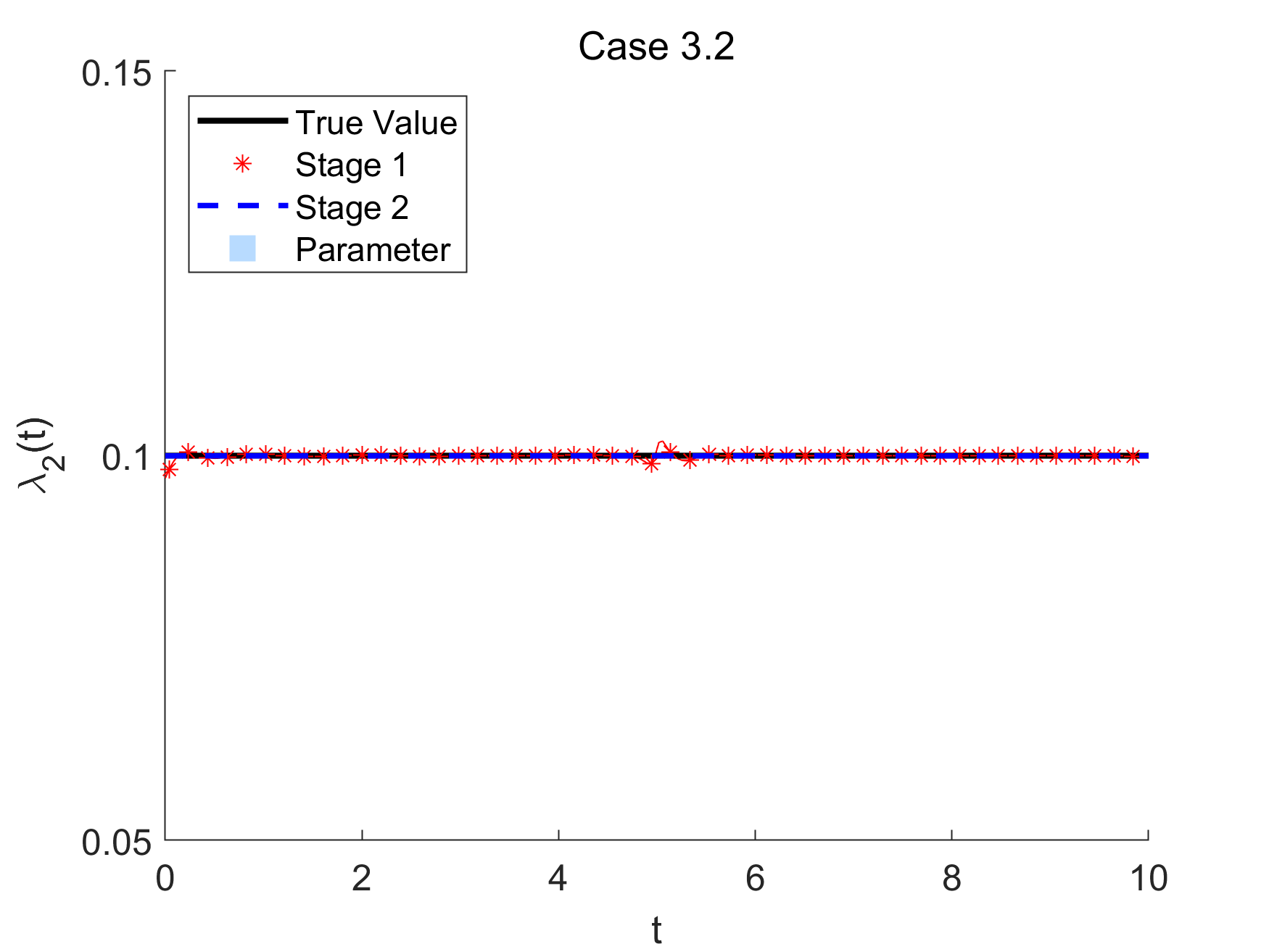}
			\end{minipage}
		}\\
		\subfloat[Reference solution]{
			\begin{minipage}[t]{0.315\linewidth}
				\includegraphics[width=1\linewidth]{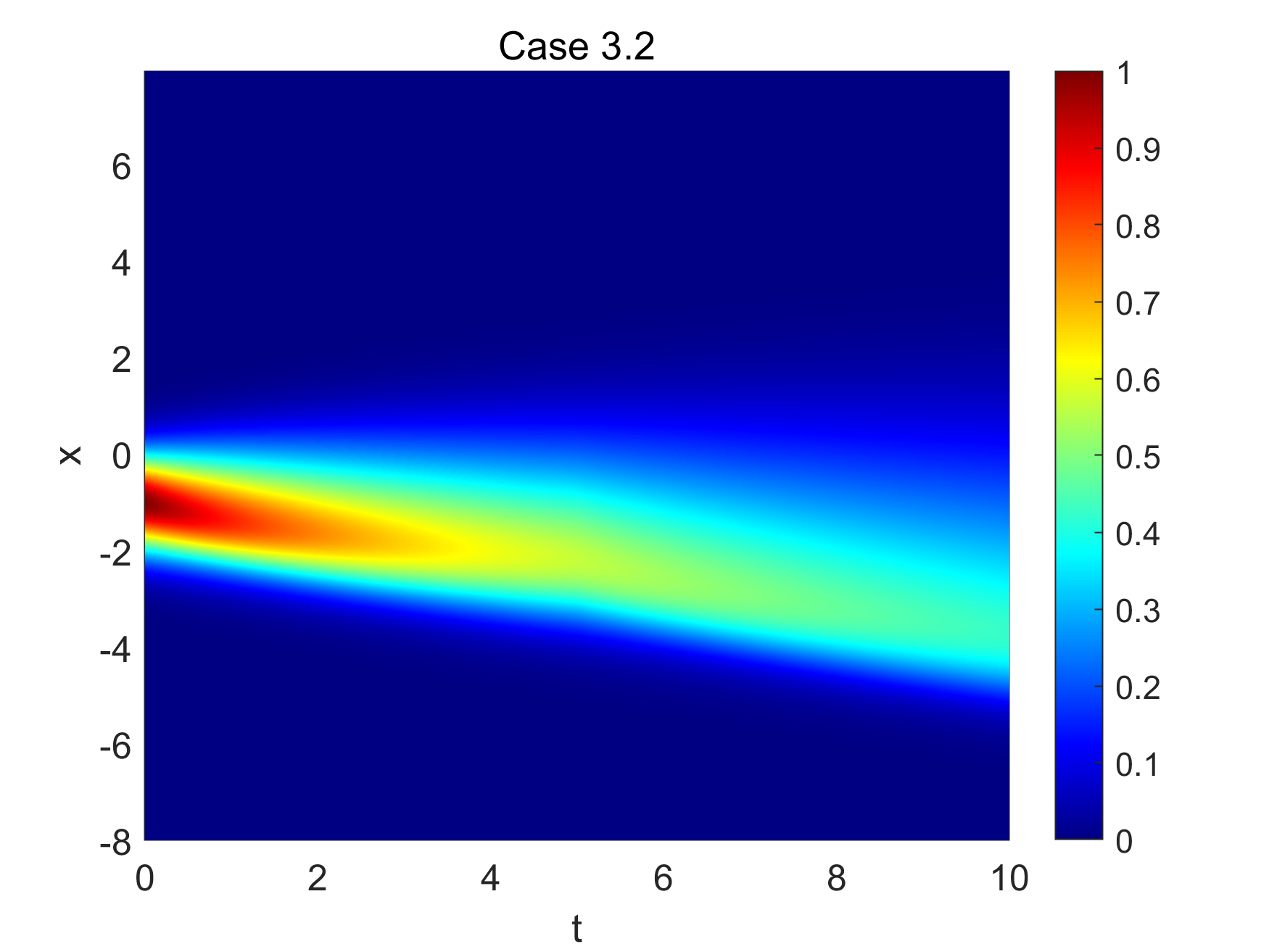}
			\end{minipage}
		}
		\subfloat[Predicted solution]{
			\begin{minipage}[t]{0.315\linewidth}
				\includegraphics[width=1\linewidth]{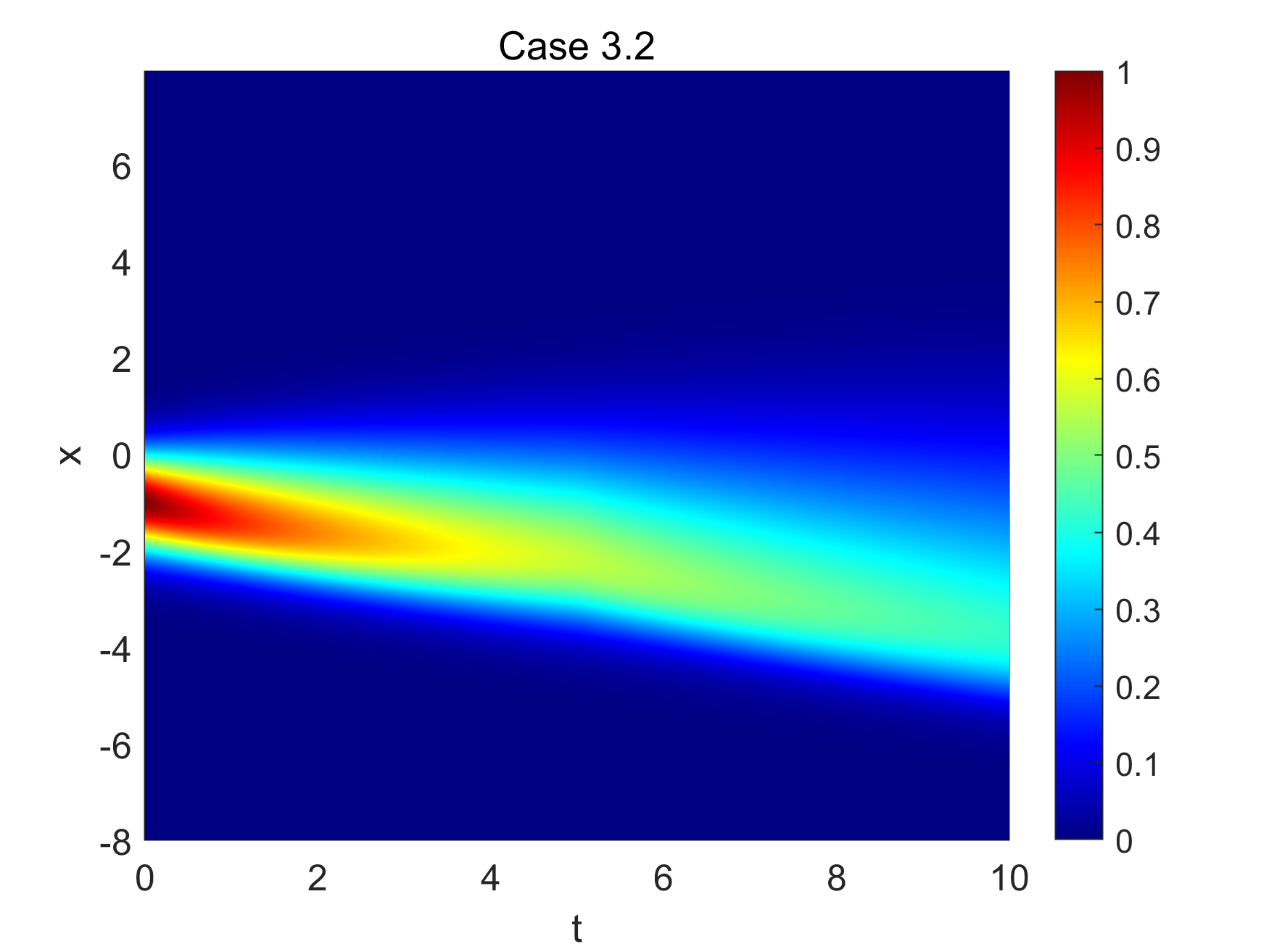}
			\end{minipage}
		}
		\subfloat[Absolute error]{
			\begin{minipage}[t]{0.315\linewidth}
				\includegraphics[width=1\linewidth]{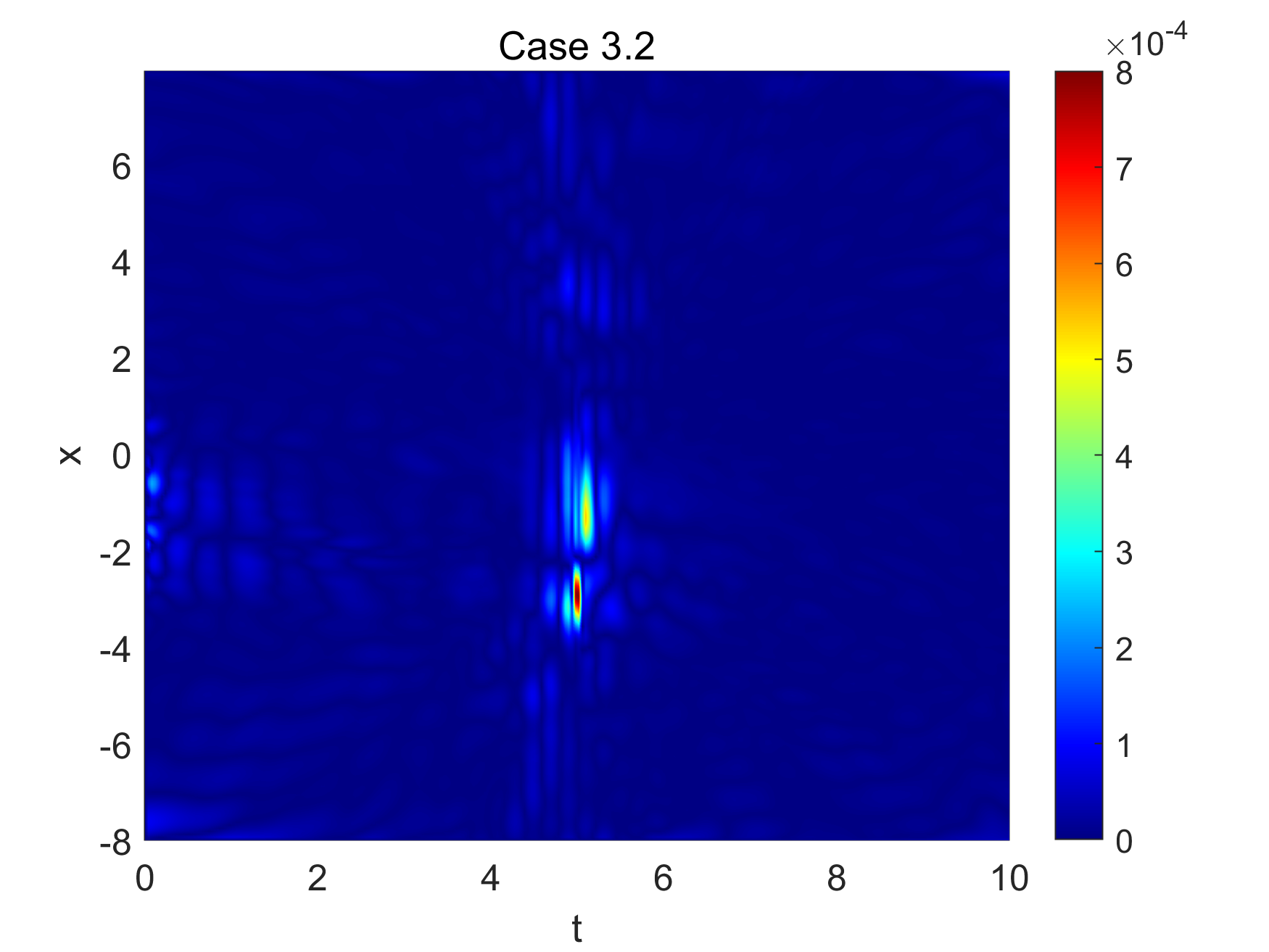}
			\end{minipage}
		}
		\caption{Numerical results for Burgers equations with discontinuously time-varying coefficients $\lambda_1(t)$ and $\lambda_2(t)$, from Case~3.1 to Case~3.2. \label{fig4}}
	\end{figure}
	
	\begin{figure}[p]
		\centering
		\subfloat[Spatiotemporal solution]{
			\begin{minipage}[t]{0.315\linewidth}
				\includegraphics[width=1\linewidth]{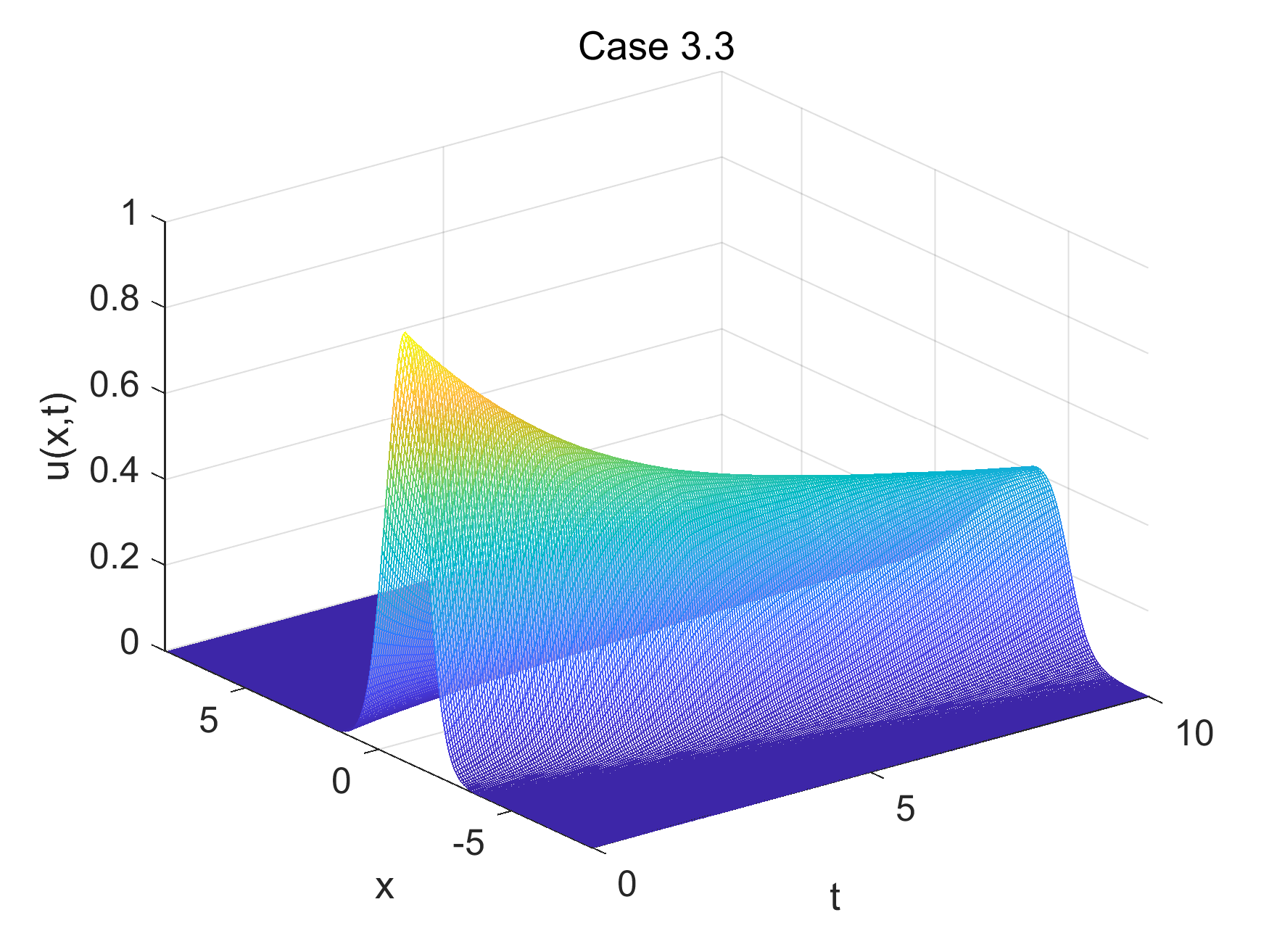}
			\end{minipage}
		}
		\subfloat[Parameter inverse result]{
			\begin{minipage}[t]{0.315\linewidth}
				\includegraphics[width=1\linewidth]{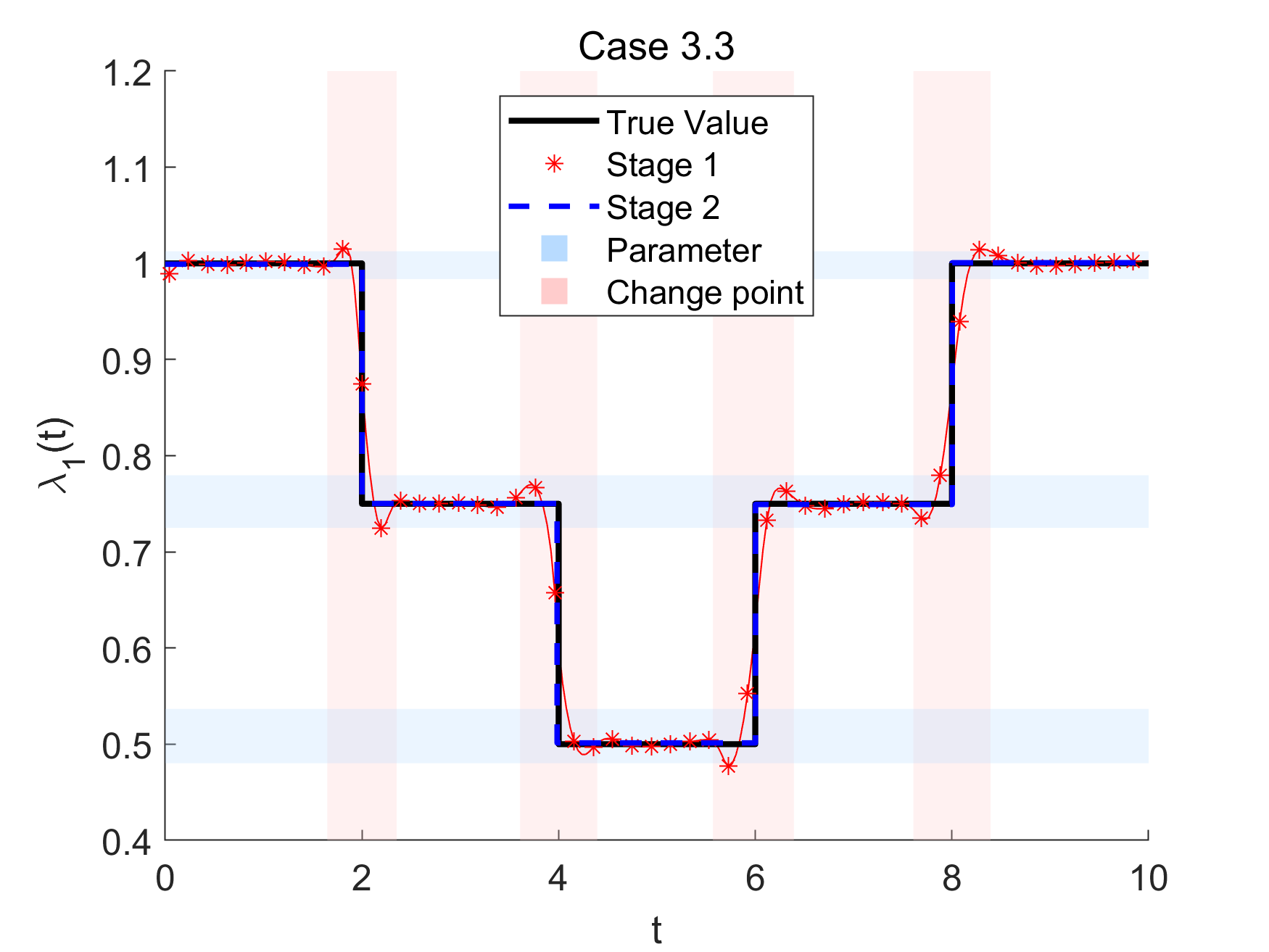}
			\end{minipage}
		}
		\subfloat[Parameter inverse result]{
			\begin{minipage}[t]{0.315\linewidth}
				\includegraphics[width=1\linewidth]{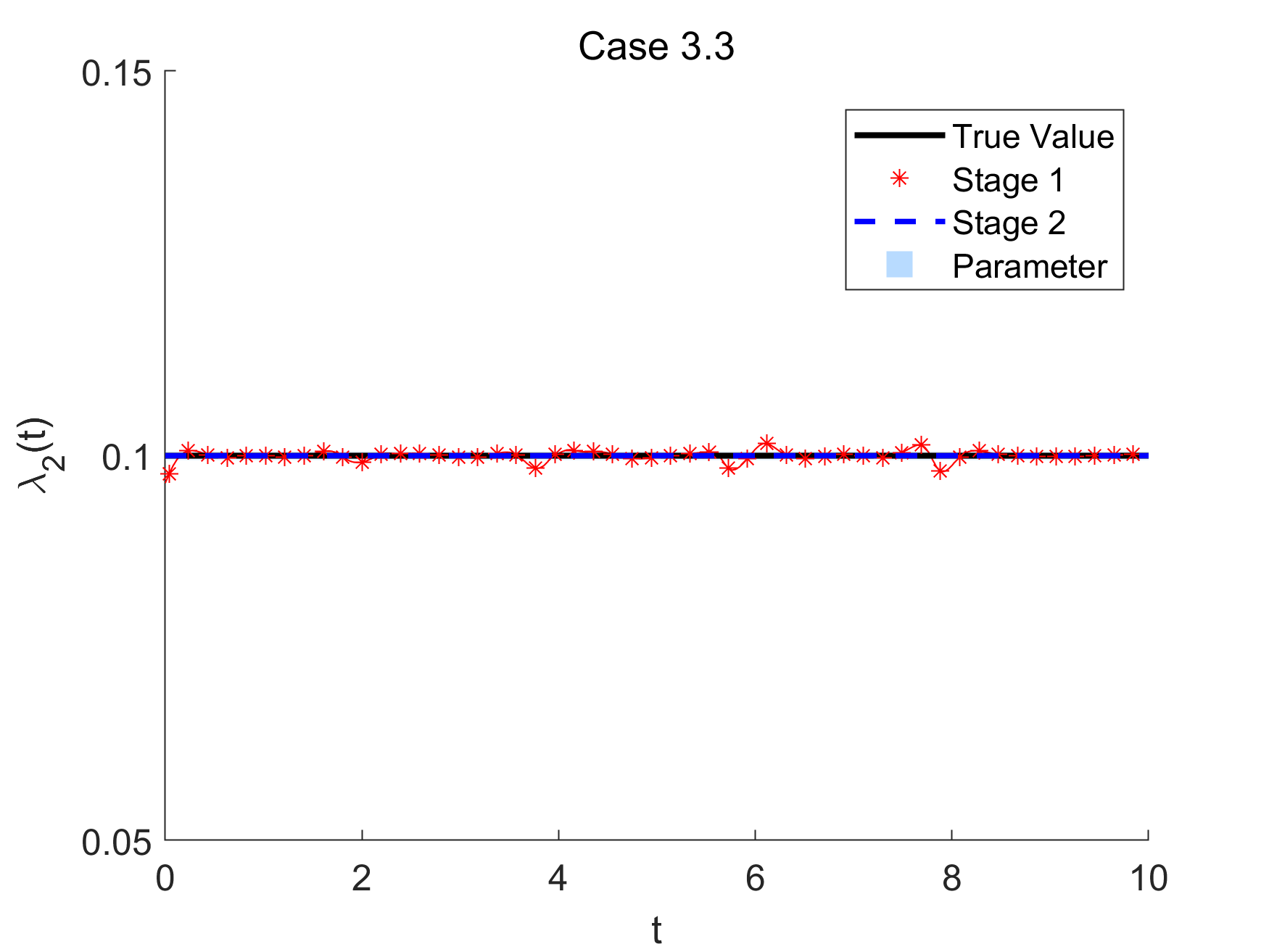}
			\end{minipage}
		}
		\\
		\subfloat[Reference solution]{
			\begin{minipage}[t]{0.315\linewidth}
				\includegraphics[width=1\linewidth]{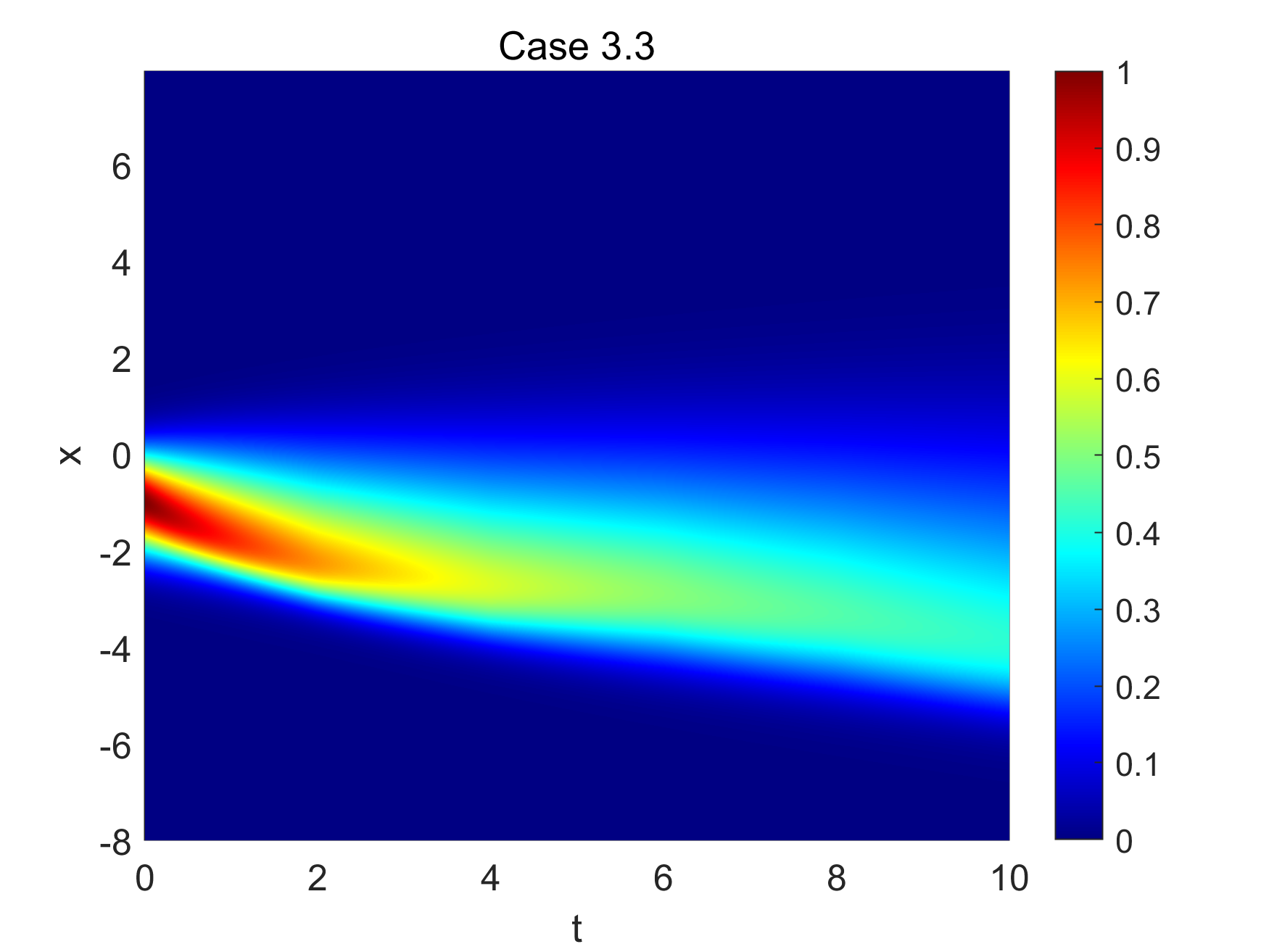}
			\end{minipage}
		}
		\subfloat[Predicted solution]{
			\begin{minipage}[t]{0.315\linewidth}
				\includegraphics[width=1\linewidth]{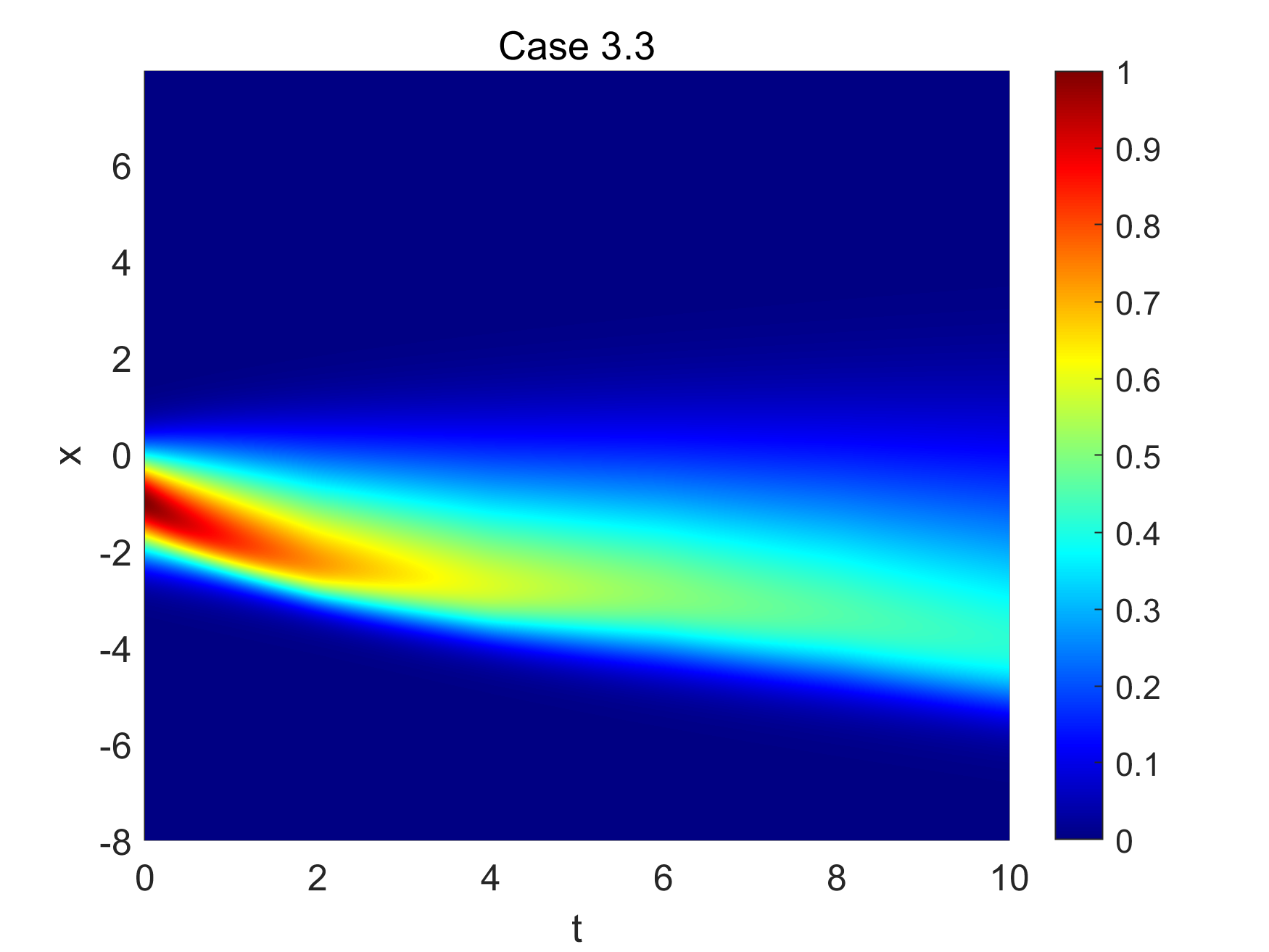}
			\end{minipage}
		}
		\subfloat[Absolute error]{
			\begin{minipage}[t]{0.315\linewidth}
				\includegraphics[width=1\linewidth]{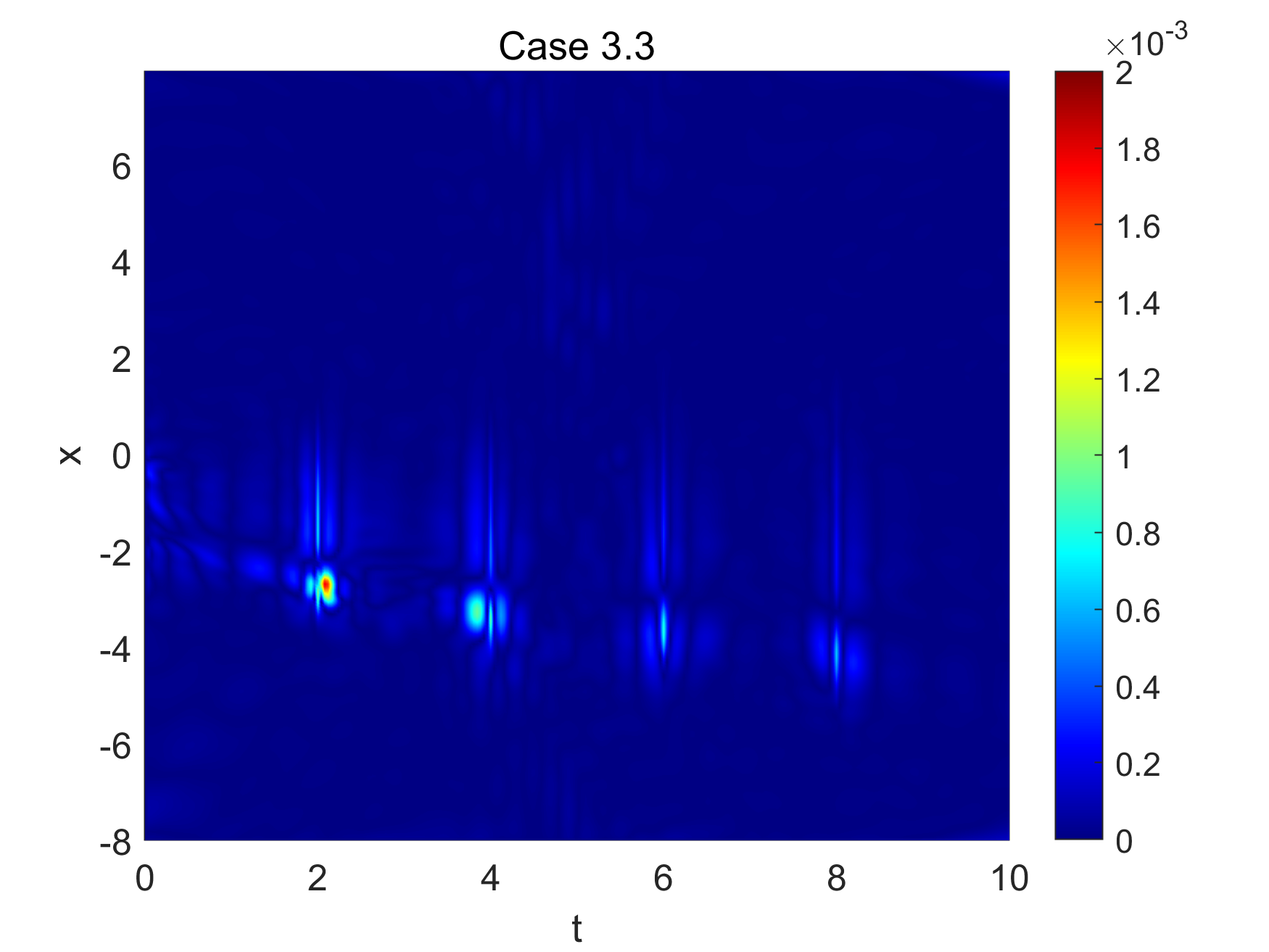}
			\end{minipage}
		}\\
		\subfloat[Spatiotemporal solution]{
			\begin{minipage}[t]{0.315\linewidth}
				\includegraphics[width=1\linewidth]{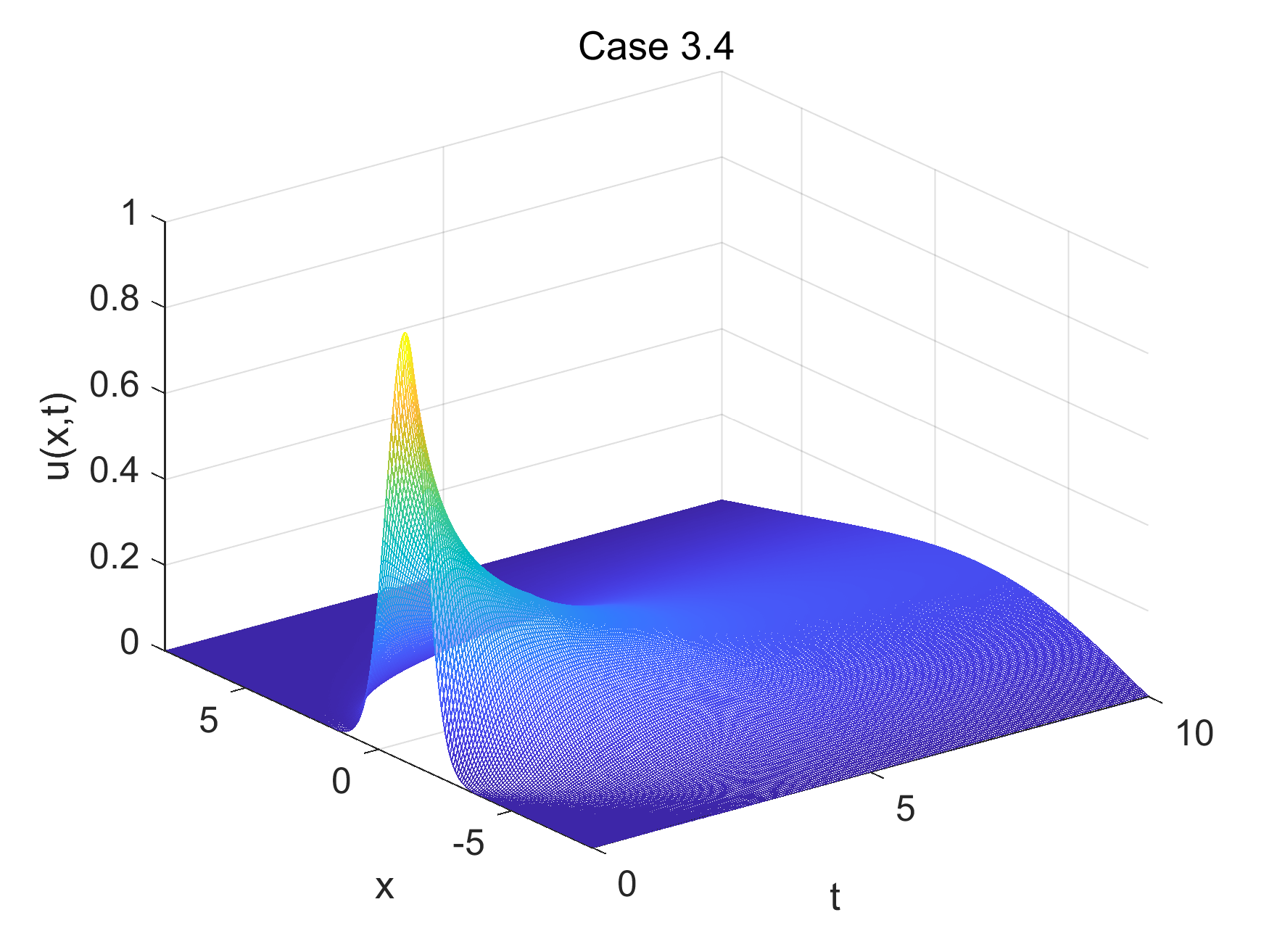}
			\end{minipage}
		}
		\subfloat[Parameter inverse result]{
			\begin{minipage}[t]{0.315\linewidth}
				\includegraphics[width=1\linewidth]{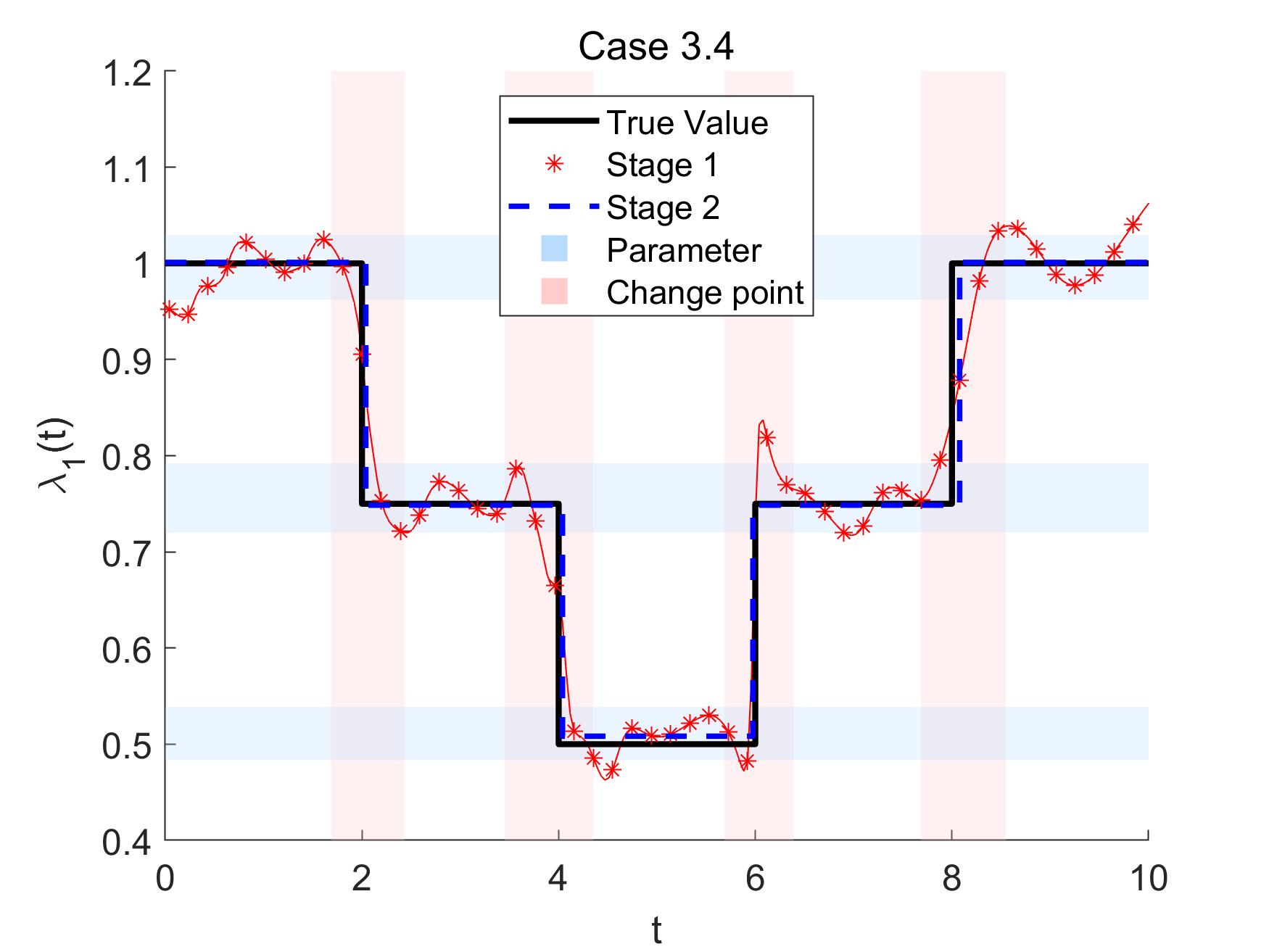}
			\end{minipage}
		}
		\subfloat[Parameter inverse result]{
			\begin{minipage}[t]{0.315\linewidth}
				\includegraphics[width=1\linewidth]{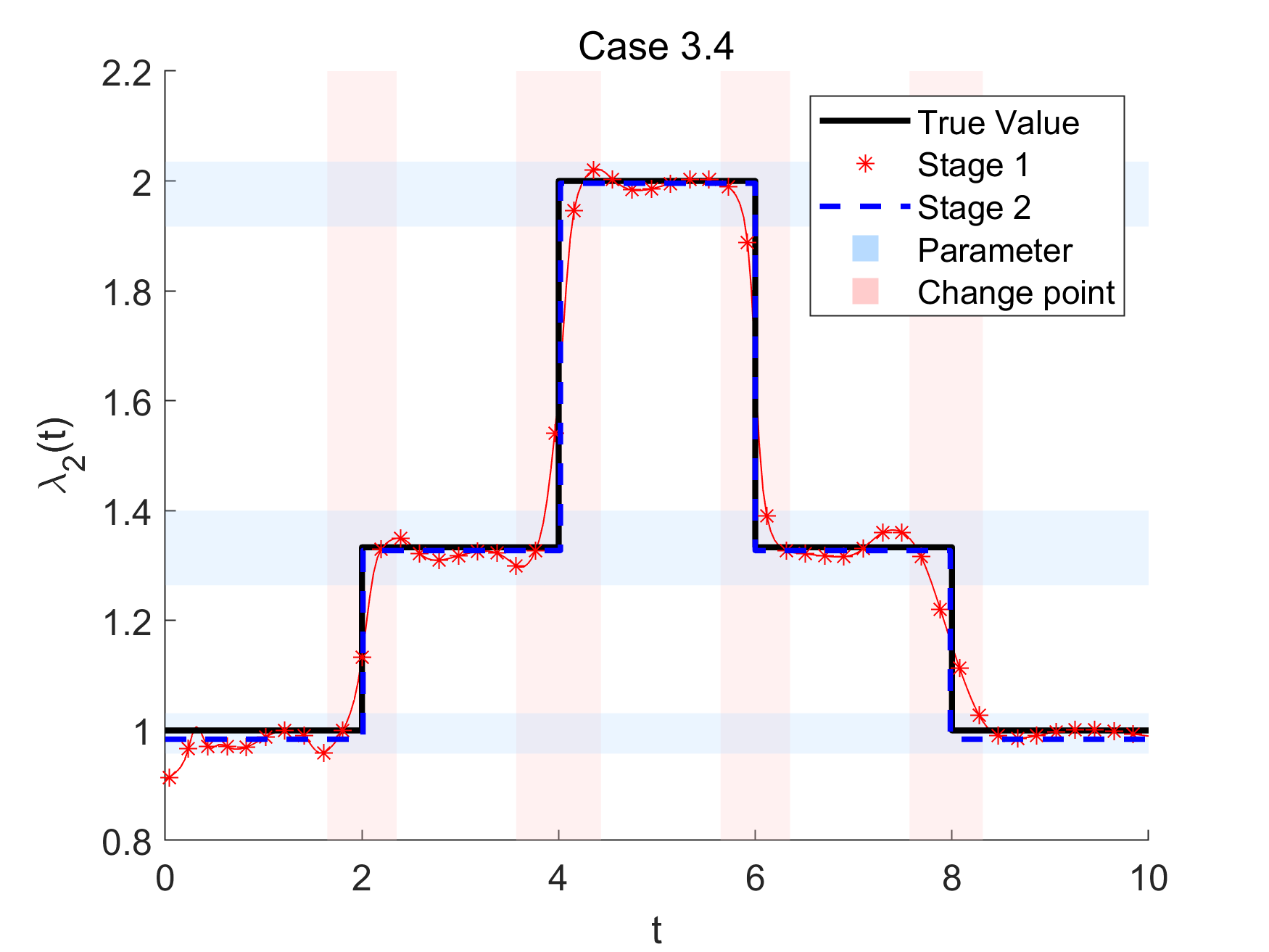}
			\end{minipage}
		}\\
		\subfloat[Reference solution]{
			\begin{minipage}[t]{0.315\linewidth}
				\includegraphics[width=1\linewidth]{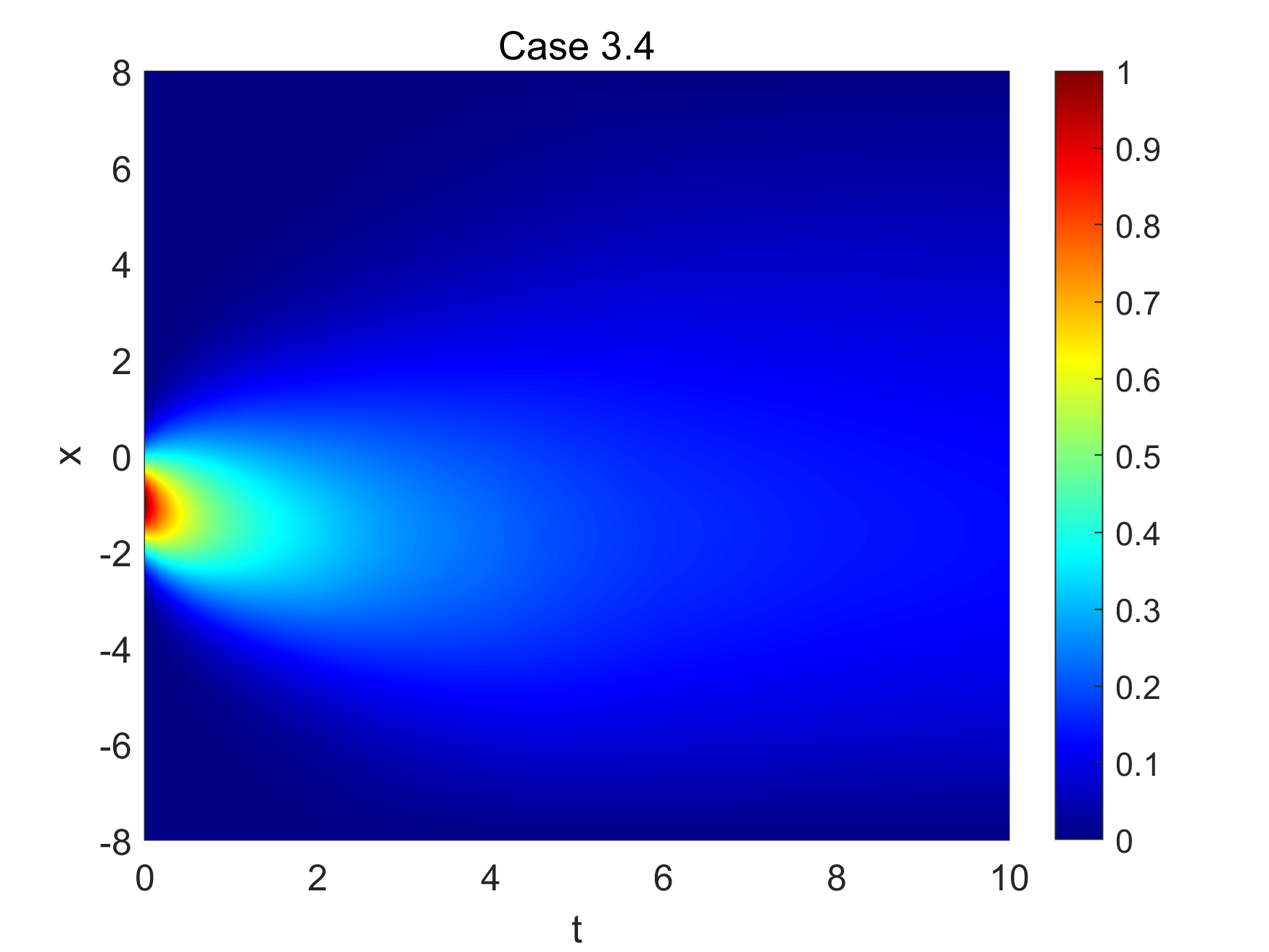}
			\end{minipage}
		}
		\subfloat[Predicted solution]{
			\begin{minipage}[t]{0.315\linewidth}
				\includegraphics[width=1\linewidth]{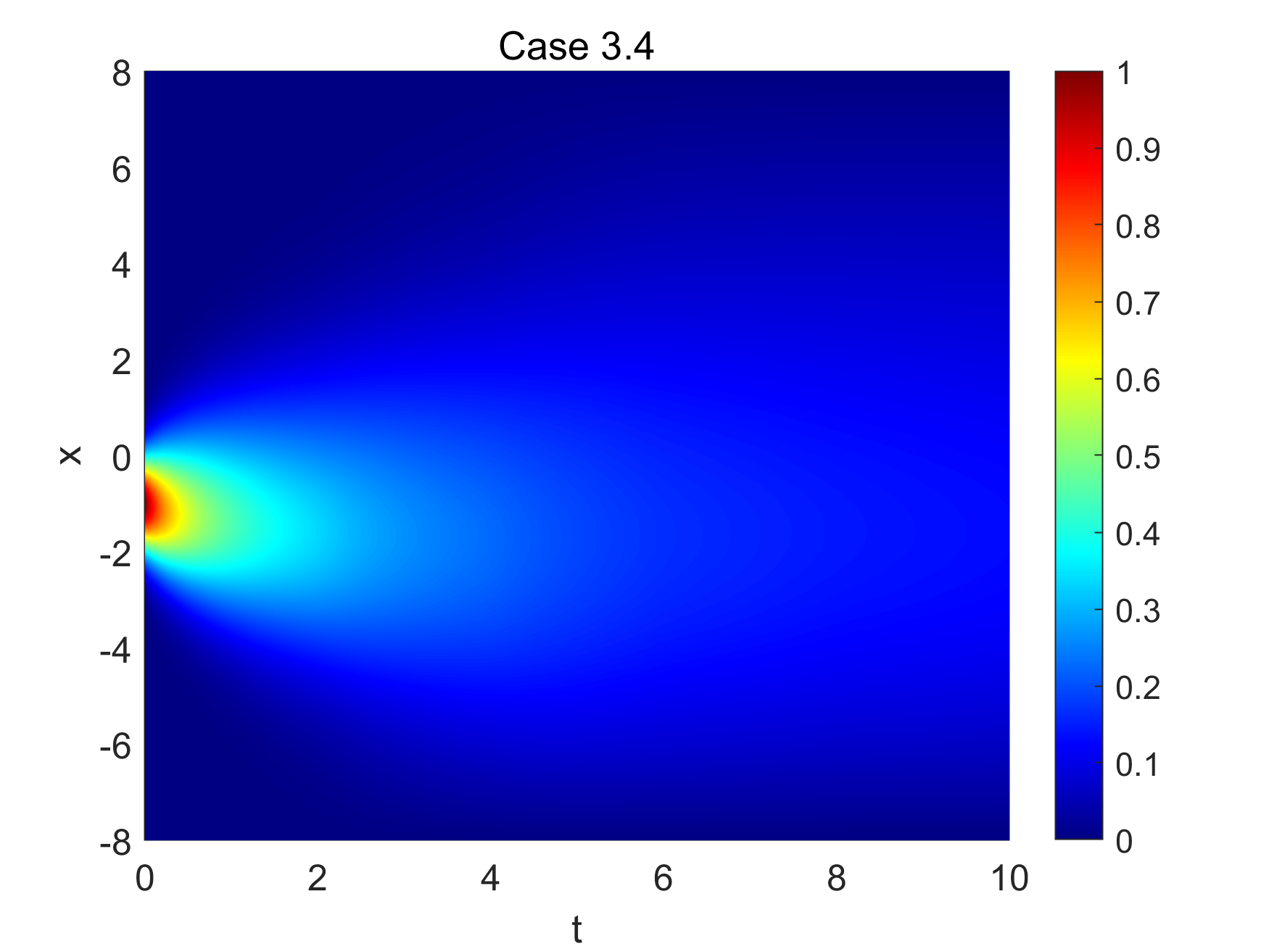}
			\end{minipage}
		}
		\subfloat[Absolute error]{
			\begin{minipage}[t]{0.315\linewidth}
				\includegraphics[width=1\linewidth]{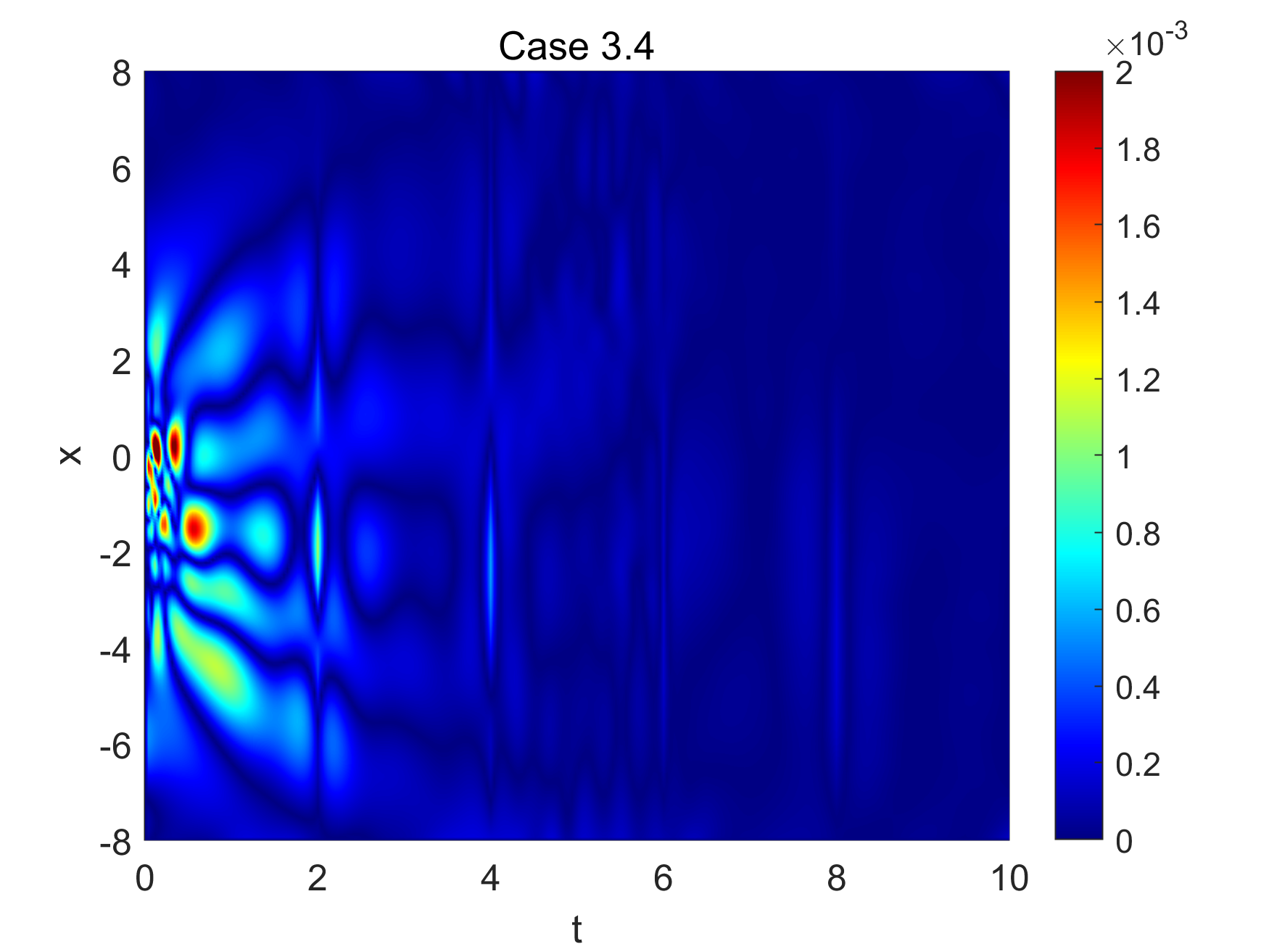}
			\end{minipage}
		}
		\caption{Numerical results for Burgers equations with discontinuously time-varying coefficients $\lambda_1(t)$ and $\lambda_2(t)$, from Case~3.3 to Case~3.4. \label{fig5}}
	\end{figure}
	
	\subsection{Burgers Equation}
	
	The Burgers equation is a nonlinear PDE that describes one-dimensional fluid motion under appropriate conditions. In this example, the coefficients $\lambda_1(t)$ and $\lambda_2(t)$ may exhibit temporal jumps, so that the physical parameter vector of the system is $\theta_p=\{\lambda_1(t),\lambda_2(t)\}$. This example is taken from \cite{mattey2022novel}, and the corresponding 1+1D PDE is given by
	\begin{equation}
		\begin{cases}
			u_t = \lambda_1(t) u u_x+\lambda_2(t) u_{xx}, & \quad (x,t)\in U \times (0,T],\\[2pt]
			u(x,t)=0, & \quad (x,t)\in \partial U \times (0,T],\\[6pt]
			u(x,0)=g(x), & \quad x \in U.
		\end{cases}
	\end{equation}
	Let $U=[-8,8]$ and $T=10$, with the initial value function
	\begin{equation}
		g(x)=\exp\big({-(x+1)^2}\big).
	\end{equation}
	
	Four scenarios are considered, in which the parameters exhibit different types of temporal behavior, including constant coefficients, single-parameter jumps, multi-state switching, and simultaneous jumps of multiple coefficients.
	
	\textbf{Case~3.1:}
	\begin{equation}
		\lambda_1(t)=1.5,\quad \lambda_2(t)=0.1,\quad t\in[0,10].
	\end{equation}
	
	\textbf{Case~3.2:}
	\begin{equation}
		\lambda_1(t)=
		\begin{cases}
			0.5, & t\in[0,5),\\[2pt]
			1, & t\in[5,10],
		\end{cases}
		\quad
		\lambda_2(t)=0.1,\quad t\in[0,10].
	\end{equation}
	
	\textbf{Case~3.3:}
	\begin{equation}
		\lambda_1(t)=
		\begin{cases}
			1, & t\in[0,2)\cup[8,10],\\[2pt]
			0.75, & t\in[2,4)\cup[6,8),\\[2pt]
			0.5, & t\in[4,6),
		\end{cases}
		\quad
		\lambda_2(t)=0.1,\quad t\in[0,10].
	\end{equation}
	
	\textbf{Case~3.4:}
	\begin{equation}
		\lambda_1(t)=
		\begin{cases}
			1, & t\in[0,2)\cup[8,10],\\[2pt]
			0.75, & t\in[2,4)\cup[6,8),\\[2pt]
			0.5, & t\in[4,6),
		\end{cases}
		\quad
		\lambda_2(t)=
		\begin{cases}
			1, & t\in[0,2)\cup[8,10],\\[2pt]
			4/3, & t\in[2,4)\cup[6,8),\\[2pt]
			2, & t\in[4,6).
		\end{cases}
	\end{equation}
	
	From Case~3.1 to Case~3.4, $\theta_p(t)=(\lambda_1(t),\lambda_2(t))$ on $(x,t)\in[-8,8]\times[0,10]$. The Stage~1 main network had layers $[2,128,128,128,128,128,128,1]$, and the coefficient network had layers $[1,40,40,40,40,2]$, Adam for $200000$ iterations with a learning rate of $10^{-3}$ for both networks and decay factors of $0.8$ and $0.9$ every $20000$ iterations. The training set used a $50\times80=4000$ residual grid, $4000$ random observations sampled with seed $314$, $100$ analytic initial-condition points, and $200$ homogeneous Dirichlet boundary points. The remaining loss weights were $\gamma_1=1$, $\gamma_2=0.5$ and $\gamma_3=0.5$. GMM-BDMC was applied separately to $\lambda_1$ and $\lambda_2$. Stage~2 used hard-step $\tilde\lambda_1(t)$ and $\tilde\lambda_2(t)$ initialized from Stage~1 stable-platform means and detector intervals. The Stage~2 included a $2500$ iteration refinement at learning rate $10^{-3}$ in both main and sub networks.
	
	The solution images for these four Burgers equation cases are shown in the first columns of Figure~\ref{fig4} and Figure~\ref{fig5}. It can be observed that directly inferring the existence and locations of temporal coefficient jumps from the solution images is not intuitive. Therefore, these examples provide a meaningful test for the proposed JVC-PINNs framework in extracting hidden jump-varying physical coefficients from observed solution data. The two-stage parameter inversion results for $\lambda_1(t)$ and $\lambda_2(t)$ are presented in the second and third columns of Figure~\ref{fig4} and Figure~\ref{fig5}. In these plots, the Stage~1 curves denote the relaxed continuous approximations generated by the GWS-PINNs coefficient sub network, while the Stage~2 curves denote the hard piecewise-constant estimators refined by CCD-PINNs. The heuristic coefficient search intervals and candidate change-point intervals are inferred by the GMM-BDMC statistical learner from the Stage~1 samples and are subsequently used as constraints in the Stage~2 refinement. Since two coefficients are involved, the statistical mixture modeling and constrained refinement are applied to $\lambda_1(t)$ and $\lambda_2(t)$ separately, allowing different parameters to have different numbers of states and different change-point structures. The reconstructed solution images from the second-stage main network $\tilde{u}$ and the corresponding absolute errors with respect to the reference solutions are displayed in the second and fourth rows of Figure~\ref{fig4} and Figure~\ref{fig5}. Similarly, the quantitative results are summarized in Table~\ref{tab1}, Table~\ref{tab2} and Table~\ref{tab4}.
	
	Case~3.1 corresponds to the constant-coefficient setting, where both $\lambda_1(t)$ and $\lambda_2(t)$ remain unchanged over the entire time interval. For both coefficients, the GMM-BDMC learner identifies $\widehat{K}=1$, indicating that no temporal change point is detected. Therefore, the second-stage CCD-PINNs refinement does not introduce any change-point variable and reduces to a constrained constant-coefficient inverse estimator. This confirms that the proposed framework can automatically distinguish constant coefficients from jump-varying coefficients and does not artificially create discontinuities when the underlying parameter is constant.
	
	Case~3.2 considers a single temporal jump in $\lambda_1(t)$, while $\lambda_2(t)$ remains constant. The GMM-BDMC learner identifies two states for $\lambda_1(t)$ and one state for $\lambda_2(t)$. Accordingly, a candidate change-point interval is introduced only for $\lambda_1(t)$, while no change point is introduced for $\lambda_2(t)$. The Stage~1 approximation of $\lambda_1(t)$ captures the transition behavior but smooths the discontinuity near the jump time. The Stage~2 refinement converts this relaxed approximation into a sharp step-function estimator and produces an accurate refined change point.
	
	Case~3.3 contains multi-state switching in $\lambda_1(t)$ and a constant $\lambda_2(t)$. The proposed framework identifies multiple coefficient states and several candidate change-point intervals for $\lambda_1(t)$, while automatically detecting that $\lambda_2(t)$ has only one constant state. The Stage~2 estimator accurately reconstructs the piecewise-constant structure of $\lambda_1(t)$ and preserves the constant behavior of $\lambda_2(t)$. This case demonstrates that the method can handle mixed situations in which only part of the physical coefficients are jump-varying.
	
	Case~3.4 further increases the difficulty by allowing both $\lambda_1(t)$ and $\lambda_2(t)$ to switch among multiple states. The statistical learner is applied to the two parameters separately and detects the corresponding state structures and candidate change-point intervals. The Stage~2 CCD-PINNs refinement then produces hard piecewise-constant reconstructions for both coefficients. The results show that the proposed method remains effective when multiple PDE coefficients undergo simultaneous discontinuous temporal variations.
	
	The previous examples covered several discontinuous coefficient-variation scenarios, including constant coefficients, a single jump in one parameter, multiple jumps in one parameter, and simultaneous jumps in multiple parameters. The proposed framework incorporates two additional continuously varying coefficient patterns, which means a piecewise-smooth transition and a trigonometric variation. These two extra scenarios of continuously time-varying coefficients in Burgers equations are set as
	
	\textbf{Case~3.5:}
	\begin{equation}
		\lambda_1(t)= \begin{cases}0.5, & 0 \leq t < \dfrac{305}{64}, \\[2pt] \dfrac{64}{65}t-\dfrac{109}{26}, & \dfrac{305}{64} \leq t < \dfrac{675}{128}, \\[2pt] 1, & \dfrac{675}{128} \leq t \leq 10. \end{cases}  \quad \lambda_2(t)=0.1, \quad t\in [0,10].
	\end{equation}
	
	\textbf{Case~3.6:}
	\begin{equation}
		\lambda_1(t)= -1-\dfrac{\sin t}{4}, \quad \lambda_2(t)=0.1,\quad t\in [0,10].
	\end{equation}
	
\begin{figure}[p]
		\centering
		\subfloat[Spatiotemporal solution]{
			\begin{minipage}[t]{0.315\linewidth}
				\includegraphics[width=1\linewidth]{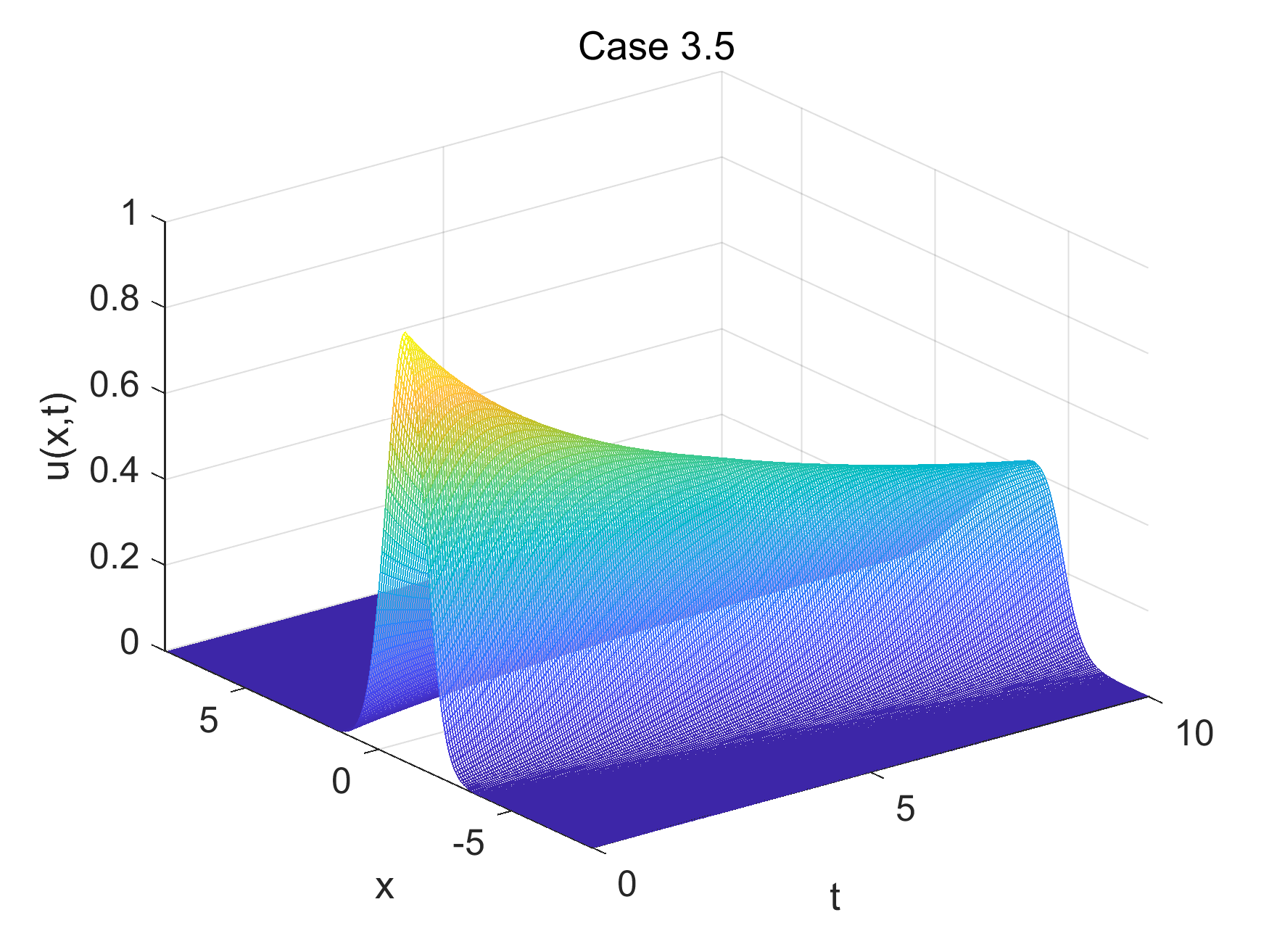}
			\end{minipage}
		}
		\subfloat[Parameter sample result]{
			\begin{minipage}[t]{0.315\linewidth}
				\includegraphics[width=1\linewidth]{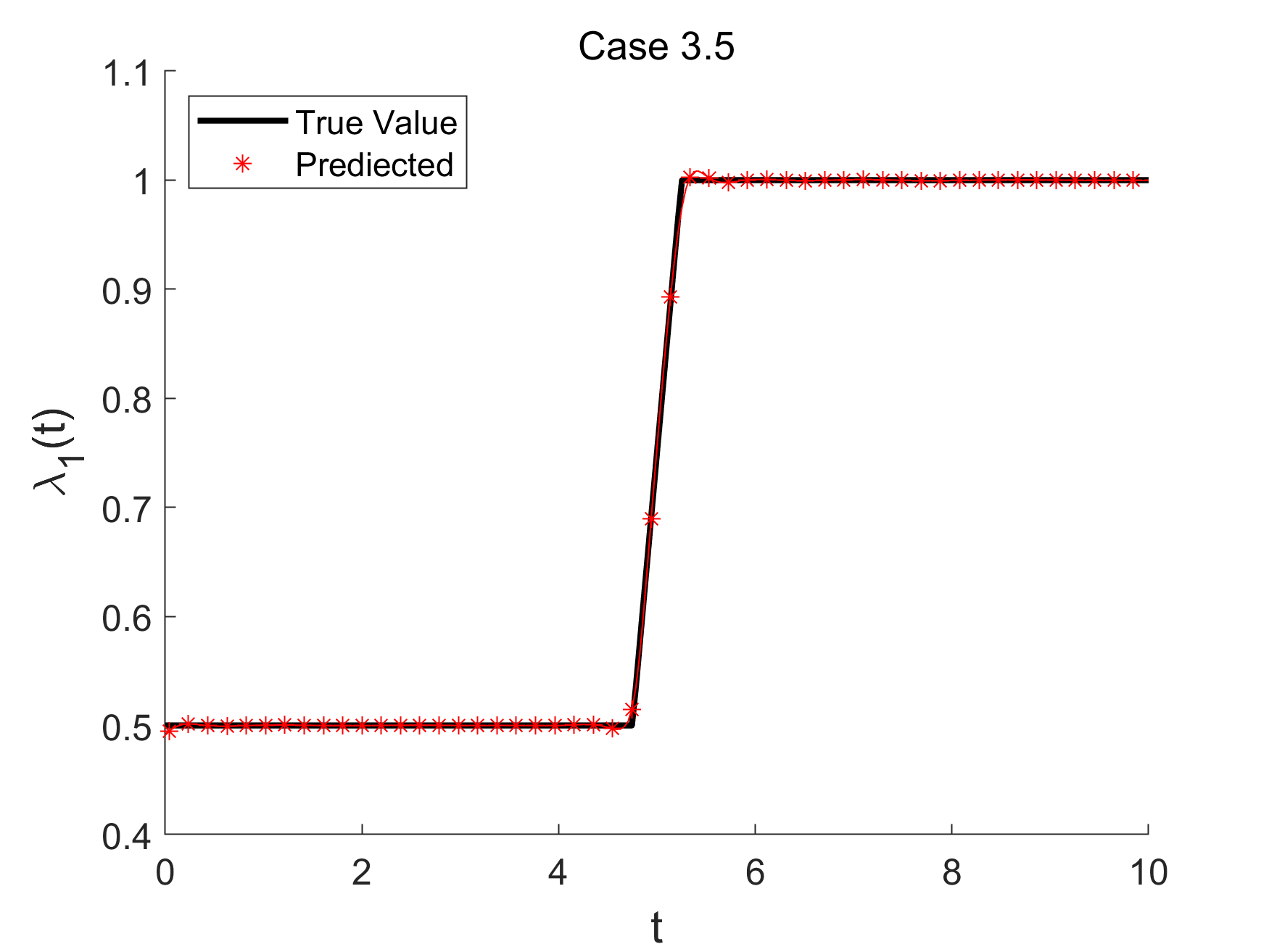}
			\end{minipage}
		}
		\subfloat[Parameter sample result]{
			\begin{minipage}[t]{0.315\linewidth}
				\includegraphics[width=1\linewidth]{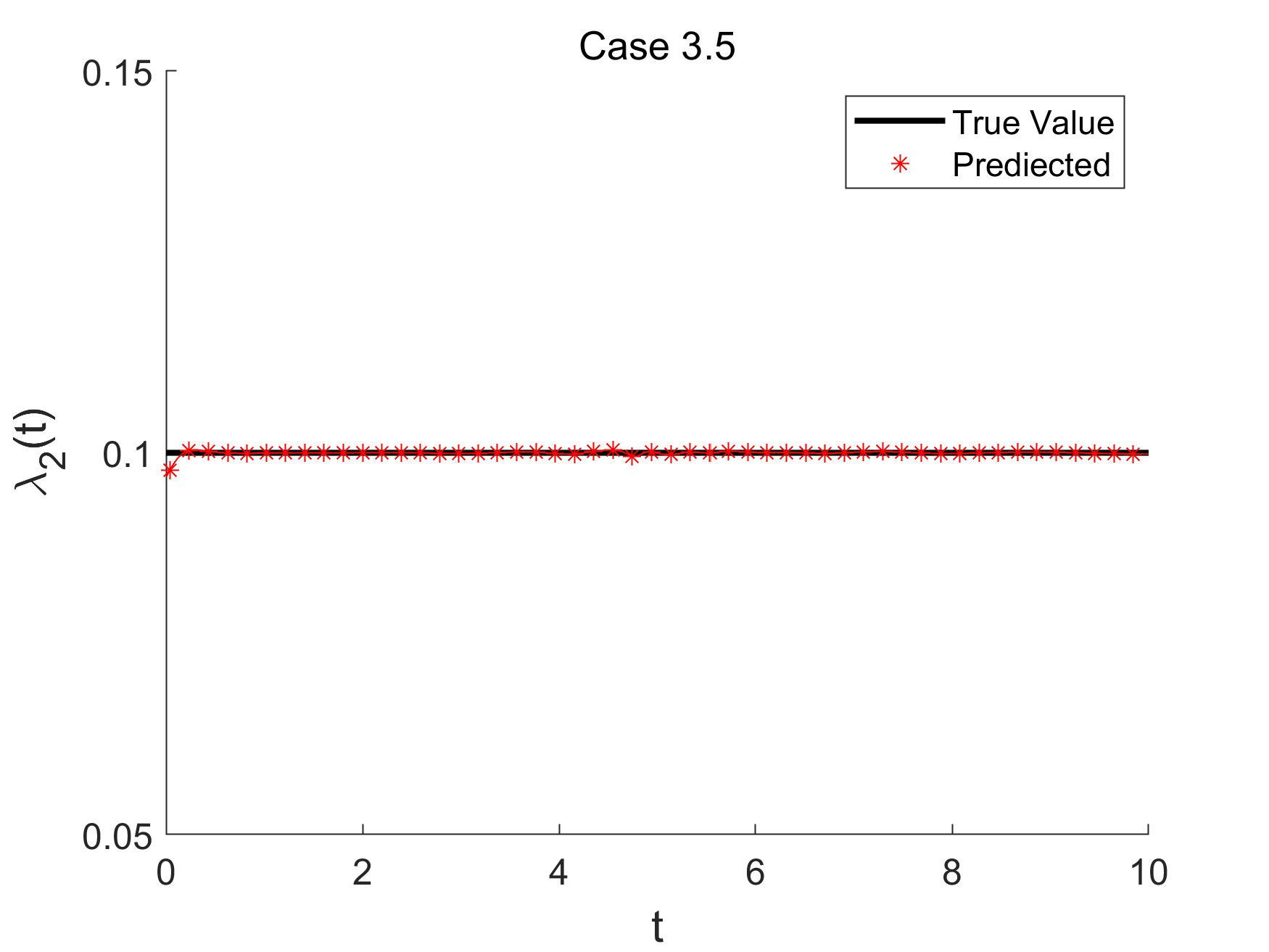}
			\end{minipage}
		}
		\\
		\subfloat[Reference solution]{
			\begin{minipage}[t]{0.315\linewidth}
				\includegraphics[width=1\linewidth]{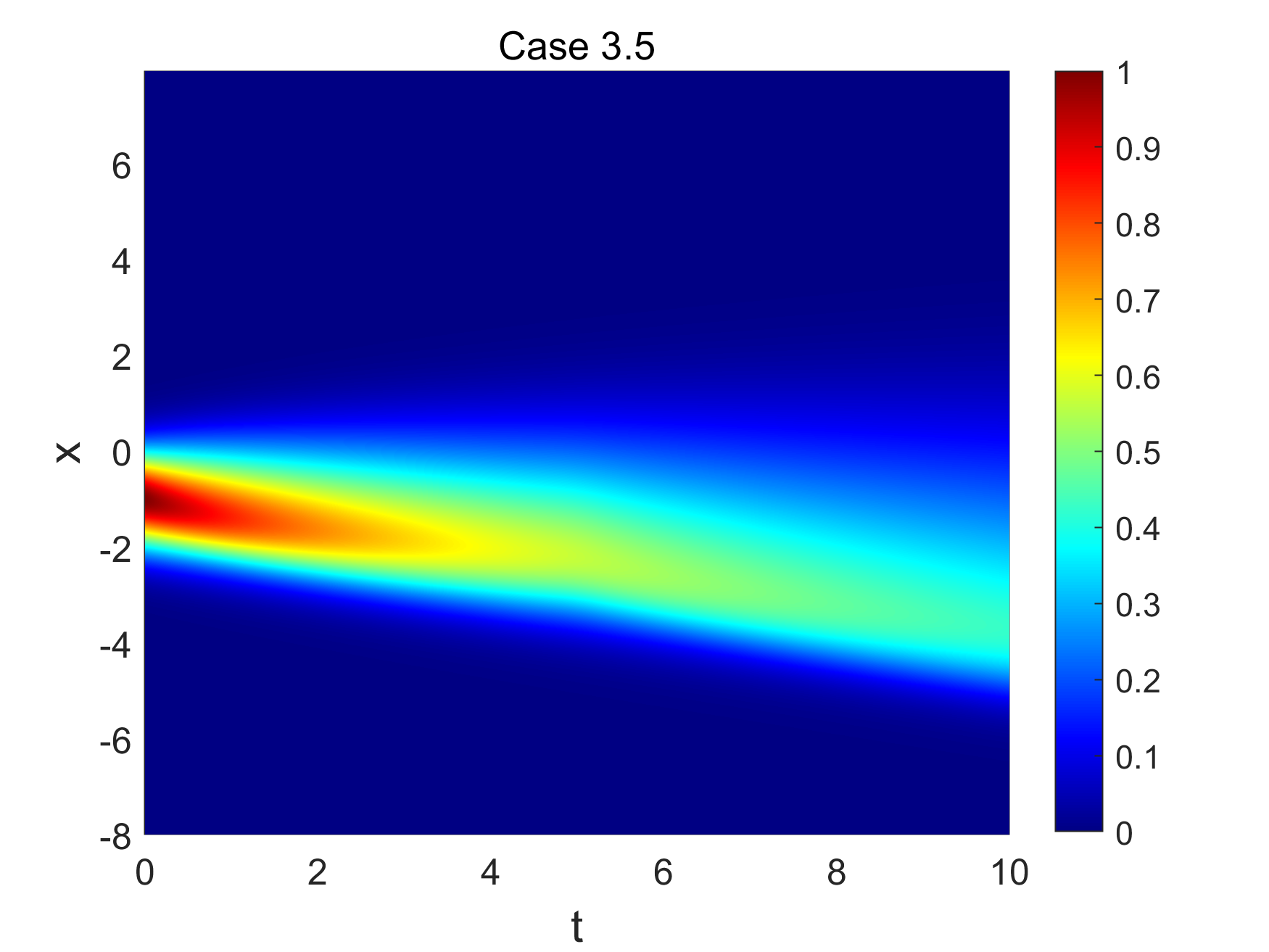}
			\end{minipage}
		}
		\subfloat[Predicted solution]{
			\begin{minipage}[t]{0.315\linewidth}
				\includegraphics[width=1\linewidth]{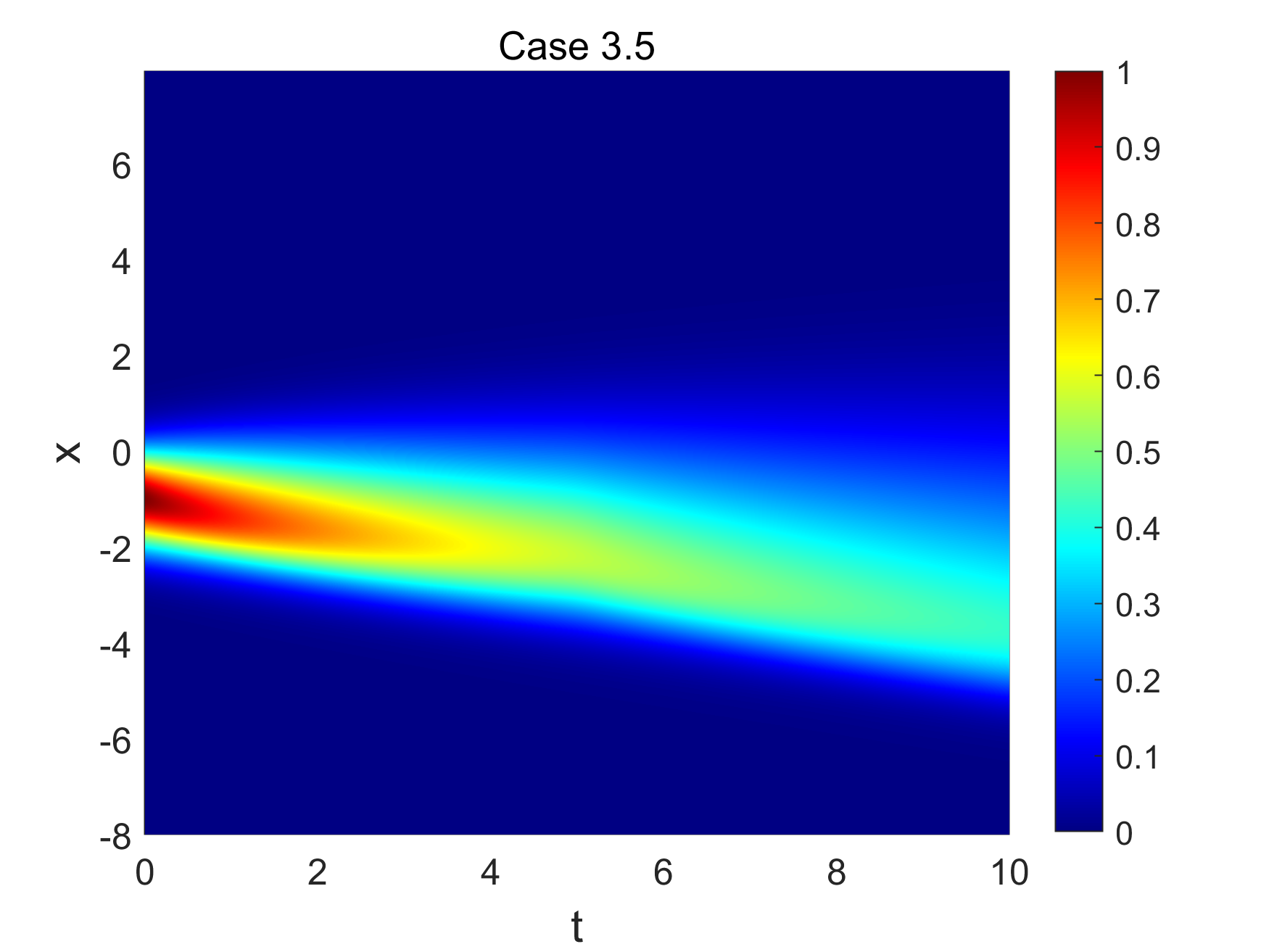}
			\end{minipage}
		}
		\subfloat[Absolute error]{
			\begin{minipage}[t]{0.315\linewidth}
				\includegraphics[width=1\linewidth]{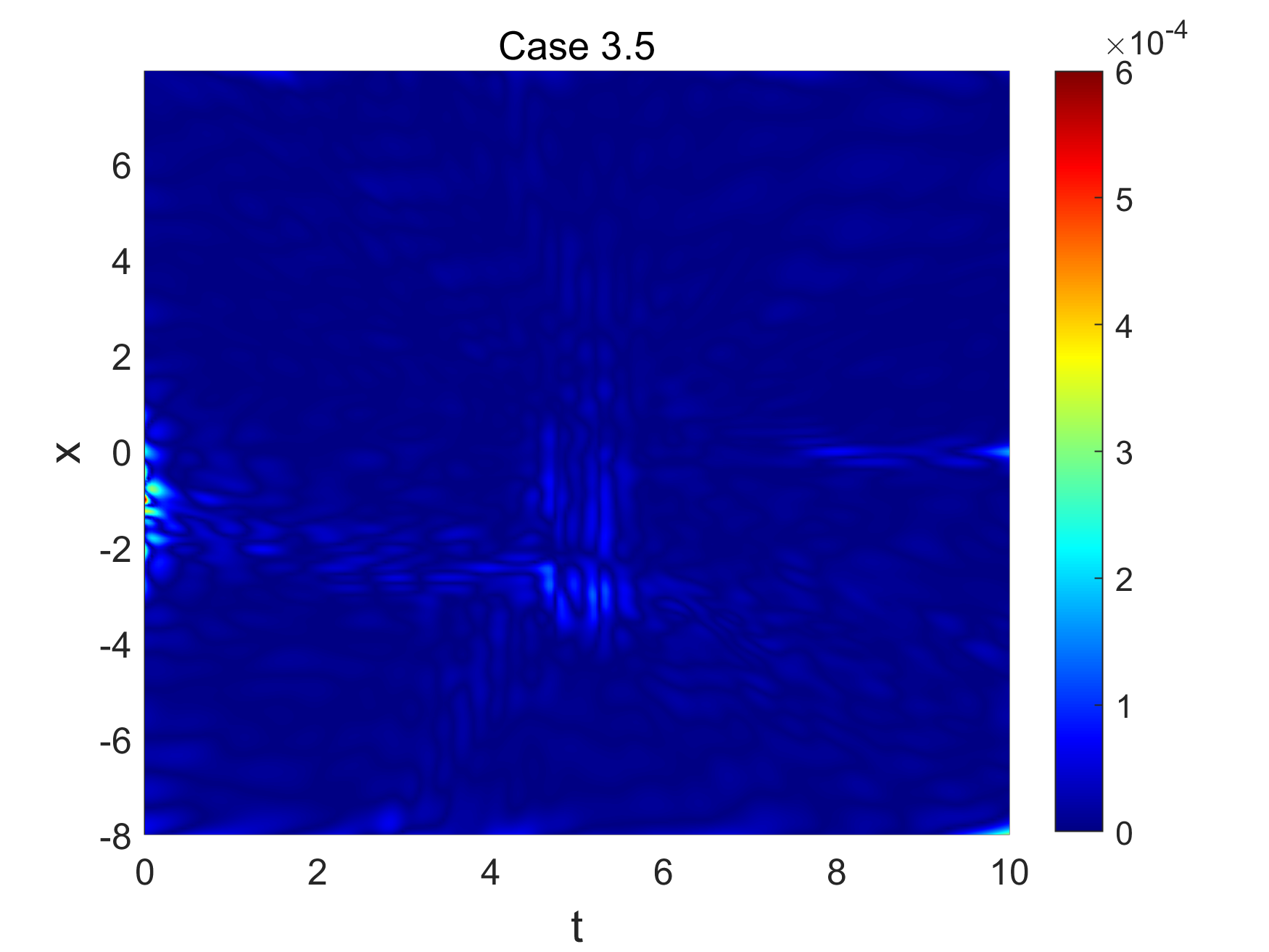}
			\end{minipage}
		}\\
		\subfloat[Spatiotemporal solution]{
			\begin{minipage}[t]{0.315\linewidth}
				\includegraphics[width=1\linewidth]{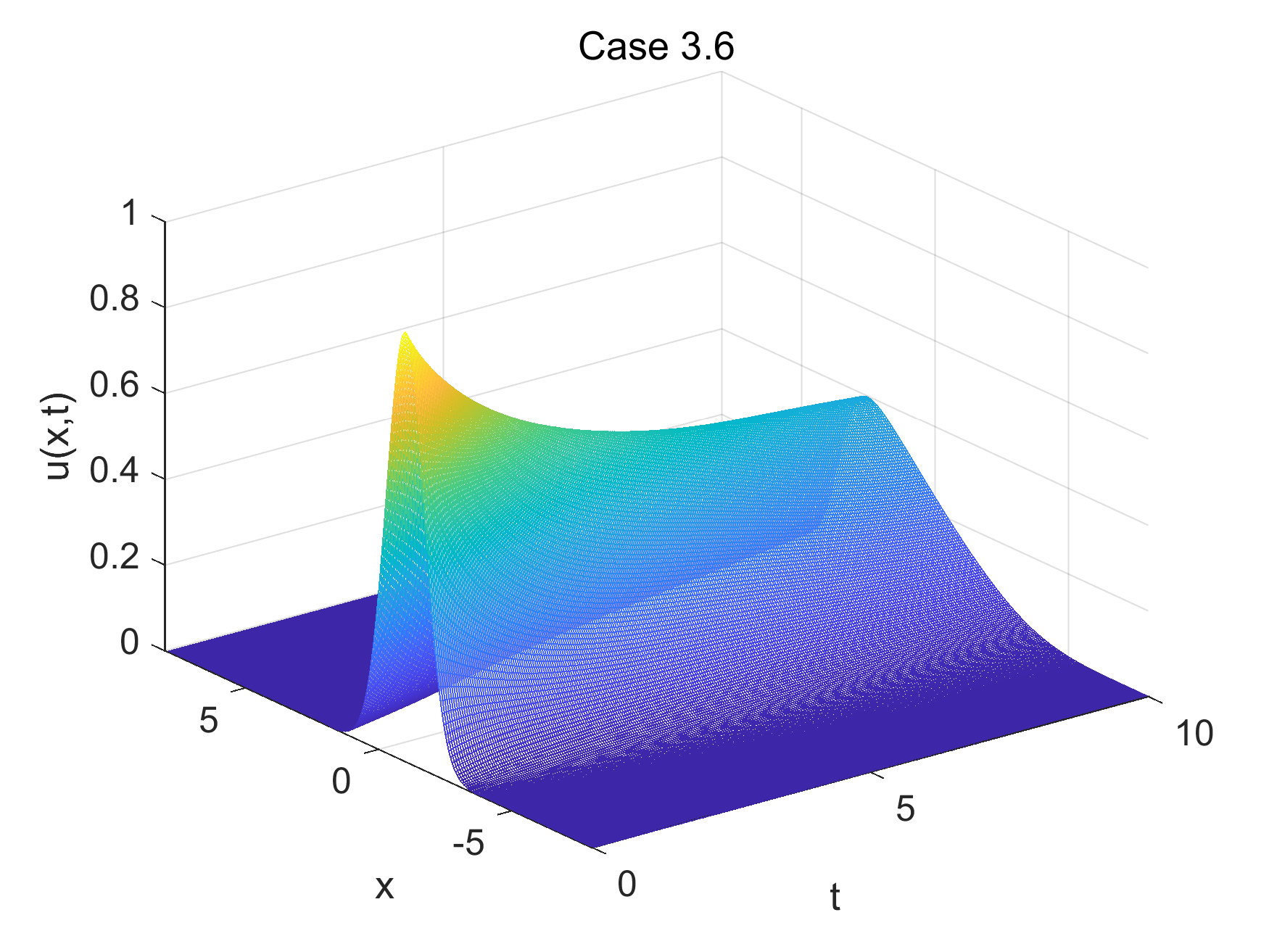}
			\end{minipage}
		}
		\subfloat[Parameter sample result]{
			\begin{minipage}[t]{0.315\linewidth}
				\includegraphics[width=1\linewidth]{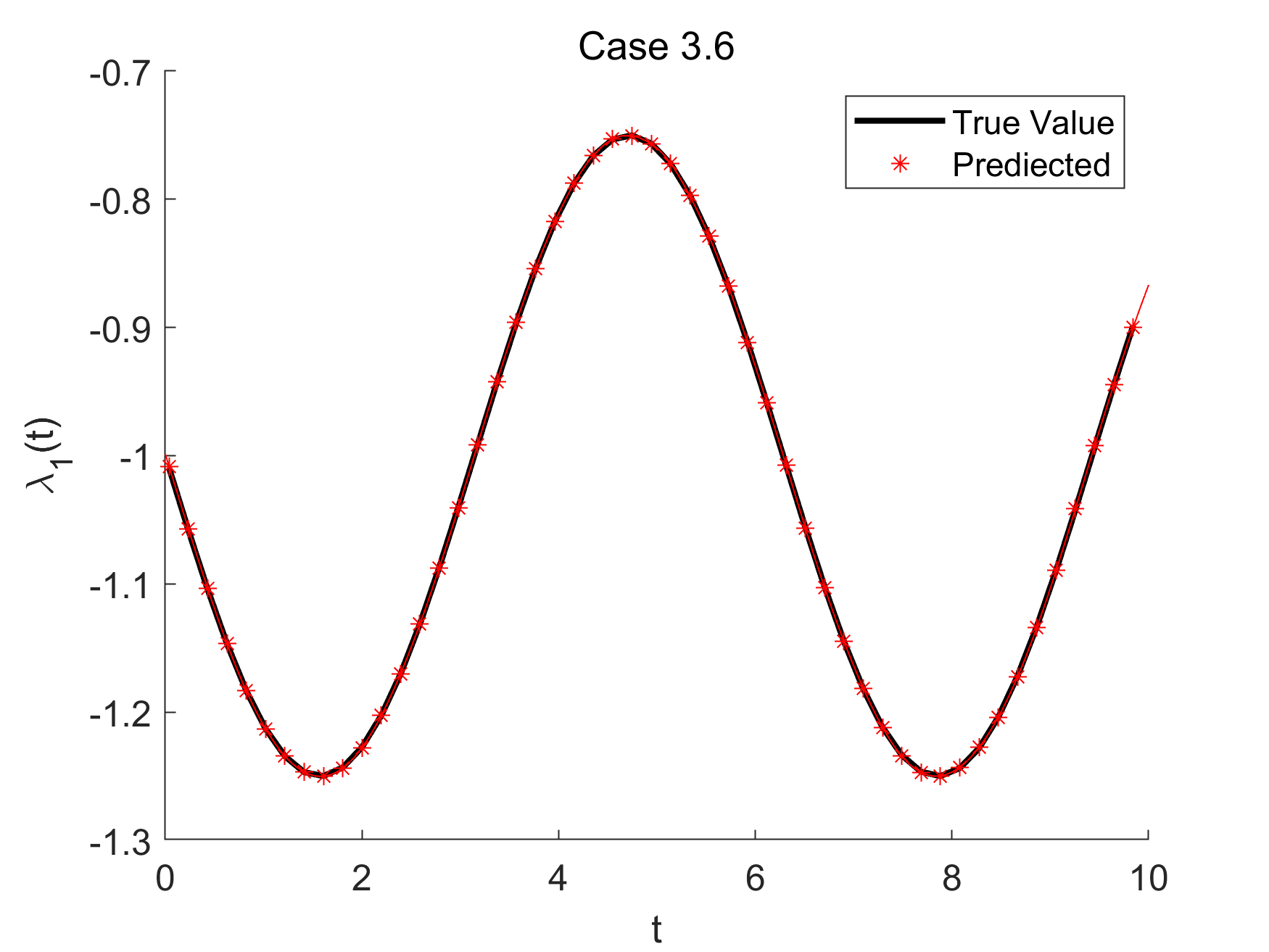}
			\end{minipage}
		}
		\subfloat[Parameter sample result]{
			\begin{minipage}[t]{0.315\linewidth}
				\includegraphics[width=1\linewidth]{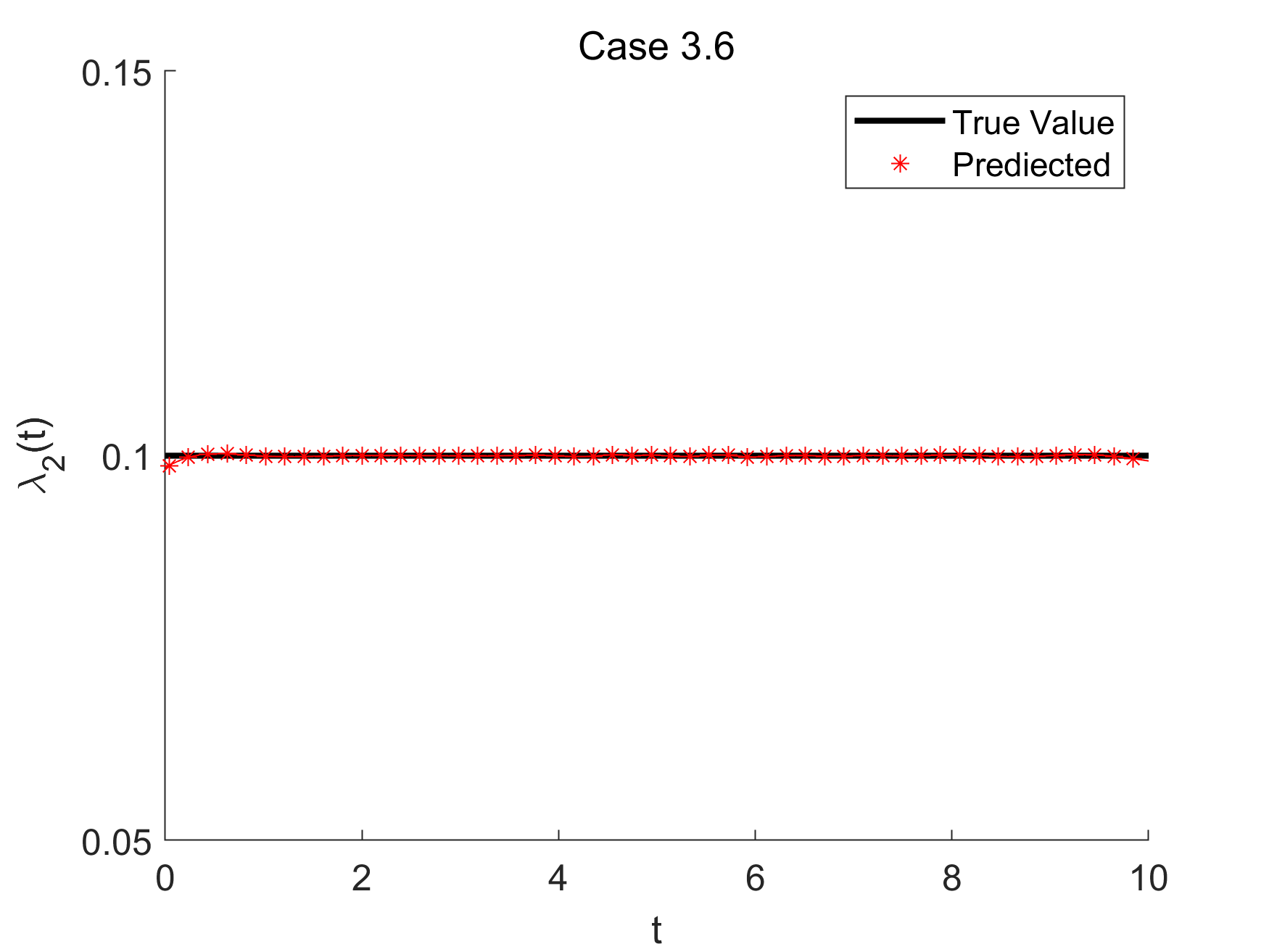}
			\end{minipage}
		}\\
		\subfloat[Reference solution]{
			\begin{minipage}[t]{0.315\linewidth}
				\includegraphics[width=1\linewidth]{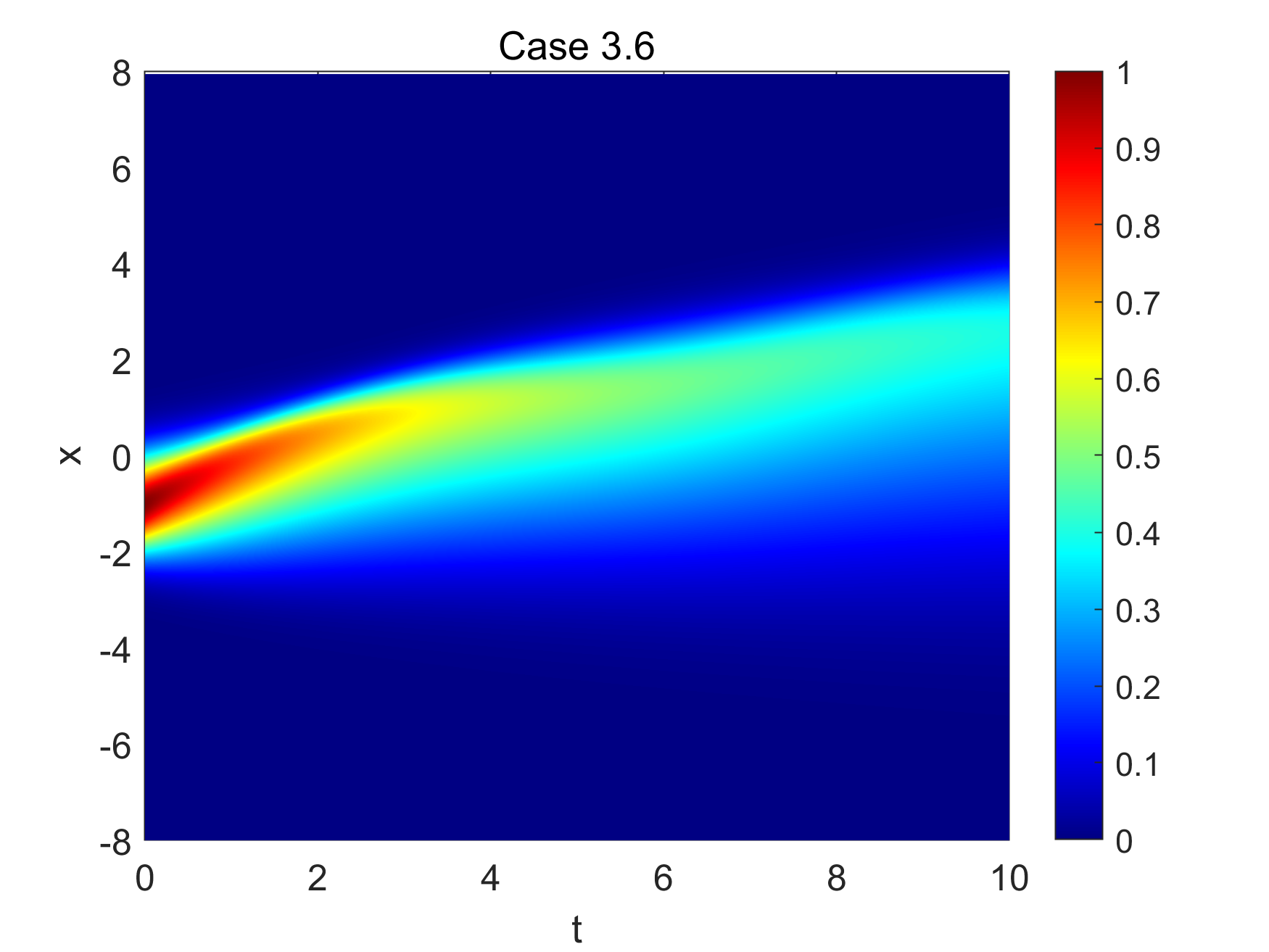}
			\end{minipage}
		}
		\subfloat[Predicted solution]{
			\begin{minipage}[t]{0.315\linewidth}
				\includegraphics[width=1\linewidth]{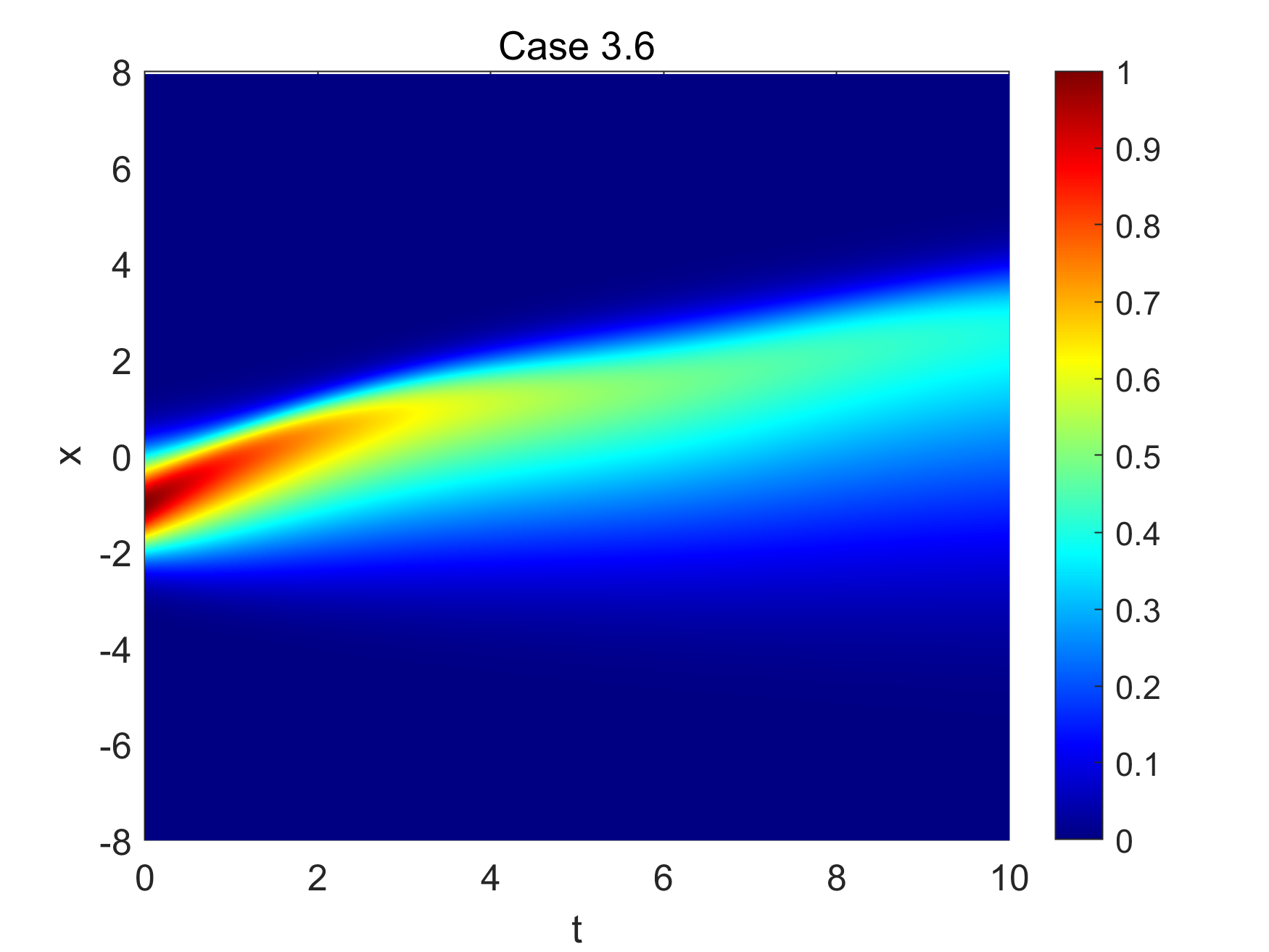}
			\end{minipage}
		}
		\subfloat[Absolute error]{
			\begin{minipage}[t]{0.315\linewidth}
				\includegraphics[width=1\linewidth]{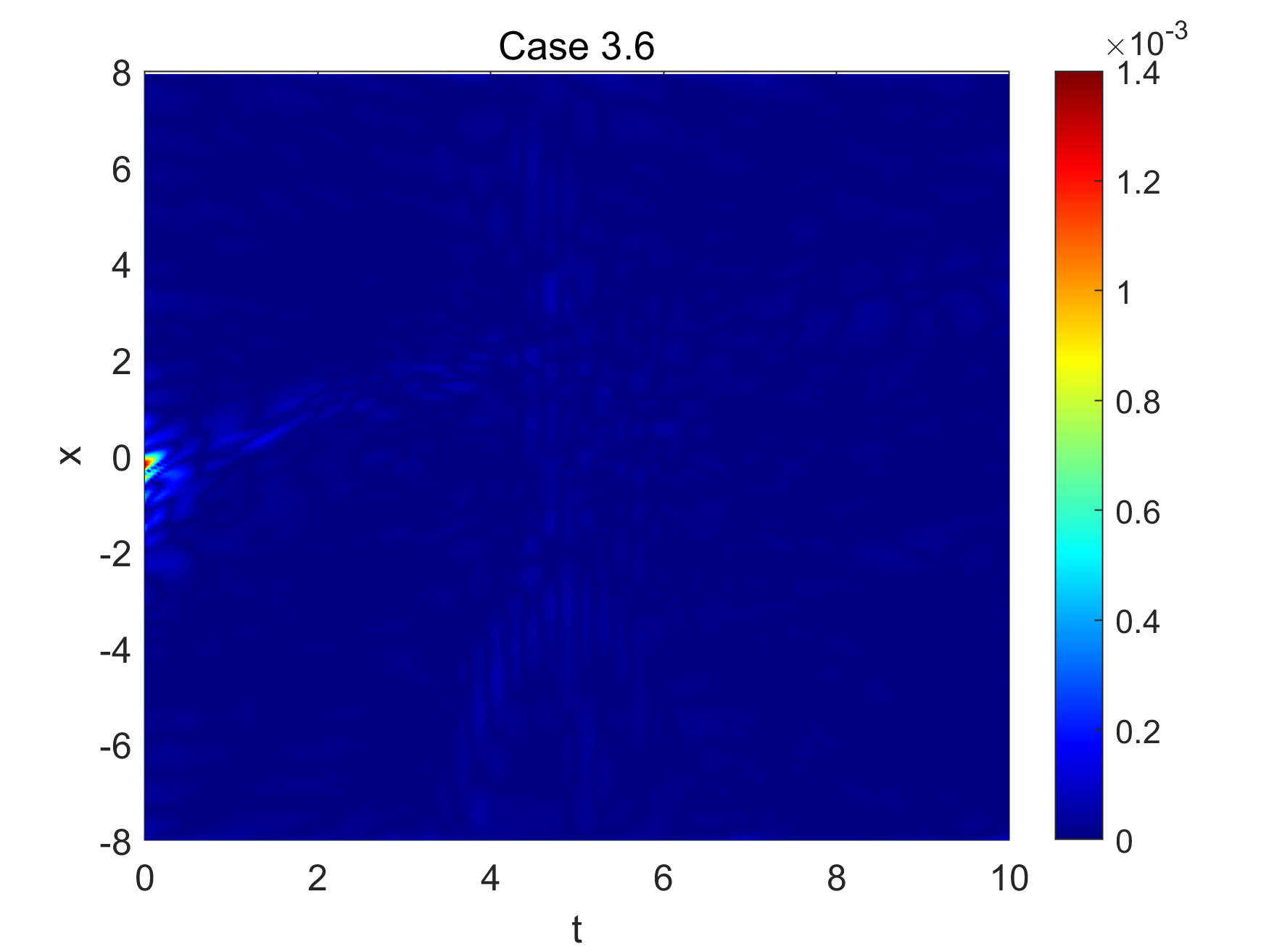}
			\end{minipage}
		}
		\caption{Numerical results for Burgers equations with continuously time-varying coefficients $\lambda_1(t)$ and $\lambda_2(t)$, from Case~3.5 to Case~3.6. \label{fig6}}
	\end{figure}

	For Case~3.5 and Case~3.6, because the coefficient fields contain no jump discontinuities, only Stage~1 was used and the experimental setup was the same as described above. The main network provided the solution approximation, and the coefficient sub network provided the parameter samples for the sampling process, without Stage~2 CCD-PINNs refinement. The GWS-PINNs sampling results for the continuously time-varying coefficient $\lambda_1(t)$ and the constant coefficient $\lambda_2(t)$ are shown in the second and third columns of the first and third rows of Figure~\ref{fig6}, respectively. Comparisons of the predictions of the solution images based on statistical inference results and the absolute errors with respect to the reference solutions are shown in the second and fourth rows of Figure~\ref{fig6}. According to the Stage~1 solution MSE reported in Table~\ref{tab4}, both cases achieved lower errors than the discontinuous-coefficient cases. Subsequent comparative studies validate that GWS-PINNs, through its improved loss function formulation, produces better sampling outcomes and deliver higher precision in coefficient estimation for both time-varying and time-invariant parameters. The adaptive weighting mechanism in the loss function enables more accurate identification of PDE coefficients with jump discontinuities, demonstrating competitive performance compared with existing PINNs approaches for variable-coefficient PDEs.
	
	It should be noted that the proposed JVC-PINNs framework is most naturally suited to inverse problems for PDE systems with piecewise-constant coefficients over different temporal or spatial regions. Cases 3.5 and 3.6 are included to clarify the behavior of the method when the target coefficient is not genuinely discontinuous. In Case~3.5, $\lambda_1(t)$ contains a smooth transition segment and is therefore close to a piecewise-constant pattern but does not contain a sharp jump. In Case~3.6, $\lambda_1(t)$ is a continuously varying trigonometric function. These two examples show that, when no true discontinuity exists, the first-stage GWS-PINNs alone can provide accurate reconstructions of both the physical solution field and the varying coefficient field. The sub-network approximation errors for these continuous time-varying coefficient cases are reported in Table~\ref{tab5}, confirming that Stage~1 GWS-PINNs can accurately recover both the varying coefficient $\lambda_1(t)$ and the constant coefficient $\lambda_2(t)$. The subsequent GMM-BDMC and CCD-PINNs components are specifically designed for piecewise-constant PDE coefficients. If they are forced onto continuously varying coefficients, they may introduce an inaccurate interpretation of the parameter variation behavior. Meanwhile, for the constant coefficient $\lambda_2(t)$ in these two cases, the GMM-BDMC inference gives $\widehat{K}=1$, indicating that the statistical learner does not incorrectly introduce change points for a constant physical parameter.
	
	Overall, the Burgers equation examples validate the effectiveness of the proposed JVC-PINNs framework for multi-parameter inverse problems with discontinuously time-varying coefficients. The Stage~1 GWS-PINNs sampler provides continuous but informative coefficient surrogates, the GMM-BDMC learner automatically determines the number of coefficient states and candidate change-point intervals, and the Stage~2 CCD-PINNs estimator refines the coefficient values and temporal discontinuities into explicit step-function representations. The accurate reconstruction of $\tilde{u}$ further confirms that the inferred jump-varying coefficients satisfy the governing PDE and are consistent with the observed dynamical behavior.
	
	\subsection{Navier-Stokes Equation}
	
	The Navier-Stokes equations are fundamental models for describing fluid motion and are widely studied in fluid mechanics, applied mathematics, and scientific computing. A 2+1D incompressible Navier-Stokes equation with a time-varying viscosity coefficient $\nu(t)$ is considered in the following form
	\begin{equation}
		\begin{cases}
			u_t + (u \cdot \nabla) u + \nabla p = \nu(t) \Delta u + f, & \quad (x,y,t) \in U \times (0,T],\\[2pt]
			\nabla \cdot u = 0, & \quad (x,y,t) \in U \times (0,T],\\[2pt]
			u(x,y,0) = u_0(x,y), & \quad (x,y) \in U,\\[2pt]
			u(x,y,t) = 0, & \quad (x,y,t) \in \partial U \times (0,T],
		\end{cases}
	\end{equation}
	where $\nu(t)>0$ denotes the viscosity coefficient, and $u(x,y,t)=(u_1(x,y,t),u_2(x,y,t))\in\mathbb{R}^2$ represents the velocity field. The initial condition is given by
	\begin{equation}
		u_0(x,y)=\left[g_1(x,y),g_2(x,y)\right]\in\mathbb{R}^2,
	\end{equation}
	where
	\begin{equation}
        \begin{aligned}
		      g_1(x,y) &= -200x^2(1-x)^2 \left[y(1-y)^2-y^2(1-y)\right],\\
            g_2(x,y) &= 200\left[x(1-x)^2-x^2(1-x)\right] y^2(1-y)^2.
     	\end{aligned}   
	\end{equation}
	The function $p(x,y,t)\in\mathbb{R}$ represents the fluid pressure, with the initial condition
	\begin{equation}
		p(x,y,0)=\cos(2\pi y)\left[\pi\sin(\pi x)-2\right].
	\end{equation}
	The external force is denoted by $f(x,y,t)\in\mathbb{R}^2$. In this experiment, $U=[0,1]\times[0,1]\subset\mathbb{R}^2$, $T=1$, and $f\equiv0$ are used. This example is taken from \cite{chen2024efficient}. The viscosity coefficient $\nu(t)$ is inversely related to the Reynolds number. Therefore, jumps in $\nu(t)$ could cause abrupt changes in the Reynolds number, introducing additional challenges for numerical computation.
	
	For the time-varying viscosity coefficient $\nu(t)$, eight cases are set as
	\begin{equation}
		\begin{aligned}
			\mathrm{\textbf{Case~4.1:}} \quad \nu(t) &= 0.01, \quad t \in [0,1],\\
			\mathrm{\textbf{Case~4.2:}} \quad \nu(t) &= 0.01, \quad t \in [0,0.5), \quad \nu(t) = 0.02, \quad t \in [0.5,1],\\
			\mathrm{\textbf{Case~4.3:}} \quad \nu(t) &= 0.01, \quad t \in [0,0.5), \quad \nu(t) = 0.03, \quad t \in [0.5,1],\\
			\mathrm{\textbf{Case~4.4:}} \quad \nu(t) &= 0.01, \quad t \in [0,0.5), \quad \nu(t) = 0.04, \quad t \in [0.5,1],\\
			\mathrm{\textbf{Case~4.5:}} \quad \nu(t) &= 0.01, \quad t \in [0,0.5), \quad \nu(t) = 0.05, \quad t \in [0.5,1],\\
			\mathrm{\textbf{Case~4.6:}} \quad \nu(t) &= 0.01, \quad t \in [0,0.5), \quad \nu(t) = 0.06, \quad t \in [0.5,1],\\
			\mathrm{\textbf{Case~4.7:}} \quad \nu(t) &= 0.01, \quad t \in [0,0.5), \quad \nu(t) = 0.08, \quad t \in [0.5,1],\\
			\mathrm{\textbf{Case~4.8:}} \quad \nu(t) &= 0.01, \quad t \in [0,0.5), \quad \nu(t) = 0.1, \:\; \quad t \in [0.5,1].
		\end{aligned}
	\end{equation}
	
	From Case~4.1 to Case~4.8, $\theta_p=\nu(t)$ for the two-dimensional Navier-Stokes system. The main network for $(u_1,u_2,p)$ used layers $[3,20,20,20,20,20,20,3]$, and the sub network used layers $[1,20,20,1]$. Using seed $314$, Adam for $20000$ iterations, learning rates $10^{-3}$ for the main network and $8\times10^{-4}$ for the sub network, $5000$ residual points, $5000$ observation points, $1024$ initial-condition points, and $4\times512$ boundary points. Data batches were refreshed every $5000$ iterations, logs were written every $100$ iterations, and full evaluation was performed every $500$ iterations. The loss weights were $\gamma_1=1$ and $\gamma_2= \gamma_3=100$. Stage~2 represented $\tilde\nu(t)$ as a hard step function, used Adam for $5000$ iterations with learning rates $5\times10^{-4}$ in main network for $\tilde u$ and $5\times10^{-3}$ for learnable change points and plateau values in sub network.
	
	\begin{figure}[p]
		\centering
		\subfloat[Parameter inverse result]{
			\begin{minipage}[t]{0.375\linewidth}
				\includegraphics[width=1\linewidth]{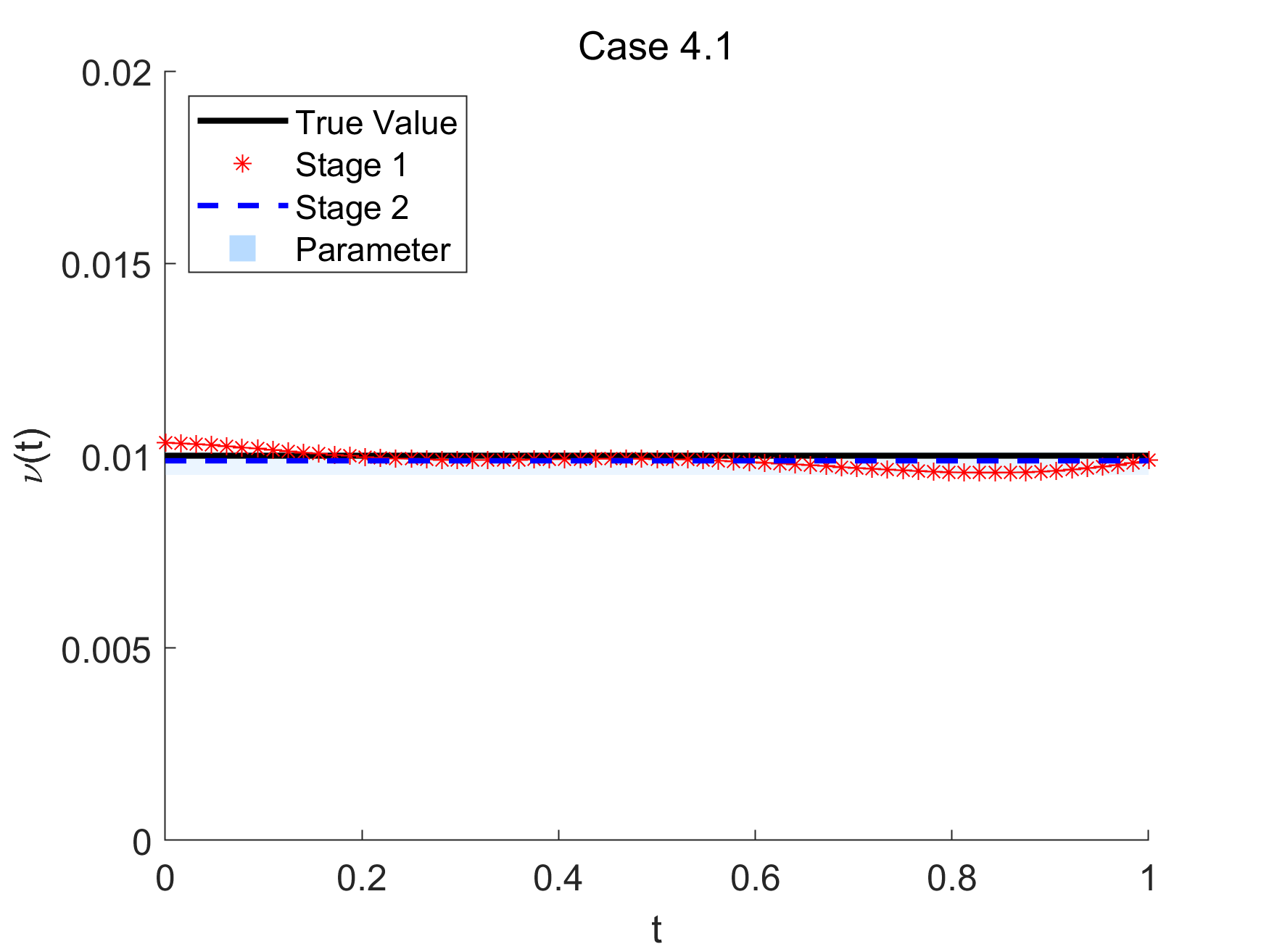}
			\end{minipage}
		}
		\subfloat[Parameter inverse result]{
			\begin{minipage}[t]{0.375\linewidth}
				\includegraphics[width=1\linewidth]{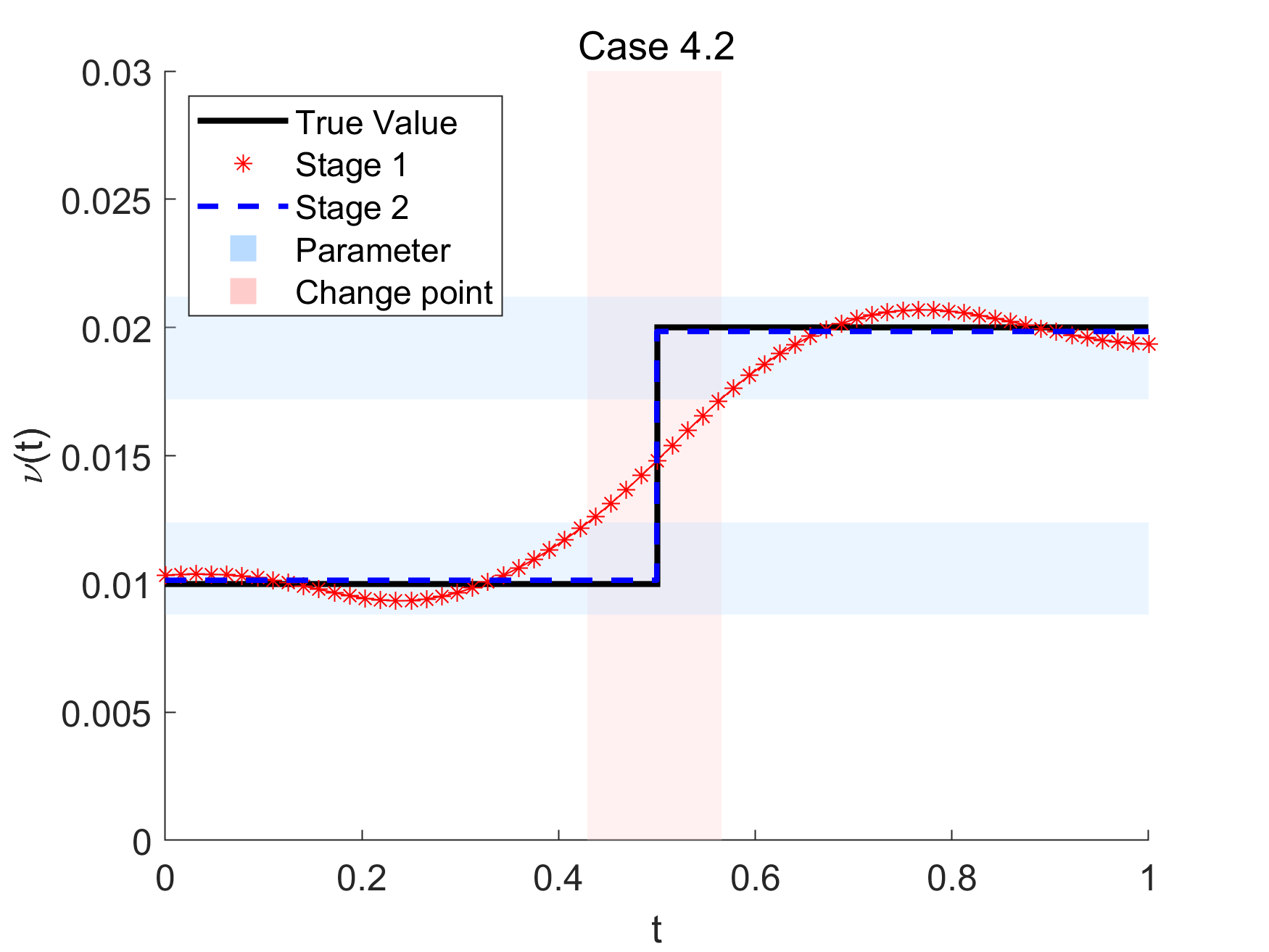}
			\end{minipage}
		}\\
		\subfloat[Parameter inverse result]{
			\begin{minipage}[t]{0.375\linewidth}
				\includegraphics[width=1\linewidth]{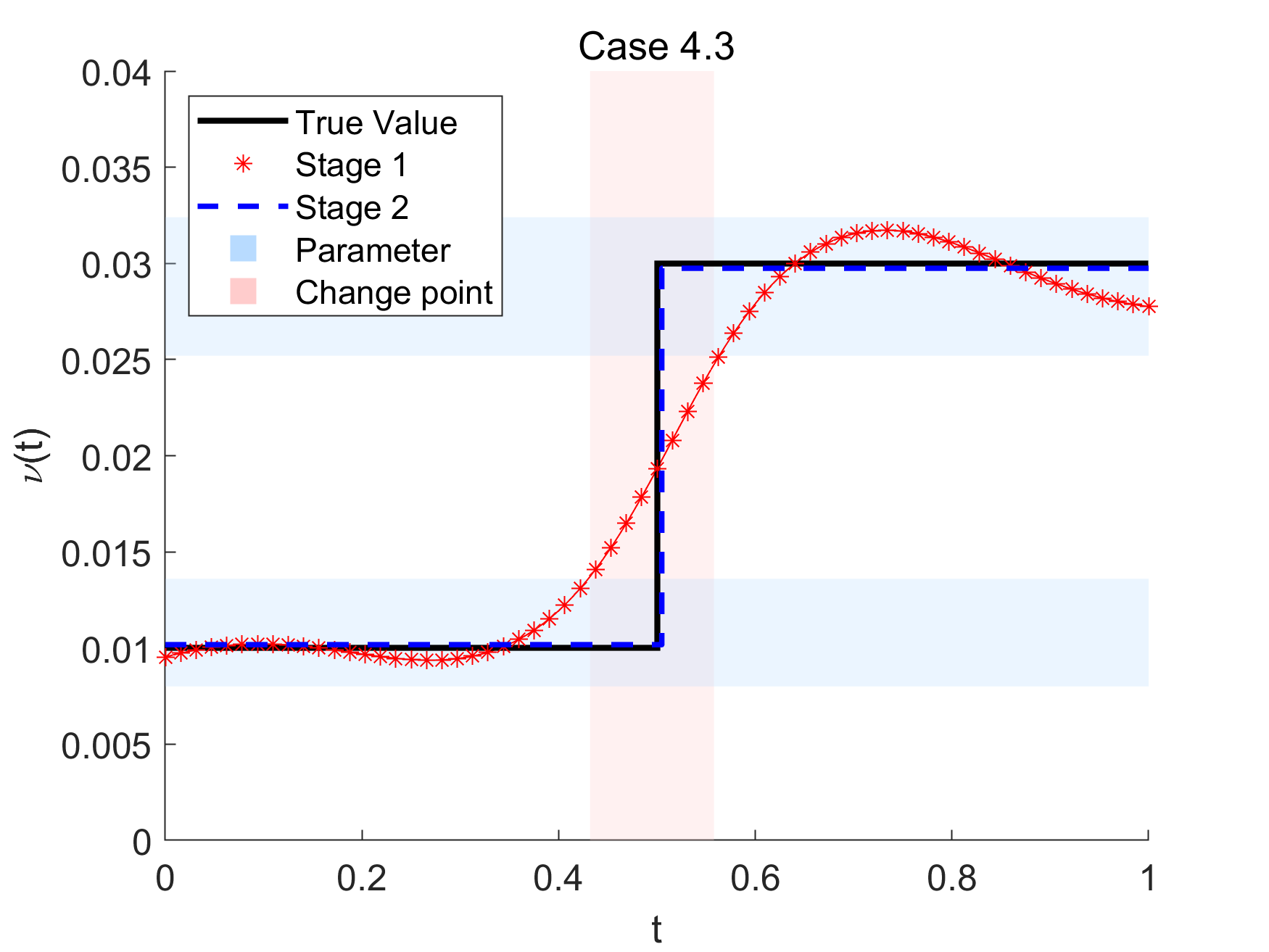}
			\end{minipage}
		}
		\subfloat[Parameter inverse result]{
			\begin{minipage}[t]{0.375\linewidth}
				\includegraphics[width=1\linewidth]{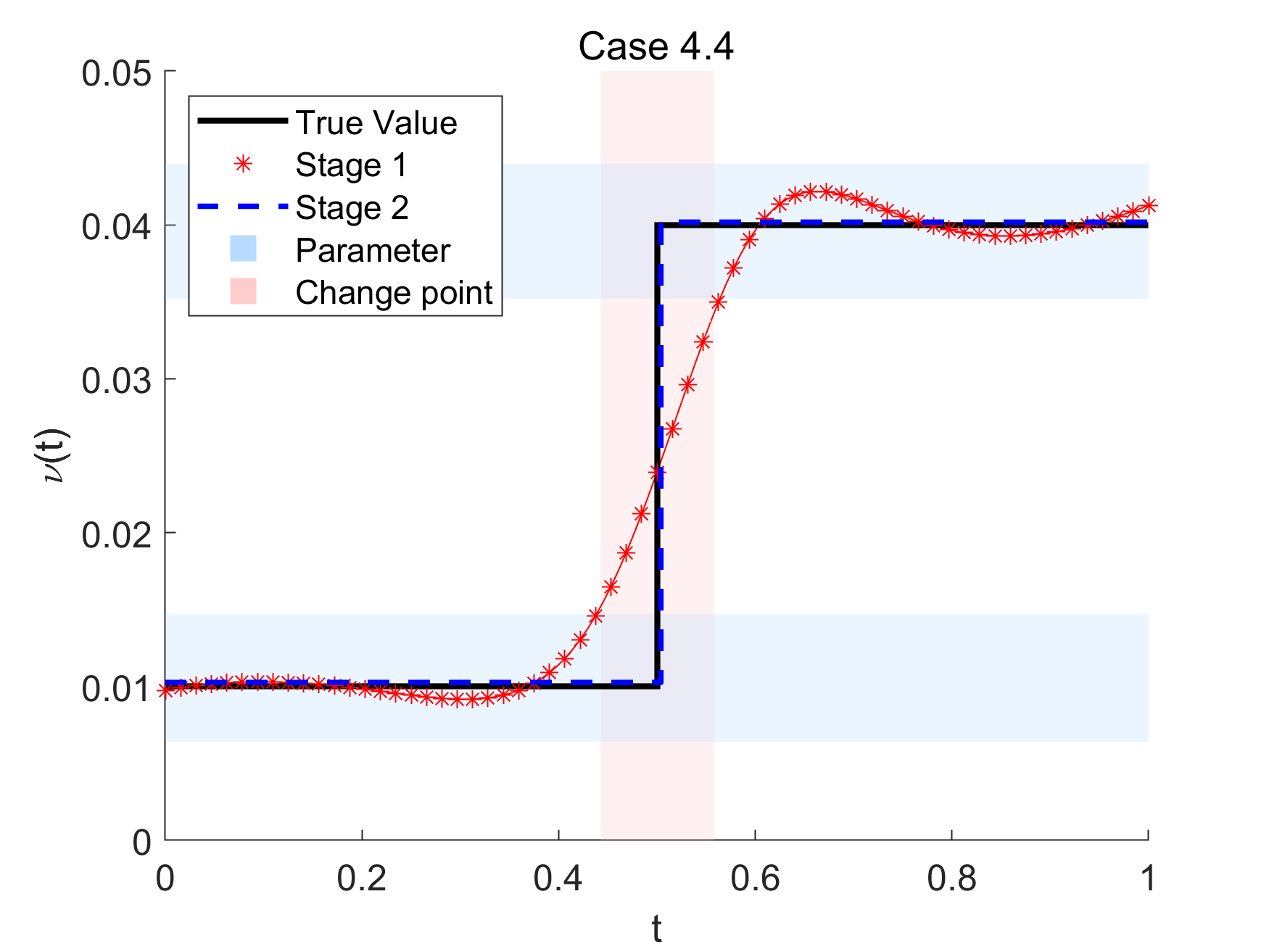}
			\end{minipage}
		}\\
		\subfloat[Parameter inverse result]{
			\begin{minipage}[t]{0.375\linewidth}
				\includegraphics[width=1\linewidth]{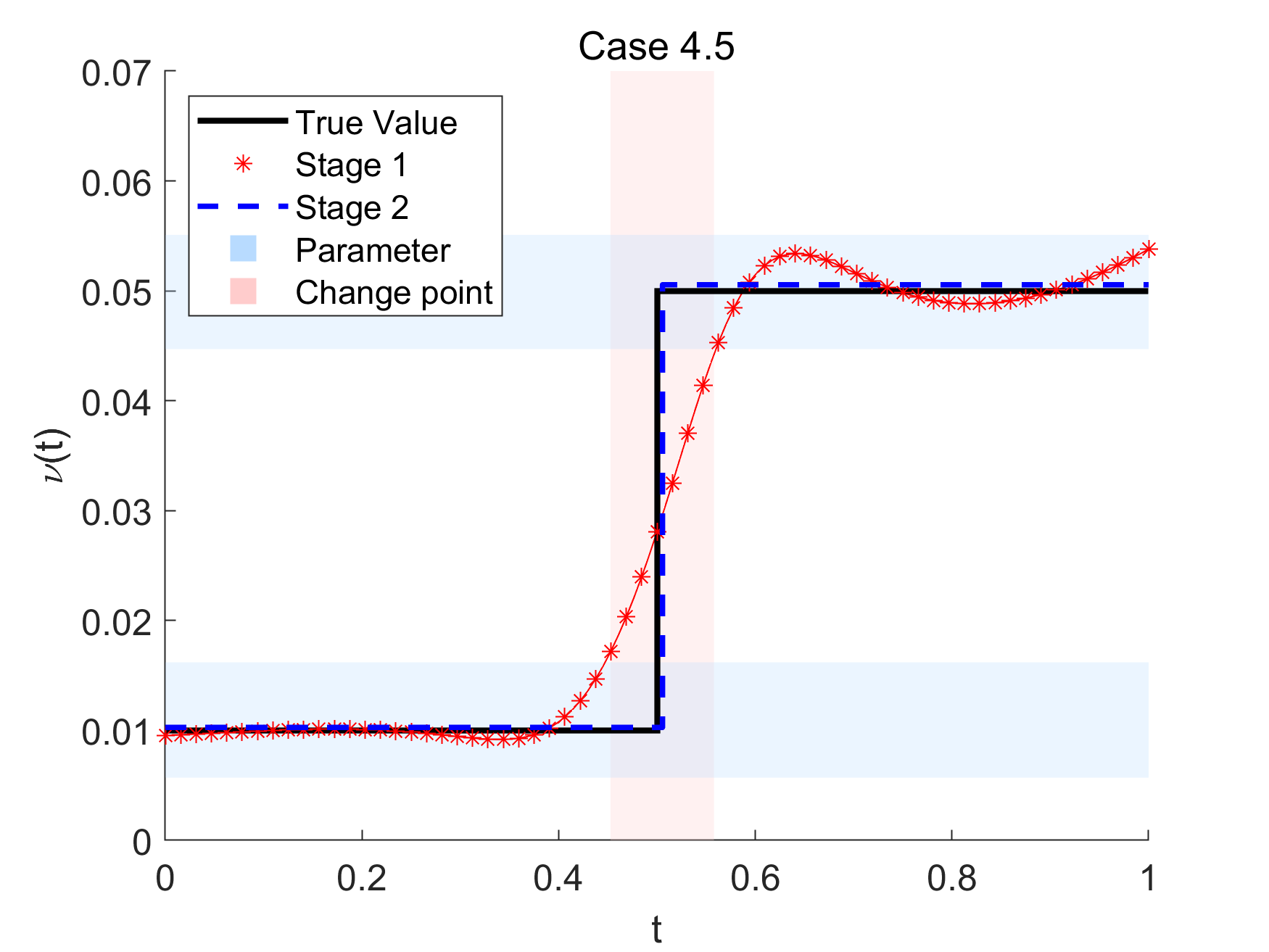}
			\end{minipage}
		}
		\subfloat[Parameter inverse result]{
			\begin{minipage}[t]{0.375\linewidth}
				\includegraphics[width=1\linewidth]{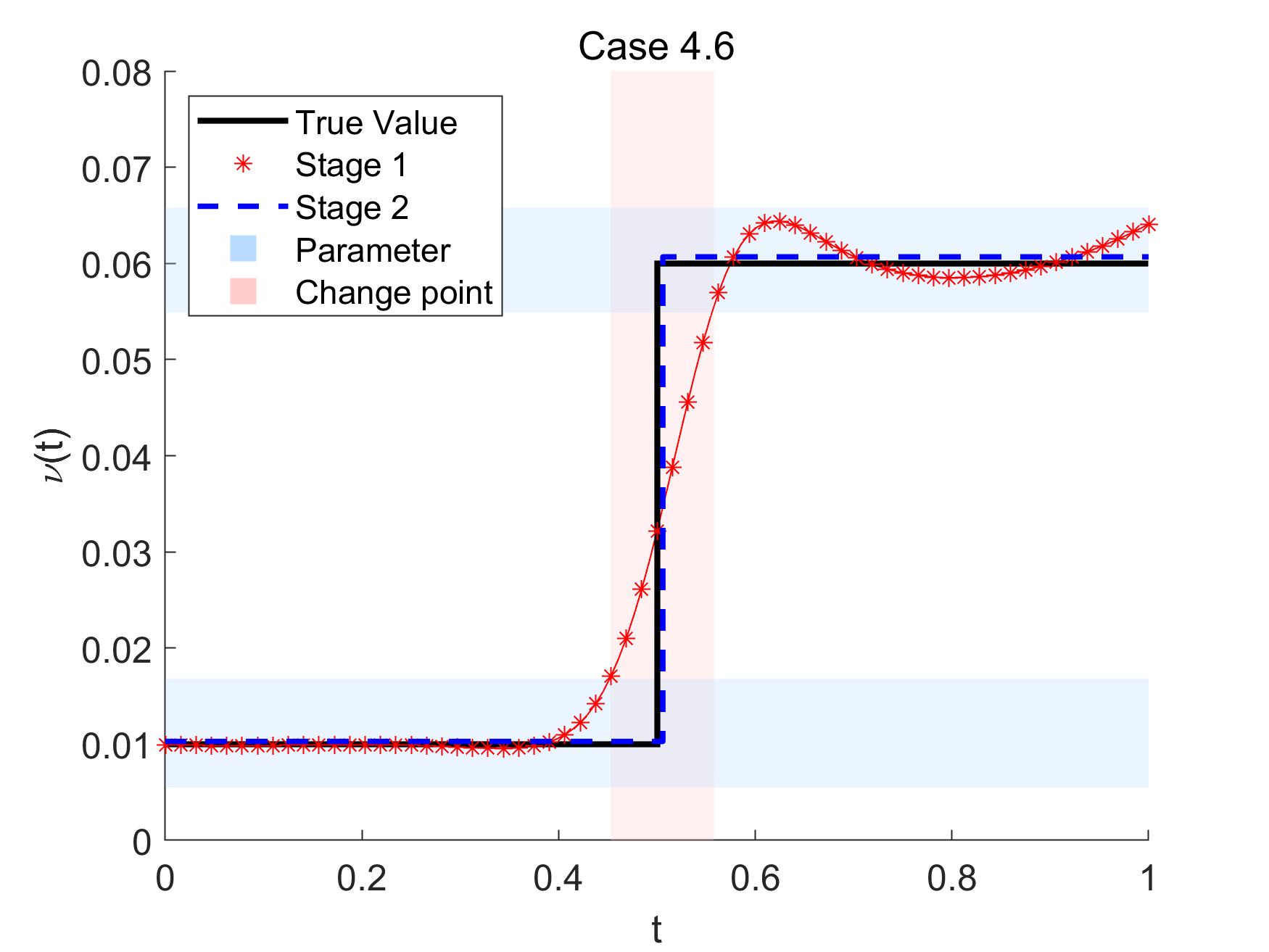}
			\end{minipage}
		}\\
		\subfloat[Parameter inverse result]{
			\begin{minipage}[t]{0.375\linewidth}
				\includegraphics[width=1\linewidth]{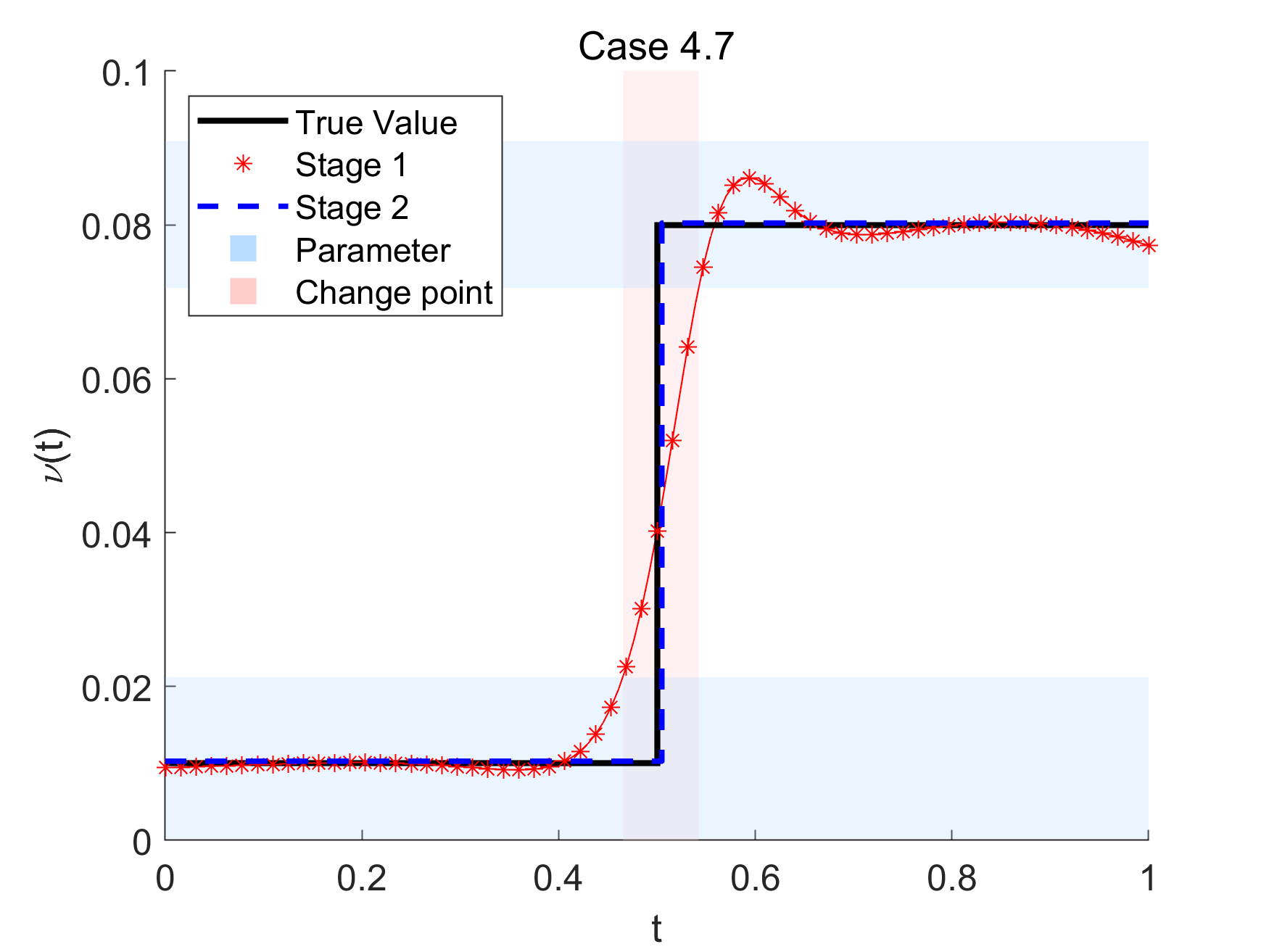}
			\end{minipage}
		}
		\subfloat[Parameter inverse result]{
			\begin{minipage}[t]{0.375\linewidth}
				\includegraphics[width=1\linewidth]{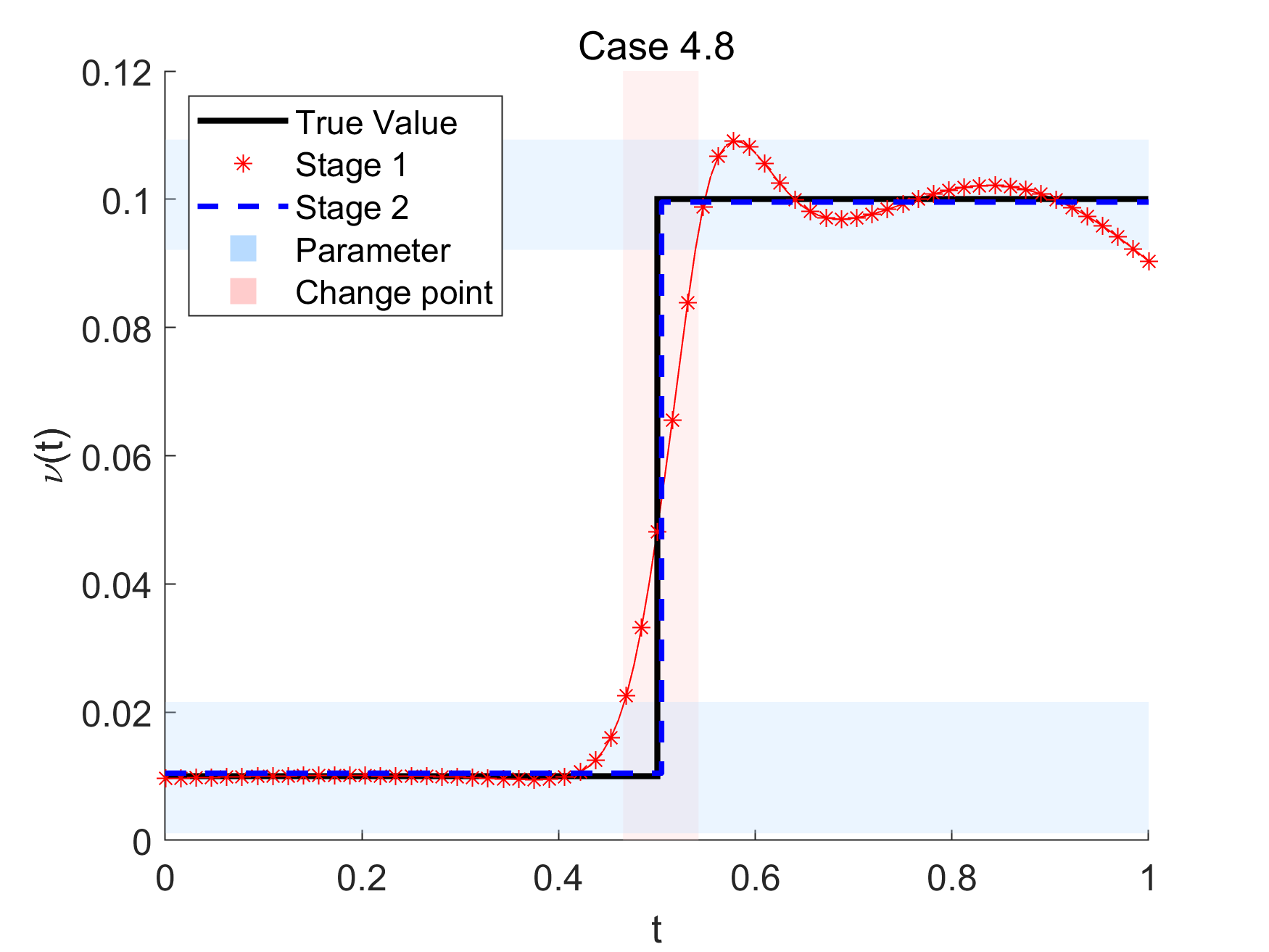}
			\end{minipage}
		}
		\caption{Numerical results for the Navier-Stokes equations with discontinuously time-varying coefficient $\nu(t)$. \label{fig7}}
	\end{figure}
	
	\begin{figure}[p]
		\centering
		\subfloat[Reference solution]{
			\begin{minipage}[t]{0.315\linewidth}
				\includegraphics[width=1\linewidth]{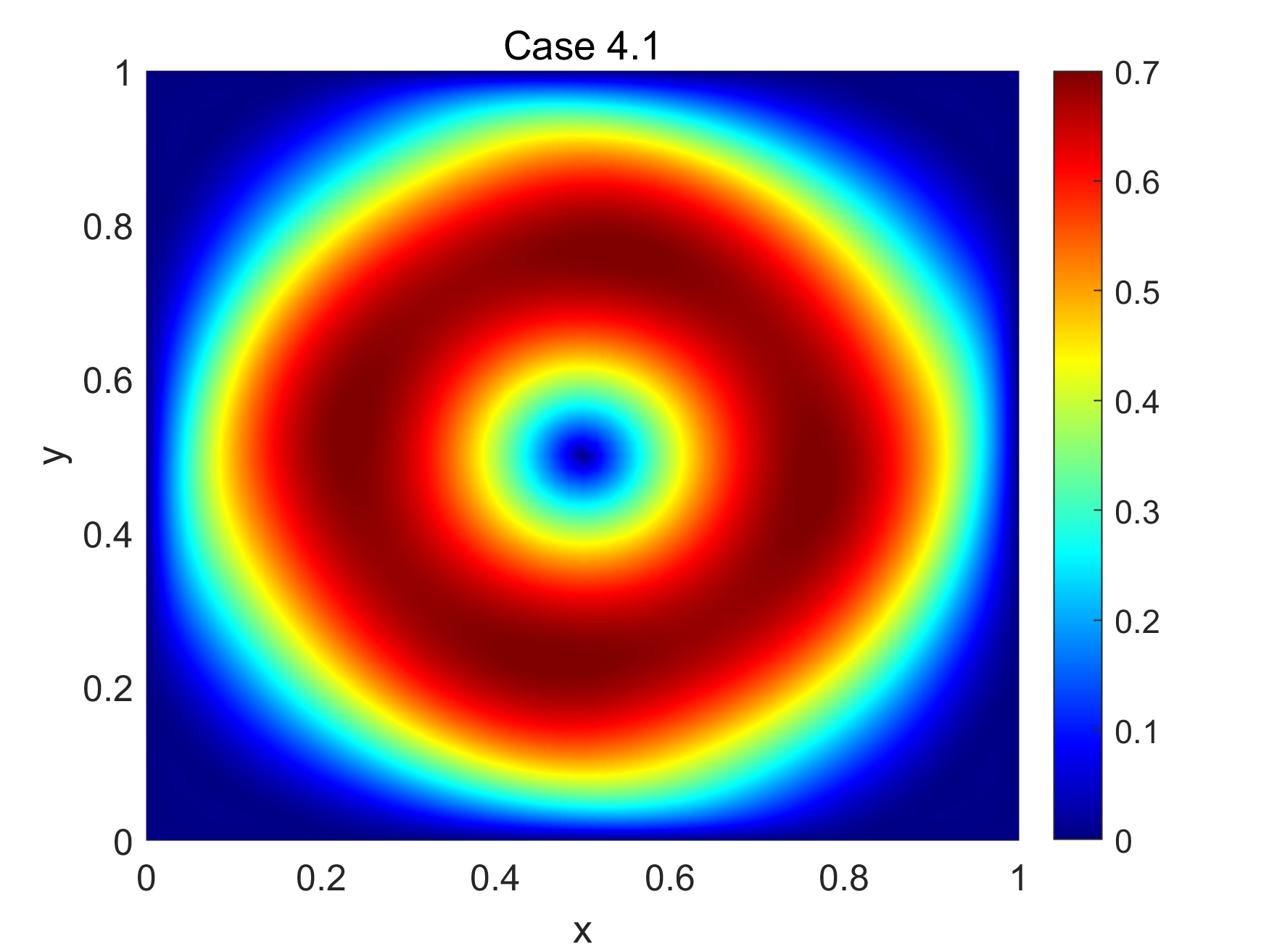}
			\end{minipage}
		}
		\subfloat[Predicted solution]{
			\begin{minipage}[t]{0.315\linewidth}
				\includegraphics[width=1\linewidth]{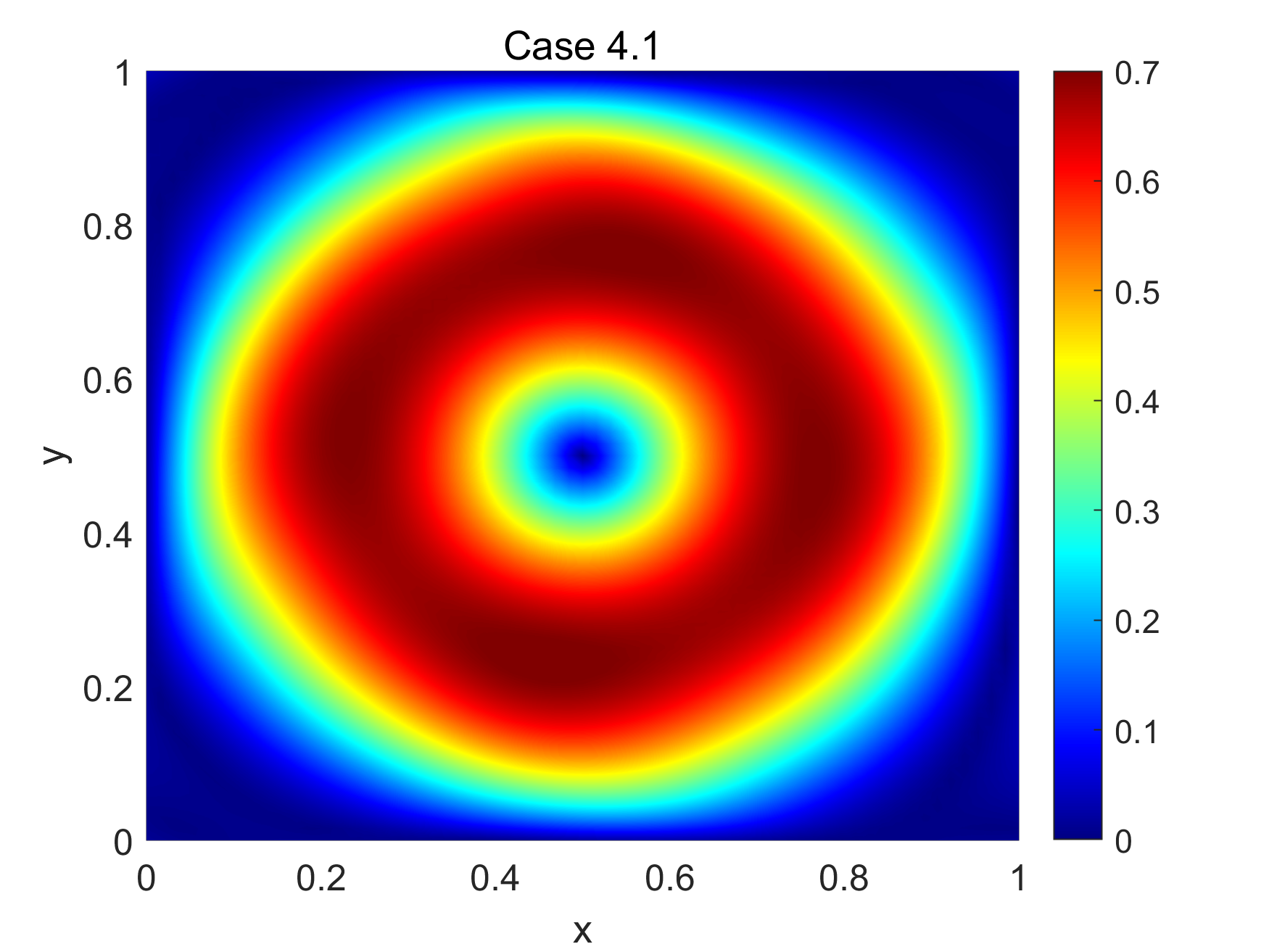}
			\end{minipage}
		}
		\subfloat[Absolute error]{
			\begin{minipage}[t]{0.315\linewidth}
				\includegraphics[width=1\linewidth]{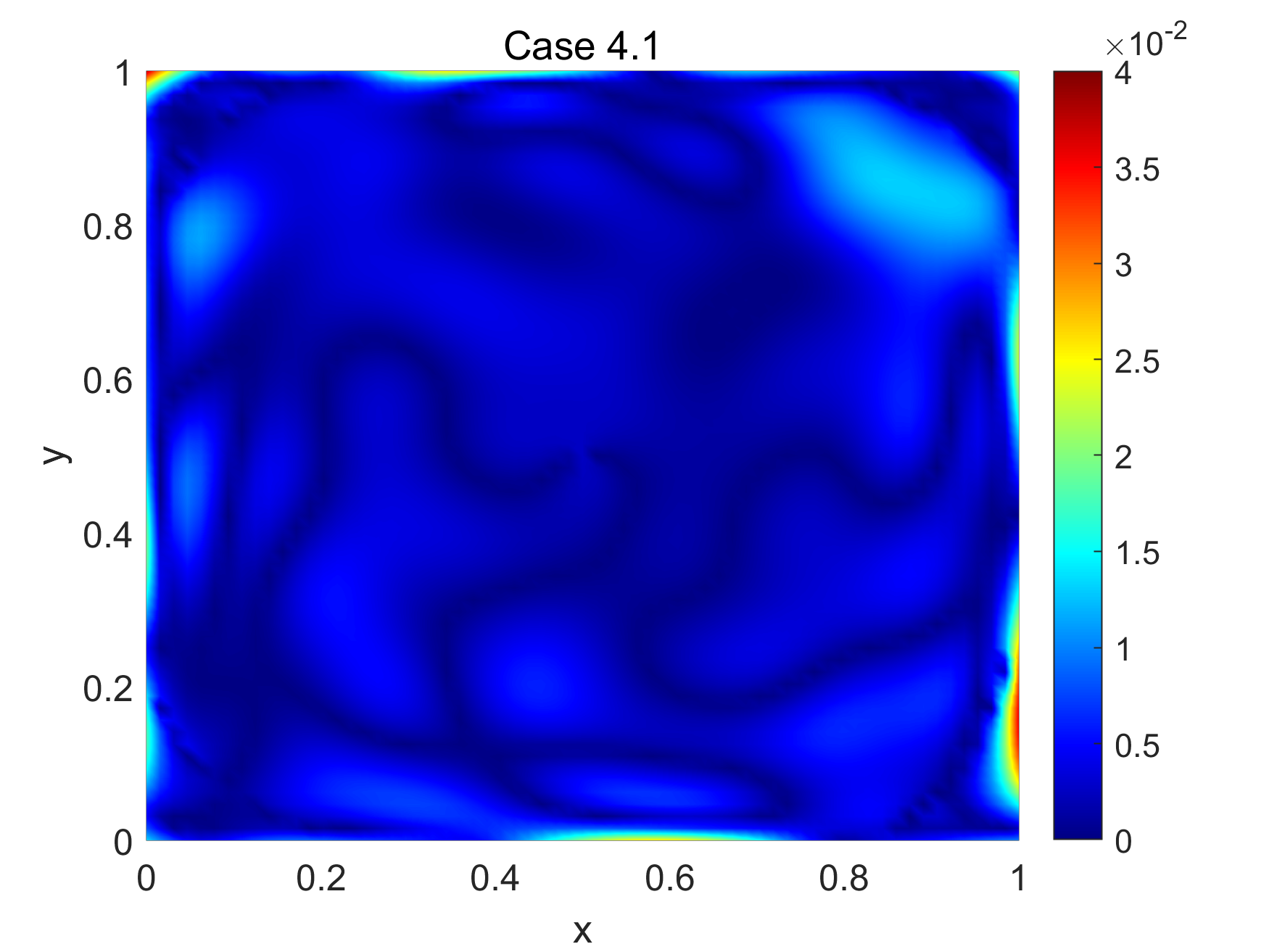}
			\end{minipage}
		}\\
		\subfloat[Reference solution]{
			\begin{minipage}[t]{0.315\linewidth}
				\includegraphics[width=1\linewidth]{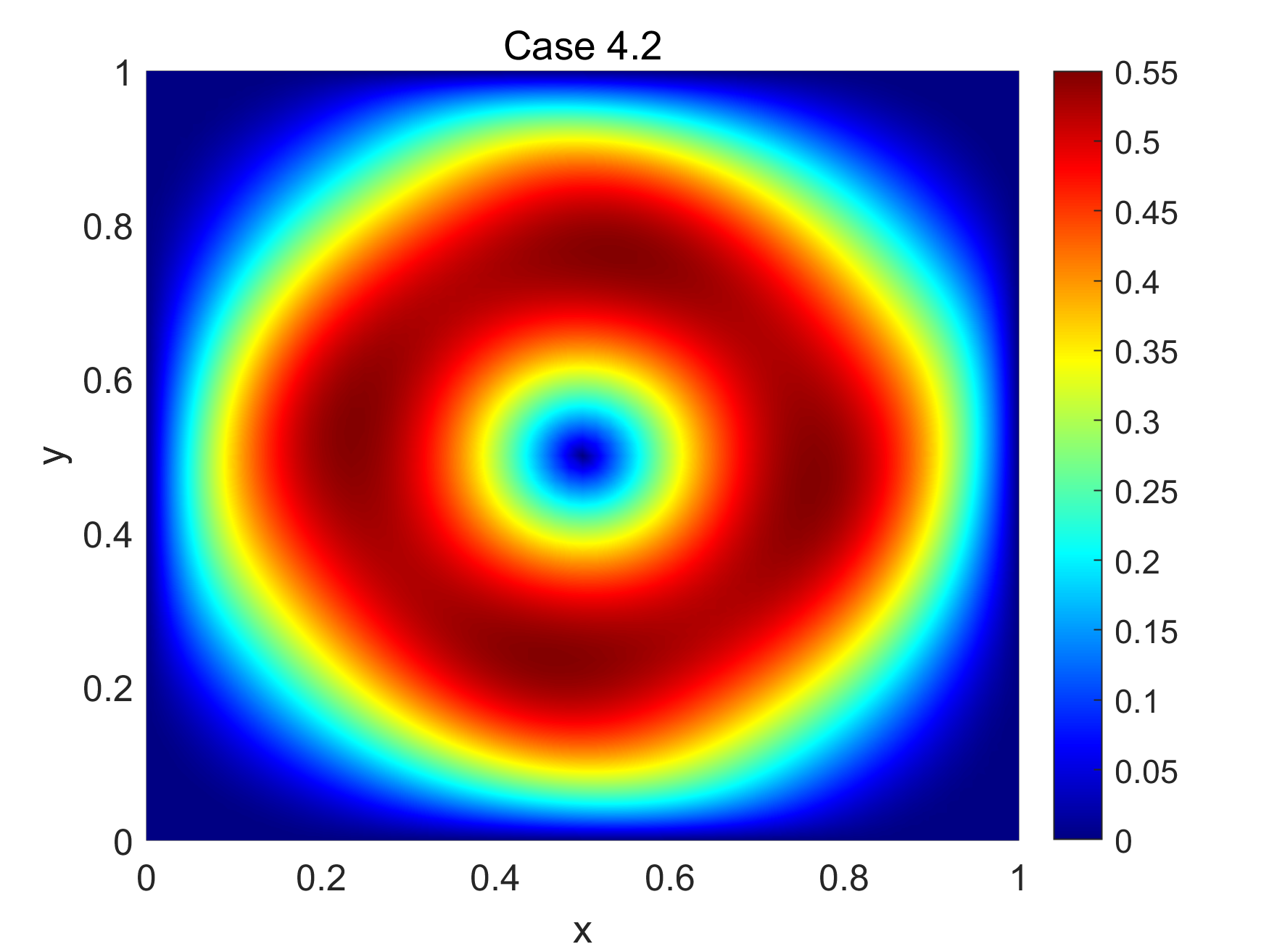}
			\end{minipage}
		}
		\subfloat[Predicted solution]{
			\begin{minipage}[t]{0.315\linewidth}
				\includegraphics[width=1\linewidth]{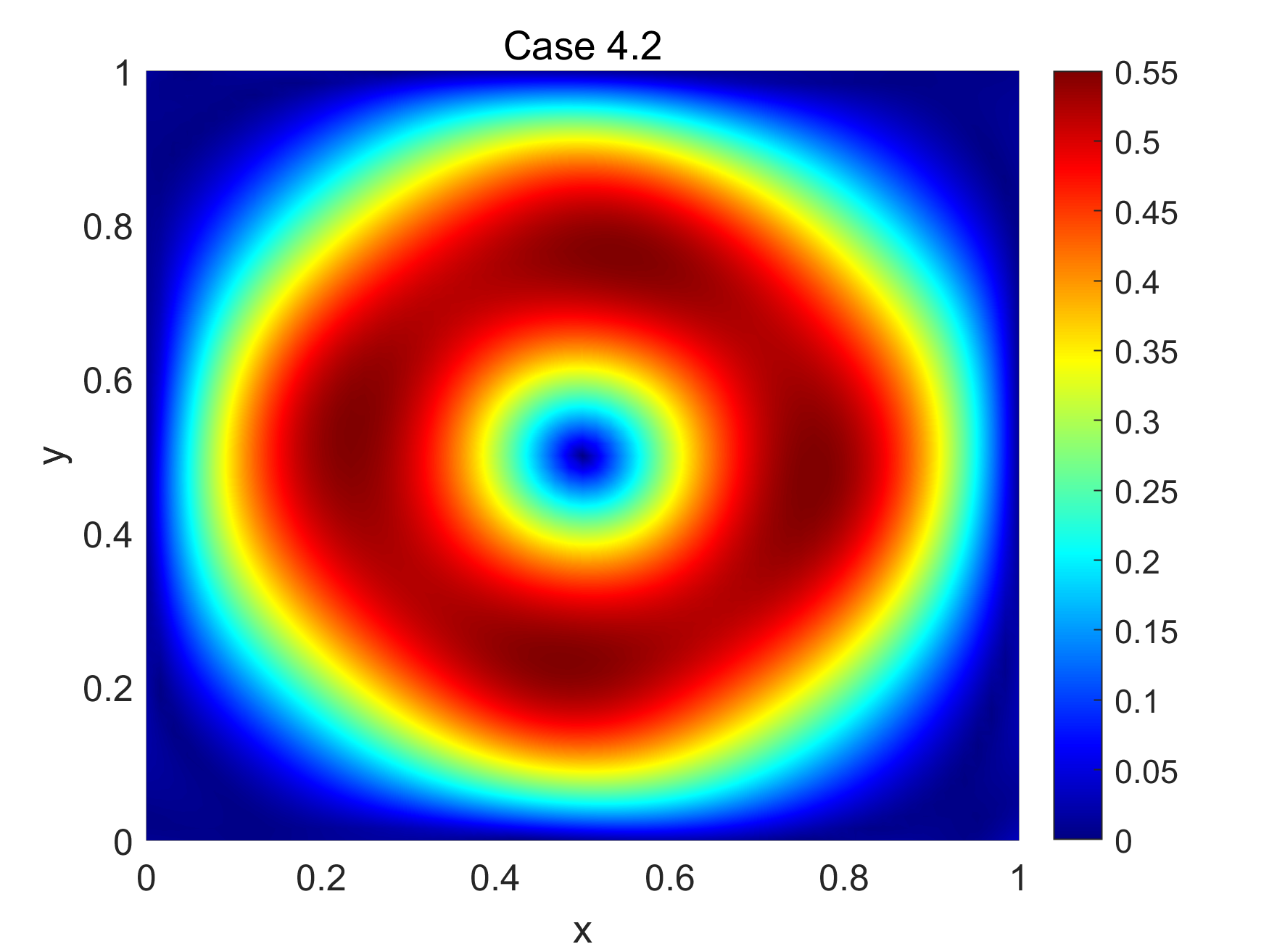}
			\end{minipage}
		}
		\subfloat[Absolute error]{
			\begin{minipage}[t]{0.315\linewidth}
				\includegraphics[width=1\linewidth]{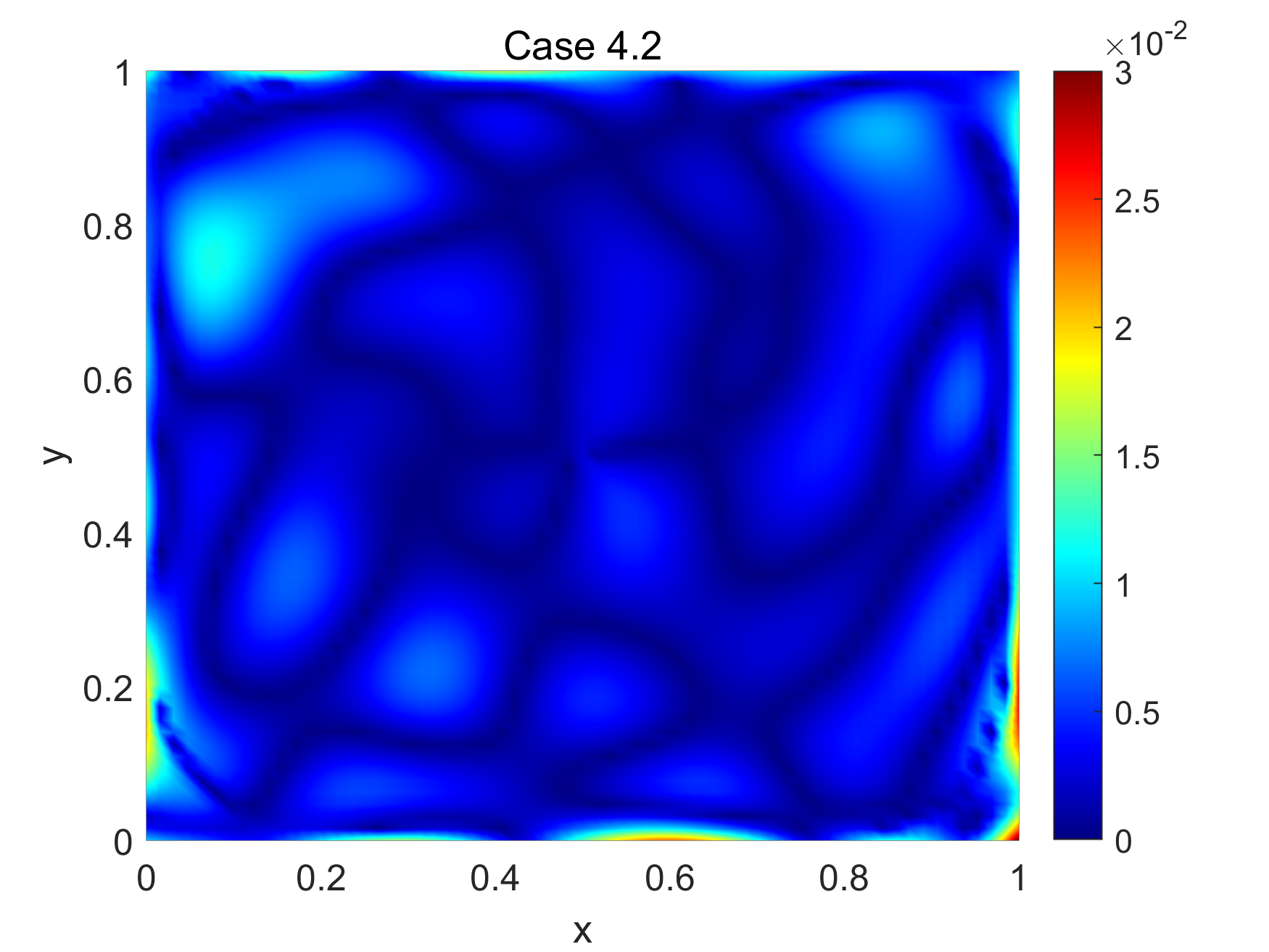}
			\end{minipage}
		}\\
		\subfloat[Reference solution]{
			\begin{minipage}[t]{0.315\linewidth}
				\includegraphics[width=1\linewidth]{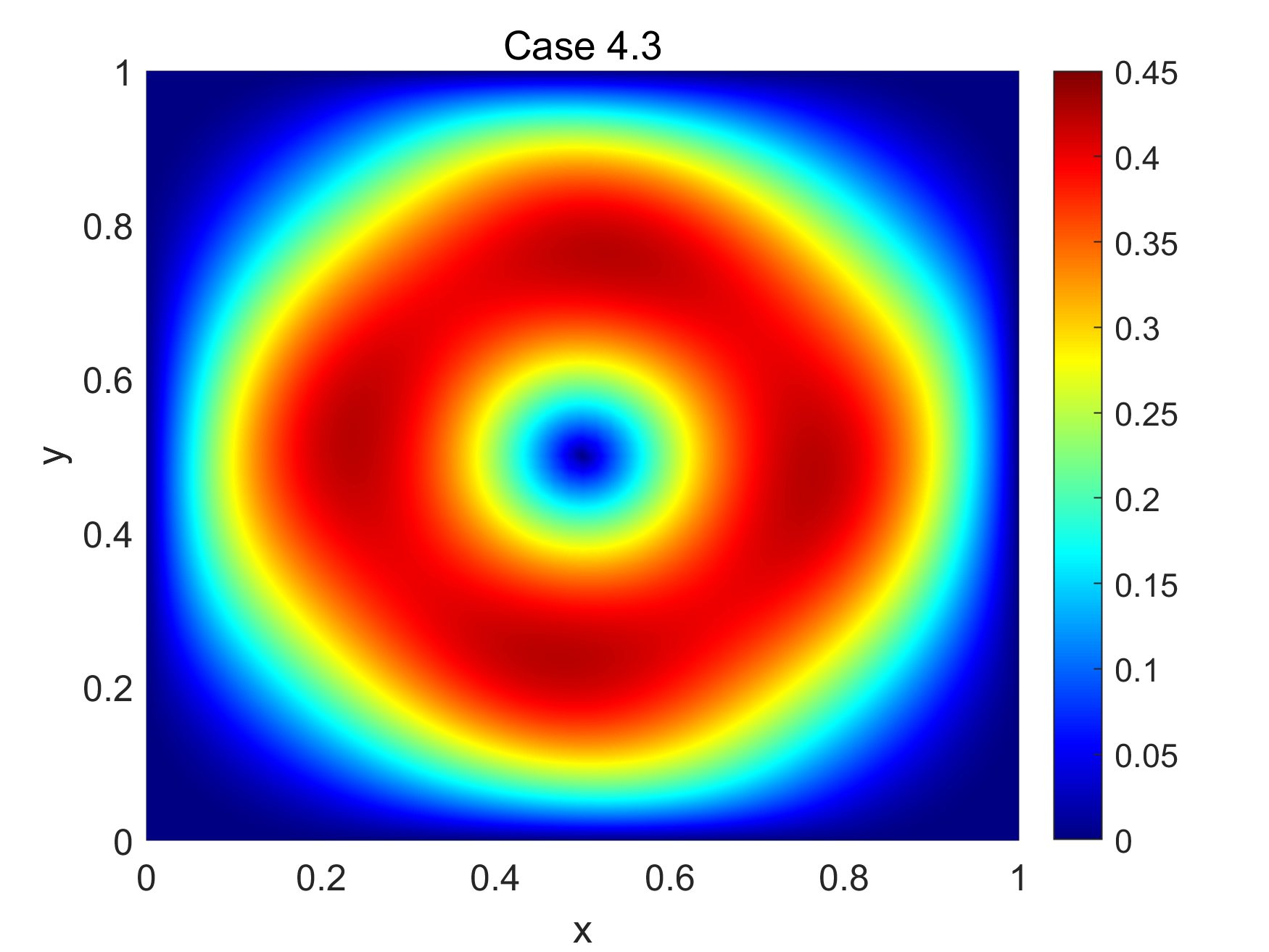}
			\end{minipage}
		}
		\subfloat[Predicted solution]{
			\begin{minipage}[t]{0.315\linewidth}
				\includegraphics[width=1\linewidth]{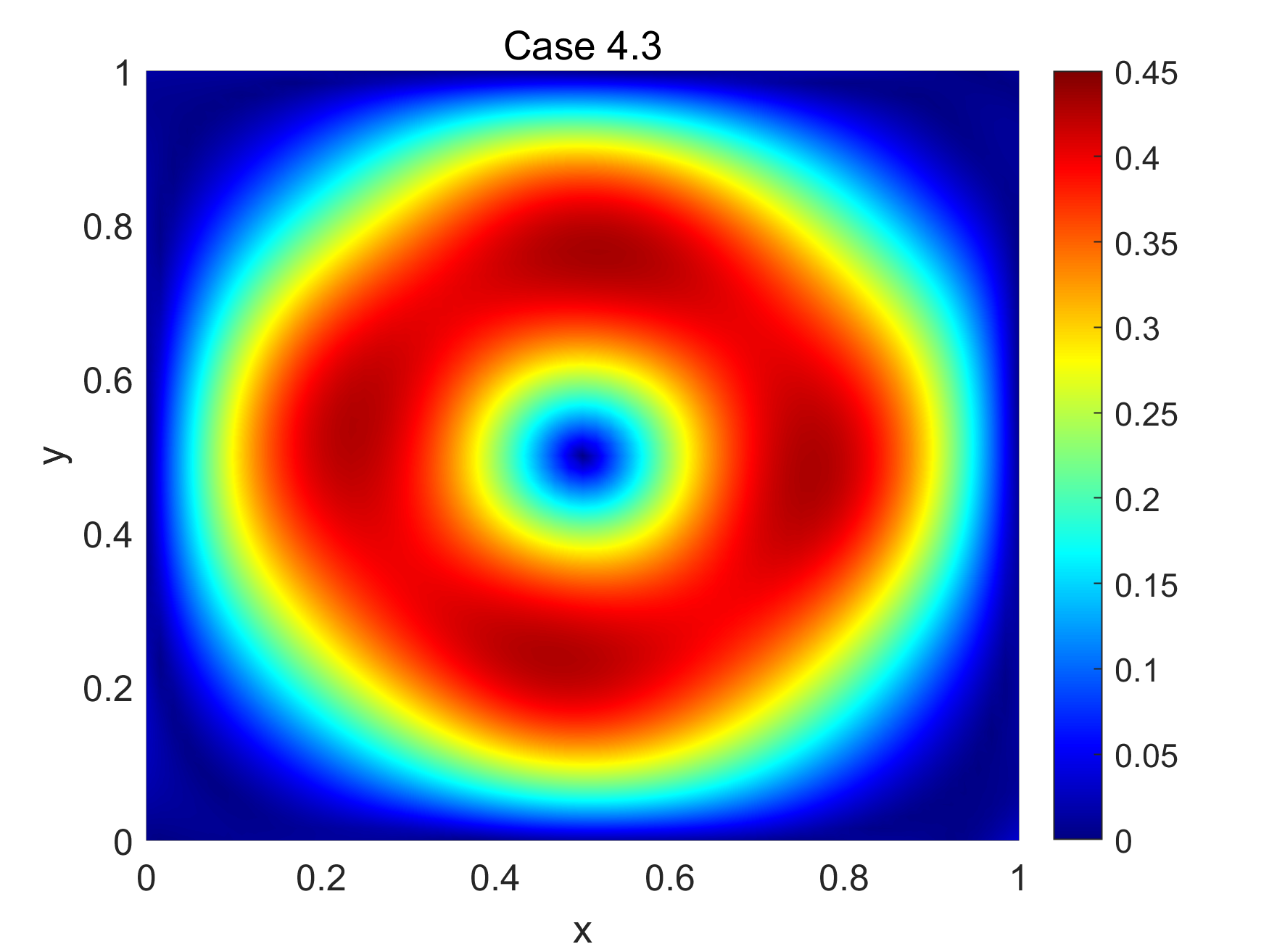}
			\end{minipage}
		}
		\subfloat[Absolute error]{
			\begin{minipage}[t]{0.315\linewidth}
				\includegraphics[width=1\linewidth]{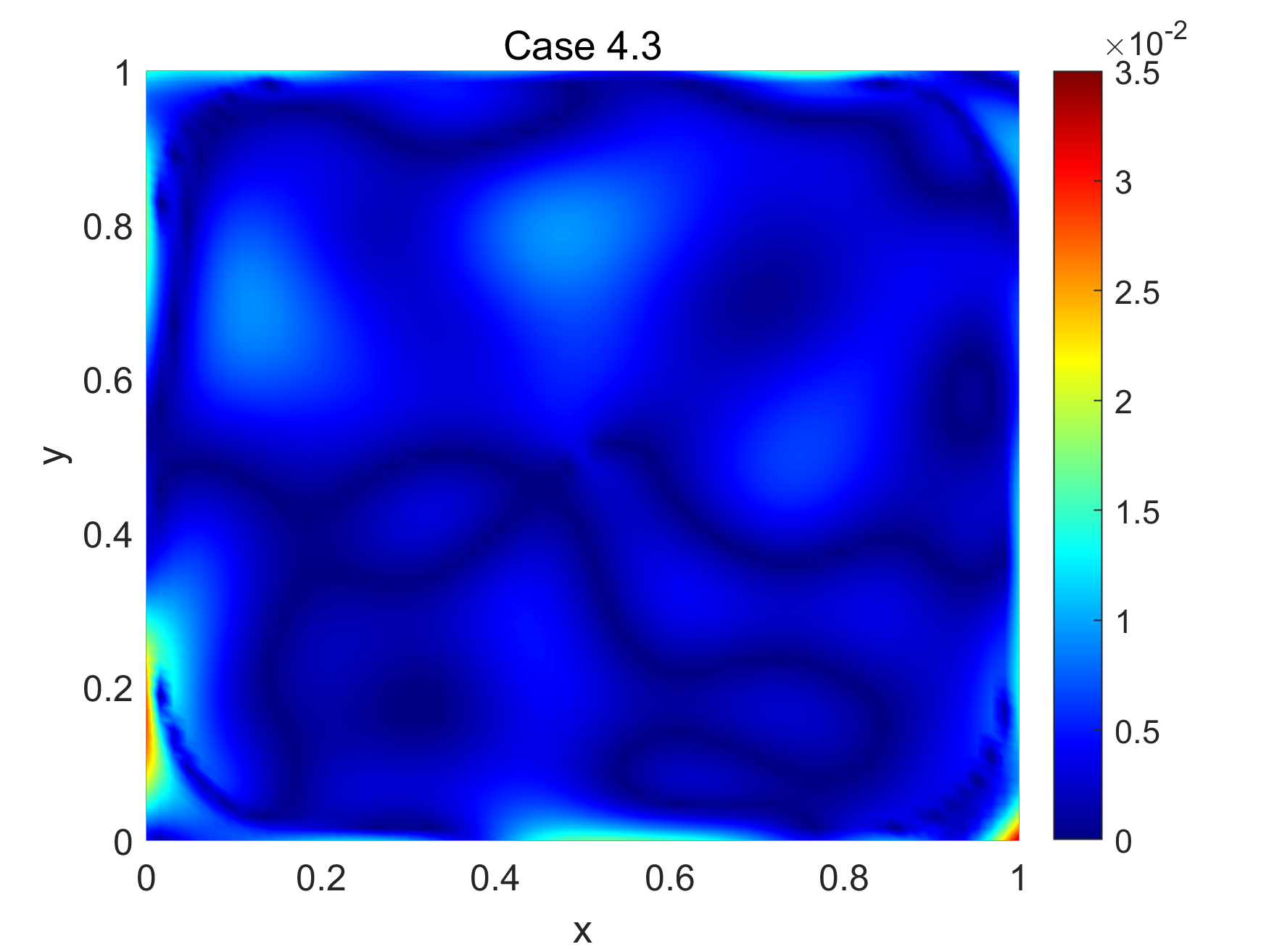}
			\end{minipage}
		}\\
		\subfloat[Reference solution]{
			\begin{minipage}[t]{0.315\linewidth}
				\includegraphics[width=1\linewidth]{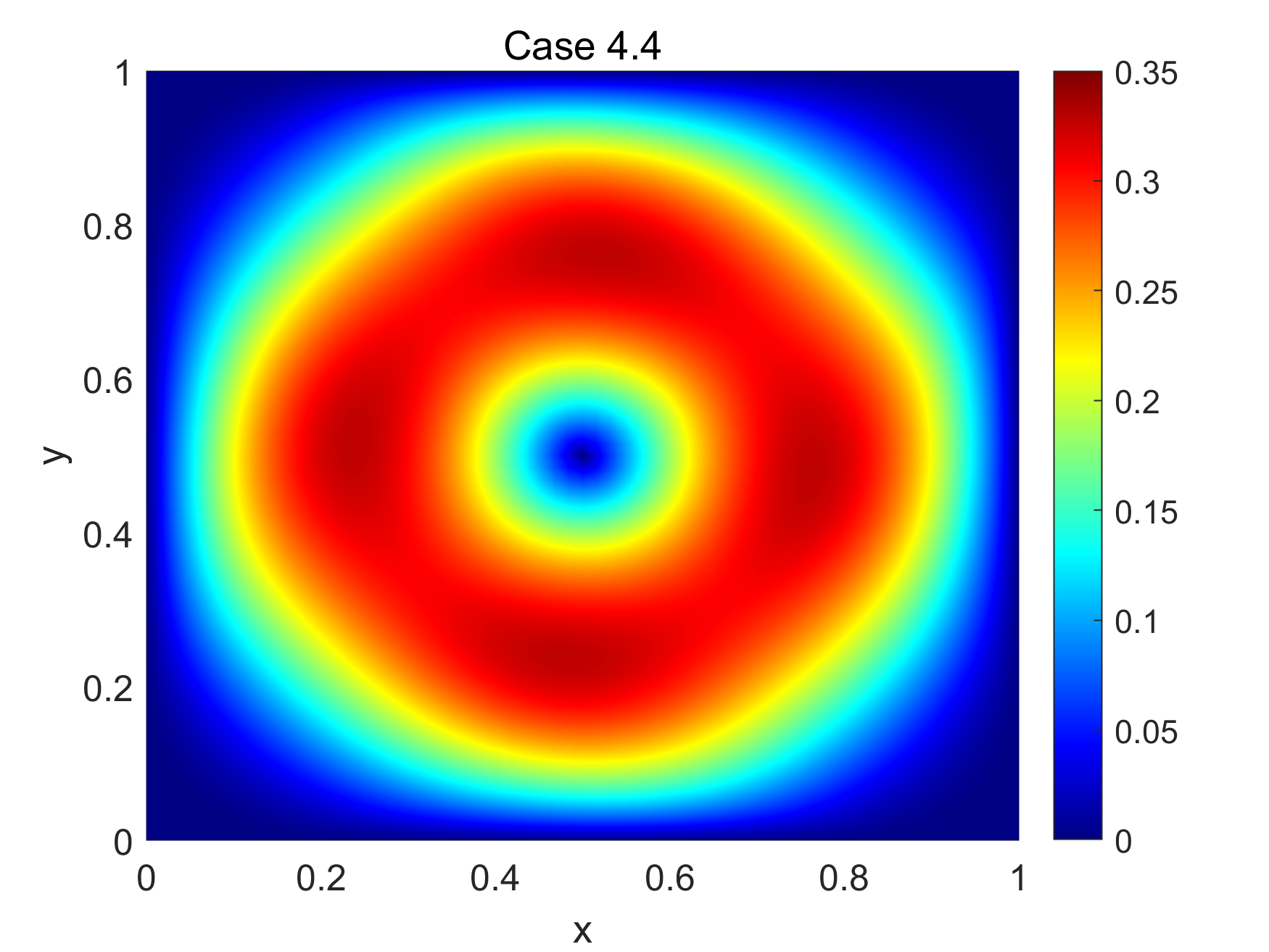}
			\end{minipage}
		}
		\subfloat[Predicted solution]{
			\begin{minipage}[t]{0.315\linewidth}
				\includegraphics[width=1\linewidth]{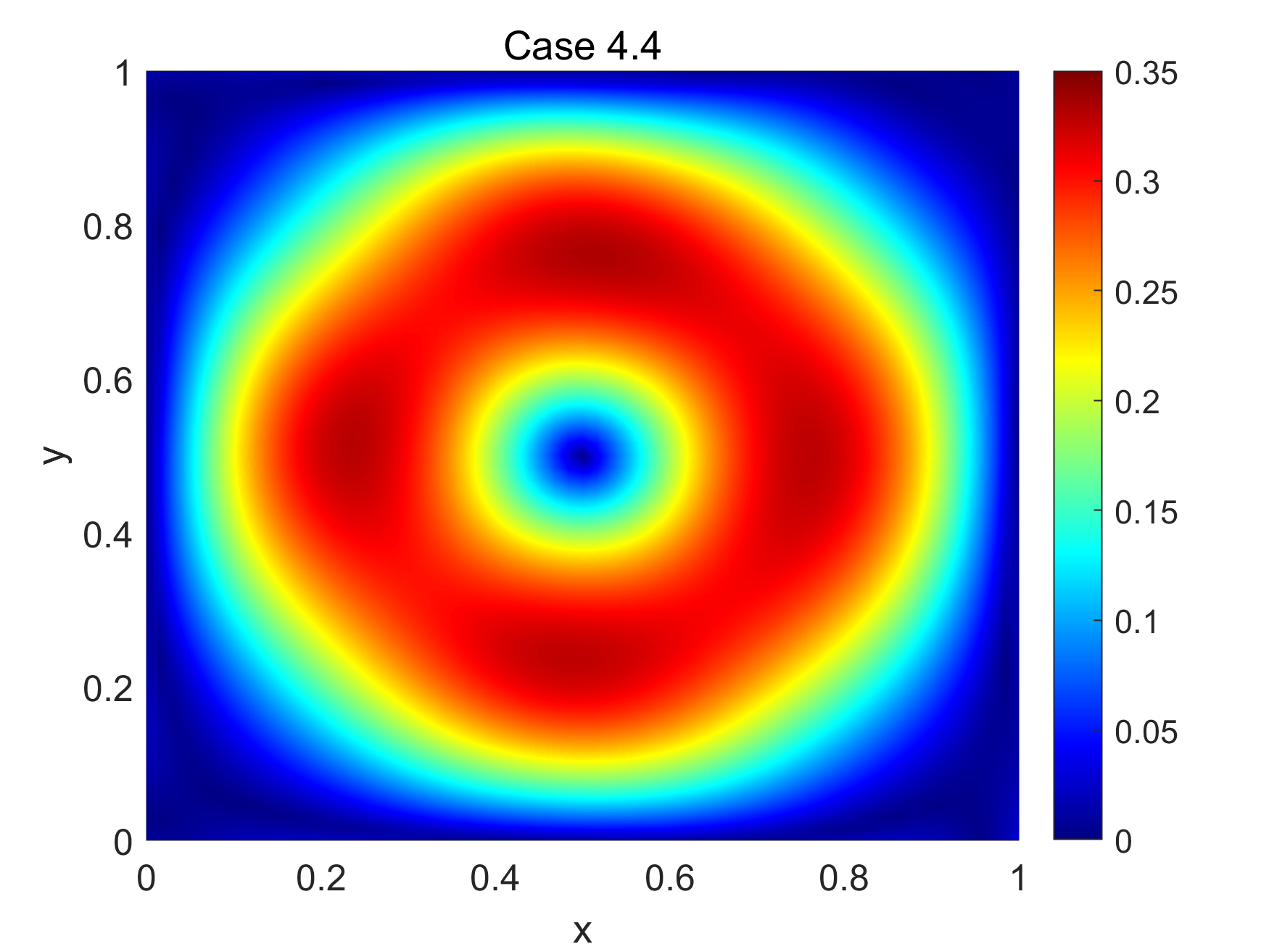}
			\end{minipage}
		}
		\subfloat[Absolute error]{
			\begin{minipage}[t]{0.315\linewidth}
				\includegraphics[width=1\linewidth]{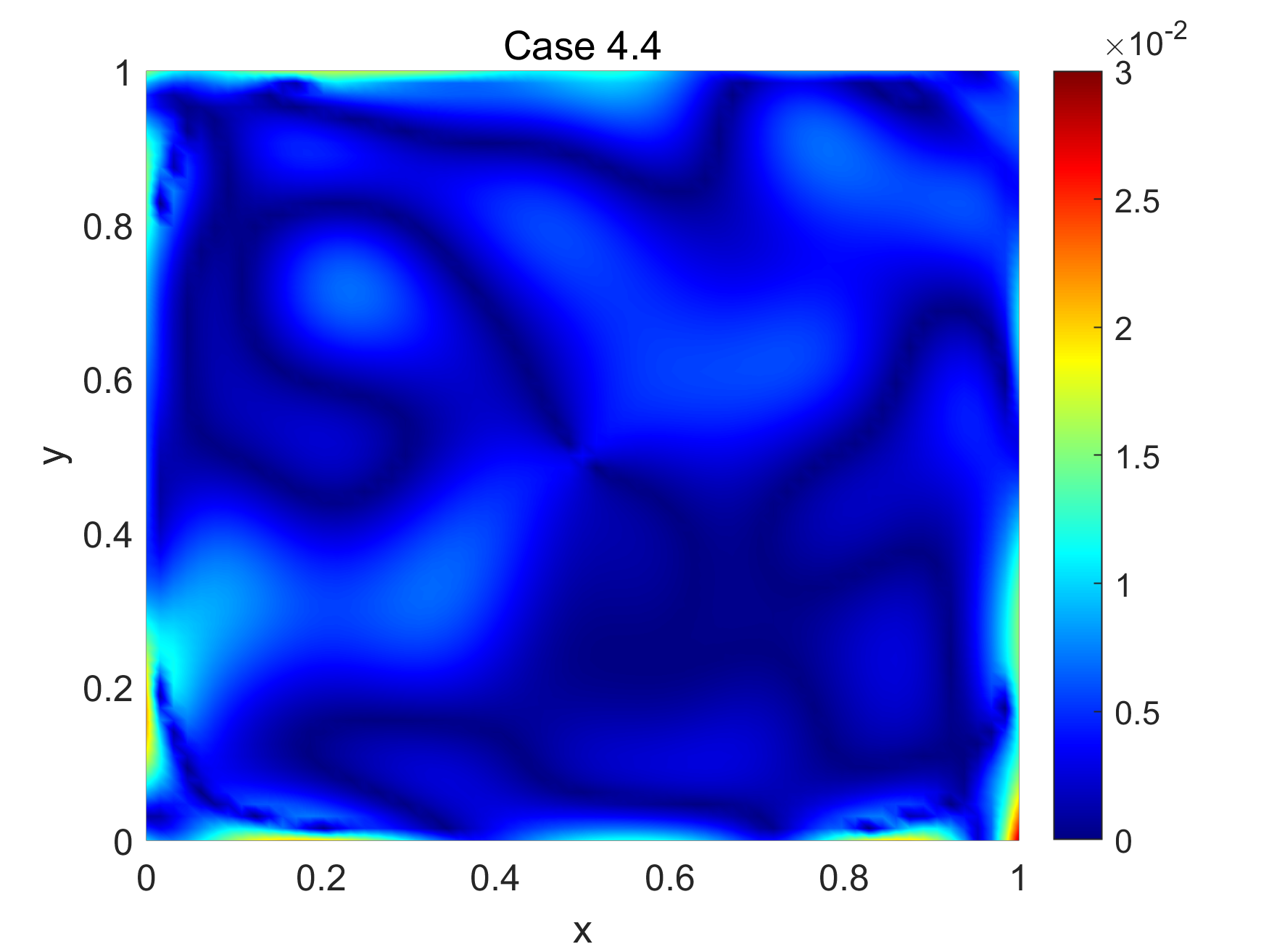}
			\end{minipage}
		}
		\caption{Numerical results for Navier-Stokes equations with discontinuously time-varying coefficients $\nu(t)$, from Case~4.1 to Case~4.4. \label{fig8}}
	\end{figure}
	
	\begin{figure}[p]
		\centering
		\subfloat[Reference solution]{
			\begin{minipage}[t]{0.315\linewidth}
				\includegraphics[width=1\linewidth]{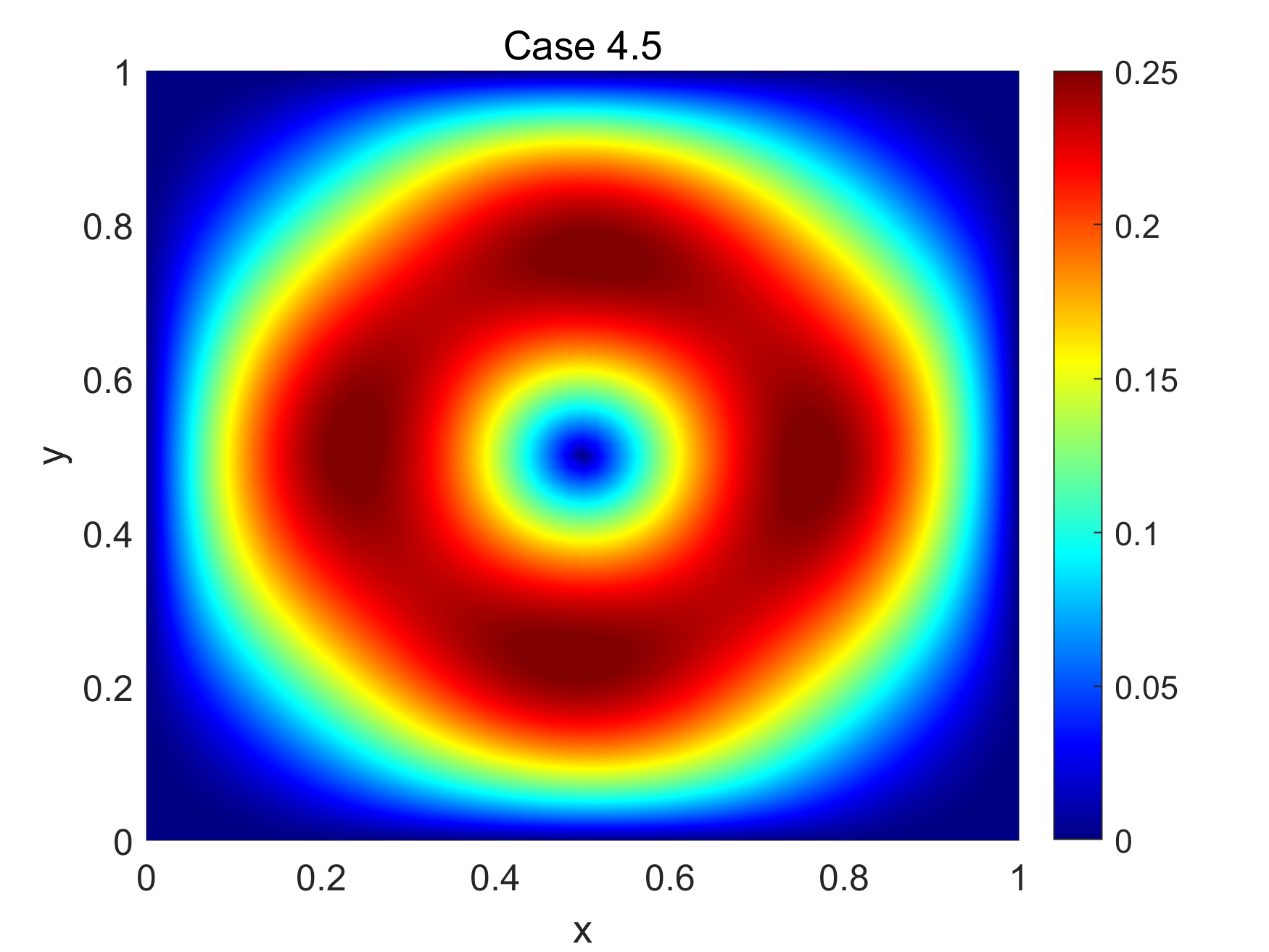}
			\end{minipage}
		}
		\subfloat[Predicted solution]{
			\begin{minipage}[t]{0.315\linewidth}
				\includegraphics[width=1\linewidth]{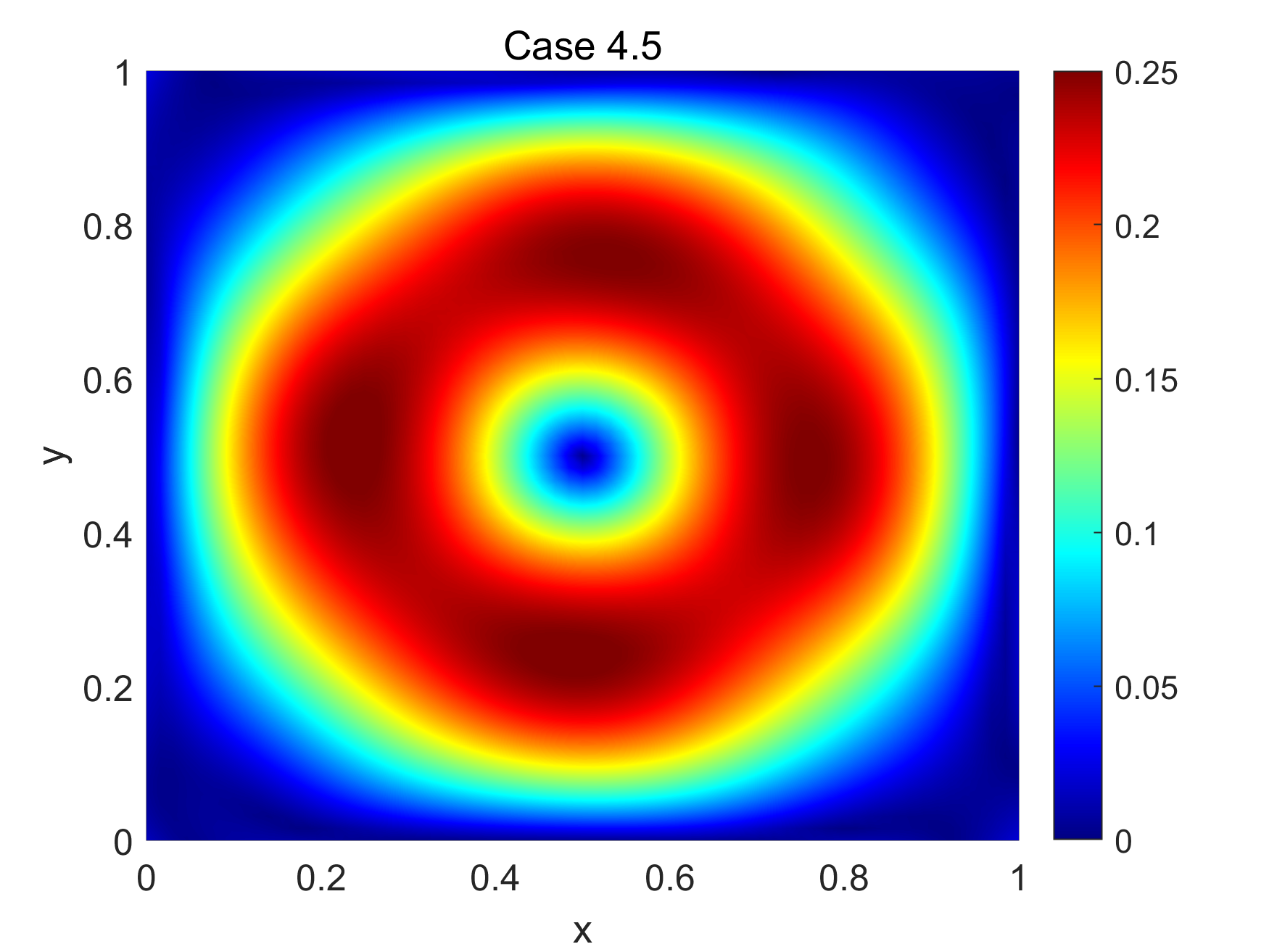}
			\end{minipage}
		}
		\subfloat[Absolute error]{
			\begin{minipage}[t]{0.315\linewidth}
				\includegraphics[width=1\linewidth]{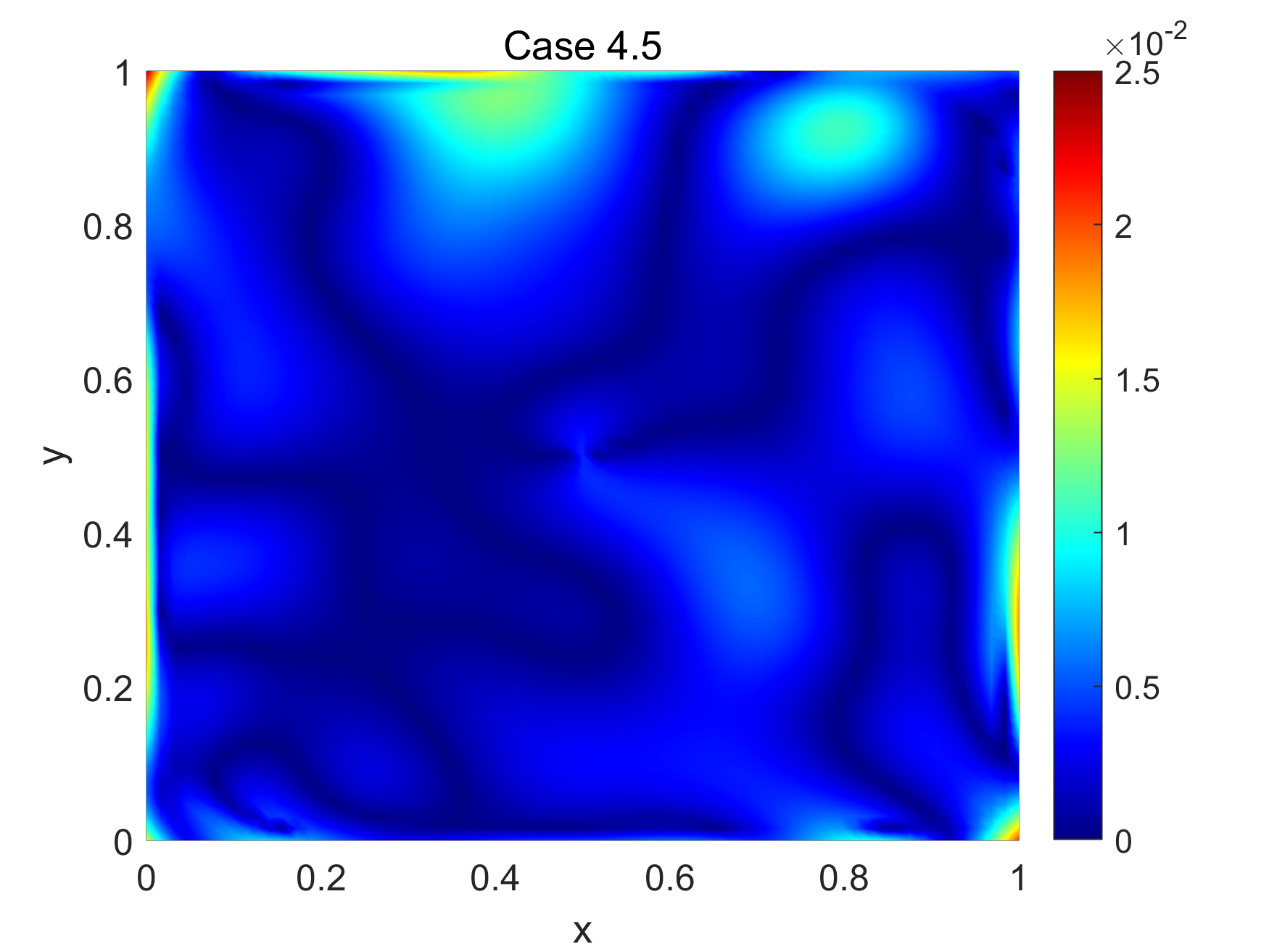}
			\end{minipage}
		}\\
		\subfloat[Reference solution]{
			\begin{minipage}[t]{0.315\linewidth}
				\includegraphics[width=1\linewidth]{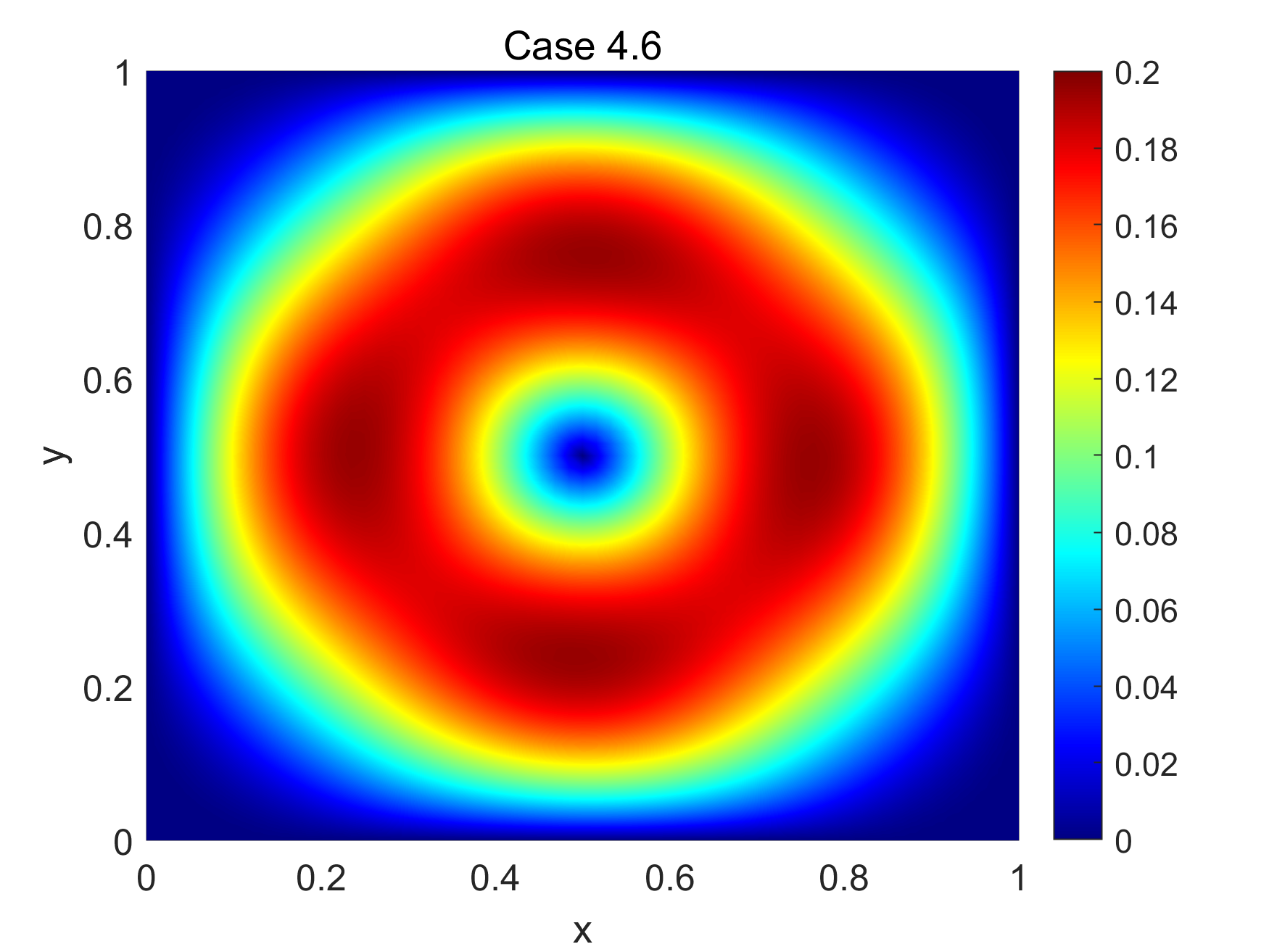}
			\end{minipage}
		}
		\subfloat[Predicted solution]{
			\begin{minipage}[t]{0.315\linewidth}
				\includegraphics[width=1\linewidth]{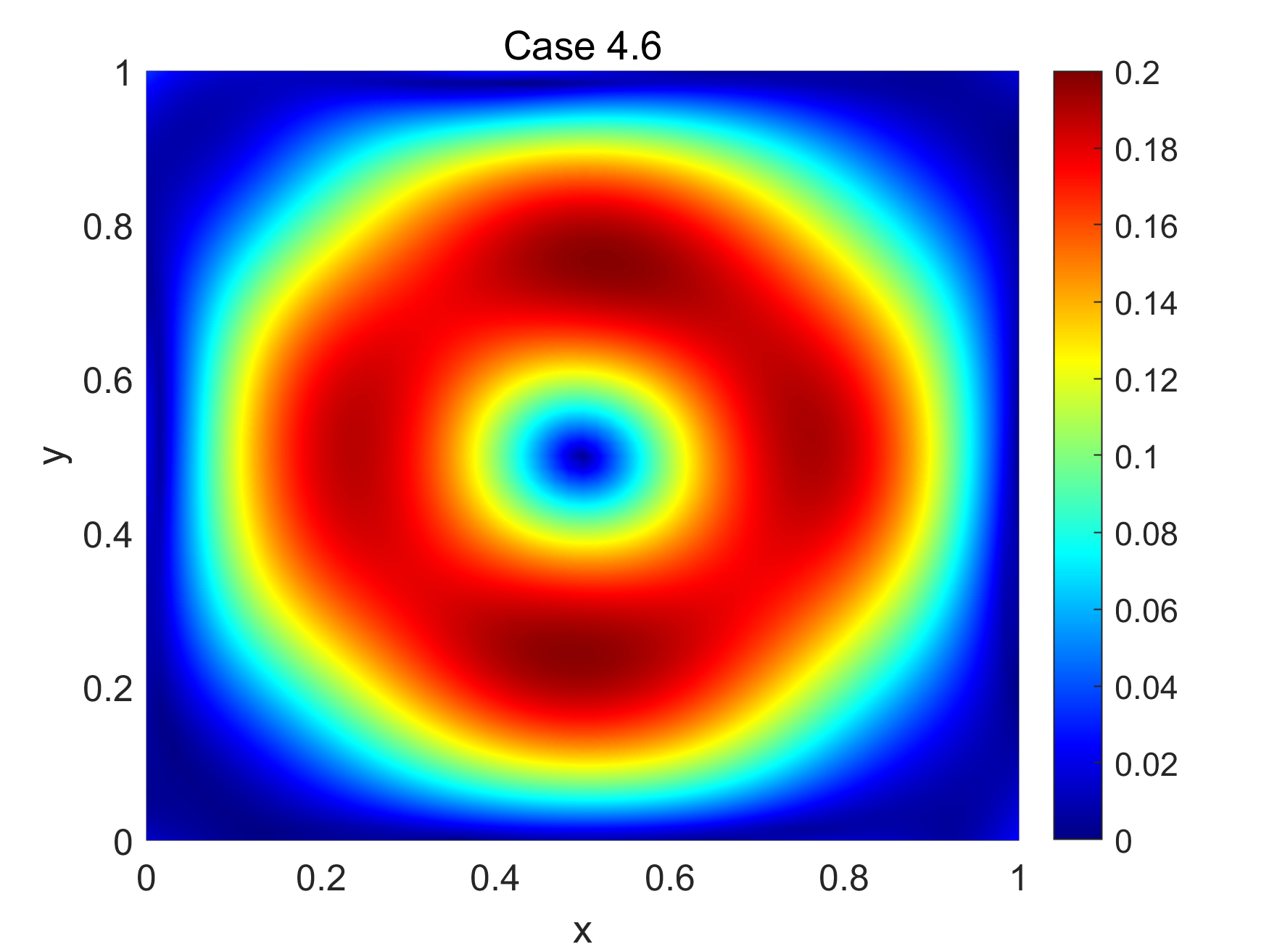}
			\end{minipage}
		}
		\subfloat[Absolute error]{
			\begin{minipage}[t]{0.315\linewidth}
				\includegraphics[width=1\linewidth]{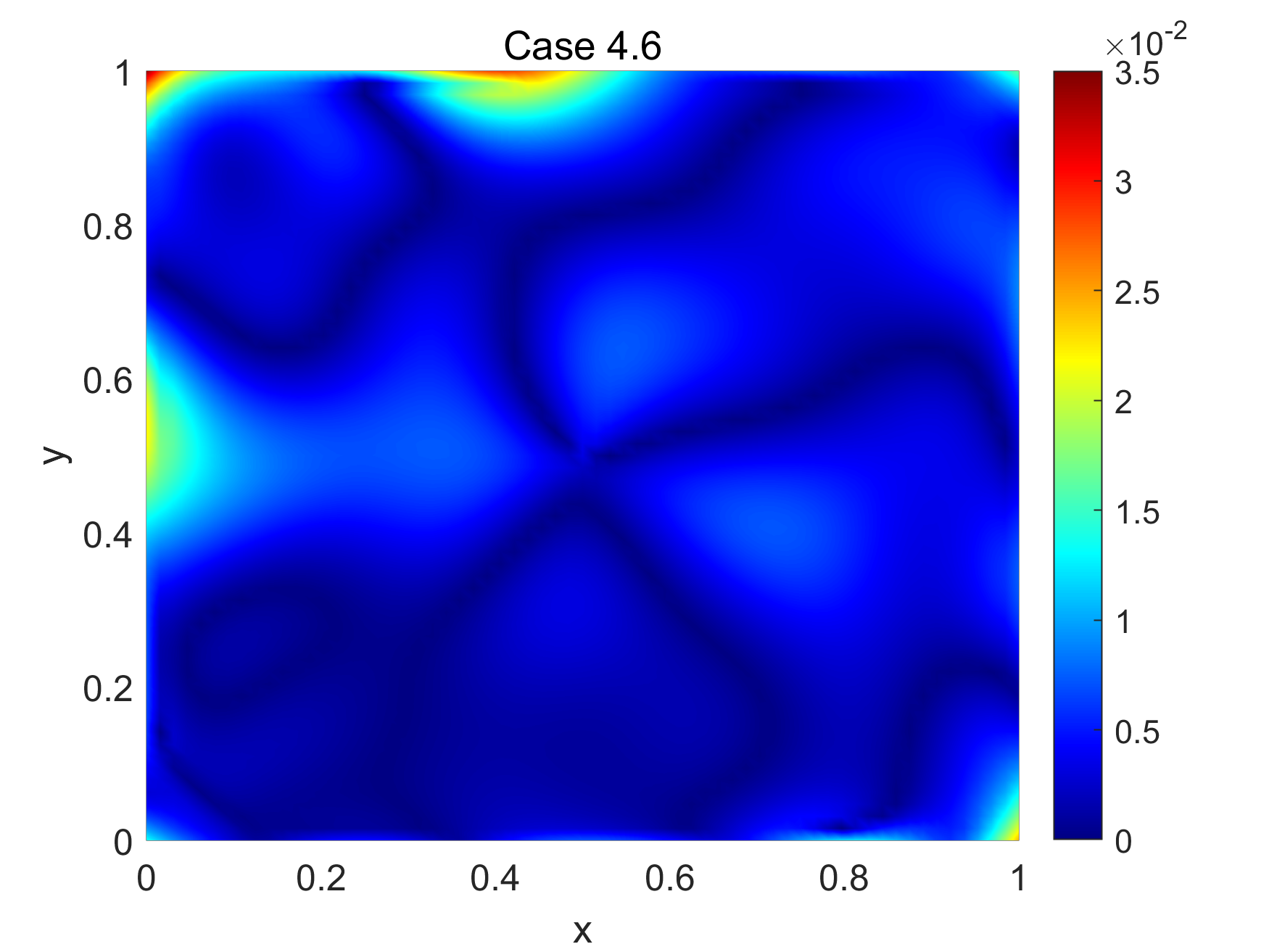}
			\end{minipage}
		}\\
		\subfloat[Reference solution]{
			\begin{minipage}[t]{0.315\linewidth}
				\includegraphics[width=1\linewidth]{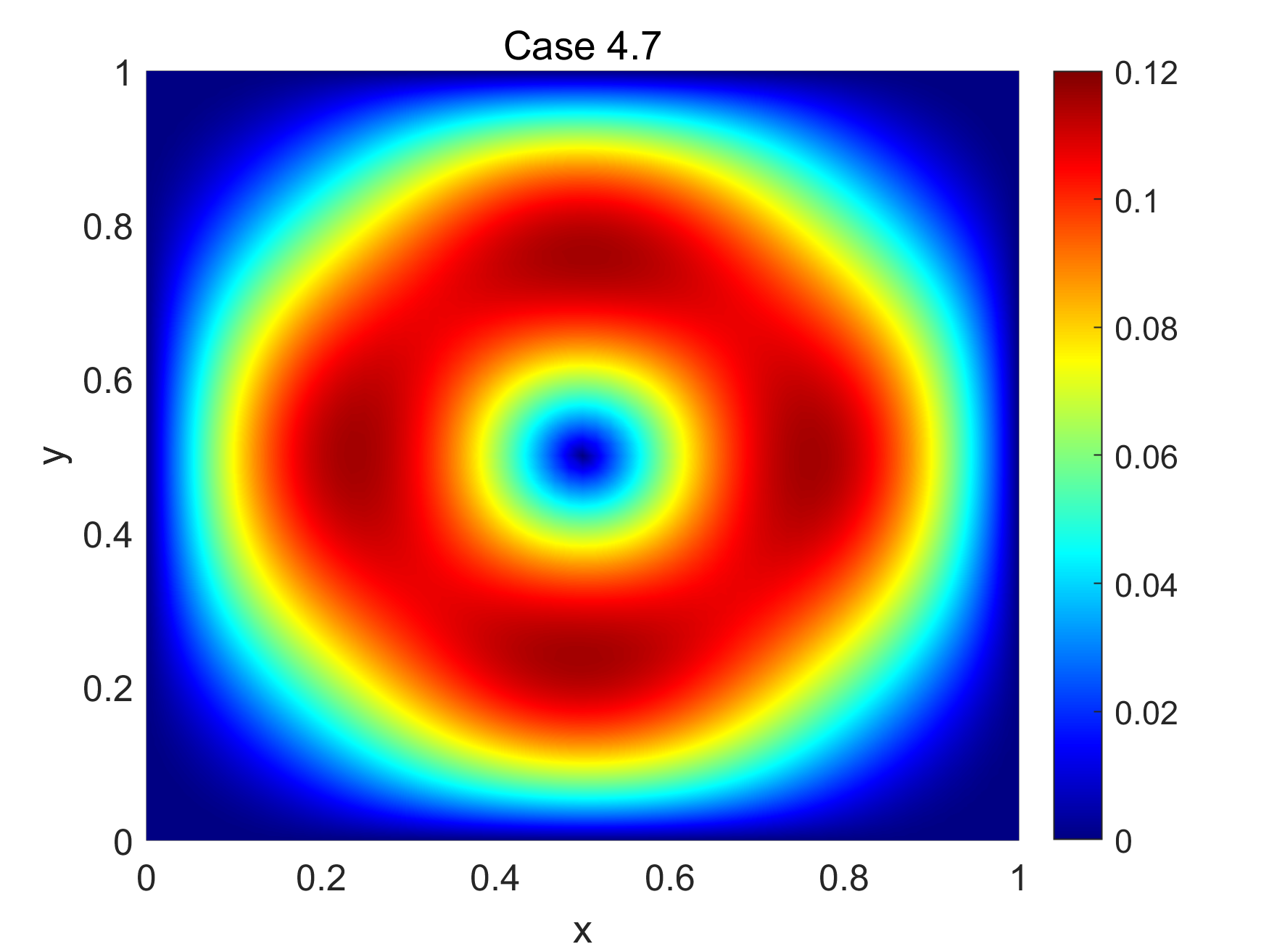}
			\end{minipage}
		}
		\subfloat[Predicted solution]{
			\begin{minipage}[t]{0.315\linewidth}
				\includegraphics[width=1\linewidth]{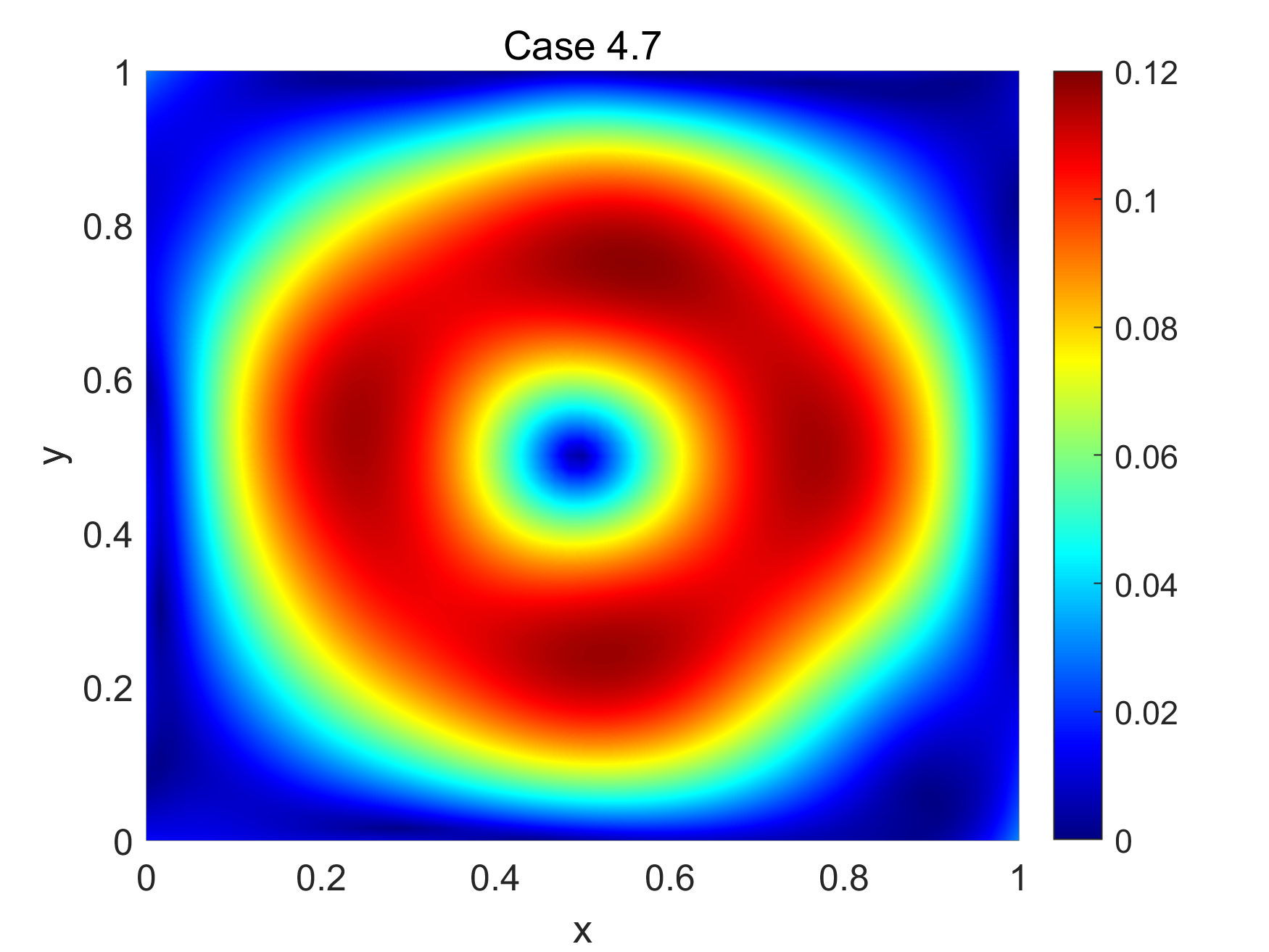}
			\end{minipage}
		}
		\subfloat[Absolute error]{
			\begin{minipage}[t]{0.315\linewidth}
				\includegraphics[width=1\linewidth]{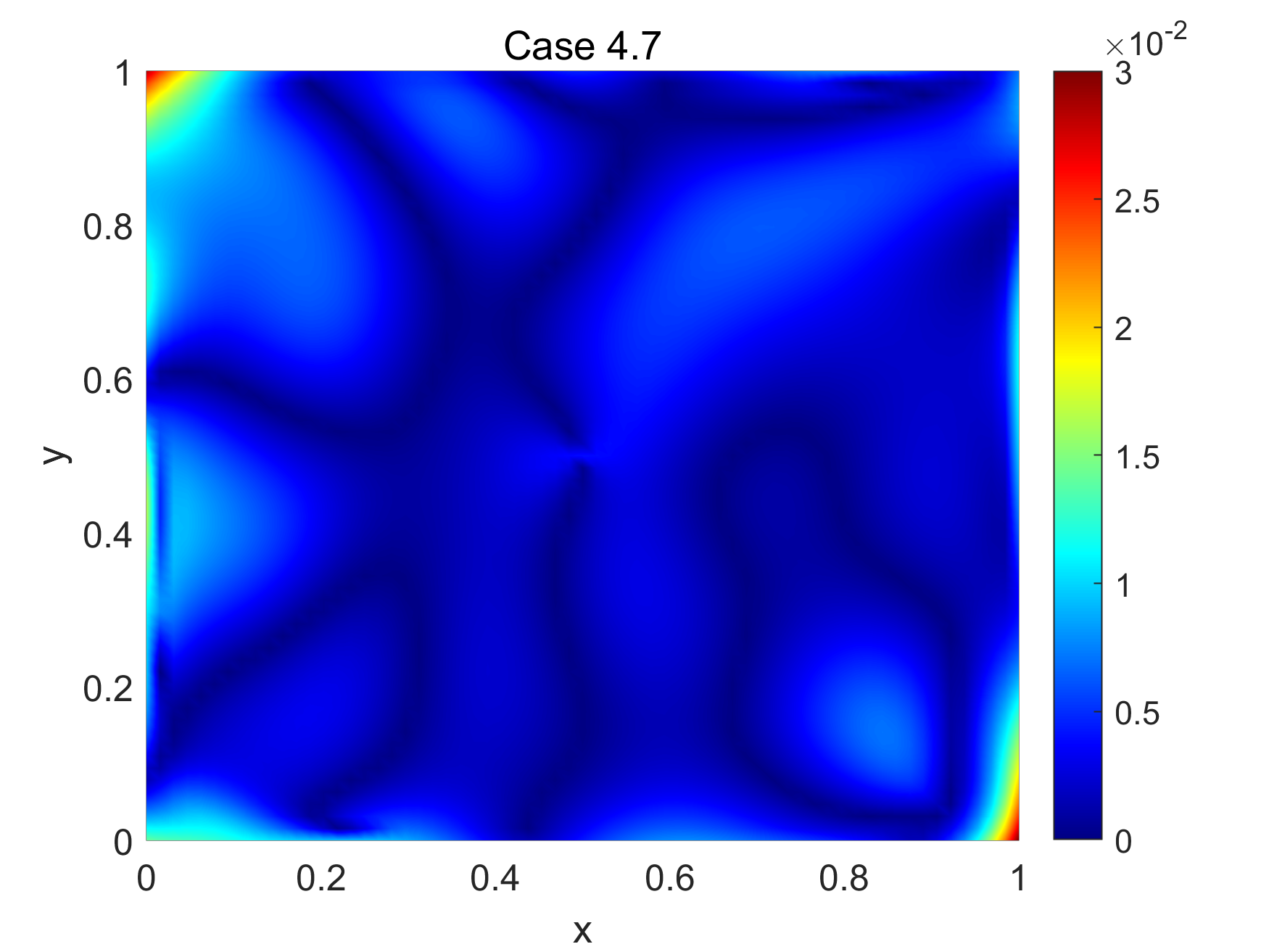}
			\end{minipage}
		}\\
		\subfloat[Reference solution]{
			\begin{minipage}[t]{0.315\linewidth}
				\includegraphics[width=1\linewidth]{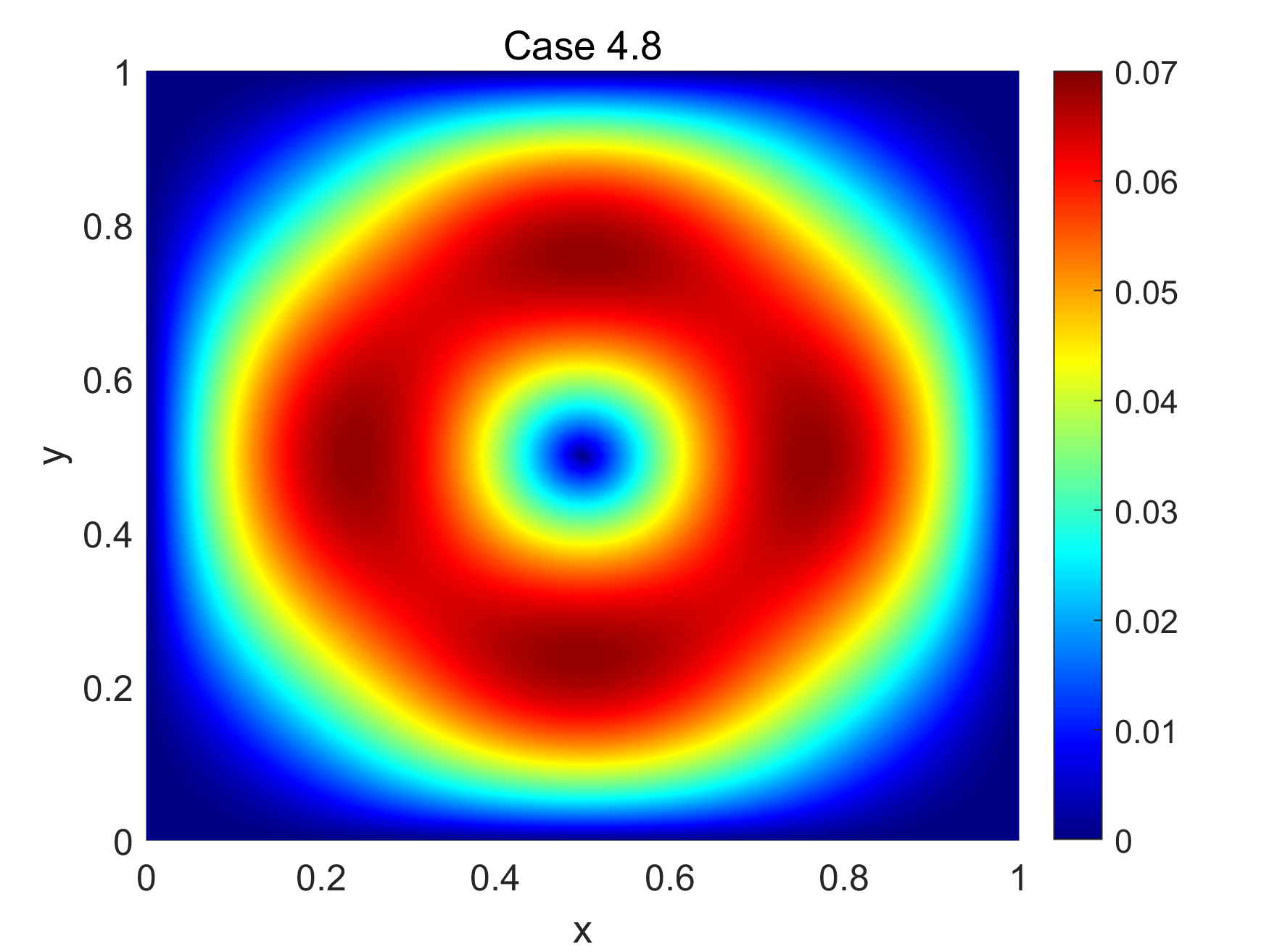}
			\end{minipage}
		}
		\subfloat[Predicted solution]{
			\begin{minipage}[t]{0.315\linewidth}
				\includegraphics[width=1\linewidth]{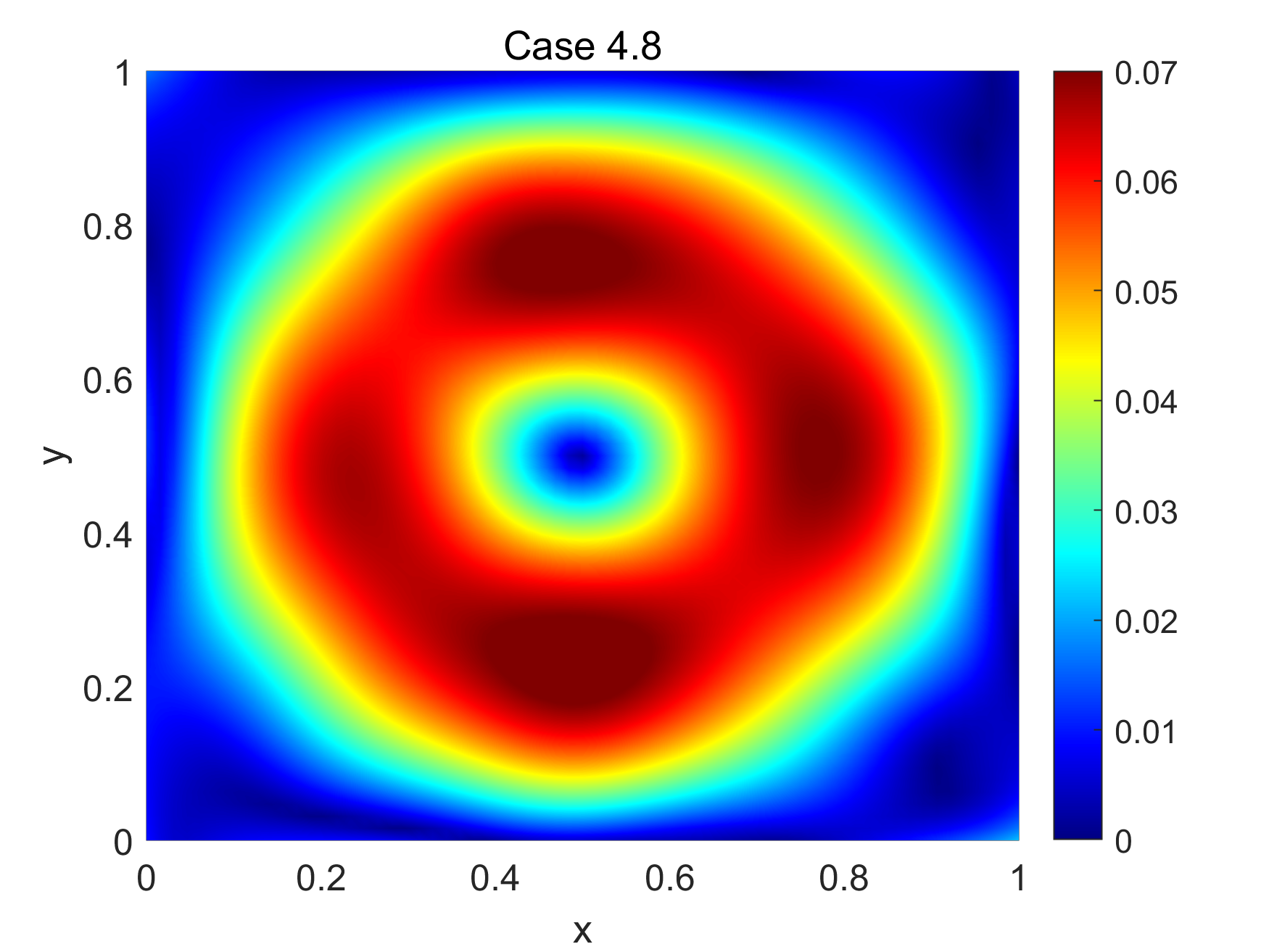}
			\end{minipage}
		}
		\subfloat[Absolute error]{
			\begin{minipage}[t]{0.315\linewidth}
				\includegraphics[width=1\linewidth]{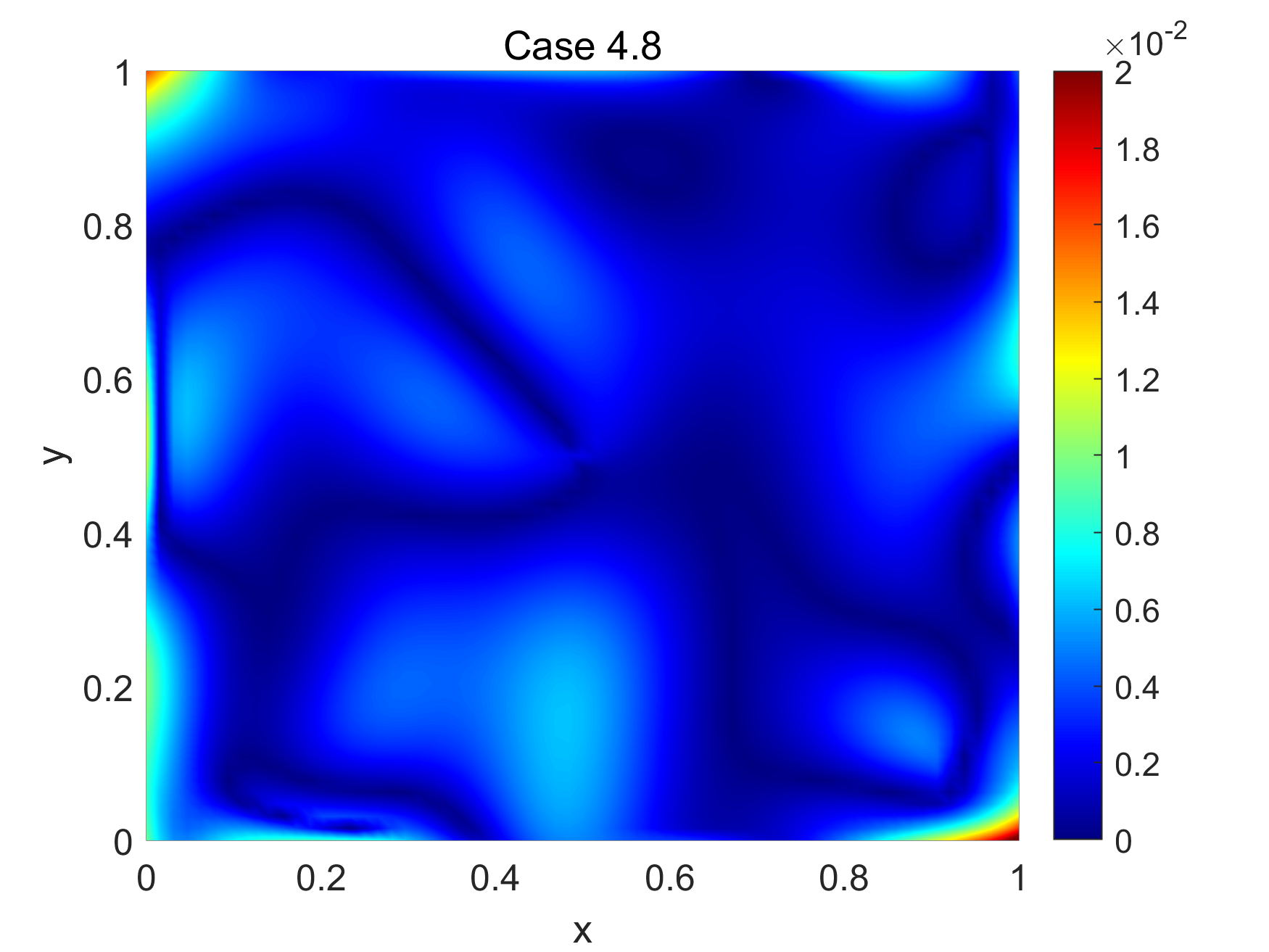}
			\end{minipage}
		}
		\caption{Numerical results for Navier-Stokes equations with discontinuously time-varying coefficient $\nu(t)$, from Case~4.5 to Case~4.8. \label{fig9}}
	\end{figure}
	
	The two-stage viscosity coefficient inversion results for these eight cases are shown in Figure~\ref{fig7}. The Stage~1 result corresponds to the relaxed continuous approximation generated by the GWS-PINNs coefficient sub network. This stage provides continuous samples of $\nu(t)$ over the temporal domain and supplies the statistical input for GMM-BDMC. The GMM-BDMC learner then automatically infers the number of viscosity states, the heuristic coefficient search intervals, and the candidate change-point interval. The Stage~2 result corresponds to the hard piecewise-constant estimator refined by CCD-PINNs under these statistical constraints.
	
	For Case~4.1, the viscosity coefficient remains constant over the whole time interval. The GMM-BDMC learner identifies $\widehat{K}=1$, indicating that no temporal change point is detected. Therefore, no change-point variable is introduced in the CCD-PINNs refinement, and the inverse problem reduces to a constrained constant-coefficient estimation task. For Cases 4.2-4.8, the viscosity coefficient switches from $0.01$ to a larger value at $t=0.5$. In these cases, GMM-BDMC identifies two viscosity states and provides a candidate change-point interval around the true transition time. The Stage~1 curves may present smooth transition layers near the jump due to the continuity of the neural approximation, while the Stage~2 results recover sharp step-function estimators and refined change points.
	
	The quantitative coefficient estimation results are reported in Table~\ref{tab1}. The temporal change-point detection results are presented in Table~\ref{tab2}, where the Stage~1 column gives the candidate change-point intervals inferred by GMM-BDMC and the Stage~2 column gives the refined change-point estimates obtained by CCD-PINNs. The solution approximation errors of the main network in both stages are summarized in Table~\ref{tab4}. For the Navier-Stokes equation, the MSE of solution reported in Table~\ref{tab4} is computed only for the velocity field $u=(u_1,u_2)$, and the pressure field $p$ is not included. These results show that the proposed framework can accurately estimate the viscosity coefficient across different jump magnitudes, from a small change in Case~4.2 to a large change in Case~4.8. For visualization of the flow field, the scalar velocity magnitude is used as
	\begin{equation}
		\|\mathbf{u}\| := \sqrt{u_1^2+u_2^2}.
	\end{equation}
	Taking into account the cumulative effect of nonlinear error propagation, a representative evolutionary end-time instant $t=1$ is selected to display the reconstructed solution fields. The solution reconstructed by the second-stage main network $\tilde{\mathbf{u}}$, the reference solution, and the corresponding absolute error are shown in Figure~\ref{fig8} and Figure~\ref{fig9}. These figures demonstrate that the refined solution network can reconstruct the velocity magnitude with small error even when the hidden viscosity coefficient undergoes discontinuous temporal changes. Meanwhile, a gradual degradation in the solution approximation accuracy can be observed as the variation amplitude of the viscosity coefficient $\nu$ increases. This trend is qualitatively consistent with the increased difficulty encountered in solving Navier–Stokes systems with strongly time-varying coefficients, since larger variations in $\nu$ alter the relative balance between convection and viscous diffusion over time and may introduce sharper temporal transitions in the solution dynamics.
	
	The Navier-Stokes cases further demonstrate the effectiveness of the proposed JVC-PINNs framework for complex nonlinear spatiotemporal systems. The Stage~1 GWS-PINNs sampler provides stable continuous approximations of the viscosity coefficient, the GMM-BDMC learner automatically distinguishes constant and jump-varying cases through the inferred component number $\widehat{K}$, and the Stage~2 CCD-PINNs estimator refines the viscosity values and the change point into an explicit piecewise-constant representation. The reconstruction results by main-network outputs for velocity field $\tilde{\mathbf{u}}$ are consistent with the observed dynamical behavior. These results suggest potential applications in fluid mechanics and aerodynamics, such as parameter inference from limited experimental data in wind tunnel simulations or flow systems modeled by Navier-Stokes equations.
	
	\subsection{Helmholtz Equation}
	
	The previous examples mainly focus on parabolic and hyperbolic PDEs. As another fundamental class of PDEs, elliptic equations are also considered to further validate the applicability of the proposed framework. Although elliptic problems do not contain a physical time variable, the spatial coefficient samples generated by the first-stage GWS-PINNs sub network can still be flattened into a one-dimensional coefficient sample set. After this transformation, the GMM-BDMC statistical learner can be applied to infer the discrete coefficient states, heuristic coefficient search intervals, and candidate discontinuity regions. The second-stage CCD-PINNs then refines the hard piecewise-constant coefficient field and its spatial jump boundary under these statistical constraints.
	
	A 2D Helmholtz equation with a 2D space-varying coefficient $k(x,y)$ is considered as a representative elliptic example
	\begin{equation}
		\Delta u + k(x,y)^2 u = f(x,y), \quad (x,y) \in U \subset \mathbb{R}^2,
	\end{equation}
	where $u(x,y)=v(x,y)+\mathrm{i}w(x,y)\in\mathbb{C}$ is the complex-valued scalar solution with $v,w:\,U\to\mathbb{R}$. This example is taken from \cite{zhang2023identification}. The Sommerfeld-type radiation condition is given by
	\begin{equation}
		\left| \frac{\partial u}{\partial r} -
		\mathrm{i}\,k(x,y)\,u \right| =
		o\bigl(r^{-1/2}\bigr), \quad r\to\infty,
	\end{equation}
	where $r=\sqrt{x^2+y^2}$. In this experiment, the computational domain is set as
	\begin{equation}
		U=[-1,1]\times[-1,1],
	\end{equation}
	and the right-hand side function is
	\begin{equation}
		f(x,y)=
		\begin{cases}
			-\dfrac{1}{2a\sqrt{\pi}}
			\exp\left[
			-\left(
			\dfrac{|(x,y)-(0.5,0)|}{a}
			\right)^2
			\right],
			& |(x,y)-(0.5,0)|<0.5,\\[2pt]
			0,
			& |(x,y)-(0.5,0)|\geq0.5,
		\end{cases}
	\end{equation}
	where $a=\pi/k_0$. In this numerical experiment, $k_0=20$ is used. A regular square open region $U_1\subset U$ is located on the left side of the spatial domain, where the coefficient differs from that in the exterior region. The spatially varying coefficient $k(x,y)$ is set as follows.
	
	\textbf{Case~5.1:}
	\begin{equation}
		k(x,y)=
		\begin{cases}
			\dfrac{k_0}{1.5}, & (x,y)\in U_1,\\[2pt]
			k_0, & (x,y)\in U\setminus U_1.
		\end{cases}
	\end{equation}
	
	For Case~5.1, $\theta_p=k(x,y)$ on $(x,y)\in[-1,1]^2$. Stage~1 networks used tanh networks. The complex-valued main network used layers $[2,60,60,60,60,60,60,60,60,2]$, and the coefficient network used layers $[2,20,20,1]$. Using seed $314$, Adam for $200000$ iterations, learning rates $10^{-3}$ for both networks, reduce factor $0.6$, $5000$ interior residual/observation points, and $2048$ boundary points. There were no separate initial-condition points, corresponding to $\gamma_1=\gamma_2=1$, $\gamma_3=0$. Stage~2 used a $32\times32$ patch-wise hard field, applied GMM-BDMC in the spatial domain, and then refined the transferred main network at learning rate $10^{-4}$ and sub network at learning rate $10^{-3}$ for $20000$ Adam iterations.
	
\begin{figure}[p]
		\centering
		\subfloat[Reference solution $v(x,y)$]{
			\begin{minipage}[t]{0.315\linewidth}
				\includegraphics[width=1\linewidth]{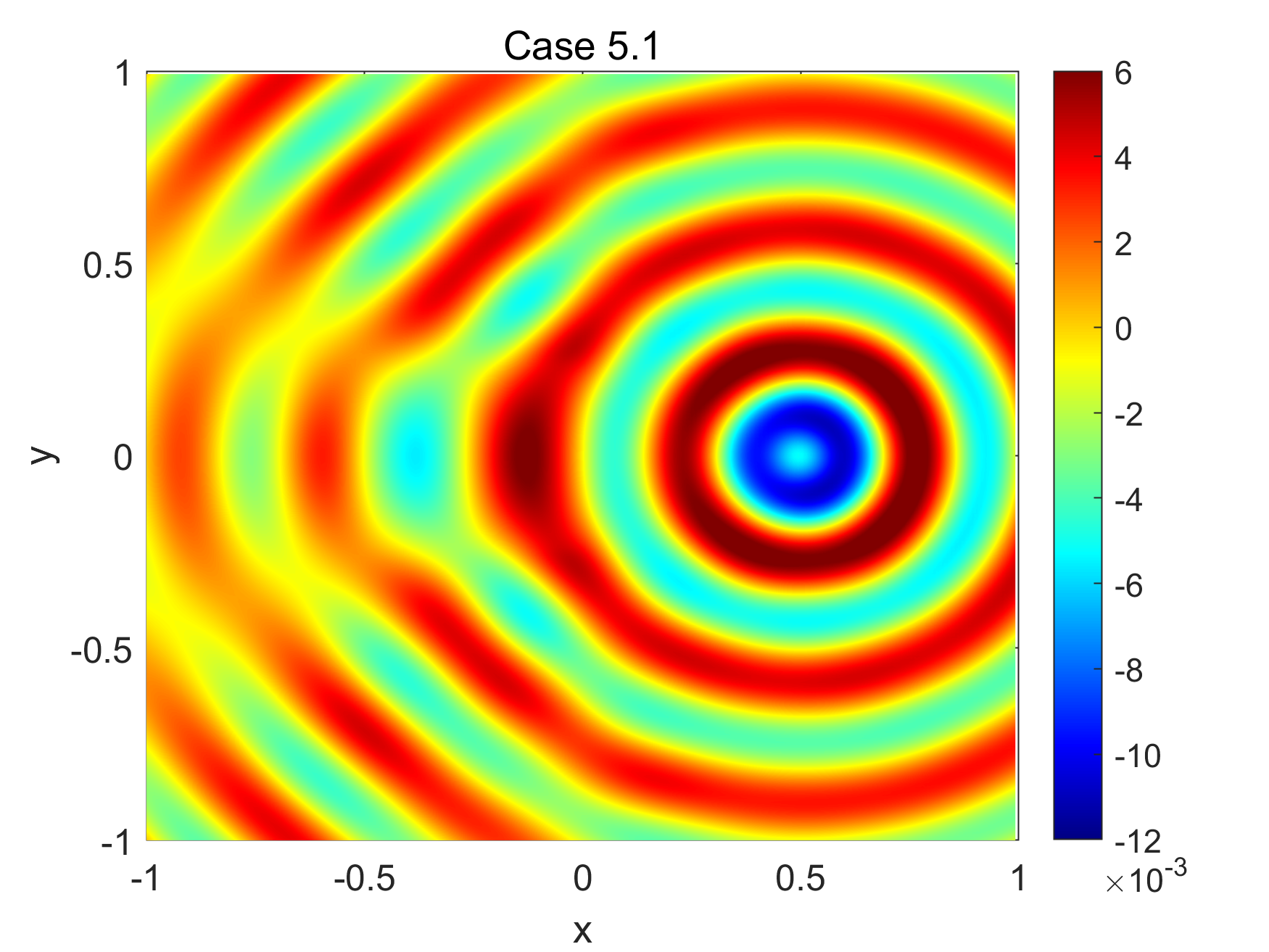}
			\end{minipage}
		}
		\subfloat[Predicted solution $v(x,y)$]{
			\begin{minipage}[t]{0.315\linewidth}
				\includegraphics[width=1\linewidth]{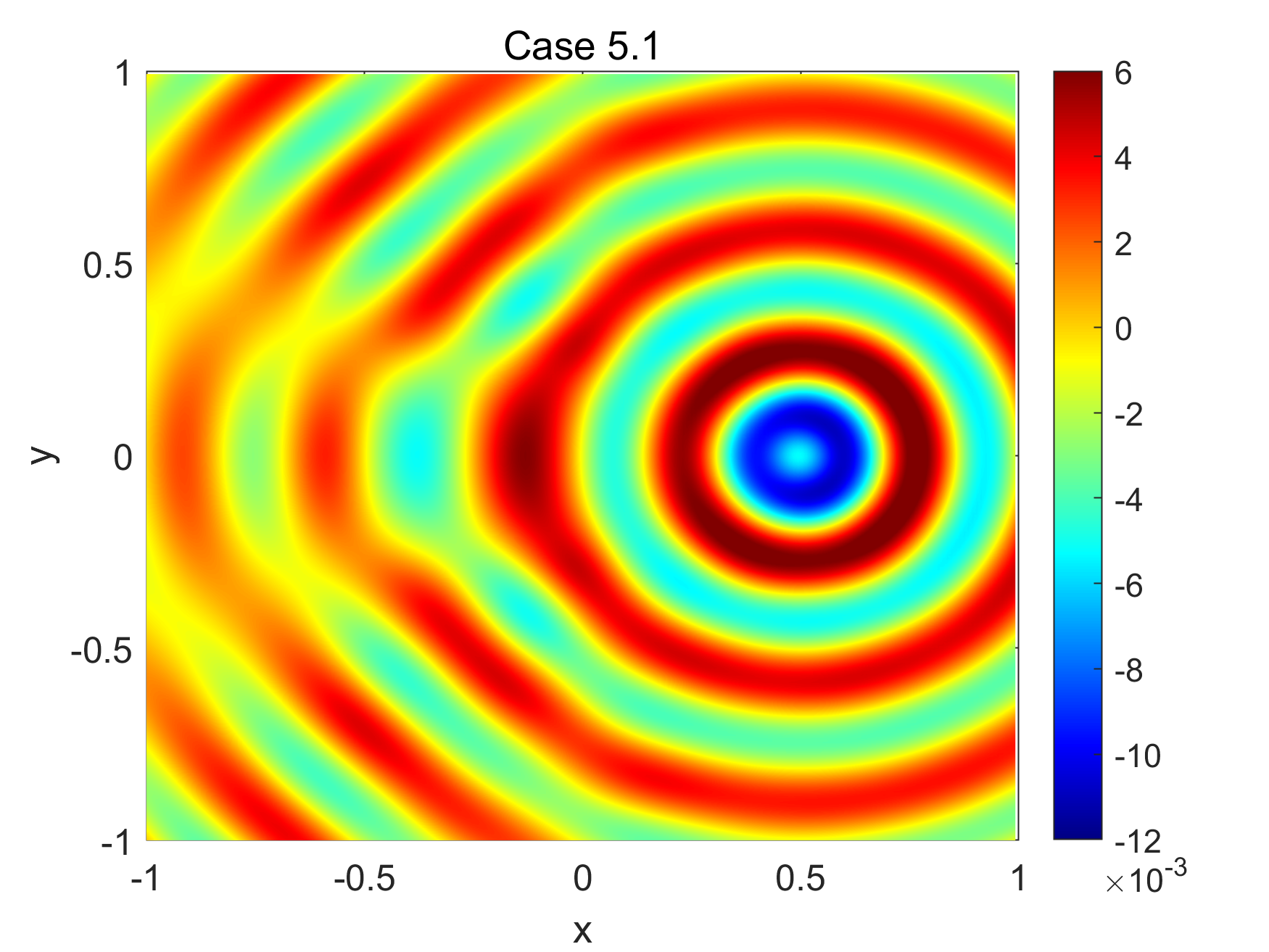}
			\end{minipage}
		}
		\subfloat[Absolute error]{
			\begin{minipage}[t]{0.315\linewidth}
				\includegraphics[width=1\linewidth]{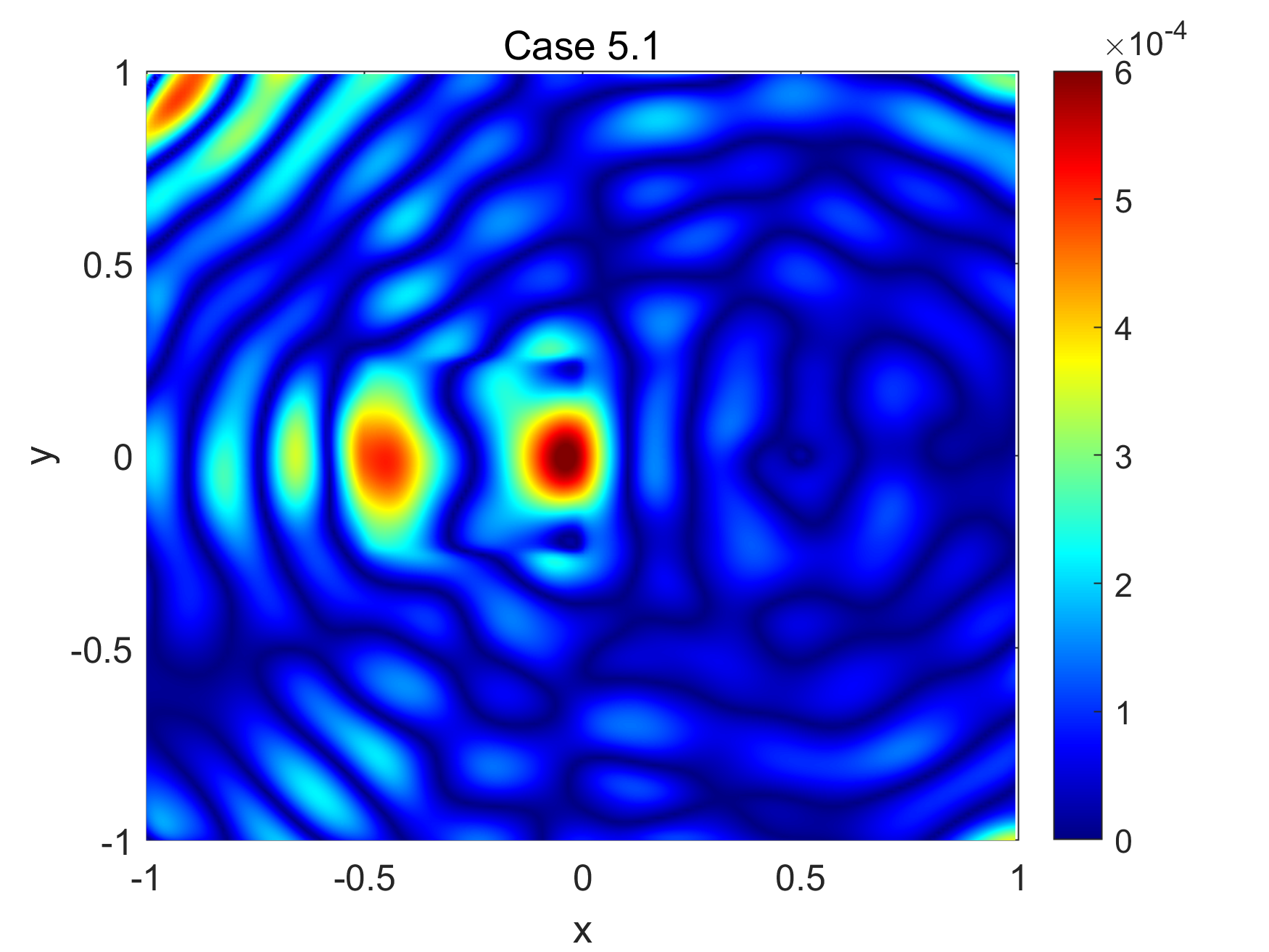}
			\end{minipage}
		}\\
		\subfloat[Reference solution $w(x,y)$]{
			\begin{minipage}[t]{0.315\linewidth}
				\includegraphics[width=1\linewidth]{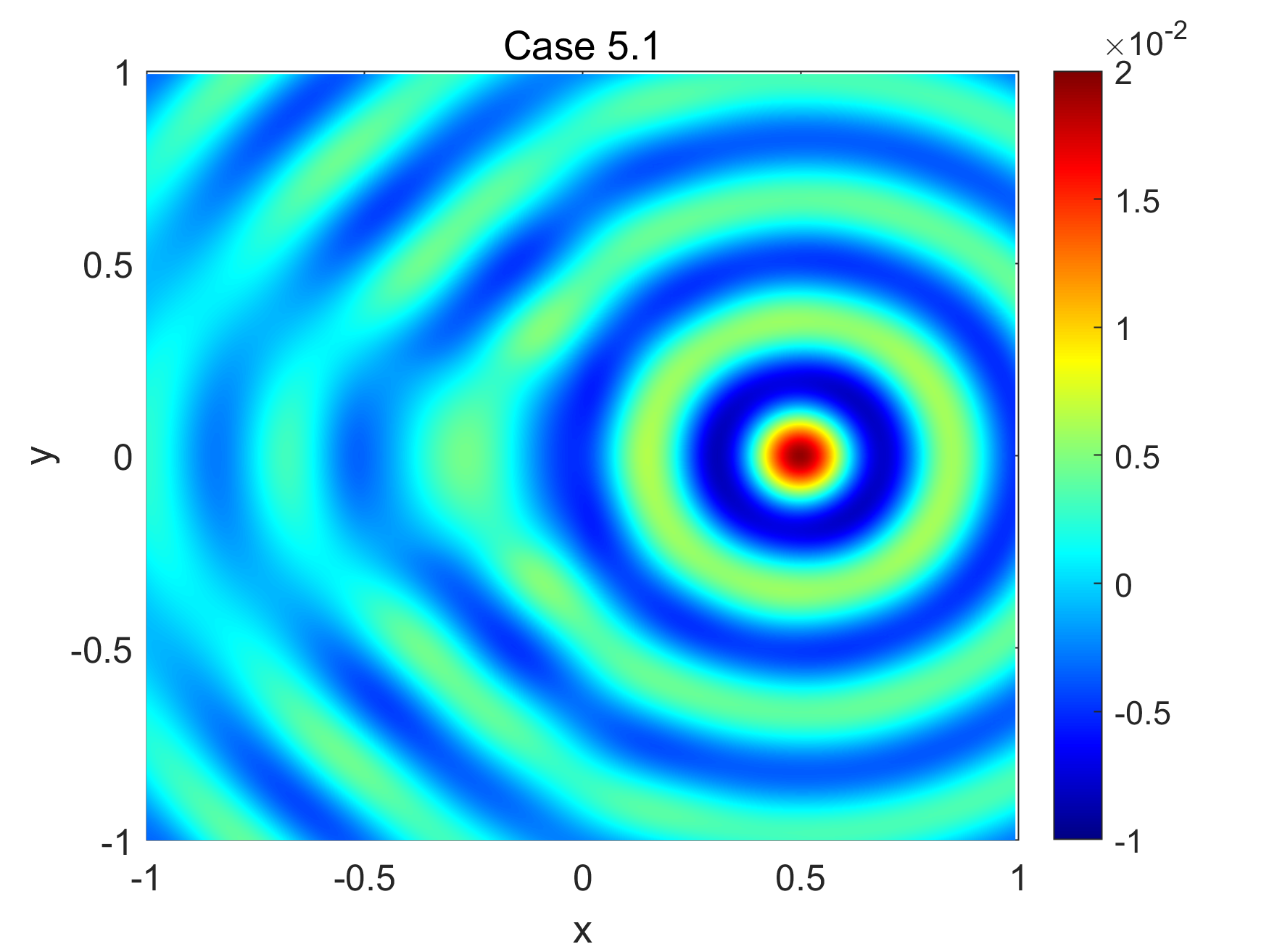}
			\end{minipage}
		}
		\subfloat[Predicted solution $w(x,y)$]{
			\begin{minipage}[t]{0.315\linewidth}
				\includegraphics[width=1\linewidth]{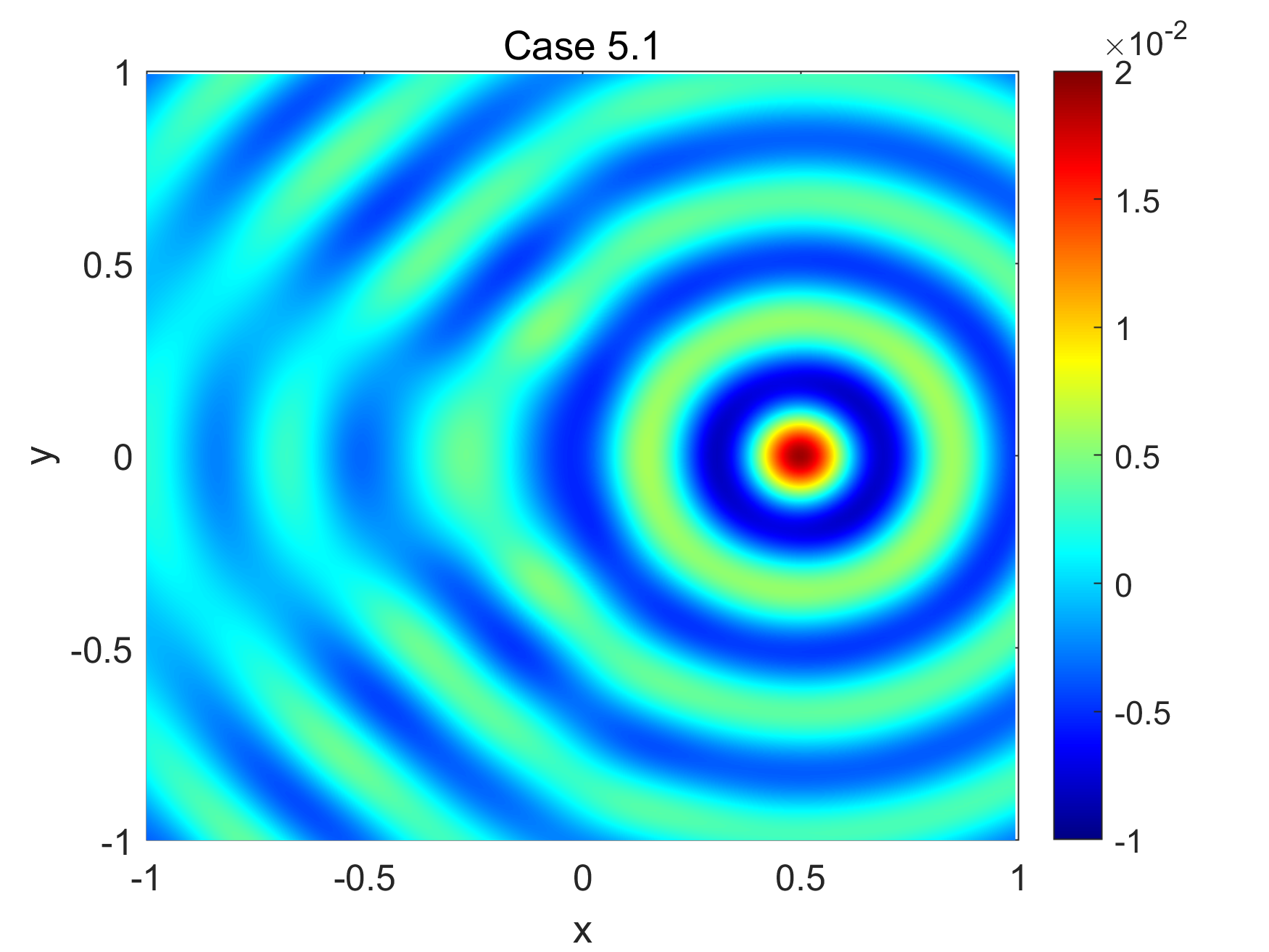}
			\end{minipage}
		}
		\subfloat[Absolute error]{
			\begin{minipage}[t]{0.315\linewidth}
				\includegraphics[width=1\linewidth]{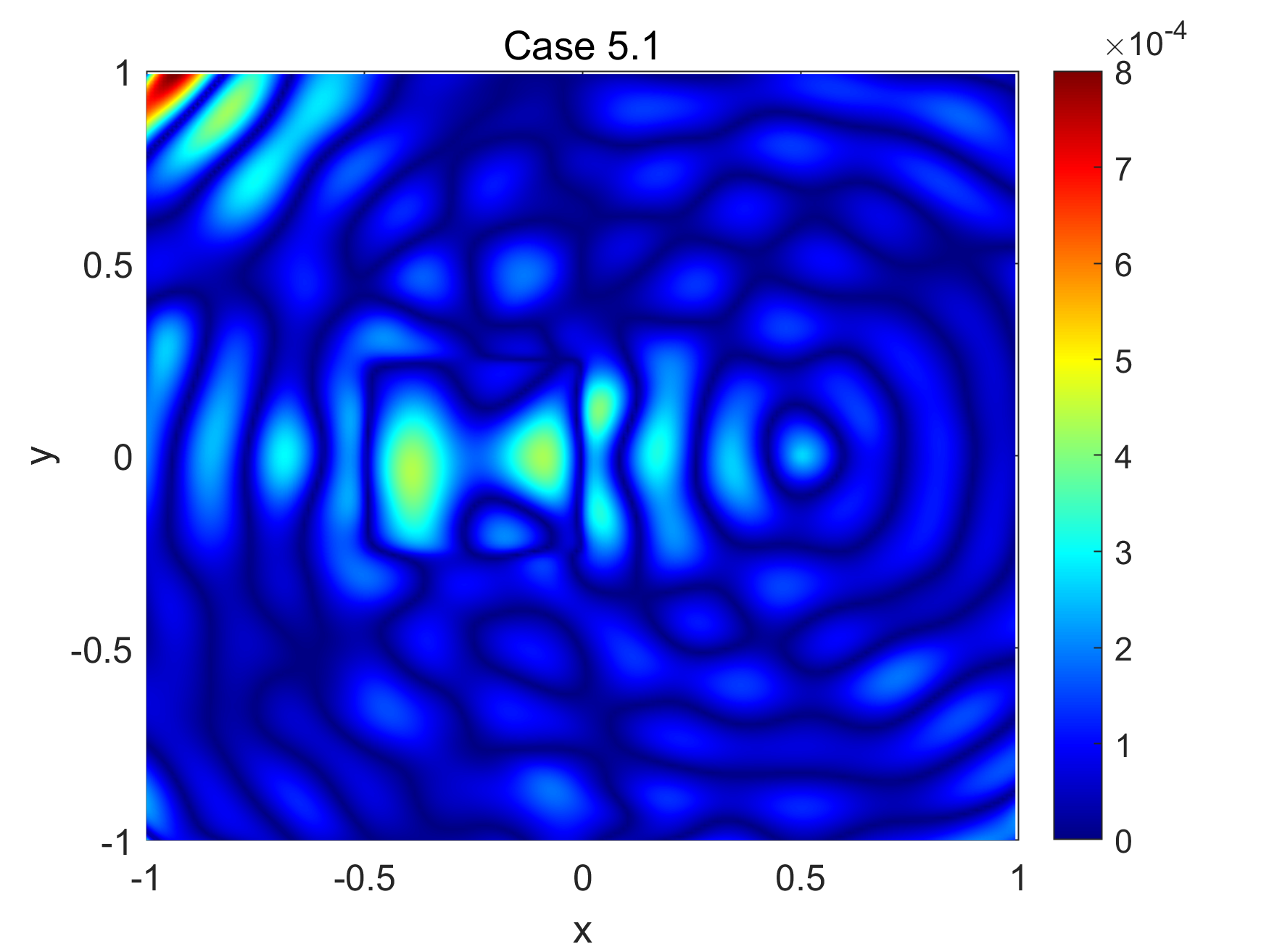}
			\end{minipage}
		}\\
		\subfloat[Reference value $k(x,y)$]{
			\begin{minipage}[t]{0.315\linewidth}
				\includegraphics[width=1\linewidth]{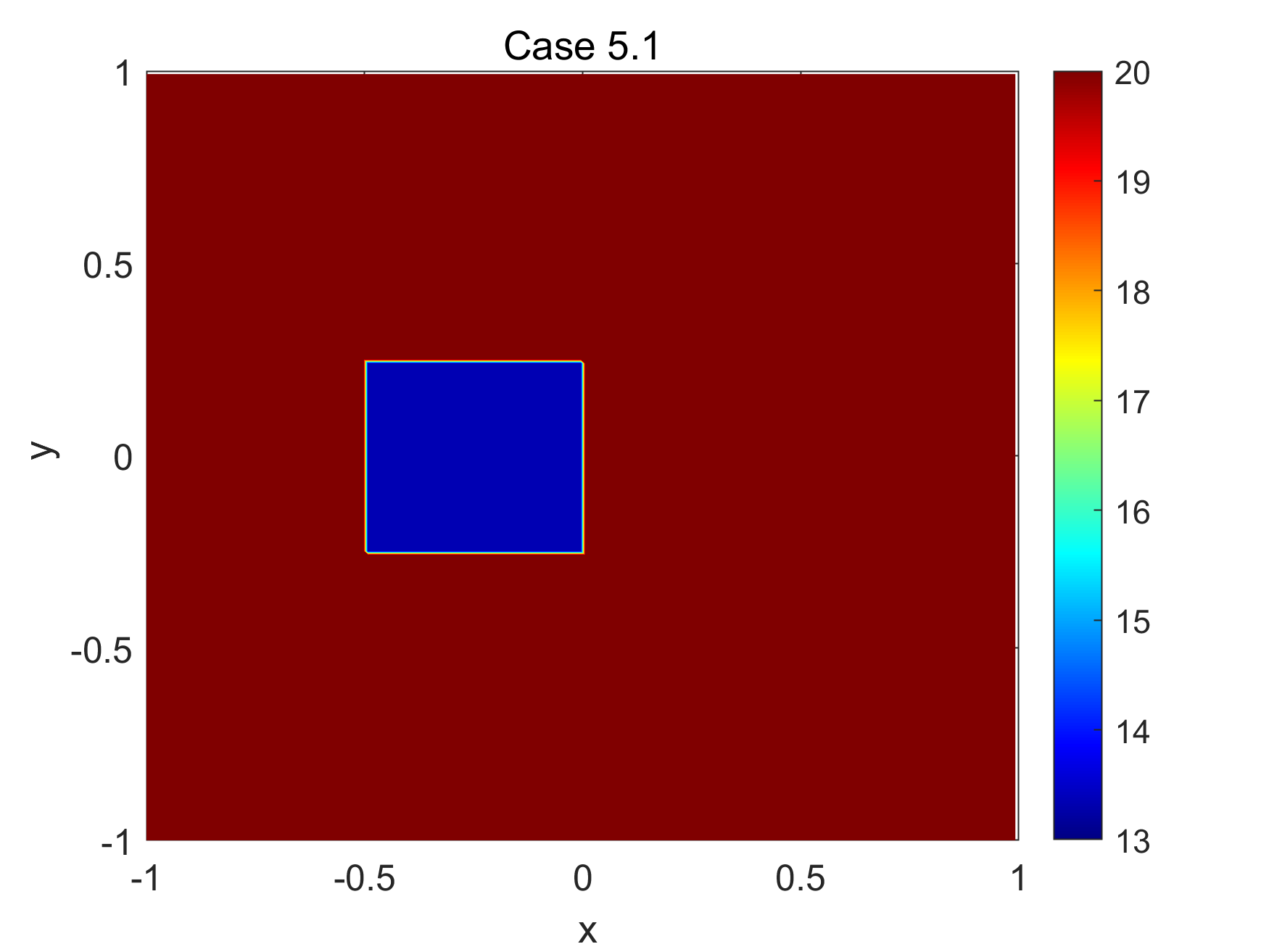}
			\end{minipage}
		}
		\subfloat[Parameter inverse result]{
			\begin{minipage}[t]{0.315\linewidth}
				\includegraphics[width=1\linewidth]{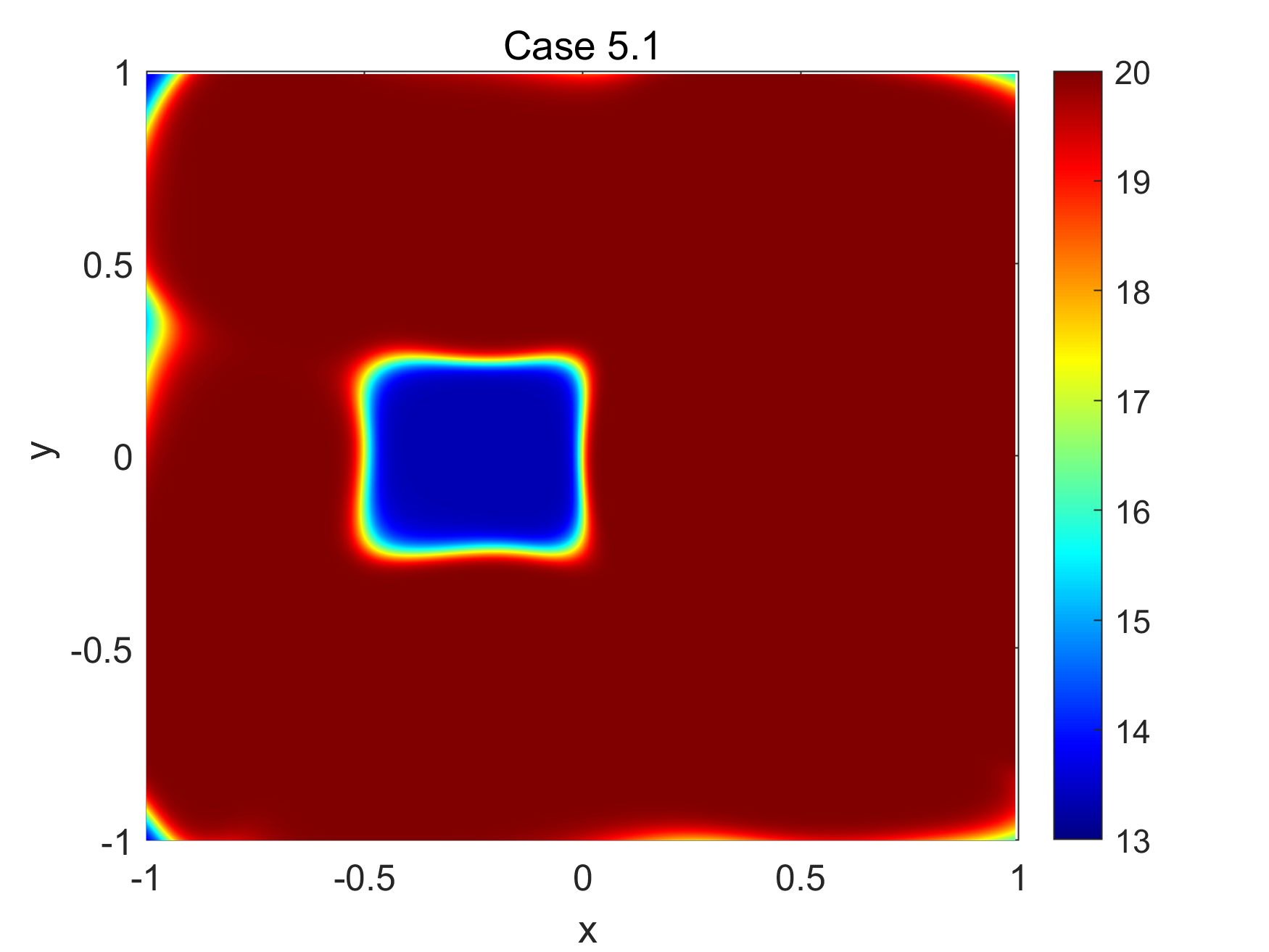}
			\end{minipage}
		}
		\subfloat[Stage~1 absolute error]{
			\begin{minipage}[t]{0.315\linewidth}
				\includegraphics[width=1\linewidth]{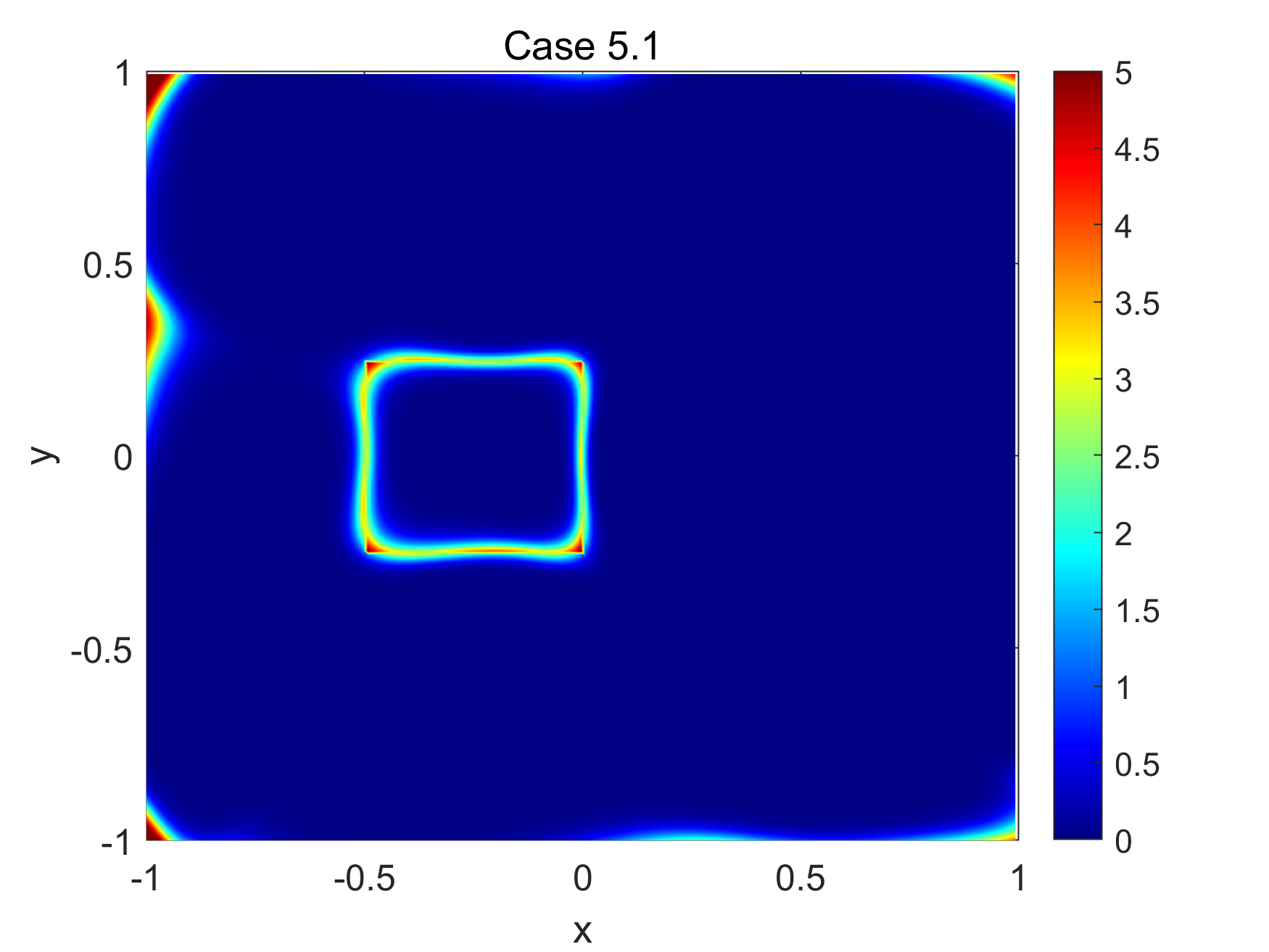}
			\end{minipage}
		}\\
		\subfloat[Parameter estimate result]{
			\begin{minipage}[t]{0.315\linewidth}
				\includegraphics[width=1\linewidth]{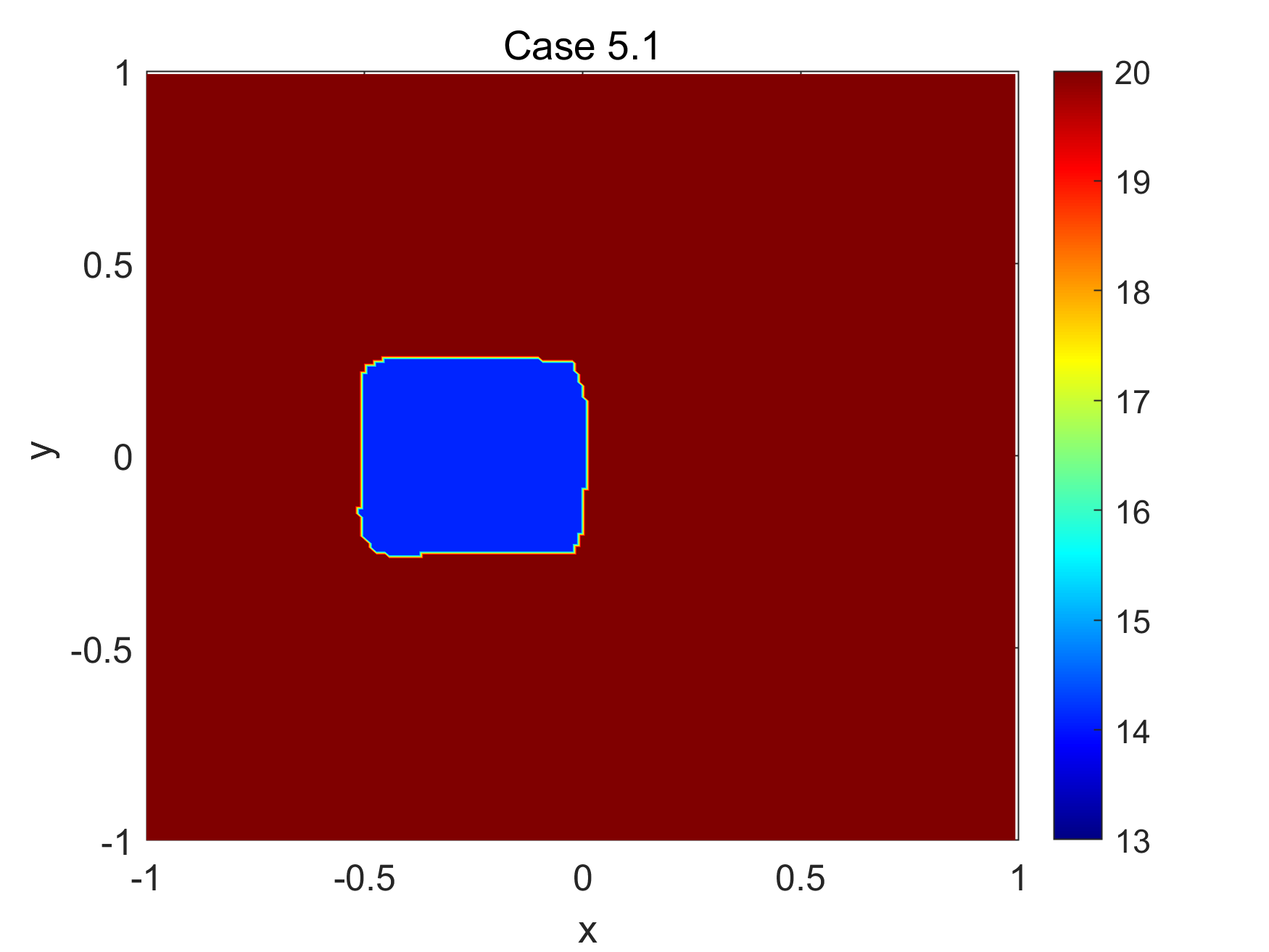}
			\end{minipage}
		}
		\subfloat[Stage~2 absolute error]{
			\begin{minipage}[t]{0.315\linewidth}
				\includegraphics[width=1\linewidth]{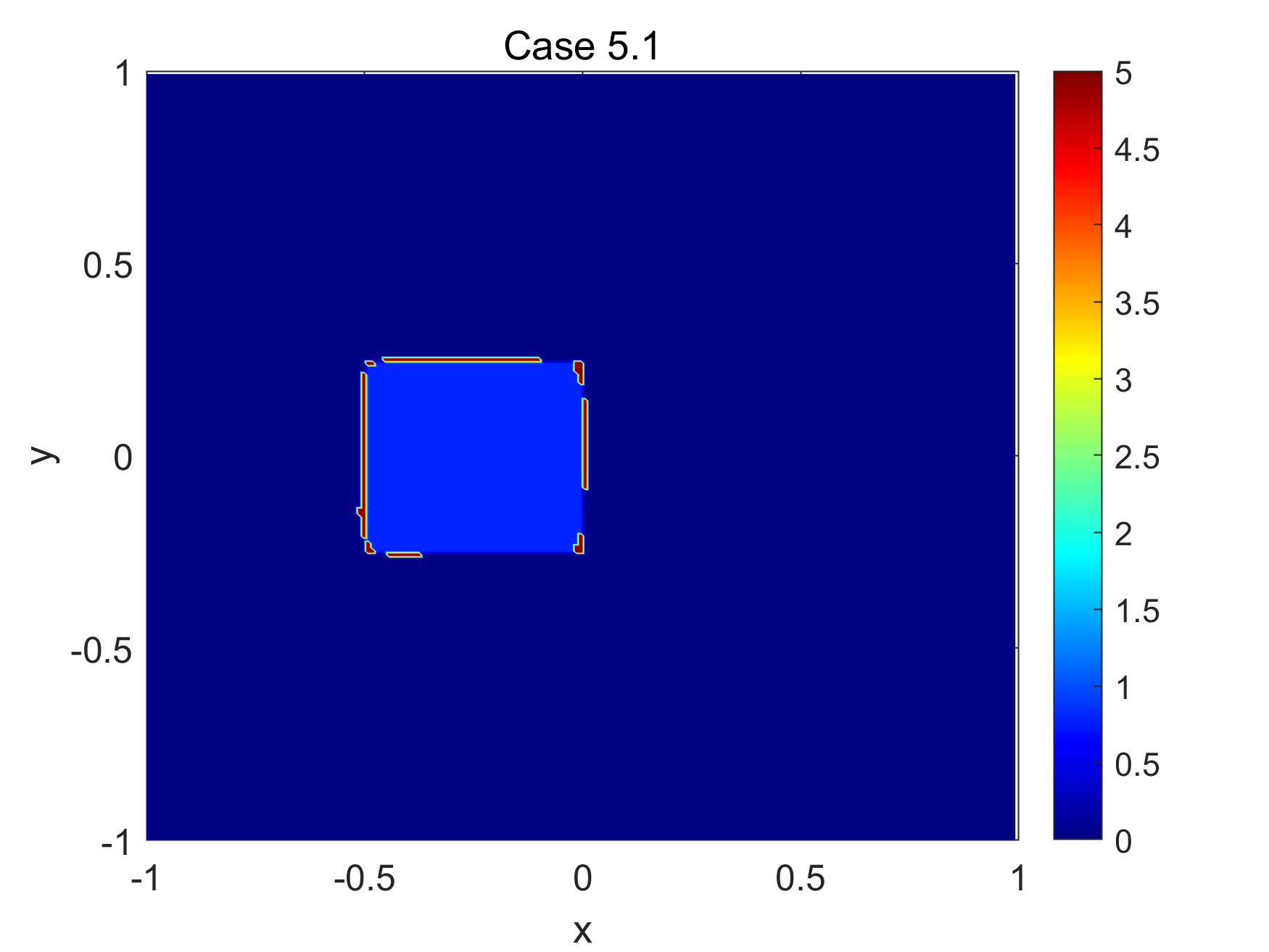}
			\end{minipage}
		}
		\subfloat[Change boundary detection]{
			\begin{minipage}[t]{0.315\linewidth}
				\includegraphics[width=1\linewidth]{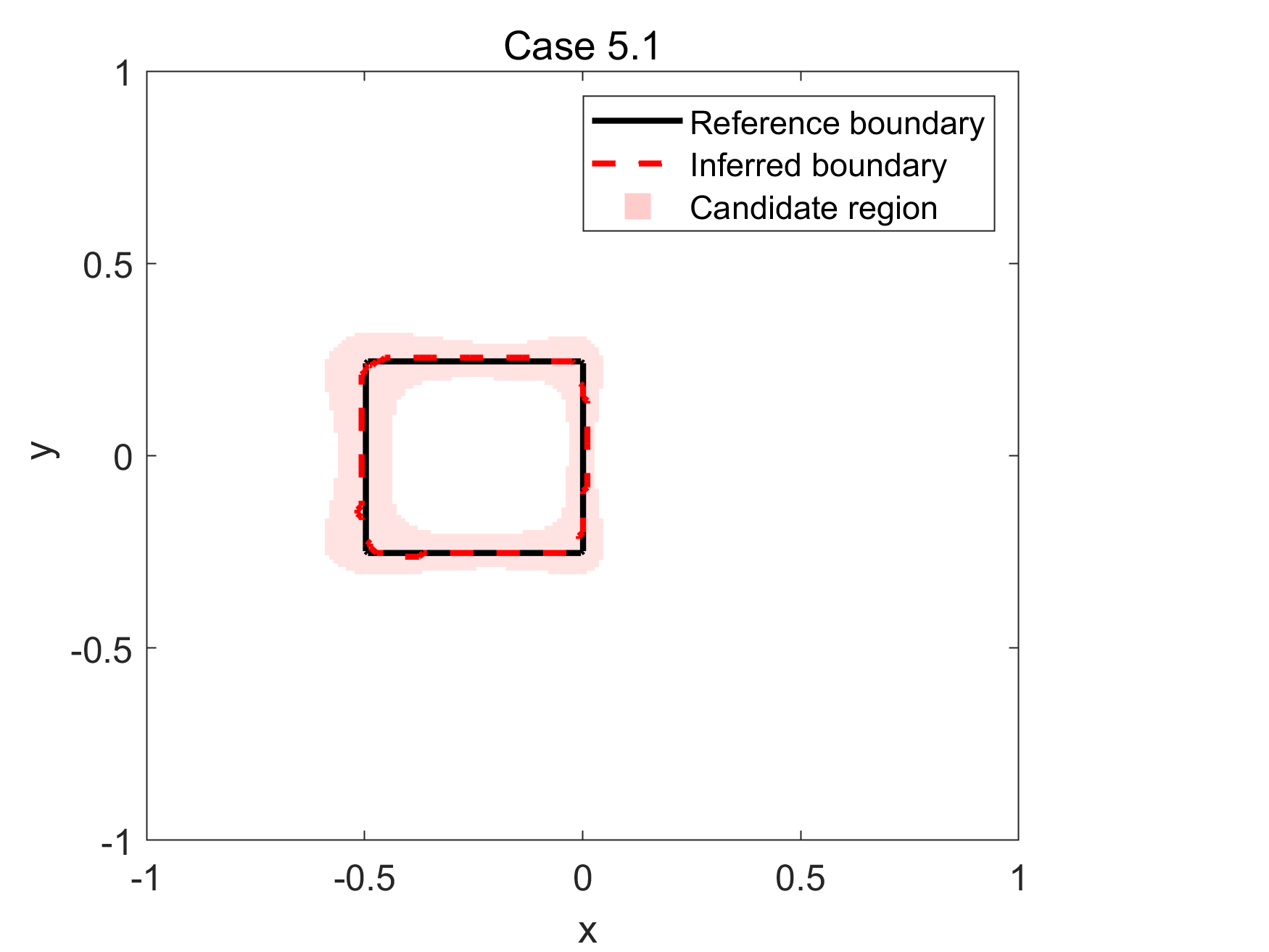}
			\end{minipage}
		}
		\caption{Numerical results for the Helmholtz equation with discontinuously 2D space-varying coefficient $k(x,y)$.  \label{fig10}}
	\end{figure}

	The comparison between the reconstructed solution from the second-stage main network and the reference solution, together with the corresponding absolute error, is shown in the first two rows of Figure~\ref{fig10}. The spatial coefficient inversion results are shown in the third and fourth rows of Figure~\ref{fig10}, including the reference coefficient field, the Stage~1 relaxed continuous approximation produced by the GWS-PINNs coefficient sub network, and the Stage~2 hard piecewise-constant reconstruction refined by CCD-PINNs. The coefficient search intervals inferred by GMM-BDMC constrain the possible values of the two coefficient states, while the candidate discontinuity region guides the refinement of the square jump boundary. The quantitative coefficient estimation results are reported in Table~\ref{tab1}, the jump-boundary detection error is reported in Table~\ref{tab3}, and the solution approximation errors of the main network in the two stages are summarized in Table~\ref{tab4}.
	
	The results show that the proposed JVC-PINNs framework can accurately reconstruct both the solution field and the discontinuously varying spatial coefficient for the Helmholtz equation. The Stage~1 GWS-PINNs coefficient sampler captures the global spatial distribution of $k(x,y)$ but smooths the discontinuity near the boundary of $U_1$. Based on the statistical constraints supplied by GMM-BDMC, the Stage~2 CCD-PINNs refinement estimator sharpens the transition and produces an explicit hard piecewise-constant coefficient estimator. The predicted jump boundary is close to the reference square interface, demonstrating the ability of the proposed framework to handle elliptic PDEs with spatially discontinuous coefficients.
	
	From the spatially varying coefficient examples, the proposed JVC-PINNs framework successfully completes parameter estimation for PDEs with spatial jump coefficients in both irregular and regular subdomains. In addition, the MSEs of the identified discontinuity boundary functions are of the same order of magnitude for both cases, indicating that the proposed method maintains comparable boundary-reconstruction accuracy under different geometric structures. These results demonstrate that the combination of GWS-PINNs coefficient sampling, GMM-BDMC statistical inference, and CCD-PINNs constrained refinement is effective for multidimensional spatial parameter identification. This capability has potential for practical applications involving heterogeneous media.
	
	\begin{table}[p]
		\centering
		\caption{Coefficient sampling and estimating results of sub network in two-stage PINNs framework. \label{tab1}}
		\scalebox{1}{
			\setlength{\tabcolsep}{2.5pt}
			\begin{tabular}{ccccccc}
				\toprule
				Equation & Case & Parameter & True & Stage~1 & Stage~2 & Relative Error \\
				\midrule
				\multirow{5}*{Wave} & 1.1 & $\alpha(t)$ & 1 & $[0.9997, 1.0001]$ & $0.9999$ & $-$0.0098\% \\
				\cmidrule(lr){2-7}
				~ & \multirow{2}*{1.2} & \multirow{2}*{$\alpha(t)$} & 0.5 & $[0.4727, 0.5362]$ & $0.4991$ & $-$0.1700\% \\
				~ & ~ & ~ & 1 & $[0.9516, 1.0222]$ & $0.9980$ & $-$0.1955\% \\
				\cmidrule(lr){2-7}
				~ & \multirow{2}*{1.3} & \multirow{2}*{$\alpha(x,y)$} & 2.5 & $[2.3359, 2.8350]$ & $2.5114$ & +0.4572\% \\
				~ & ~ & ~ & 3 & $[2.9669, 3.0164]$ & $3.0000$ & +0.0000\% \\
				\midrule
				
				\multirow{5}*{Heat} & \multirow{2}*{2.1} & \multirow{2}*{$c(t)$} & 0.1 & $[0.0904, 0.1095]$ & $0.0980$ & $-$2.0034\% \\
				~ & ~ & ~ & 0.2 & $[0.1875, 0.2044]$ & $0.1981$ & $-$0.9534\% \\
				\cmidrule(lr){2-7}
				~ & \multirow{3}*{2.2} & \multirow{3}*{$c(t)$} & 0.05 & $[0.0462, 0.0592]$ & $0.0498$ & $-$0.3936\% \\
				~ & ~ & ~ & 0.1 & $[0.0861, 0.1143]$ & $0.0988$ & $-$1.1741\% \\
				~ & ~ & ~ & 0.2 & $[0.1792, 0.2077]$ & $0.2000$ & +0.0168\% \\
				\midrule
				
				\multirow{15}*{Burgers} & \multirow{2}*{3.1} & $\lambda_1(t)$ & 1.5 & $[1.4957, 1.5036]$ & $1.4996$ & $-$0.0249\% \\
				\cmidrule(lr){3-7}
				~ & ~ & $\lambda_2(t)$ & 0.1 & $[0.0994, 0.1006]$ & $0.1000$ & +0.0021\% \\
				\cmidrule(lr){2-7}
				~ & \multirow{3}*{3.2} & \multirow{2}*{$\lambda_1(t)$} & 0.5 & $[0.4856, 0.5172]$ & $0.4998$ & $-$0.0387\% \\
				~ & ~ & ~ & 1 & $[0.9833, 1.0140]$ & $0.9998$ & $-$0.0152\% \\
				\cmidrule(lr){3-7}
				~ & ~ & $\lambda_2(t)$ & 0.1 & $[0.0997, 0.1003]$ & $0.1000$ & $-$0.0183\% \\
				\cmidrule(lr){2-7}
				~ & \multirow{4}*{3.3} & \multirow{3}*{$\lambda_1(t)$} & 0.5 & $[0.4802, 0.5367]$ & $0.5013$ & +0.2570\% \\
				~ & ~ & ~ & 0.75 & $[0.7250, 0.7796]$ & $0.7498$ & $-$0.0313\% \\
				~ & ~ & ~ & 1 & $[0.9837, 1.0126]$ & $0.9999$ & $-$0.0113\% \\
				\cmidrule(lr){3-7}
				~ & ~ & $\lambda_2(t)$ & 0.1 & $[0.0994, 0.1006]$ & $0.1000$ & $-$0.0041\% \\
				\cmidrule(lr){2-7}
				~ & \multirow{6}*{3.4} & \multirow{3}*{$\lambda_1(t)$} & 0.5 & $[0.4833, 0.5388]$ & $0.5083$ & +1.6584\% \\
				~ & ~ & ~ & 0.75 & $[0.7201, 0.7921]$ & $0.7483$ & $-$0.2311\% \\
				~ & ~ & ~ & 1 & $[0.9620, 1.0296]$ & $1.0003$ & +0.0329\% \\
				\cmidrule(lr){3-7}
				~ & ~ & \multirow{3}*{$\lambda_2(t)$} & 1 & $[0.9578, 1.0314]$ & $0.9842$ & $-$1.5758\% \\
				~ & ~ & ~ & 1.3333 & $[1.2642, 1.4001]$ & $1.3274$ & $-$0.4474\% \\
				~ & ~ & ~ & 2 & $[1.9169, 2.0353]$ & $1.9958$ & $-$0.2116\% \\
				\midrule
				
				\multirow{15}*{Navier-Stokes} & 4.1 & $\nu(t)$ & 0.01 & $[0.0095, 0.0101]$ & $0.0098$ & $-$2.0083\% \\
				\cmidrule{2-7}
				~ & \multirow{2}*{4.2} & \multirow{2}*{$\nu(t)$} & 0.01 & $[0.0088, 0.0117]$ & $0.0101$ & +1.4376\% \\
				~ & ~ & ~ & 0.02 & $[0.0180, 0.0212]$ & $0.0198$ & $-$0.7740\% \\
				\cmidrule{2-7}
				~ & \multirow{2}*{4.3} & \multirow{2}*{$\nu(t)$} & 0.01 & $[0.0080, 0.0130]$ & $0.0102$ & +1.7357\% \\
				~ & ~ & ~ & 0.03 & $[0.0264, 0.0324]$ & $0.0297$ & $-$0.8514\% \\
				\cmidrule{2-7}
				~ & \multirow{2}*{4.4} & \multirow{2}*{$\nu(t)$} & 0.01 & $[0.0064, 0.0144]$ & $0.0103$ & +2.6291\% \\
				~ & ~ & ~ & 0.04 & $[0.0352, 0.0440]$ & $0.0402$ & +0.4476\% \\
				\cmidrule{2-7}
				~ & \multirow{2}*{4.5} & \multirow{2}*{$\nu(t)$} & 0.01 & $[0.0057, 0.0156]$ & $0.0103$ & +2.7598\% \\
				~ & ~ & ~ & 0.05 & $[0.0449, 0.0551]$ & $0.0506$ & +1.1087\% \\
				\cmidrule{2-7}
				~ & \multirow{2}*{4.6} & \multirow{2}*{$\nu(t)$} & 0.01 & $[0.0055, 0.0165]$ & $0.0103$ & +2.8712\% \\
				~ & ~ & ~ & 0.06 & $[0.0541, 0.0658]$ & $0.0607$ & +1.1240\% \\
				\cmidrule{2-7}
				~ & \multirow{2}*{4.7} & \multirow{2}*{$\nu(t)$} & 0.01 & $[0.0002, 0.0209]$ & $0.0103$ & +2.8796\% \\
				~ & ~ & ~ & 0.08 & $[0.0698, 0.0909]$ & $0.0802$ & +0.3077\% \\
				\cmidrule{2-7}
				~ & \multirow{2}*{4.8} & \multirow{2}*{$\nu(t)$} & 0.01 & $[0.0011, 0.0218]$ & $0.0105$ & +4.7731\% \\
				~ & ~ & ~ & 0.1 & $[0.0891, 0.1093]$ & $0.0996$ & $-$0.4467\% \\
				\midrule
				
				\multirow{2}*{Helmholtz} & \multirow{2}*{5.1} & \multirow{2}*{$k(x,y)$} & 13.3333 & $[13.3333, 13.5019]$ & $13.3333$ & +0.0001\% \\
				~ & ~ & ~ & 20 & $[19.6164, 20.0000]$ & $19.9734$ & $-$0.1329\% \\
				\bottomrule
		\end{tabular}}
	\end{table}
	
	\begin{table}[p]
		\centering
		\caption{Change-point detection results for 1D temporal coefficients of sub network in two-stage PINNs framework. \label{tab2}}
		\scalebox{1}{
			\setlength{\tabcolsep}{2.5pt}
			\begin{tabular}{cccccccc}
				\toprule
				Equation & Case & Parameter & Change Point & True & Stage~1 & Stage~2 & Relative Error \\
				\midrule
				Wave & 1.2 & $\alpha(t)$ & $\tau_1$ & 5 & $[4.7750, 5.3950]$ & $5.0413$ & +0.8257\% \\
				\midrule
				
				\multirow{5}*{Heat} 
				& 2.1 & $c(t)$ & $\tau_1$ & 5 & $[4.6400, 5.3300]$ & $4.9846$ & $-$0.3074\% \\
				\cmidrule(lr){2-8}
				~ & \multirow{4}*{2.2} & \multirow{4}*{$c(t)$} & $\tau_1$ & 2 & $[1.6700, 2.3600]$ & $2.0416$ & +2.0781\% \\
				~ & ~ & ~ & $\tau_2$ & 4 & $[3.8000, 4.4900]$ & $4.0787$ & +1.9678\% \\
				~ & ~ & ~ & $\tau_3$ & 6 & $[5.6700, 6.3500]$ & $5.9211$ & $-$1.3145\% \\
				~ & ~ & ~ & $\tau_4$ & 8 & $[7.6300, 8.3300]$ & $7.9585$ & $-$0.5187\% \\
				\midrule
				
				\multirow{13}*{Burgers} 
				& 3.2 & $\lambda_1(t)$ & $\tau_1$ & 5 & $[4.4706, 5.3725]$ & $4.9995$ & $-$0.0093\% \\
				\cmidrule(lr){2-8}
				~ & \multirow{4}*{3.3} & \multirow{4}*{$\lambda_1(t)$} & $\tau_1$ & 2 & $[1.6471, 2.3529]$ & $2.0002$ & +0.0099\% \\
				~ & ~ & ~ & $\tau_2$ & 4 & $[3.6078, 4.3922]$ & $3.9838$ & $-$0.4054\% \\
				~ & ~ & ~ & $\tau_3$ & 6 & $[5.5686, 6.3922]$ & $6.0021$ & +0.0345\% \\
				~ & ~ & ~ & $\tau_4$ & 8 & $[7.6078, 8.3922]$ & $8.0029$ & +0.0368\% \\
				\cmidrule(lr){2-8}
				~ & \multirow{8}*{3.4} & \multirow{4}*{$\lambda_1(t)$} & $\tau_1$ & 2 & $[1.6863, 2.4314]$ & $2.0392$ & +1.9608\% \\
				~ & ~ & ~ & $\tau_2$ & 4 & $[3.4510, 4.3529]$ & $4.0392$ & +0.9804\% \\
				~ & ~ & ~ & $\tau_3$ & 6 & $[5.6863, 6.3922]$ & $5.9749$ & $-$0.4179\% \\
				~ & ~ & ~ & $\tau_4$ & 8 & $[7.6863, 8.5490]$ & $8.0784$ & +0.9804\% \\
				\cmidrule(lr){3-8}
				~ & ~ & \multirow{4}*{$\lambda_2(t)$} & $\tau_1$ & 2 & $[1.6471, 2.3529]$ & $2.0105$ & +0.5236\% \\
				~ & ~ & ~ & $\tau_2$ & 4 & $[3.5686, 4.4314]$ & $4.0184$ & +0.4612\% \\
				~ & ~ & ~ & $\tau_3$ & 6 & $[5.6471, 6.3529]$ & $5.9986$ & $-$0.0242\% \\
				~ & ~ & ~ & $\tau_4$ & 8 & $[7.5686, 8.3137]$ & $7.9827$ & $-$0.2159\% \\
				\midrule
				
				\multirow{7}*{Navier-Stokes} 
				& 4.2 & $\nu(t)$ & $\tau_1$ & 0.5 & $[0.4492, 0.5508]$ & $0.4998$ & $-$0.0489\% \\
				\cmidrule(lr){2-8}
				~ & 4.3 & $\nu(t)$ & $\tau_1$ & 0.5 & $[0.4570, 0.5430]$ & $0.5049$ & +0.9883\% \\
				\cmidrule(lr){2-8}
				~ & 4.4 & $\nu(t)$ & $\tau_1$ & 0.5 & $[0.4648, 0.5430]$ & $0.5040$ & +0.8037\% \\
				\cmidrule(lr){2-8}
				~ & 4.5 & $\nu(t)$ & $\tau_1$ & 0.5 & $[0.4727, 0.5430]$ & $0.5056$ & +1.1242\% \\
				\cmidrule(lr){2-8}
				~ & 4.6 & $\nu(t)$ & $\tau_1$ & 0.5 & $[0.4727, 0.5430]$ & $0.5060$ & +1.2078\% \\
				\cmidrule(lr){2-8}
				~ & 4.7 & $\nu(t)$ & $\tau_1$ & 0.5 & $[0.4805, 0.5273]$ & $0.5050$ & +0.9965\% \\
				\cmidrule(lr){2-8}
				~ & 4.8 & $\nu(t)$ & $\tau_1$ & 0.5 & $[0.4805, 0.5273]$ & $0.5047$ & +0.9405\% \\
				\bottomrule
		\end{tabular}}
		
		\vspace{12pt}
		
		\caption{Jump boundary detection MSE results for 2D spatial coefficients of sub network in CCD-PINNs. \label{tab3}}
		\scalebox{0.9}{
			\setlength{\tabcolsep}{2.5pt}
			\begin{tabular}{@{}cccc@{\hspace{6pt}}cccc@{}}
				\toprule
				Equation & Case & Parameter & MSE & Equation & Case & Parameter & MSE \\
				\midrule
				Wave & 1.3 & $\alpha(x,y)$ & 1.2592e-4 & Helmholtz & 5.1 & $k(x,y)$ & 9.3683e-5 \\
				\bottomrule
		\end{tabular}}
		
		\vspace{12pt}
		
		\caption{MSE results of PDEs solution approximation of main network in two-stage PINNs framework. \label{tab4}}
		\scalebox{1}{
			\setlength{\tabcolsep}{2.5pt}
			\begin{tabular}{@{}cccc@{\hspace{6pt}}cccc@{}}
				\toprule
				Equation & Case & Stage~1 & Stage~2 & Equation & Case & Stage~1 & Stage~2 \\
				\midrule
				Wave & 1.1 & 1.3684e-6 & 1.3482e-6 & Burgers & 3.6 & 5.6432e-10 & -- \\
				Wave & 1.2 & 9.3052e-6 & 5.1713e-6 & Navier-Stokes & 4.1 & 1.7214e-5 & 1.5892e-5 \\
				Wave & 1.3 & 1.1071e-4 & 1.0886e-4 & Navier-Stokes & 4.2 & 2.1269e-5 & 2.1226e-5 \\
				Heat & 2.1 & 6.4613e-6 & 6.3578e-6 & Navier-Stokes & 4.3 & 2.9634e-5 & 2.0634e-5 \\
				Heat & 2.2 & 1.8888e-6 & 1.7903e-6 & Navier-Stokes & 4.4 & 3.4843e-5 & 2.3094e-5 \\
				Burgers & 3.1 & 1.3476e-8 & 4.8732e-9 & Navier-Stokes & 4.5 & 4.2367e-5 & 2.6175e-5 \\
				Burgers & 3.2 & 7.8007e-10 & 7.5590e-10 & Navier-Stokes & 4.6 & 4.8325e-5 & 2.3883e-5 \\
				Burgers & 3.3 & 3.6277e-9 & 3.5863e-9 & Navier-Stokes & 4.7 & 5.4288e-5 & 3.0744e-5 \\
				Burgers & 3.4 & 4.0232e-8 & 3.9129e-8 & Navier-Stokes & 4.8 & 5.6403e-5 & 3.0631e-5 \\
				Burgers & 3.5 & 8.5891e-10 & -- & Helmholtz & 5.1 & 1.3884e-9 & 7.9232e-10 \\
				\bottomrule
		\end{tabular}}
		
	\end{table}
	
	\begin{table}[t]
		\centering
		\caption{Sub-network approximation error of GWS-PINNs for continuous time-varying coefficients in Burgers equations. \label{tab5}}
		\scalebox{1}{
			\setlength{\tabcolsep}{2.5pt}
			\begin{tabular}{@{}cccc@{\hspace{6pt}}cccc@{}}
				\toprule
				Equation & Case & Parameter & MSE & Equation & Case & Parameter & MSE \\
				\midrule
				\multirow{2}*{Burgers} & \multirow{2}*{3.5} & $\lambda_1(t)$ & 7.5088e-6 & \multirow{2}*{Burgers} & \multirow{2}*{3.6} & $\lambda_1(t)$ & 3.2288e-7  \\
				~ & ~ & $\lambda_2(t)$ & 8.3945e-8 & ~ & ~ & $\lambda_2(t)$ & 3.7786e-8  \\
				\bottomrule
		\end{tabular}}
		
		\vspace{12pt}
		
		\caption{Computational time of GWS-PINNs, GMM-BDMC and CCD-PINNs in discontinuous-coefficient PDE cases. \label{tab6}}
		\setlength{\tabcolsep}{2.5pt}
		\scalebox{1}{
			\begin{tabular}{cccccc}
				\toprule
				Equation & Case & GWS-PINNs & GMM-BDMC & CCD-PINNs & Total \\
				\midrule
				\multirow{3}*{Wave}
				& 1.1 & 2417.4 & 10.1 & 98.6 & 2526.1 \\
				\cmidrule(lr){2-6}
				~ & 1.2 & 2481.7 & 19.5 & 103.8 & 2605.1 \\
				\cmidrule(lr){2-6}
				~ & 1.3 & 12600.3 & 37.2 & 105.4 & 12742.9 \\
				\midrule
				\multirow{2}*{Heat}
				& 2.1 & 3320.0 & 20.2 & 175.7 & 3515.9 \\
				\cmidrule(lr){2-6}
				~ & 2.2 & 13265.0 & 26.2 & 181.4 & 13472.6 \\
				\midrule
				\multirow{4}*{Burgers}
				& 3.1 & 8064.8 & 19.8 & 107.8 & 8192.4 \\
				\cmidrule(lr){2-6}
				~ & 3.2 & 8082.1 & 20.4 & 99.6 & 8202.1 \\
				\cmidrule(lr){2-6}
				~ & 3.3 & 8076.4 & 20.2 & 101.1 & 8197.7 \\
				\cmidrule(lr){2-6}
				~ & 3.4 & 8108.0 & 20.2 & 102.0 & 8230.2 \\
				\midrule
				\multirow{8}*{Navier-Stokes}
				& 4.1 & 4160.3 & 17.7 & 704.0 & 4882.0 \\
				\cmidrule(lr){2-6}
				~ & 4.2 & 4255.1 & 17.5 & 725.6 & 4998.2 \\
				\cmidrule(lr){2-6}
				~ & 4.3 & 4209.3 & 17.7 & 674.6 & 4901.6 \\
				\cmidrule(lr){2-6}
				~ & 4.4 & 4290.4 & 17.6 & 694.2 & 5002.2 \\
				\cmidrule(lr){2-6}
				~ & 4.5 & 4269.1 & 17.9 & 697.6 & 4984.5 \\
				\cmidrule(lr){2-6}
				~ & 4.6 & 4194.9 & 17.6 & 755.2 & 4967.7 \\
				\cmidrule(lr){2-6}
				~ & 4.7 & 4246.9 & 18.0 & 858.3 & 5123.2 \\
				\cmidrule(lr){2-6}
				~ & 4.8 & 4194.0 & 17.8 & 614.5 & 4826.3 \\
				\midrule
				Helmholtz & 5.1 & 7581.1 & 84.2 & 941.7 & 8606.9 \\
				\bottomrule
		\end{tabular}}
	\end{table}
	
	Furthermore, Table~\ref{tab6} reports the computational time of the three main components of JVC-PINNs, including the first-stage GWS-PINNs sampler, the GMM-BDMC statistical learner, the second-stage CCD-PINNs estimator, and the total time computed using the unrounded values. The results show that, although JVC-PINNs adopts a two-stage framework with both solution and coefficient networks, its total computational cost remains comparable to that of a single-stage neural-network solver. Most of the computational cost is incurred during the first-stage joint physics-informed training, while GMM-BDMC incurs only a small overhead since it operates on coefficient samples rather than the full PDE solution space. The second-stage CCD-PINNs is also efficient because it uses transfer learning from the trained GWS-PINNs solution network and restricts the optimization of coefficients and jump locations to the search intervals and candidate regions inferred by GMM-BDMC. Therefore, the proposed framework improves coefficient identification and discontinuity localization with only moderate additional computational cost, supporting its practical applicability to PDE inverse problems with jump-varying coefficients.
	
	In summary, the core idea of this framework is to first use the PINNs with a coefficient sub network and gradient adaptive weighting to learn a smooth approximation of a jump-varying coefficient, then apply GMM and BDMC to automatically identify the discrete coefficient states and candidate jump regions, and finally build the constrained piecewise-constant PINNs to refine the coefficient values and discontinuity locations. The main research value lies in the integrated framework that combines PINNs based sampling, Bayesian mixture model detection, and constrained PINNs refinement, enabling accurate identification and reconstruction of unknown jump coefficients in PDEs while avoiding the introduction of erroneous discontinuities in parameters. The following sections and experiments will further illustrate the framework’s performance and other applications.
	
	\section{Ablation Study}\label{sec4}
	This section evaluates the contribution of each component in the proposed two-stage framework. Starting from standard PINNs, the JVC-PINNs framework progressively adds the coefficient sub network, the gradient-adaptive residual weighting strategy, the GMM-BDMC regime extractor, the classification-based refinement mechanism, and the second-stage constrained parameter refinement. In addition, a direct Stage~2 variant is included to examine whether the statistical information provided by Stage~1 is necessary for stable piecewise-constant reconstruction.
	
	The ablation groups are defined as follows. A0 corresponds to the standard PINNs, where the unknown coefficient is treated as a global trainable constant. This group serves as the baseline for parameter inversion and illustrates the limitation of standard PINNs when the true coefficient exhibits jump discontinuities. A1 augments A0 with the coefficient sub network $\hat{\theta}_p(\mathbf{x},t)$, allowing the model to learn a continuous spatiotemporally varying coefficient field. Comparing A1 with A0 isolates the effect of replacing a global constant parameter with a learnable coefficient function. A2 further introduces the gradient-adaptive residual weighting strategy in the physics loss. Comparing A2 with A1 isolates the contribution of stabilizing coefficient sampling and reducing the influence of high-gradient transition regions. A3 incorporates the GMM-BDMC statistical learner and the CCD-PINNs refinement stage, but removes the classification loss $\mathcal{L}_{\mathrm{cec}}$. The coefficient values are refined using the heuristic search intervals $\{\mathcal{I}_k\}$ and candidate discontinuity regions $\{\mathcal{X}_l\}$ inferred from Stage~1, while the final piecewise-constant reconstruction is obtained without classification-based sharpening. Comparing A3 with A2 evaluates the contribution of the statistical-learning-guided constrained refinement framework. A4 corresponds to the complete proposed JVC-PINNs framework. In addition to all components in A3, the classification loss is activated to refine the region-to-state assignment and sharpen discontinuity boundaries. Comparing A4 with A3 directly quantifies the contribution of the classification-based refinement mechanism and evaluates whether explicit discontinuity classification improves coefficient reconstruction and solution accuracy. A5 uses only the CCD-PINNs stage and does not use any information obtained from Stage~1. The piecewise-constant estimator is trained with known $K$ directly without Stage~1 initialization, heuristic coefficient search intervals, or candidate discontinuity regions. This group tests whether the first-stage GWS-PINNs and GMM-BDMC analysis constitute an essential preprocessing step rather than merely an optional initialization strategy.
	
	The numerical results are summarized in Table~\ref{tab7} and the symbol $^{\star}$ in tables denotes the method used in JVC-PINNs configuration proposed in this study. To ensure a fair comparison, all ablation groups are evaluated under the same experimental setting in Section~\ref{sec3}. Specifically, the observation data, initial and boundary data, residual-point budget, neural-network architecture, optimizer, learning-rate schedule, and training epochs are kept identical for each ablation case. Therefore, the differences in Table~\ref{tab7} mainly reflect the effect of the removed or added algorithmic component rather than changes in data density or training configuration.
	
	Compared with A0, A1 significantly reduces the solution MSE in all tested cases. This confirms that representing the unknown coefficient by a sub network is essential for PDEs with jump-varying parameters. Since A0 can only learn a single global constant, it fails to describe the regime-dependent coefficient structure and therefore produces biased solution approximations. By contrast, A1 provides a continuous coefficient surrogate over the spatiotemporal domain, which substantially improves the flexibility of the inverse model. The comparison between A1 and A2 shows the effect of the gradient-adaptive residual weighting strategy. In the heat, Burgers, Navier-Stokes, and Helmholtz cases, A2 reduces the solution MSE relative to A1. In the wave case, however, A2 produces slightly higher MSEs. These results indicate that the benefit of gradient-adaptive weighting is problem dependent and may not translate directly into a lower solution MSE for every PDE. Overall, A2 provides a more reliable Stage~1 sampler for subsequent statistical learning.
	
	A3 further introduces the GMM-BDMC statistical learner and the second-stage constrained refinement, but without the classification loss. Compared with A2, A3 improves the MSE in most cases, especially for the Navier-Stokes and Helmholtz equations. This shows that replacing the smooth coefficient surrogate with a constrained piecewise-constant representation is beneficial for discontinuous coefficient inverse problems. The statistical learner provides coefficient search intervals and candidate jump regions, which reduce the search space of the second-stage inverse problem and improve the physical consistency of the recovered coefficient. A4 corresponds to the complete JVC-PINNs framework. It achieves the best or nearly best MSE in all tested cases. The improvement from A3 to A4 demonstrates the contribution of the classification-based refinement mechanism. By activating the classification loss, the model sharpens the region-to-state assignment and improves the localization of discontinuity boundaries. This effect is particularly important when the Stage~1 coefficient surrogate contains smoothed transition zones, because classification refinement converts these soft transitions into a hard piecewise-constant structure.
	
	\begin{table}[t]
		\centering
		\caption{Ablation results of neural-network components on different PDE cases about MSE and computational time. \label{tab7}}
		\scalebox{1}{
			\begin{tabular}{cccccccc}
				\toprule
				MSE & Case & A0 & A1 & A2 & A3 & A4$^{\star}$ & A5 \\
				\midrule
				Wave & 1.2 & 1.0290e-4 & 6.2964e-6 & 9.3052e-6 & 6.0437e-6 & 5.1713e-6 & 1.5737e-3 \\
				Heat & 2.1 & 4.9993e-5 & 6.6846e-6 & 6.4613e-6 & 6.7663e-6 & 6.3578e-6 & 7.2919e-6 \\
				Burgers & 3.2 & 2.0570e-5 & 5.2924e-9 & 7.8007e-10 & 8.3660e-10 & 7.5590e-10 & 8.6359e-8 \\
				Navier-Stokes & 4.8 & 6.2474e-4 & 2.8990e-4 & 5.6403e-5 & 3.5349e-5 & 3.0631e-5 & 2.6029e-4 \\
				Helmholtz & 5.1 & 3.3242e-8 & 2.9684e-9 & 1.3884e-9 & 9.3906e-10 & 7.9232e-10 & 1.8144e-8 \\
				\midrule
				Time/s & Case & A0 & A1 & A2 & A3 & A4$^{\star}$ & A5 \\
				\midrule
				Wave & 1.2 & 2306.6 & 2373.7 & 2481.7 & 2569.8 & 2605.1 & 2320.0 \\
				Heat & 2.1 & 2560.0 & 2960.0 & 3320.0 & 3382.2 & 3515.9 & 2965.8 \\
				Burgers & 3.2 & 7347.6 & 7825.0 & 8082.1 & 8184.0 & 8202.1 & 7706.2 \\
				Navier-Stokes & 4.8 & 3928.9 & 4100.2 & 4194.0 & 4746.9 & 4826.3 & 3948.0 \\
				Helmholtz & 5.1 & 6910.7 & 7089.3 & 7581.1 & 8447.0 & 8606.9 & 7004.2 \\
				\bottomrule
		\end{tabular}}
		
		\vspace{12pt}
		
		\caption{Sensitivity to the numbers of observation data and physics-residual points for Burgers Case~3.2.}
		\label{tab8}
		\scalebox{1}{
			\begin{tabular}{cccccccc}
				\toprule
				Equation & Case & Data Points & Residual Points & MSE & $\lambda_1$ & $\lambda_2$ & $\tau_1$  \\
				\midrule
				Burgers & 3.2 & $2000$  & $4000$ & $1.6248\mathrm{e}{-9}$  & $0.4993/1.0006$ & $0.0992$ & $4.9979$ \\
				Burgers & 3.2 & $4000^{\star}$ & $4000^{\star}$ & $7.5590\mathrm{e}{-10}$ & $0.4998/0.9998$ & $0.1000$ & $4.9995$ \\
				Burgers & 3.2 & $10000$ & $4000$ & $4.9824\mathrm{e}{-10}$ & $0.4901/0.9984$ & $0.1000$ & $4.9996$ \\
				Burgers & 3.2 & $20000$ & $4000$ & $4.7316\mathrm{e}{-10}$ & $0.5233/1.0122$ & $0.1000$ & $4.9997$ \\
				\midrule
				Burgers & 3.2 & $4000$ & $2000$ & $1.0123\mathrm{e}{-9}$  & $0.4968/1.0032$ & $0.0991$ & $4.9864$ \\
				Burgers & 3.2 & $4000^{\star}$ & $4000^{\star}$ & $7.5590\mathrm{e}{-10}$ & $0.4998/0.9998$ & $0.1000$ & $4.9995$ \\
				Burgers & 3.2 & $4000$ & $10000$ & $7.6881\mathrm{e}{-10}$ & $0.4998/0.9998$ & $0.1000$ & $4.9995$ \\
				Burgers & 3.2 & $4000$ & $20000$ & $1.1126\mathrm{e}{-9}$ & $0.4996/0.9980$ & $0.1000$ & $4.9994$ \\
				\bottomrule
		\end{tabular}}
	\end{table}
	
	The A5 results further demonstrate the necessity of Stage~1 sampling and statistical guidance. Although A5 uses the CCD-PINNs formulation, it removes the Stage~1 initialization, coefficient search intervals, and candidate discontinuity regions. Its performance is consistently worse than A4 and can even be worse than A1 or A2 in complex cases. These results indicate that direct training of the piecewise-constant estimator without Stage~1 guidance is less reliable under the tested settings, because the optimizer must simultaneously search for the coefficient values and discontinuity locations over a substantially larger nonconvex space. Thus, Stage~1 serves as more than an initialization step. It provides structural information that is important for reliable constrained refinement in the tested cases.
	
	The computational time in Table~\ref{tab7} shows that adding components gradually increases the training cost. A1 is slightly more expensive than A0 due to the coefficient sub network, A2 introduces additional cost from gradient computation in the adaptive weights, and A3-A4 require extra statistical inference and second-stage refinement. However, the increase in computational cost is moderate compared with the gain in accuracy and interpretability. In particular, A4 only requires a small additional cost over A3, but consistently improves the final reconstruction. This indicates that the classification-based refinement is computationally efficient.
	
	Table~\ref{tab8} further reports the sensitivity of JVC-PINNs to the numbers of observation and residual points using Burgers Case~3.2. The baseline setting, marked by $\star$, uses $4000$ data points and $4000$ residual points. When the number of data points is reduced to $2000$, the MSE increases and the coefficient estimates become slightly less accurate, indicating that insufficient observations weaken the data constraint. Increasing the number of data points to $10000$ or $20000$ further reduces the solution MSE, but the improvement in parameter estimation is not monotonic, which suggests that simply adding observation points does not always lead to better coefficient recovery once the data constraint is already sufficient.
	
	A similar phenomenon is observed for residual points. Reducing the number of residual points to $2000$ leads to a less accurate change-point estimate, while increasing them to $10000$ maintains stable parameter and change-point recovery. However, further increasing the number to $20000$ does not produce additional improvement and may slightly increase the MSE. This indicates that excessive residual points are not necessarily beneficial, because the optimization balance between data fitting and physics residuals may change. Overall, among the tested configurations, the baseline setting provides a reasonable balance among solution accuracy, coefficient recovery, change-point localization, and computational cost.

    Moreover, to examine the sensitivity of candidate-region construction to the interval width, this section generalizes the coefficient-state interval in equation~\eqref{eq:param_interval} as
    \begin{equation}
        \mathcal{I}_k(\rho)=\left[\widehat{\mu}_k-\rho\widehat{\sigma}_k,\: \widehat{\mu}_k+\rho\widehat{\sigma}_k\right],\quad \rho>0, \label{eq:param_interval_rho}
    \end{equation}
    where $\rho=1$ corresponds to the setting used in the preceding experiments. The interval width must balance the inclusion of the true coefficient levels in the feasible parameter ranges against the separation of adjacent coefficient states. When the scaling factor is reduced to $\rho=0.5$, the resulting intervals no longer contain the true coefficient values in Navier-Stokes Case~4.1 and Helmholtz Case~5.1. Because the Stage~2 coefficient estimates are strictly constrained to remain within these intervals, excessively narrow intervals may exclude the target values from the feasible sets and consequently prevent CCD-PINNs from recovering the corresponding coefficient states.
    
    Conversely, excessively large values of $\rho$ may cause intervals associated with adjacent states to overlap, reducing the coefficient-space gaps used to identify candidate transition samples. For example, the two spatial coefficient intervals in Wave Case~1.3 overlap at $\rho=2$, while the intervals associated with the temporal coefficients in Heat Case~2.2 overlap at $\rho=2.5$. Such overlap reduces, and may eliminate, the coefficient-space gap between adjacent states. Consequently, fewer transition samples may fall outside the union of the state intervals, weakening or even removing the candidate-region information supplied to Stage~2.

    For the cases examined here, these results support $\rho=1$ as a practical baseline because the resulting intervals contain the reference coefficient values while remaining sufficiently separated to produce nonempty candidate transition regions. This empirical choice is not claimed to be statistically optimal or universally applicable. More adaptive alternatives, such as data-dependent scaling factors or component-specific interval widths, may improve candidate region construction in other settings and warrant further investigation.
    
	These ablation results demonstrate the contribution of the main components of JVC-PINNs under the tested settings. The coefficient sub network substantially improves representational capacity relative to a globally constant parameter. The gradient-adaptive weighting strategy improves Stage~1 accuracy in several of the tested PDE cases. The GMM-BDMC module supplies coefficient-state and candidate-region information that reduces the search space of Stage~2, while the constrained refinement enforces an explicit piecewise-constant representation. Finally, the classification loss consistently reduces the solution MSE relative to A3. Overall, the complete A4 configuration achieves the lowest solution MSE across all five representative cases, with a moderate increase in computational cost.
	
	\section{Comparison with Existing Method}\label{sec5}
	This section compares the proposed JVC-PINNs framework with representative existing methods from three perspectives. First, comparing neural-network and Bayesian inverse methods for recovering PDE coefficients from observed solution data, with an emphasis on their abilities to approximate the solution and identify jump-varying parameters. Second, comparing statistical learning methods for extracting discrete coefficient regimes from the coefficient samples, focusing on state estimation and candidate transition-region detection. Third, analyzing the computational efficiency of different model-selection strategies, including the Akaike information criterion (AIC), the Bayesian information criterion (BIC), and the proposed GMM-BDMC procedure. These comparisons demonstrate that JVC-PINNs integrates accurate neural-network sampling, statistical regime identification, and constrained discontinuity refinement for PDE inverse problems with discontinuous coefficients.
	
	\begin{figure}[p]
		\centering
		\subfloat[SBL]{
			\begin{minipage}[t]{0.245\linewidth}
				\includegraphics[width=1\linewidth]{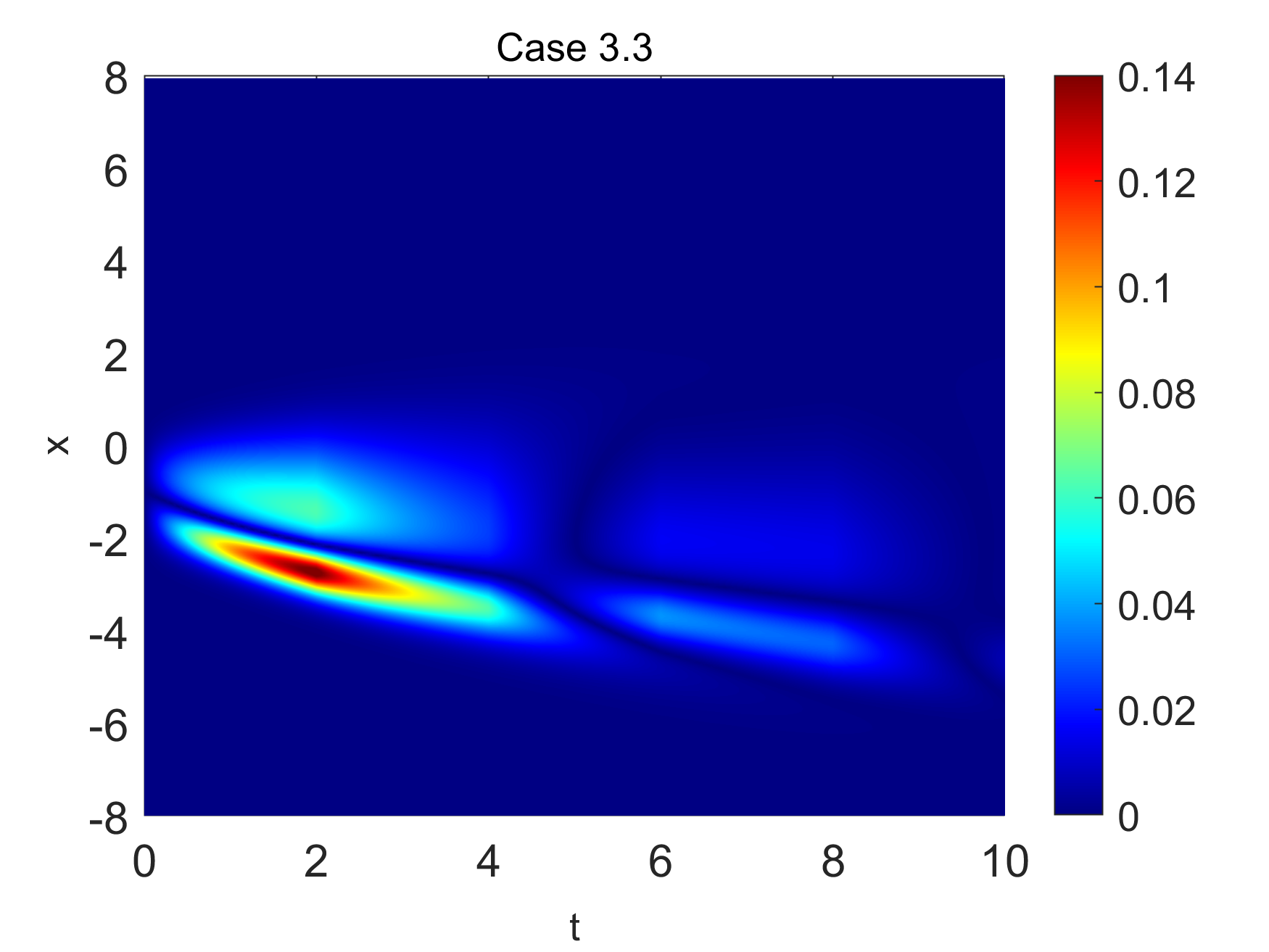}
			\end{minipage}
		}
		\subfloat[DSBL]{
			\begin{minipage}[t]{0.245\linewidth}
				\includegraphics[width=1\linewidth]{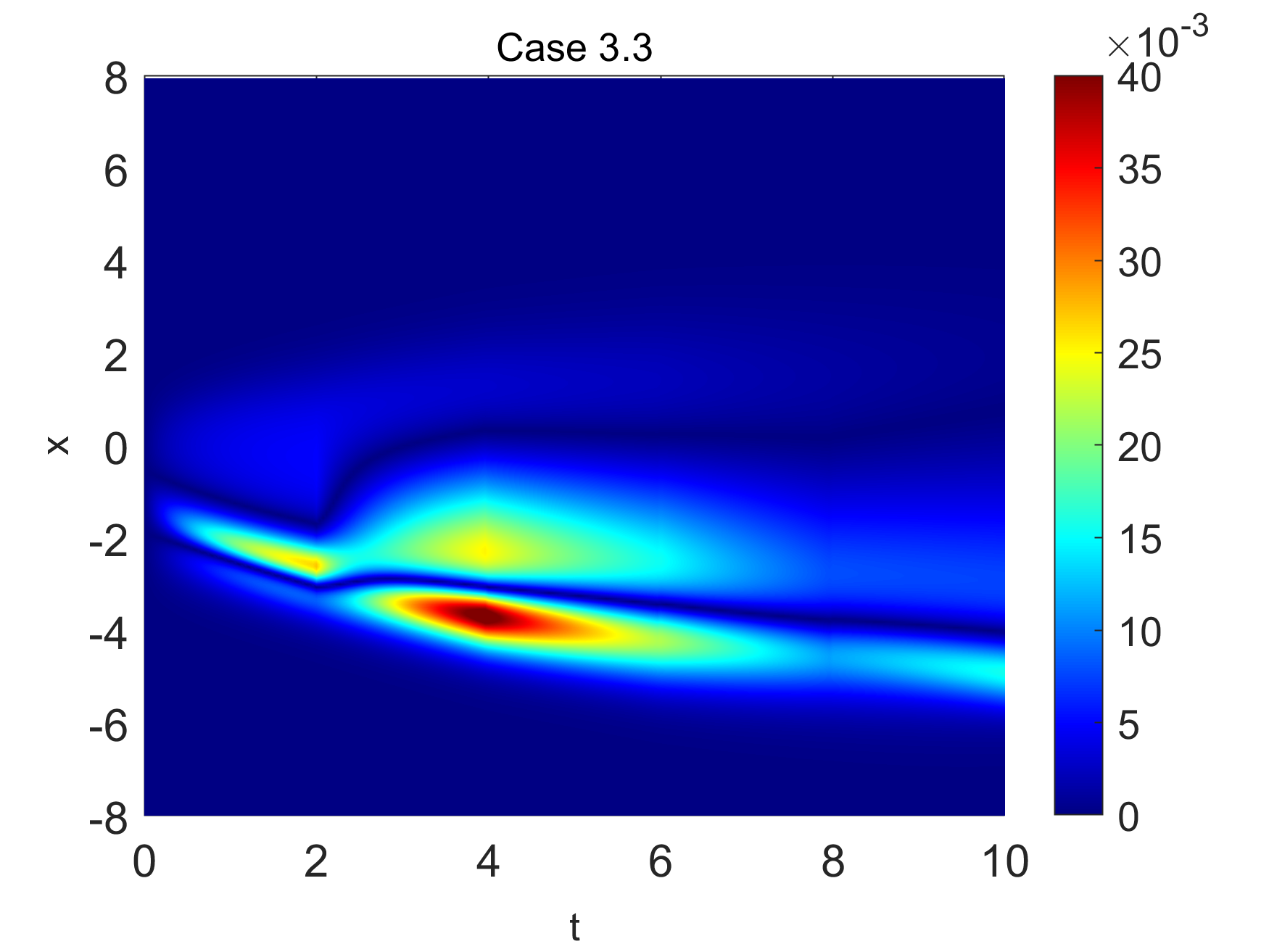}
			\end{minipage}
		}
		\subfloat[STD-PINNs]{
			\begin{minipage}[t]{0.245\linewidth}
				\includegraphics[width=1\linewidth]{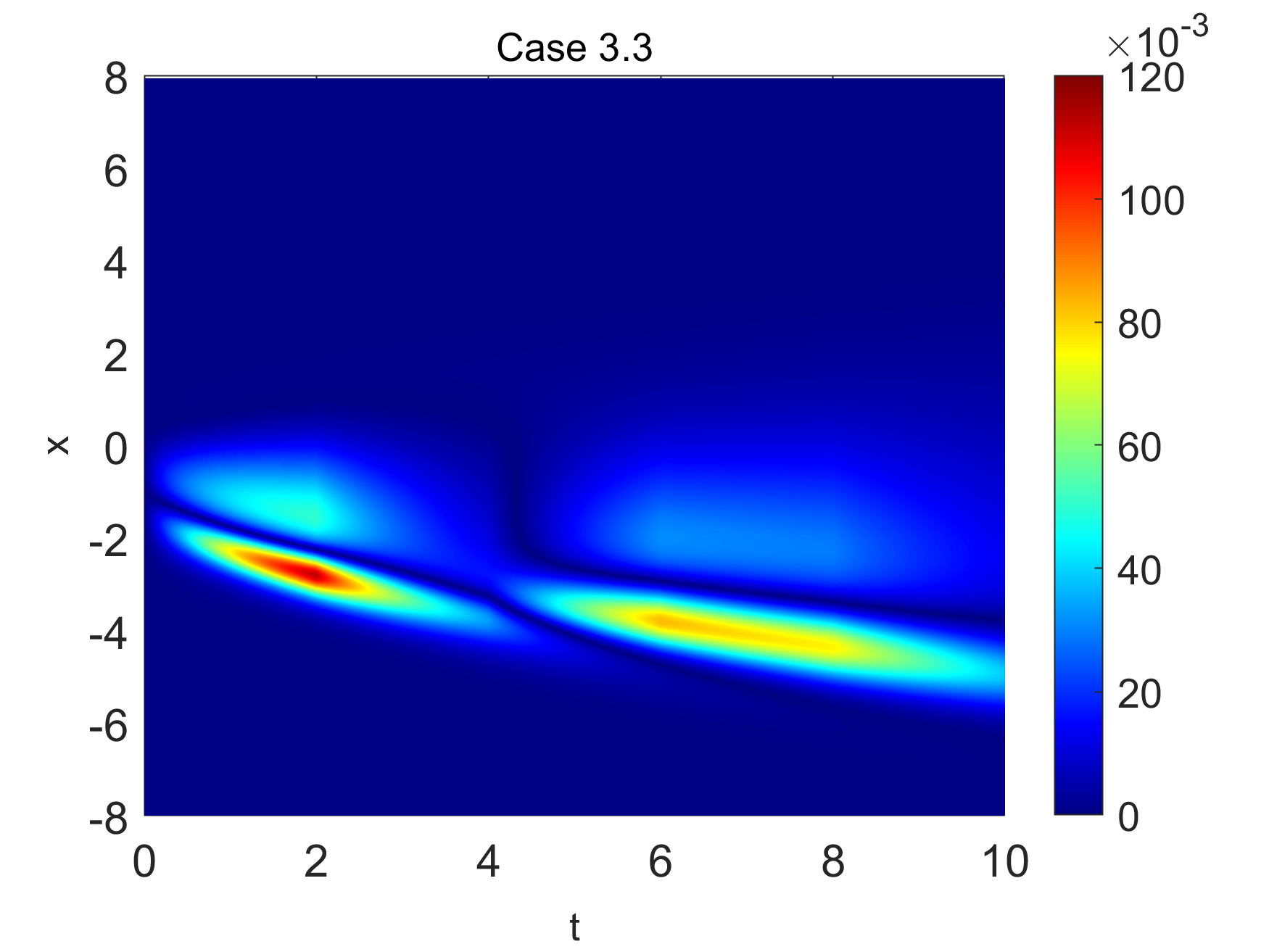}
			\end{minipage}
		}
		\subfloat[CP-PINNs]{
			\begin{minipage}[t]{0.245\linewidth}
				\includegraphics[width=1\linewidth]{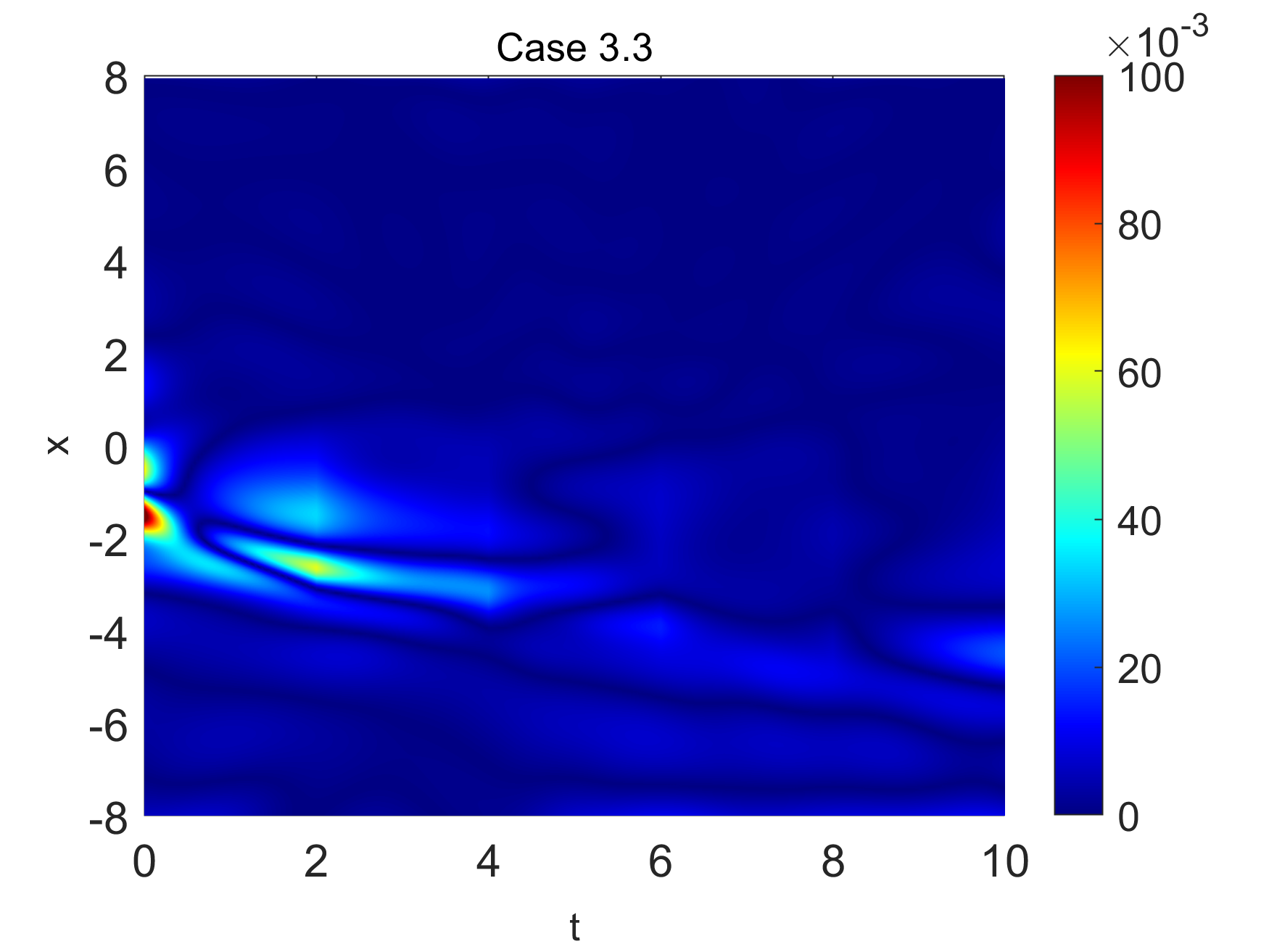}
			\end{minipage}
		}\\
		\subfloat[BC-PINNs]{
			\begin{minipage}[t]{0.245\linewidth}
				\includegraphics[width=1\linewidth]{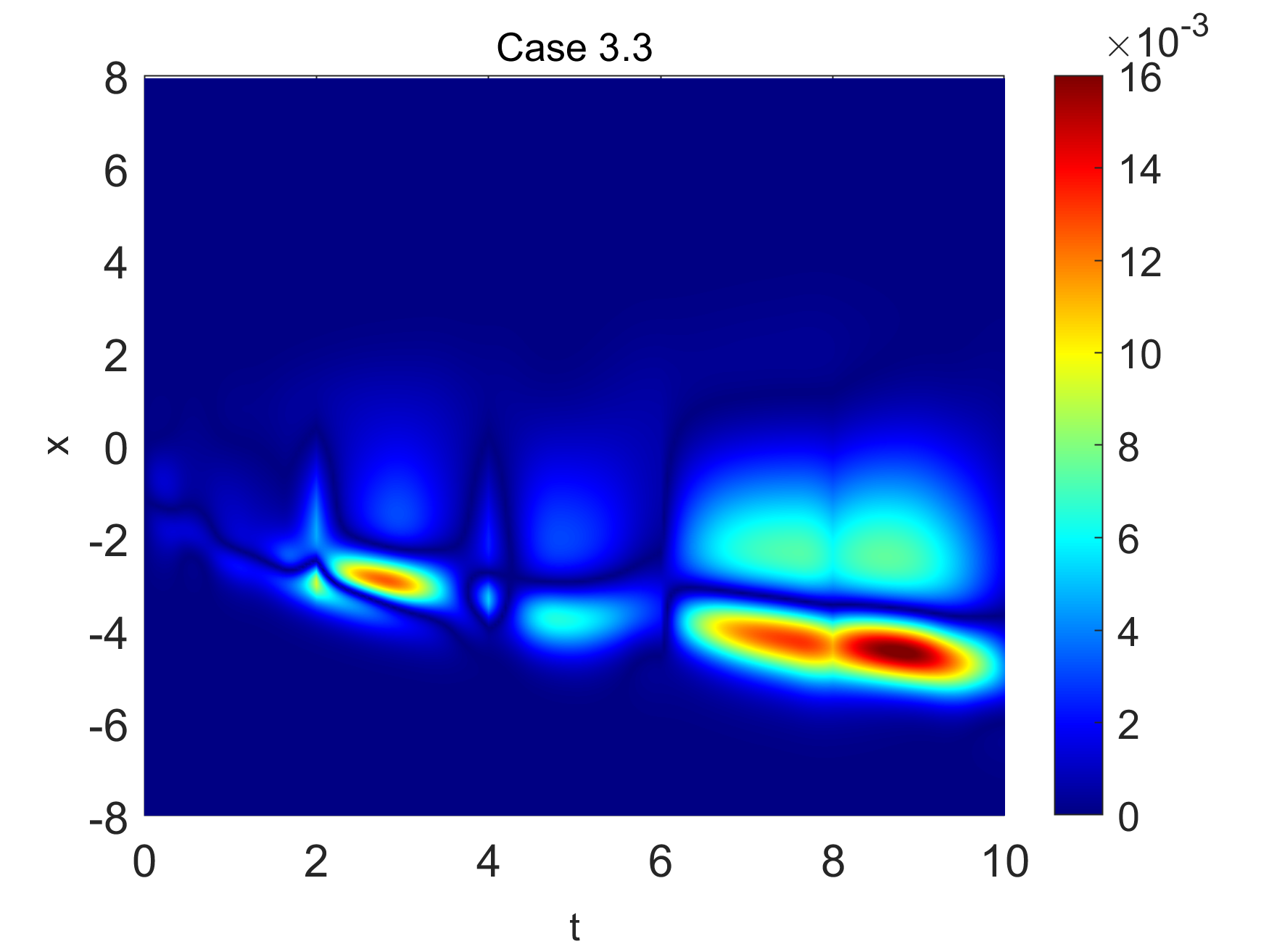}
			\end{minipage}
		}
		\subfloat[VC-PINNs]{
			\begin{minipage}[t]{0.245\linewidth}
				\includegraphics[width=1\linewidth]{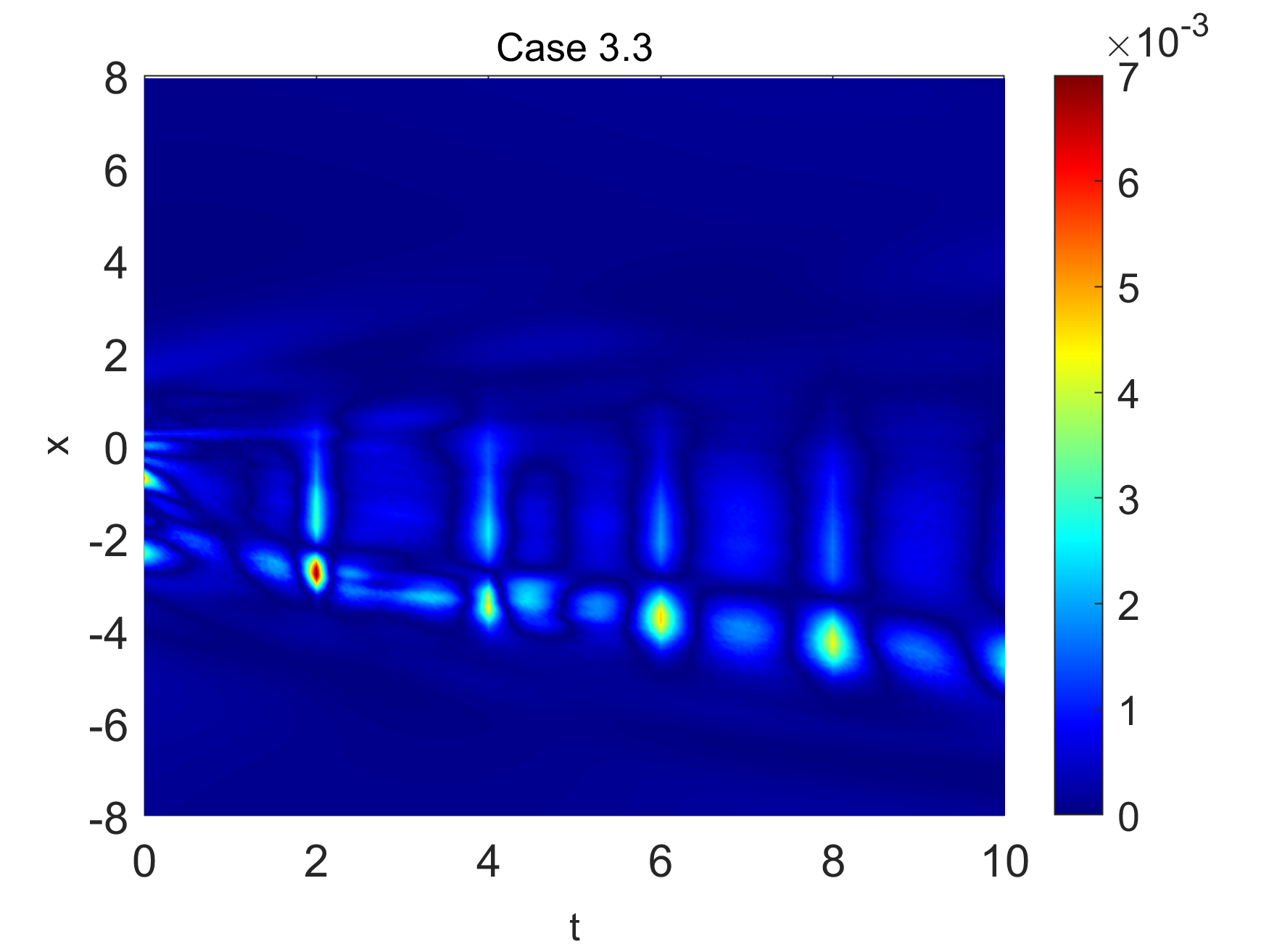}
			\end{minipage}
		}
		\subfloat[GWS-PINNs]{
			\begin{minipage}[t]{0.245\linewidth}
				\includegraphics[width=1\linewidth]{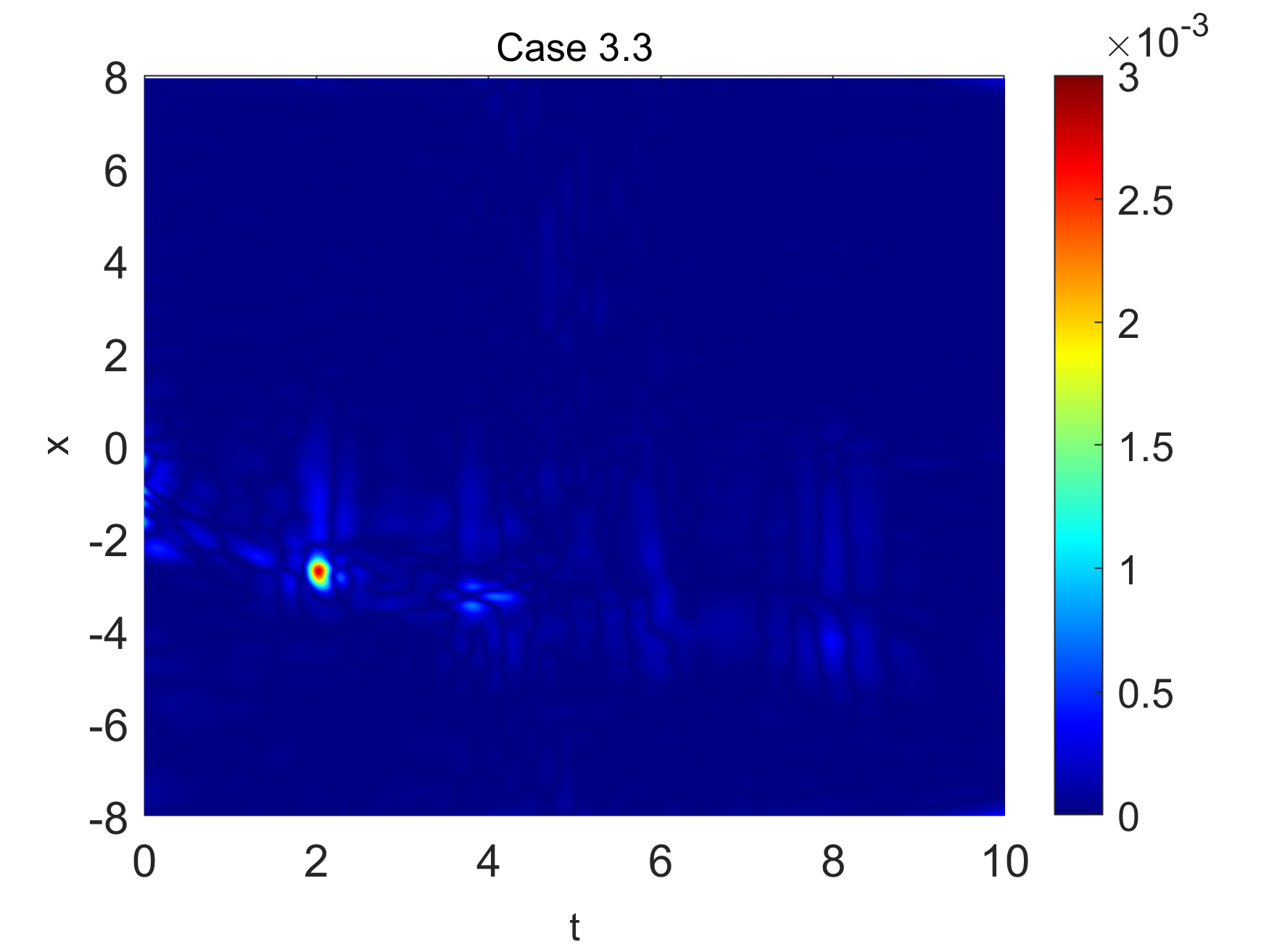}
			\end{minipage}
		}
		\subfloat[JVC-PINNs]{
			\begin{minipage}[t]{0.245\linewidth}
				\includegraphics[width=1\linewidth]{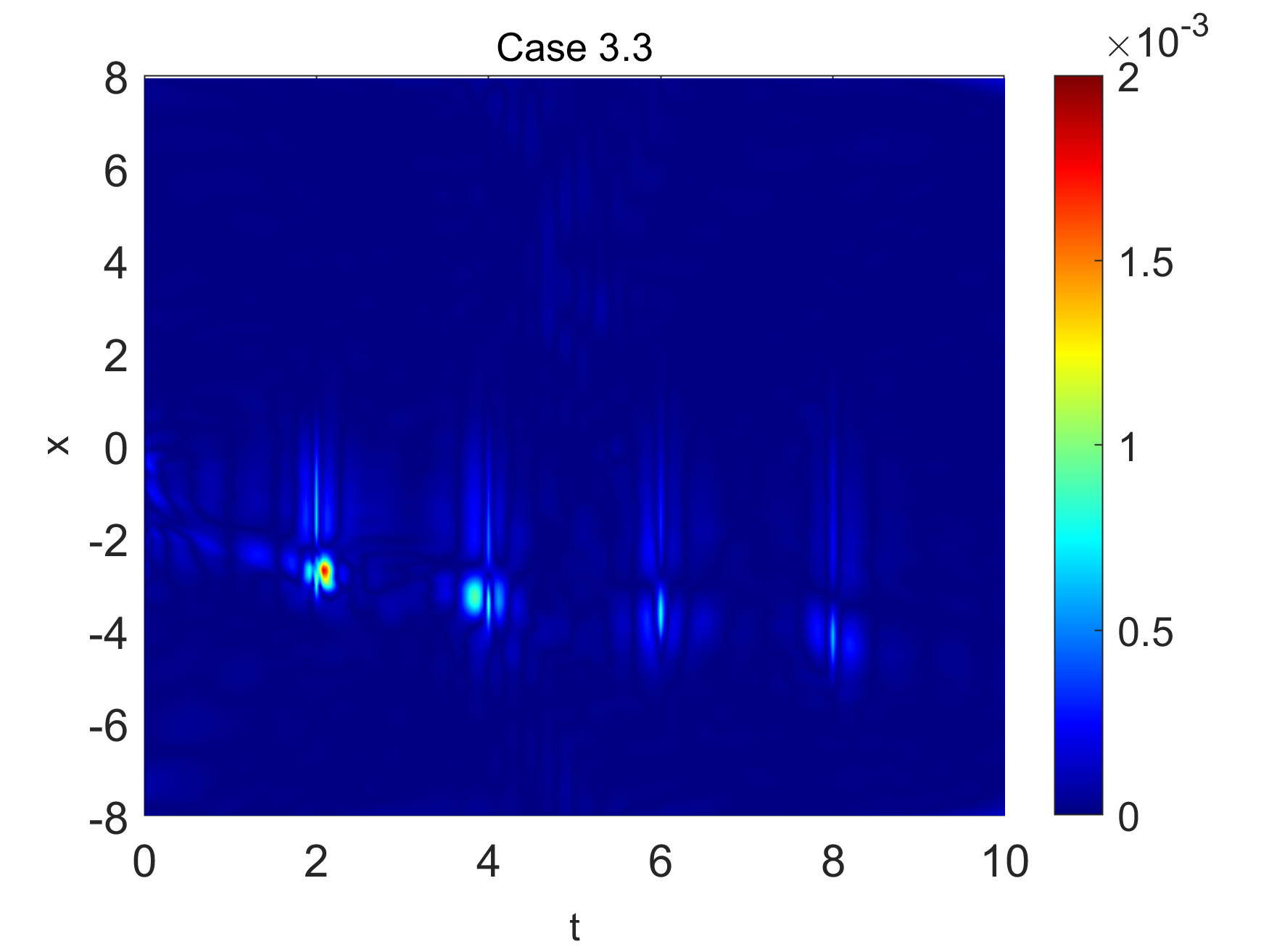}
			\end{minipage}
		}\\
		\caption{Comparison of absolute errors between predicted solution and reference solution by different inverse methods for time-varying coefficients of Burgers equation in Case~3.3. \label{fig11}}
		\subfloat[Parameter $\lambda_1$]{
			\begin{minipage}[t]{0.485\linewidth}
				\includegraphics[width=1\linewidth]{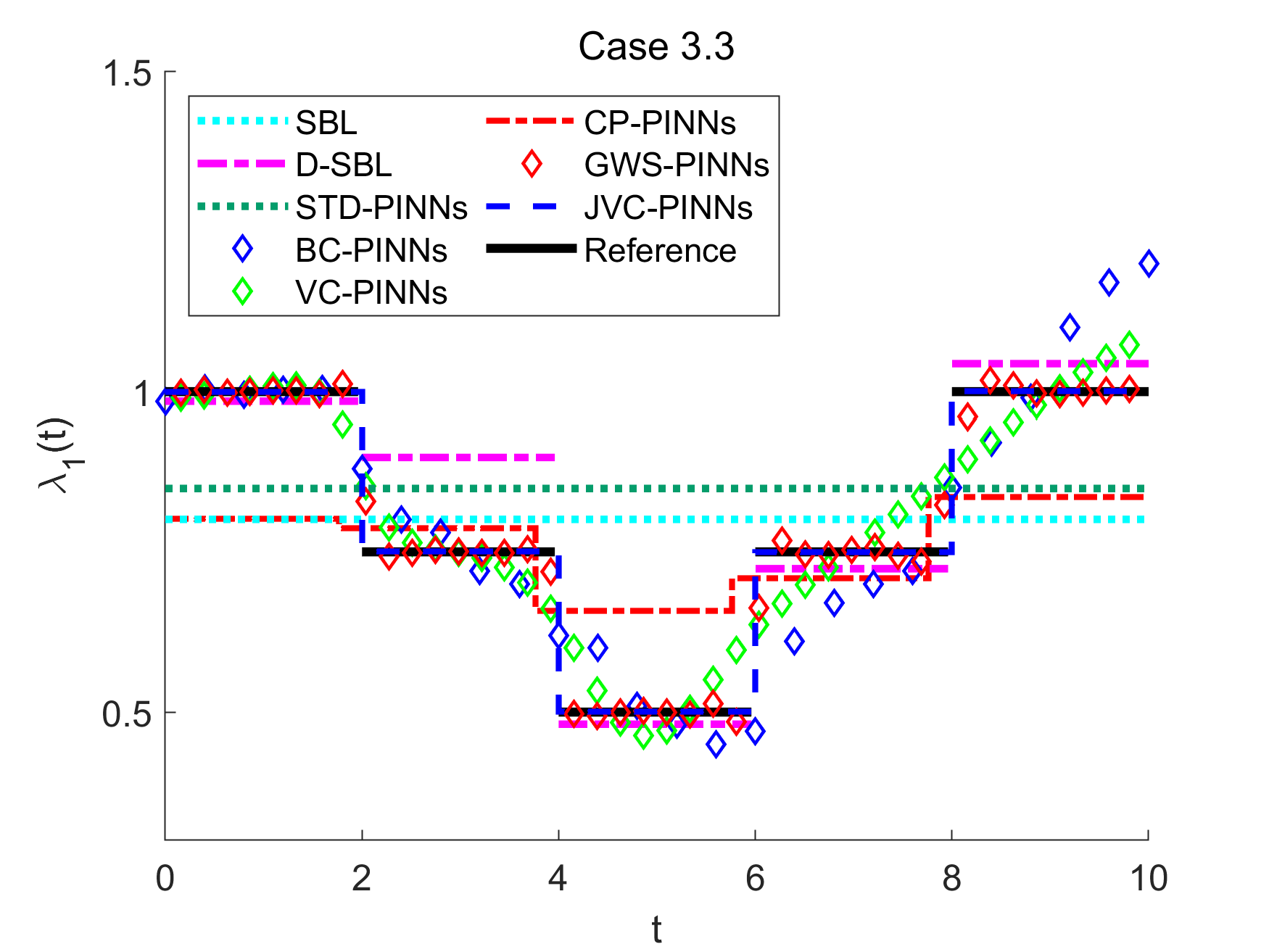}
			\end{minipage}
		}
		\subfloat[Parameter $\lambda_2$]{
			\begin{minipage}[t]{0.485\linewidth}
				\includegraphics[width=1\linewidth]{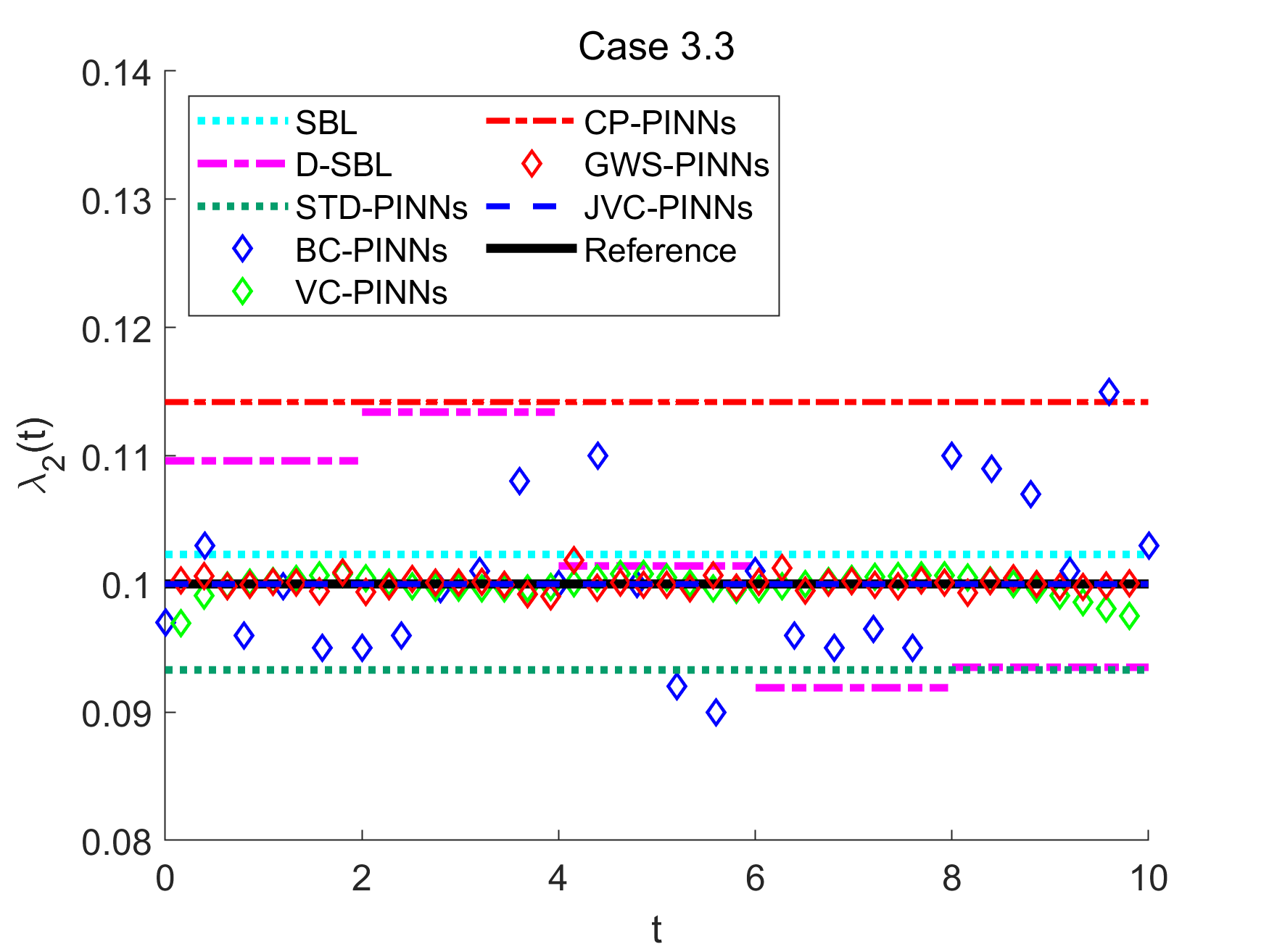}
			\end{minipage}
		}
		\caption{Comparison of JVC-PINNs sampling results for time-varying coefficients of Burgers equation in Case~3.3 with other methods. \label{fig12}}
	\end{figure}
	
	\subsection{Bayesian and Neural Network}
	
	First, this part compares the proposed framework with representative methods for PDE parameter inverse problems, including sparse Bayesian learning (SBL) \cite{yuan2023machine}, discrete sparse Bayesian learning (DSBL) \cite{li2023diffusion}, standard physics-informed neural networks (STD-PINNs) \cite{raissi2019physics}, change-point detection physics-informed neural networks (CP-PINNs) \cite{dong2024cp}, backward compatible physics-informed neural networks (BC-PINNs) \cite{mattey2022novel}, variable coefficient physics-informed neural networks (VC-PINNs) \cite{miao2023vc}, the first-stage GWS-PINNs sampler, and the complete JVC-PINNs framework proposed in this work.
    
    To ensure a fair comparison, all methods use observations generated from the same reference solution. SBL and DSBL use the same number of available observations. For the PINNs-based methods, the numbers of observation, initial, boundary, and physics-residual points are kept fixed. The network architectures, optimizer settings, learning rates, and training budgets are also kept consistent whenever the corresponding methods permit the same configuration. Thus, the observed differences primarily reflect the algorithmic designs rather than differences in data availability or training budgets.
	
	The SBL and STD-PINNs are primarily formulated for inverse problems with constant coefficients. When the true coefficient varies over time or space, these methods tend to recover only an effective averaged parameter, which leads to large errors in different coefficient regimes. DSBL improves the classical Bayesian framework by introducing a discrete switching mechanism, but it requires prior information about possible switching times or state transitions. Therefore, its applicability is limited when the number and locations of jumps are unknown.
	
	\begin{table}[t]
		\centering
		\caption{Comparison estimating results for identifying time-varying coefficients in Burgers equation Case~3.3 with other Bayesian and neural networks methods. \label{tab9}}
		\scalebox{1}{
			\setlength{\tabcolsep}{2.5pt}
			\begin{tabular}{ccccccc}
				\toprule
				Algorithm & MSE & Parameter & Reference & Predict & Relative Error & Time/s \\
				\midrule
				\multirow{4}*{SBL} & \multirow{4}*{2.3418e-4} & \multirow{3}*{$\lambda_1(t)$} & 0.5 & 0.8007 & +60.1447\% & \multirow{4}*{2880.7} \\
				~ & ~ & ~ & 0.75 & 0.8007 & +6.7632\% & ~ \\
				~ & ~ & ~ & 1 & 0.8007 & $-$19.9276\% & ~ \\
				\cmidrule(lr){3-6}
				~ & ~ & $\lambda_2(t)$ & 0.1 & 0.1023 & +2.2785\% & ~ \\
				\cmidrule(lr){1-7}
				
				\multirow{4}*{DSBL} & \multirow{4}*{4.0347e-5} & \multirow{3}*{$\lambda_1(t)$} & 0.5 & 0.4812 & $-$3.7561\% & \multirow{4}*{5726.6} \\
				~ & ~ & ~ & 0.75 & 0.8105 & +8.0635\% & ~ \\
				~ & ~ & ~ & 1 & 1.0143 & +1.4286\% & ~ \\
				\cmidrule(lr){3-6}
				~ & ~ & $\lambda_2(t)$ & 0.1 & 0.1019 & +1.9235\% & ~ \\
				\cmidrule(lr){1-7}
				
				\multirow{4}*{STD-PINNs} & \multirow{4}*{2.6651e-4} & \multirow{3}*{$\lambda_1(t)$} & 0.5 & 0.8488 & +69.7526\% & \multirow{4}*{7432.4} \\
				~ & ~ & ~ & 0.75 & 0.8488 & +13.1684\% & ~ \\
				~ & ~ & ~ & 1 & 0.8488 & $-$15.1237\% & ~ \\
				\cmidrule(lr){3-6}
				~ & ~ & $\lambda_2(t)$ & 0.1 & 0.0933 & $-$6.6713\% & ~ \\
				\cmidrule(lr){1-7}
				
				\multirow{4}*{BC-PINNs} & \multirow{4}*{5.8597e-6} & \multirow{3}*{$\lambda_1(t)$} & 0.5 & 0.4770 & $-$4.5963\% & \multirow{4}*{8278.5} \\
				~ & ~ & ~ & 0.75 & 0.6825 & $-$8.9944\% & ~ \\
				~ & ~ & ~ & 1 & 0.9895 & $-$1.0474\% & ~ \\
				\cmidrule(lr){3-6}
				~ & ~ & $\lambda_2(t)$ & 0.1 & 0.1004 & +0.3825\% & ~ \\
				\cmidrule(lr){1-7}
				
				\multirow{4}*{VC-PINNs} & \multirow{4}*{2.1545e-7} & \multirow{3}*{$\lambda_1(t)$} & 0.5 & 0.4804 & $-$3.9147\% & \multirow{4}*{8012.7} \\
				~ & ~ & ~ & 0.75 & 0.7378 & $-$1.6251\% & ~ \\
				~ & ~ & ~ & 1 & 1.0252 & +2.5184\% & ~ \\
				\cmidrule(lr){3-6}
				~ & ~ & $\lambda_2(t)$ & 0.1 & 0.0998 & $-$0.2035\% & ~ \\
				\cmidrule(lr){1-7}
				
				\multirow{4}*{CP-PINNs} & \multirow{4}*{4.7447e-5} & \multirow{3}*{$\lambda_1(t)$} & 0.5 & 0.6574 & +31.4877\% & \multirow{4}*{7714.7} \\
				~ & ~ & ~ & 0.75 & 0.7477 & $-$0.3084\% & ~ \\
				~ & ~ & ~ & 1 & 0.8184 & $-$18.1609\% & ~ \\
				\cmidrule(lr){3-6}
				~ & ~ & $\lambda_2(t)$ & 0.1 & 0.1142 & +14.1632\% & ~ \\
				\cmidrule(lr){1-7}
				
				\multirow{4}*{GWS-PINNs} & \multirow{4}*{3.6277e-9} & \multirow{3}*{$\lambda_1(t)$} & 0.5 & 0.4994 & $-$0.1297\% & \multirow{4}*{8076.4} \\
				~ & ~ & ~ & 0.75 & 0.7512 & +0.1634\% & ~ \\
				~ & ~ & ~ & 1 & 1.0010 & +0.0962\% & ~ \\
				\cmidrule(lr){3-6}
				~ & ~ & $\lambda_2(t)$ & 0.1 & 0.1000 & $-$0.0150\% & ~ \\
				\cmidrule(lr){1-7}
				
				\multirow{4}*{JVC-PINNs$^{\star}$} & \multirow{4}*{3.5863e-9} & \multirow{3}*{$\lambda_1(t)$} & 0.5 & 0.5013 & +0.2570\% & \multirow{4}*{8197.7} \\
				~ & ~ & ~ & 0.75 & 0.7498 & $-$0.0313\% & ~ \\
				~ & ~ & ~ & 1 & 0.9999 & $-$0.0113\% & ~ \\
				\cmidrule(lr){3-6}
				~ & ~ & $\lambda_2(t)$ & 0.1 & 0.1000 & $-$0.0041\% & ~ \\
				\bottomrule
		\end{tabular}}
	\end{table}
	
	Among PINNs-based methods, BC-PINNs and VC-PINNs can approximate time-varying or spatially varying coefficients to some extent. However, their coefficient representations are continuous or locally smooth in the implementations considered here, which may limit their ability to represent sharp jumps explicitly. BC-PINNs also relies on a prescribed time-domain decomposition and too few subdomains may lead to inaccurate jump localization, whereas too many subdomains increase the computational burden causing unstable local estimates. CP-PINNs explicitly introduces change-point variables for temporal switching problems, but its performance is sensitive to initialization and prespecified number of change points. As a result, although it can introduce a discontinuous parameter representation, the estimated jump-varying coefficient function is not sufficiently accurate.
	
	Compared with these methods, GWS-PINNs introduces an auxiliary coefficient network to learn a continuous surrogate of the unknown coefficient field and further incorporates a gradient-adaptive weighting strategy. This design improves the reliability of coefficient sampling in approximately constant regimes and suppresses the adverse influence of high-gradient transition regions. Therefore, GWS-PINNs is more suitable as a neural-network sampler for discontinuously varying coefficients in PDEs. The complete JVC-PINNs framework subsequently converts the smooth coefficient surrogate produced by GWS-PINNs into an explicit hard piecewise-constant estimator. After first-stage sampling, the GMM-BDMC module infers the number of coefficient states, estimates coefficient search intervals, and identifies candidate jump regions. The second-stage CCD-PINNs then refines the coefficient values and discontinuity locations under these constraints. In this way, JVC-PINNs retains the flexible sampling capability of GWS-PINNs while providing an explicit piecewise-constant representation and refined change-point estimates.
	
	The comparison is conducted on the Burgers equation in Case~3.3, where $\lambda_1(t)$ is a jump-varying coefficient and $\lambda_2(t)$ is a constant coefficient. The absolute errors of the predicted solution fields are shown in Figure~\ref{fig11}, and the corresponding coefficient recovery results are shown in Figure~\ref{fig12}. Quantitative results are reported in Table~\ref{tab9}. The results show that SBL and STD-PINNs produce large errors because they cannot represent the temporal switching of $\lambda_1(t)$. DSBL, BC-PINNs, VC-PINNs, and CP-PINNs reduce the error to different degrees, but their errors remain concentrated near coefficient jump regions. In contrast, GWS-PINNs provides more accurate coefficient samples, and JVC-PINNs achieves the lowest solution MSE among the compared methods, with only a moderate increase in computational time relative to GWS-PINNs.
	
	For the solution field, JVC-PINNs achieves the best approximation accuracy among all compared methods for PDEs with jump-varying parameters. In terms of parameter identification, it accurately recovers the three plateau values of $\lambda_1(t)$ and the constant value of $\lambda_2(t)$. In addition, except for CP-PINNs, JVC-PINNs is the only method among the compared neural-network approaches that can directly output accurate estimates of both coefficient values and change points. These results demonstrate that the proposed two-stage framework is effective not only for solution reconstruction, but also for interpretable identification of discontinuous coefficients and their switching locations.
	
	\subsection{Statistical Clustering Learning}
	The second comparison focuses on statistical learning methods for extracting discrete coefficient regimes from the first-stage GWS-PINNs samples. After the neural-network sampler produces the empirical coefficient values, the statistical module is required to identify the coefficient states, estimate within-regime dispersion, and provide candidate jump regions for the second-stage CCD-PINNs refinement. This part compares several representative clustering estimators, including K-nearest neighbor (KNN) \cite{agarwal2021ml}, K-means \cite{chawla2013k}, the expectation maximization (EM) \cite{yang2012robust} for GMMs, the MCMC \cite{fruhwirth2006finite} for GMMs, the generative adversarial networks (GANs) for clustering \cite{mukherjee2019clustergan} as a deep learning method, and the proposed GMM-BDMC model-selection strategy. For KNN, K-means, and GAN-based clustering, the reported mean and standard deviation are computed as the empirical average and empirical standard deviation of the samples assigned to each estimated cluster. In contrast, the GMM-based methods naturally provide these quantities through the mean and variance parameters of each Gaussian component.
	
	For known-$K$ estimation, Table~\ref{tab10} reports the estimated means, standard deviations, and computational time for different clustering or mixture-learning algorithms. The BDMC can estimate $K$ automatically, whereas other methods are supplied with the reference value of $K$. The KNN and K-means are computationally efficient, but their accuracy depends strongly on the separability of the sampled coefficient values. When transition samples or sampling noise are present, distance-based classification may merge nearby regimes or enlarge the estimated variance. The EM is also efficient, but it is sensitive to initialization and may converge to local optima, especially when the mixture components are unbalanced or the first-stage samples contain biased transition values. The GAN-based clustering provides competitive estimates in several cases, but it requires training additional neural networks and therefore introduces higher computational cost. The MCMC improves robustness by posterior sampling, but its efficiency decreases when the sampled distribution is complex.
	
	The proposed BDMC-based mixture learner achieves a favorable balance between accuracy and computational cost. In most cases in Table~\ref{tab10}, BDMC gives coefficient means close to the reference values and maintains moderate variance estimates. More importantly, unlike KNN, K-means, EM, and GAN-based clustering with a prescribed number of clusters, BDMC is naturally compatible with unknown-$K$ inference. This property is essential in the proposed JVC-PINNs framework, because the number of coefficient regimes is generally unavailable before solving the inverse problem.
	
	\begin{table}[p]
		\centering
		\caption{Known-$K$ comparison of coefficient-regime estimators. \label{tab10}}
		\scalebox{1}{
			\setlength{\tabcolsep}{2.5pt}
			\begin{tabular}{@{}c@{\hspace{12pt}}ccccc@{\hspace{12pt}}ccccc@{}}
				\toprule
				Case & Algorithm & Reference & Mean & Sigma & Time/s & Algorithm & Reference & Mean & Sigma & Time/s \\
				\midrule
				\multirow{6}*{1.3} & \multirow{2}*{KNN} & 2.5 & 2.7162 & 0.1476 & \multirow{2}*{4.2} & \multirow{2}*{MCMC} & 2.5 & 2.8007 & 0.1676 & \multirow{2}*{34.6} \\
				~ & ~ & 3 & 2.9948 & 0.0163 & ~ & ~ & 3 & 2.9998 & 0.0041 & ~ \\
				\cmidrule(lr){2-11}
				~ & \multirow{2}*{K-means} & 2.5 & 2.6318 & 0.1163 & \multirow{2}*{5.3} & \multirow{2}*{BDMC$^{\star}$} & 2.5 & 2.5854 & 0.2495 & \multirow{2}*{37.2} \\
				~ & ~ & 3 & 2.9844 & 0.0393 & ~ & ~ & 3 & 2.9916 & 0.0248 & ~ \\
				\cmidrule(lr){2-11}
				~ & \multirow{2}*{EM} & 2.5 & 2.9051 & 0.1606 & \multirow{2}*{6.9} & \multirow{2}*{GANs} & 2.5 & 2.6272 & 0.1799 & \multirow{2}*{60.1} \\
                ~ & ~ & 3 & 2.9722 & 0.0789 & ~ & ~ & 3 & 2.9872 & 0.0337 & ~ \\
				\midrule
				\multirow{9}*{2.2} & \multirow{3}*{KNN} & 0.05 & 0.0517 & 0.0048 & \multirow{3}*{3.1} & \multirow{3}*{MCMC} & 0.05 & 0.0509 & 0.0053 & \multirow{3}*{24.4} \\
				~ & ~ & 0.1 & 0.0984 & 0.0148 & ~ & ~ & 0.1 & 0.0965 & 0.0175 & ~ \\
				~ & ~ & 0.2 & 0.1930 & 0.0149 & ~ & ~ & 0.2 & 0.1932 & 0.0154 & ~ \\
				\cmidrule(lr){2-11}
				~ & \multirow{3}*{K-means} & 0.05 & 0.0531 & 0.0070 & \multirow{3}*{3.9} & \multirow{3}*{BDMC$^{\star}$} & 0.05 & 0.0527 & 0.0065 & \multirow{3}*{26.2} \\
				~ & ~ & 0.1 & 0.1002 & 0.0130 & ~ & ~ & 0.1 & 0.1002 & 0.0141 & ~ \\
				~ & ~ & 0.2 & 0.1926 & 0.0155 & ~ & ~ & 0.2 & 0.1935 & 0.0142 & ~ \\
				\cmidrule(lr){2-11}
				~ & \multirow{3}*{EM} & 0.05 & 0.0702 & 0.0267 & \multirow{3}*{5.9} & \multirow{3}*{GANs} & 0.05 & 0.0528 & 0.0163 & \multirow{3}*{43.3} \\
                ~ & ~ & 0.1 & 0.0831 & 0.0358 & ~ & ~ & 0.1 & 0.1005 & 0.0262 & ~ \\
                ~ & ~ & 0.2 & 0.1666 & 0.0465 & ~ & ~ & 0.2 & 0.1933 & 0.0307 & ~ \\
				\midrule
				\multirow{18}*{3.4} & \multirow{6}*{KNN} & 0.5 & 0.5087 & 0.0225 & \multirow{6}*{2.3} & \multirow{6}*{MCMC} & 0.5 & 0.5098 & 0.0444 & \multirow{6}*{18.6} \\
				~ & ~ & 0.75 & 0.7560 & 0.0397 & ~ & ~ & 0.75 & 0.7559 & 0.0458 & ~ \\
				~ & ~ & 1 & 0.9970 & 0.0318 & ~ & ~ & 1 & 0.9971 & 0.0409 & ~ \\
				\cmidrule(lr){3-5}\cmidrule(lr){8-10}
				~ & ~ & 1 & 0.9961 & 0.0398 & ~ & ~ & 1 & 0.9954 & 0.0623 & ~ \\
				~ & ~ & 1.3333 & 1.3339 & 0.0660 & ~ & ~ & 1.3333 & 1.3283 & 0.0753 & ~ \\
				~ & ~ & 2 & 1.9761 & 0.0592 & ~ & ~ & 2 & 1.9648 & 0.1057 & ~ \\
				\cmidrule(lr){2-11}
				~ & \multirow{6}*{K-means} & 0.5 & 0.5111 & 0.0278 & \multirow{6}*{3.8} & \multirow{6}*{BDMC$^{\star}$} & 0.5 & 0.5111 & 0.0277 & \multirow{6}*{20.2} \\
				~ & ~ & 0.75 & 0.7561 & 0.0360 & ~ & ~ & 0.75 & 0.7561 & 0.0360 & ~ \\
				~ & ~ & 1 & 0.9958 & 0.0338 & ~ & ~ & 1 & 0.9958 & 0.0338 & ~ \\
				\cmidrule(lr){3-5}\cmidrule(lr){8-10}
				~ & ~ & 1 & 0.9961 & 0.0398 & ~ & ~ & 1 & 0.9946 & 0.0368 & ~ \\
				~ & ~ & 1.3333 & 1.3307 & 0.0577 & ~ & ~ & 1.3333 & 1.3321 & 0.0679 & ~ \\
				~ & ~ & 2 & 1.9699 & 0.0735 & ~ & ~ & 2 & 1.9761 & 0.0592 & ~ \\
				\cmidrule(lr){2-11}
				~ & \multirow{6}*{EM} & 0.5 & 0.6247 & 0.1398 & \multirow{6}*{5.2} & \multirow{6}*{GANs} & 0.5 & 0.5106 & 0.0846 & \multirow{6}*{40.6} \\
                ~ & ~ & 0.75 & 0.8011 & 0.1555 & ~ & ~ & 0.75 & 0.7563 & 0.0937 & ~ \\
                ~ & ~ & 1 & 0.9385 & 0.1125 & ~ & ~ & 1 & 0.9959 & 0.0737 & ~ \\
                \cmidrule(lr){3-5}\cmidrule(lr){8-10}
                ~ & ~ & 1 & 1.1287 & 0.1871 & ~ & ~ & 1 & 0.9947 & 0.1146 & ~ \\
                ~ & ~ & 1.3333 & 1.2291 & 0.2498 & ~ & ~ & 1.3333 & 1.3325 & 0.1504 & ~ \\
                ~ & ~ & 2 & 1.8287 & 0.2874 & ~ & ~ & 2 & 1.9758 & 0.1818 & ~ \\
				\midrule
				\multirow{6}*{4.4} & \multirow{2}*{KNN} & 0.01 & 0.0122 & 0.0044 & \multirow{2}*{2.0} & \multirow{2}*{MCMC} & 0.01 & 0.0116 & 0.0042 & \multirow{2}*{16.6} \\
				~ & ~ & 0.04 & 0.0384 & 0.0046 & ~ & ~ & 0.04 & 0.0378 & 0.0058 & ~ \\
				\cmidrule(lr){2-11}
				~ & \multirow{2}*{K-means} & 0.01 & 0.0120 & 0.0041 & \multirow{2}*{2.9} & \multirow{2}*{BDMC$^{\star}$} & 0.01 & 0.0122 & 0.0044 & \multirow{2}*{17.6} \\
				~ & ~ & 0.04 & 0.0382 & 0.0049 & ~ & ~ & 0.04 & 0.0384 & 0.0046 & ~ \\
				\cmidrule(lr){2-11}
				~ & \multirow{2}*{EM} & 0.01 & 0.0145 & 0.0087 & \multirow{2}*{4.7} & \multirow{2}*{GANs} & 0.01 & 0.0127 & 0.0043 & \multirow{2}*{35.7} \\
                ~ & ~ & 0.04 & 0.0349 & 0.0099 & ~ & ~ & 0.04 & 0.0377 & 0.0047 & ~ \\
				\midrule
				\multirow{6}*{5.1} & \multirow{2}*{KNN} & 13.3333 & 13.9595 & 0.9486 & \multirow{2}*{7.1} & \multirow{2}*{MCMC} & 13.3333 & 17.8638 & 2.5606 & \multirow{2}*{76.8} \\
				~ & ~ & 20 & 19.8903 & 0.3681 & ~ & ~ & 20 & 19.9898 & 0.0210 & ~ \\
				\cmidrule(lr){2-11}
				~ & \multirow{2}*{K-means} & 13.3333 & 14.0549 & 1.0651 & \multirow{2}*{8.4} & \multirow{2}*{BDMC$^{\star}$} & 13.3333 & 13.4176 & 0.0843 & \multirow{2}*{84.2} \\
				~ & ~ & 20 & 19.8971 & 0.3387 & ~ & ~ & 20 & 19.8082 & 0.1918 & ~ \\
				\cmidrule(lr){2-11}
				~ & \multirow{2}*{EM} & 13.3333 & 19.3572 & 1.7473 & \multirow{2}*{9.4} & \multirow{2}*{GANs} & 13.3333 & 13.5547 & 1.7348 & \multirow{2}*{145.4} \\
                ~ & ~ & 20 & 19.7141 & 1.0963 & ~ & ~ & 20 & 19.8255 & 0.1926 & ~ \\
				\bottomrule
		\end{tabular}}
	\end{table}
	
	For unknown-$K$ estimation, the AIC and BIC methods select the same number of components in all reported cases, and their common estimates are reported as AIC/BIC. Table~\ref{tab11} compares AIC/BIC-based GMM selection with GMM-BDMC in multidimensional settings and under the one-dimensional flattening strategy. Both methods correctly identify the number of coefficient states in the tested cases, but they differ significantly in estimation stability and computational cost. AIC/BIC requires fitting multiple independent GMMs with different component numbers and then selecting the optimal model according to the information criterion. In contrast, BDMC treats the number of components as a random variable and explores different models through birth and death moves. The cumulative number of model parameters processed when independently fitting all candidate GMMs up to $K_{\max}$ can be summarized as
	\begin{equation}
		\begin{aligned}
			&N_{\mathrm{AIC/BIC}}^{1D}=\sum_{k=1}^{K_{\max}}(3k-1)=\mathcal{O}(K_{\max}^2),\\
            &N_{\mathrm{BDMC}}^{1D}=4K_{\max}=\mathcal{O}(K_{\max}),\\
            &N_{\mathrm{AIC/BIC}}^{dD}=\sum_{k=1}^{K_{\max}}\left(k\left(\frac{d(d+1)}{2}+d+1\right)-1\right)=\mathcal{O}(d^2K_{\max}^2),\\
			&N_{\mathrm{BDMC}}^{dD}=K_{\max}\left(\frac{d(d+1)}{2}+d+2\right)=\mathcal{O}(d^2K_{\max}).
		\end{aligned}
	\end{equation}
	Here, the terms include the component means, variances or covariance matrices, mixture weights, and the additional birth-death rates in BDMC. Therefore, the use of BDMC reduces the GMM model-selection complexity in terms of the total number of fitted component parameters from quadratic to linear with respect to the maximum number of candidate regimes.
	
	\begin{table}[t]
		\centering
		\caption{Unknown-$K$ comparison of AIC/BIC-based GMM estimators with BDMC. \label{tab11}}
		\scalebox{0.92}{
			\setlength{\tabcolsep}{2.5pt}
			\begin{tabular}{@{}c@{\hspace{12pt}}cccccc@{\hspace{12pt}}cccccc@{}}
				\toprule
				Case & GMM Method & $\hat{K}$ & True & Mean & Sigma & Time/s & GMM Method & $\hat{K}$ & True & Mean & Sigma & Time/s \\
				\midrule
				\multirow{4}*{1.3} & \multirow{2}*{1D AIC/BIC} & \multirow{2}*{2} & 2.5 & 2.9217 & 0.1482 & \multirow{2}*{148.6} & \multirow{2}*{1D BDMC$^{\star}$} & \multirow{2}*{2} & 2.5 & 2.5854 & 0.2495 & \multirow{2}*{37.2} \\
				~ & ~ & ~ & 3 & 2.9586 & 0.1035 & ~ & ~ & ~ & 3 & 2.9916 & 0.0248 & ~ \\
				\cmidrule(lr){2-13}
				~ & \multirow{2}*{2D AIC/BIC} & \multirow{2}*{2} & 2.5 & 2.5915 & 0.2594 & \multirow{2}*{301.3} & \multirow{2}*{2D BDMC} & \multirow{2}*{2} & 2.5 & 2.5854 & 0.4367 & \multirow{2}*{76.8} \\
				~ & ~ & ~ & 3 & 2.9780 & 0.1811 & ~ & ~ & ~ & 3 & 2.9916 & 0.0433 & ~ \\
				\midrule
				\multirow{3}*{2.2} & \multirow{3}*{1D AIC/BIC} & \multirow{3}*{3} & 0.05 & 0.0700 & 0.0269 & \multirow{3}*{132.9} & \multirow{3}*{1D BDMC$^{\star}$} & \multirow{3}*{3} & 0.05 & 0.0527 & 0.0065 & \multirow{3}*{26.2} \\
				~ & ~ & ~ & 0.1 & 0.0845 & 0.0375 & ~ & ~ & ~ & 0.1 & 0.1002 & 0.0141 & ~ \\
				~ & ~ & ~ & 0.2 & 0.1628 & 0.0488 & ~ & ~ & ~ & 0.2 & 0.1935 & 0.0142 & ~ \\
				\midrule
				\multirow{6}*{3.4} & \multirow{6}*{1D AIC/BIC} & \multirow{6}*{3} & 0.5 & 0.6265 & 0.1410 & \multirow{6}*{101.9} & \multirow{6}*{1D BDMC$^{\star}$} & \multirow{6}*{3} & 0.5 & 0.5111 & 0.0277 & \multirow{6}*{20.2} \\
				~ & ~ & ~ & 0.75 & 0.7987 & 0.1567 & ~ & ~ & ~ & 0.75 & 0.7561 & 0.0360 & ~ \\
				~ & ~ & ~ & 1 & 0.9359 & 0.1144 & ~ & ~ & ~ & 1 & 0.9958 & 0.0338 & ~ \\
				\cmidrule(lr){4-6}\cmidrule(lr){10-12}
				~ & ~ & ~ & 1 & 1.1235 & 0.1856 & ~ & ~ & ~ & 1 & 0.9946 & 0.0368 & ~ \\
				~ & ~ & ~ & 1.3333 & 1.2465 & 0.2650 & ~ & ~ & ~ & 1.3333 & 1.3321 & 0.0679 & ~ \\
				~ & ~ & ~ & 2 & 1.8034 & 0.3058 & ~ & ~ & ~ & 2 & 1.9761 & 0.0592 & ~ \\
				\midrule
				\multirow{2}*{4.4} & \multirow{2}*{1D AIC/BIC} & \multirow{2}*{2} & 0.01 & 0.0147 & 0.0089 & \multirow{2}*{70.1} & \multirow{2}*{1D BDMC$^{\star}$} & \multirow{2}*{2} & 0.01 & 0.0122 & 0.0044 & \multirow{2}*{17.6} \\
				~ & ~ & ~ & 0.04 & 0.0352 & 0.0097 & ~ & ~ & ~ & 0.04 & 0.0384 & 0.0046 & ~ \\
				\midrule
				\multirow{4}*{5.1} & \multirow{2}*{1D AIC/BIC} & \multirow{2}*{2} & 13.3333 & 19.5033 & 1.5254 & \multirow{2}*{347.5} & \multirow{2}*{1D BDMC$^{\star}$} & \multirow{2}*{2} & 13.3333 & 13.4176 & 0.0843 & \multirow{2}*{84.2} \\
				~ & ~ & ~ & 20 & 19.5572 & 1.4305 & ~ & ~ & ~ & 20 & 19.8082 & 0.1918 & ~ \\
				\cmidrule(lr){2-13}
				~ & \multirow{2}*{2D AIC/BIC} & \multirow{2}*{2} & 13.3333 & 14.4090 & 2.6695 & \multirow{2}*{657.9} & \multirow{2}*{2D BDMC} & \multirow{2}*{2} & 13.3333 & 13.4176 & 0.1475 & \multirow{2}*{170.8} \\
				~ & ~ & ~ & 20 & 19.9165 & 2.5034 & ~ & ~ & ~ & 20 & 19.8082 & 0.3357 & ~ \\
				\bottomrule
		\end{tabular}}
	\end{table}
	
	The numerical results in Table~\ref{tab11} are consistent with this analysis. For the one-dimensional GMM setting, 1D BDMC achieves much shorter computational time than 1D AIC/BIC in all tested cases. For example, in Case~5.1, the 1D AIC/BIC procedure requires repeated model fitting and takes substantially longer, whereas 1D BDMC identifies $\widehat{K}=2$ with accurate coefficient estimates and much lower cost. Similar improvements are observed in Case~2.2, Case~3.4, and Case~4.4. These results indicate that the BDMC method is more suitable for the statistical connection between the first-stage sampler and the second-stage constrained estimator.
	
	Because the one-dimensional GMM uses only the empirical distribution of the coefficient values, a bijective permutation of the sample ordering does not change the mixture likelihood. Provided that the original spatial coordinates are retained and every inferred sample index is mapped back correctly, the resulting value-based candidate set is also invariant to the ordering. For the multidimensional discontinuous coefficient problems considered in Case~1.3 and Case~5.1, additional numerical tests verify that row-major flattening, column-major flattening, and random flattening produce identical estimates of the number of coefficient states $\widehat{K}$, the heuristic coefficient search intervals ${\mathcal{I}_k}$, and the candidate discontinuity regions ${\mathcal{X}_l}$ obtained by the GMM-BDMC module. Therefore, the statistical information passed to the second-stage CCD-PINNs remains unchanged under different flattening strategies, indicating that the proposed framework is insensitive to the ordering of sampled coefficients.
	
\begin{figure}[t]
		\centering
		\subfloat[1D/2D GMM result]{
			\begin{minipage}[t]{0.485\linewidth}
				\includegraphics[width=1\linewidth]{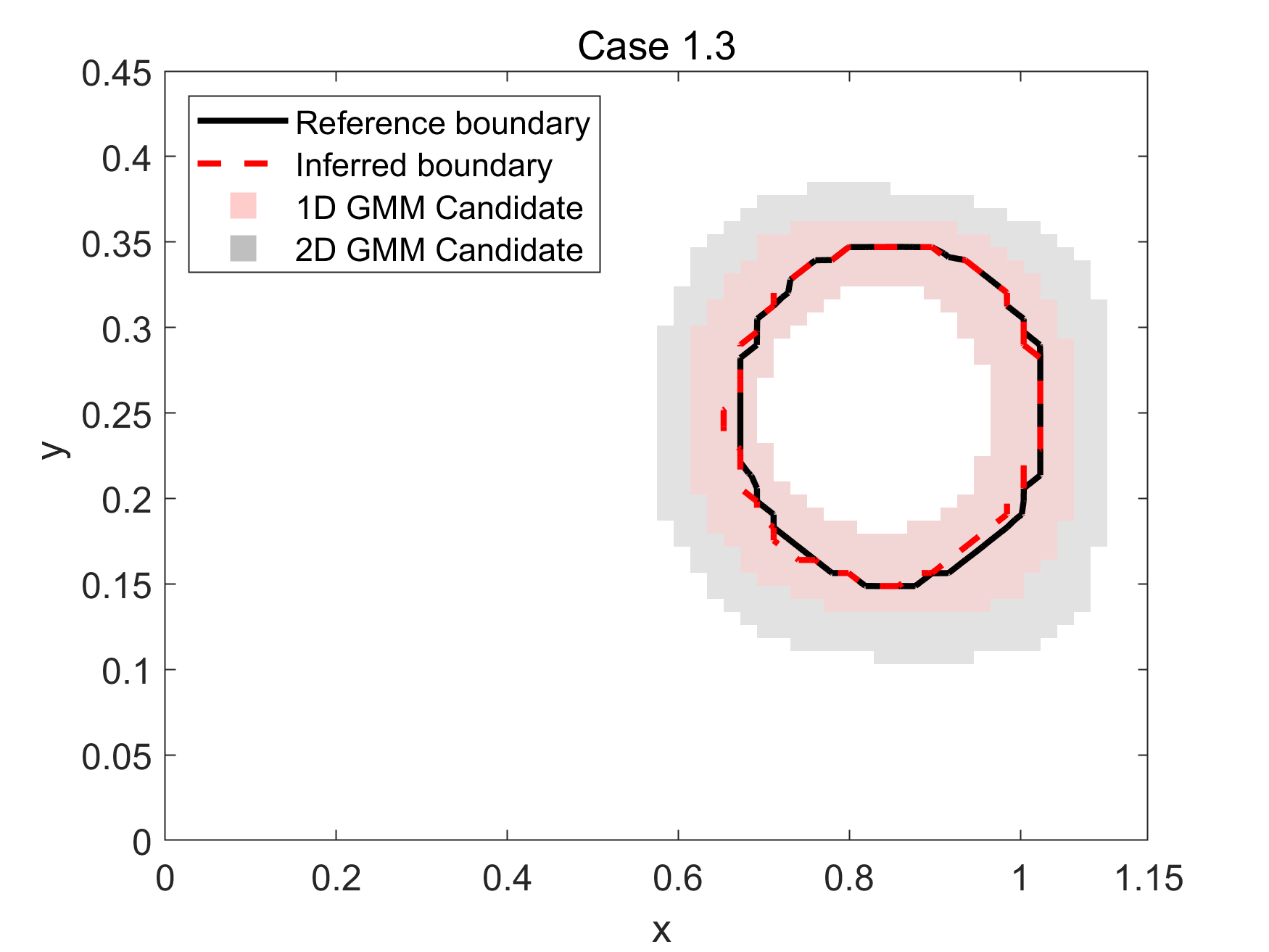}
			\end{minipage}
		}
		\subfloat[1D/2D GMM result]{
			\begin{minipage}[t]{0.485\linewidth}
				\includegraphics[width=1\linewidth]{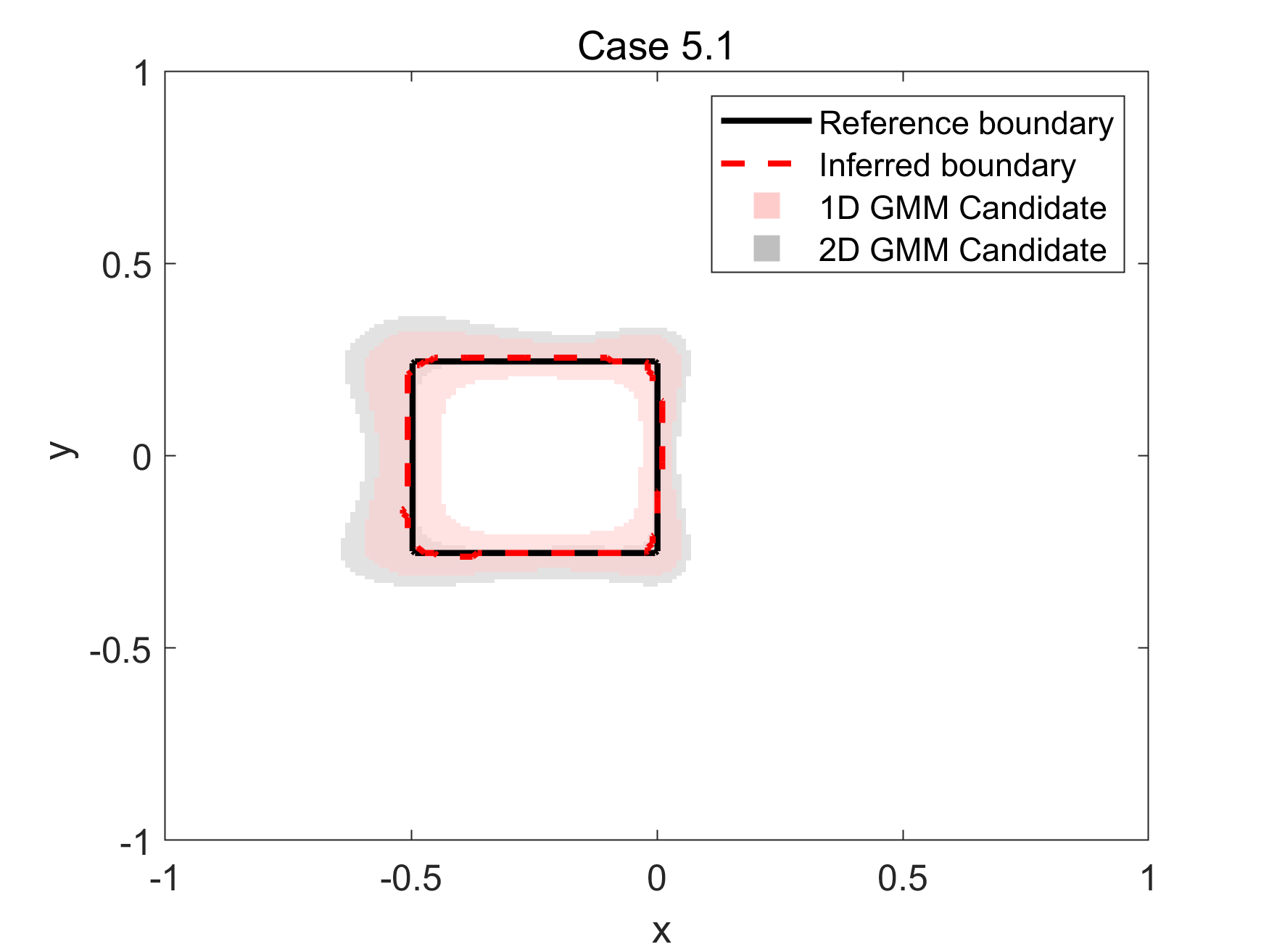}
			\end{minipage}
		}
		\caption{Comparison of 1D and 2D GMM-BDMC inference results for 2D space-varying coefficients of wave equation Case~1.3 and Helmholtz equation Case~5.1. \label{fig13}}
	\end{figure}

	For multidimensional spatial coefficients, two implementation strategies are considered. The first strategy flattens the sampled coefficient field into a one-dimensional sequence and then applies 1D GMM-BDMC. The second strategy directly applies a 2D GMM-BDMC to the spatial samples without flattening. Figure~\ref{fig13} compares these two strategies for Case~1.3 and Case~5.1. The results show that direct 2D GMM inference is feasible and can avoid the flattening operation. However, its estimates are generally less stable than those of the flattened 1D GMM-BDMC. Although the estimated means are close to the reference coefficient values, the estimated variances become larger. For the 2D GMM, each component contains a covariance matrix, and the reported marginal standard deviations are obtained by taking the square roots of the corresponding diagonal entries of the covariance matrix.
	
	Since the candidate regions are determined by the inferred GMM components, the multidimensional GMM leads to wider visualized jump candidates. This phenomenon can be observed in the $\{\mathcal{X}_l\}_{l=1}^{\widehat{L}}$ estimates for Case~1.3 and Case~5.1 in Figure~\ref{fig13}. Nevertheless, this does not necessarily degrade the final JVC-PINNs result. The statistical learner is only required to provide candidate regions that contain the true discontinuity interfaces. Then the second-stage CCD-PINNs can further refine the discontinuity location and recover an accurate hard piecewise-constant coefficient function if the true jump boundary is included in the candidate set. However, this still increases the risk of model estimation mistakes in constrained parameter learning of CCD-PINNs.
	
	Direct multidimensional mixture modeling can preserve geometric information, but it introduces additional covariance parameters and may produce less stable dispersion estimates when the available samples are limited. In the tested spatial cases, the flattened one-dimensional GMM-BDMC procedure provides more stable coefficient-state estimates and lower computational cost. The direct multidimensional formulation remains feasible, but its feature definition, covariance regularization, and candidate-region construction require additional care.

    Overall, the comparisons show that GMM-BDMC is well suited to the statistical connection between GWS-PINNs and CCD-PINNs. It automatically estimates the number of coefficient states, provides heuristic coefficient search intervals for the coefficient values, and avoids independently fitting every candidate GMM required by exhaustive AIC/BIC-based selection. Accordingly, the flattened one-dimensional GMM-BDMC procedure is used as the default statistical learner in JVC-PINNs. Direct multidimensional mixture modeling may be considered when preserving spatial geometry is important and sufficient data are available to estimate the additional covariance parameters reliably.
	
	\begin{figure}[p] 
		\centering
		\includegraphics[width=1\linewidth]{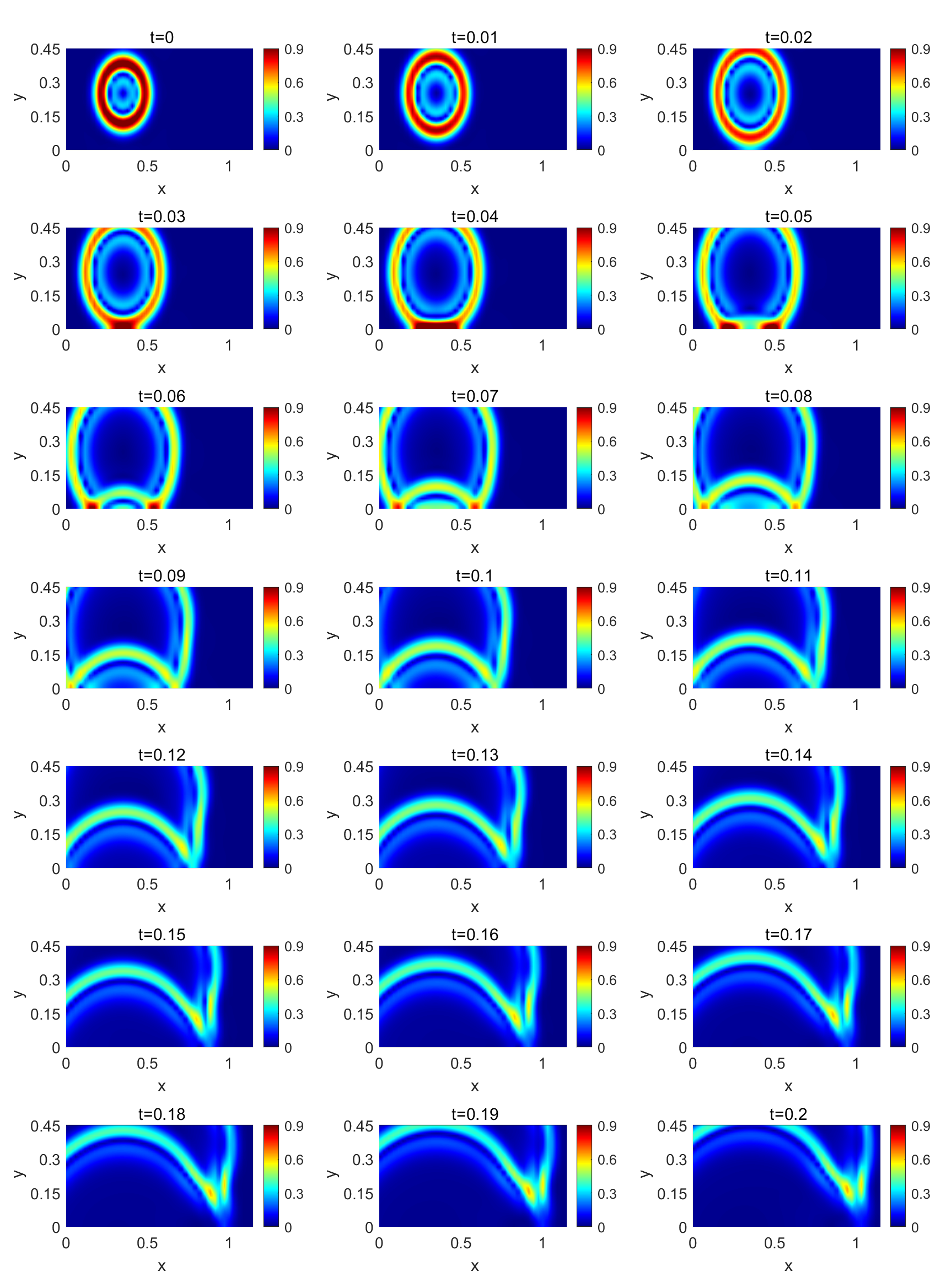}
		\caption{Physical-field reconstruction for wave equation Case~1.3 with discontinuously space-varying coefficient $\alpha(x,y)$. \label{fig14}}
	\end{figure}
	
	\begin{figure}[p]
		\centering
		\includegraphics[width=1\linewidth]{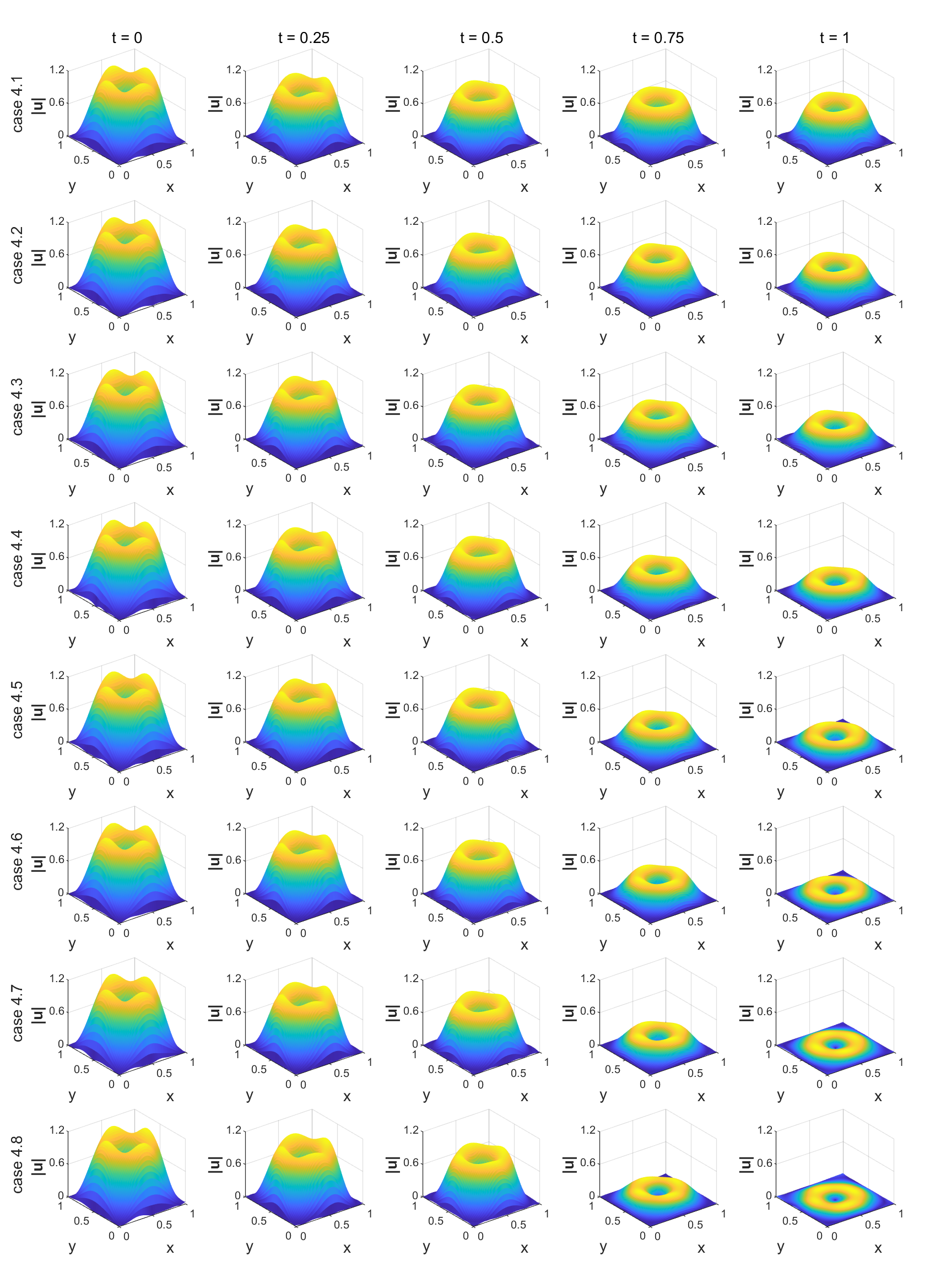}
		\caption{Physical-field reconstruction for Navier-Stokes equation Case~4.1-4.8 with discontinuously time-varying coefficient $\nu(t)$. \label{fig15}}
	\end{figure}
	
	\section{Physical Field Reconstruction}\label{sec6}
	Based on the two-stage JVC-PINNs framework established in Section~\ref{sec2} and the numerical results reported in Section~\ref{sec3}, this section demonstrates the applicability of the proposed method to the physical-field reconstruction of PDE solution fields. The reconstruction is performed using the refined main network $\tilde{u}(\mathbf{x},t)$ obtained from the second-stage CCD-PINNs model. By evaluating this trained neural surrogate on a dense spatiotemporal grid, higher resolution solution fields compared to observations can be reconstructed from sparse or discrete observations. This framework incorporates not only the physical information imposed through the PDE residual, observation data, and initial and boundary conditions, but also the refined discontinuous coefficient structure obtained by relaxed continuous approximation and statistical mixture modeling.

    Unlike conventional interpolation of discrete solution data, the reconstruction produced by $\tilde{u}(\mathbf{x},t)$ is physics-informed. The solution values on the dense grid are generated by the trained neural surrogate rather than by a prescribed interpolation stencil. Consequently, the reconstructed physical field is constrained by the governing PDE, the initial and boundary conditions, and the inferred discontinuous coefficient field. This section considers the wave equation in Case~1.3 with a spatially varying coefficient and the Navier-Stokes equations in Cases~4.1-4.8 with time-varying viscosity coefficients. The corresponding quantitative results are reported in Tables~\ref{tab1}-\ref{tab4}.

    For the wave equation Case~1.3, Figure~\ref{fig14} shows the reconstructed solution fields over the time interval $[0,0.2]$. The training data contain solution observations at only three time instances $t=0$, $t=0.015$, and $t=0.15$. In this example, a reflective boundary condition is imposed on the lower boundary, while the other boundaries are treated as open boundaries. At the initial time, a localized wave is generated within the domain. As time evolves, the wave propagates through the heterogeneous medium, leaves the domain through the open boundaries, and is reflected by the lower boundary. Because the reconstruction is generated by the second-stage network $\tilde{u}(\mathbf{x},t)$ using the refined discontinuous coefficient estimator $\tilde{\theta}_p(\mathbf{x},t)$, it reflects the recovered piecewise-constant structure of the heterogeneous medium.

    For the Navier-Stokes equation Cases~4.1-4.8, Figure~\ref{fig15} presents the reconstructed solution fields over the time interval $[0,1]$. As expected from the experimental setup, the reconstructed fields for Cases~4.1-4.8 are nearly identical before $t=0.5$, because the viscosity coefficient has the same value in all cases over this interval. After $t=0.5$, the solution fields exhibit different dynamics because the post-transition viscosity values differ among the cases. In particular, the difference between Case~4.1 and Case~4.8 illustrates how a large viscosity jump affects the subsequent evolution of the velocity field. This comparison shows that the proposed framework can capture the effects of discontinuous coefficient changes on complex spatiotemporal physical fields.

    This reconstruction capability may be useful in fluid mechanics, aerodynamics, and other engineering applications in which high-resolution experimental data are expensive or difficult to obtain. For example, data collected in wind-tunnel experiments are often sparse because of measurement costs and sensor limitations. By combining a single observation campaign with prior information from the governing PDE, the proposed framework can estimate hidden coefficient fields with temporal or spatial discontinuities from limited solution data. The inferred coefficient field can then be incorporated into a subsequent forward simulation under modified initial or boundary conditions, using either a conventional numerical solver or another deep-learning model. Thus, the proposed framework provides not only a mechanism for identifying latent coefficient structures, but also a high-resolution physical-field reconstruction of the observed physical field that can support forward prediction and scenario analysis in heterogeneous dynamical systems.
	
	\section{Robustness to Observation Noise}\label{sec7}
	This section investigates the robustness of JVC-PINNs to noisy observations, which arises from the combination of physics-informed learning, statistical inference, and constrained refinement. In Stage~1, GWS-PINNs mitigates the influence of observational noise by enforcing the governing PDE and the initial and boundary conditions, so that the learned solution and coefficient surrogates are not determined solely by the noisy observations. The gradient-adaptive residual weighting further reduces the contribution of unstable high-gradient transition regions, where noise-induced fluctuations may be difficult to distinguish from jump-induced nonsmoothness. In the statistical step, GMM-BDMC summarizes the noisy coefficient samples in terms of discrete regimes and provides data-driven search intervals for the coefficient values and candidate transition regions. In Stage~2, CCD-PINNs refines the coefficient values and change points within these inferred intervals and regions, thereby reducing the parameter search space of the inverse problem.

    To further assess the variability of the reconstructed solution and estimated parameters under observation noise, this study introduces a lightweight mechanism inspired by dropout technology and Bayesian PINNs \cite{gal2016dropout, yang2021b}, in which neural-network parameters are treated as random variables. Then they are used to obtain statistical information about PDE solutions and model parameters. Monte Carlo (MC) dropout is introduced into the JVC-PINNs framework as an efficient perturbation-based sensitivity diagnostic. This method is applied to the hidden-layer activations of both the main network and the sub network in Stage~2. The main network represents the solution field, whereas the sub network outputs a finite-dimensional parameter vector that defines the hard piecewise-constant coefficient field. For a temporal coefficient, this vector contains the plateau values and change-point locations. For a spatial coefficient, it contains the parameters that determine the coefficient states and discontinuity interfaces. Therefore, different dropout masks perturb both the solution output and parameter estimation, producing stochastic estimates of the solution field, coefficient values, and discontinuity locations. The Stage~1 GWS-PINNs and GMM-BDMC settings remain unchanged. Introducing dropout in Stage~2 mildly perturbs the trained solution and coefficient surrogates and allows their predictive variability to be assessed without retraining a full Bayesian neural network.

    Three cases with time-varying coefficients, namely Cases~2.1, 3.2, and 4.2, are selected for the noise-robustness study. To account for differences in scale among the PDE solutions, relative additive Gaussian noise is imposed on the clean reference solution data according to $u_i^{\mathrm{noisy}} = u_i + \delta\,\mathrm{std}(u)\epsilon_i$, where $\epsilon_i \sim \mathcal{N}(0,1)$, $\delta$ denotes the relative noise level and $\operatorname{std}(u)$ is the standard deviation of the clean observations. The tested noise levels range from $0.01$ to $0.25$, allowing us to examine both the stable operating range of the proposed method and its degradation under severe noise contamination. 

\begin{figure}[t]
		\centering
		\subfloat[Mean field estimation]{
			\begin{minipage}[t]{0.315\linewidth}
				\includegraphics[width=1\linewidth]{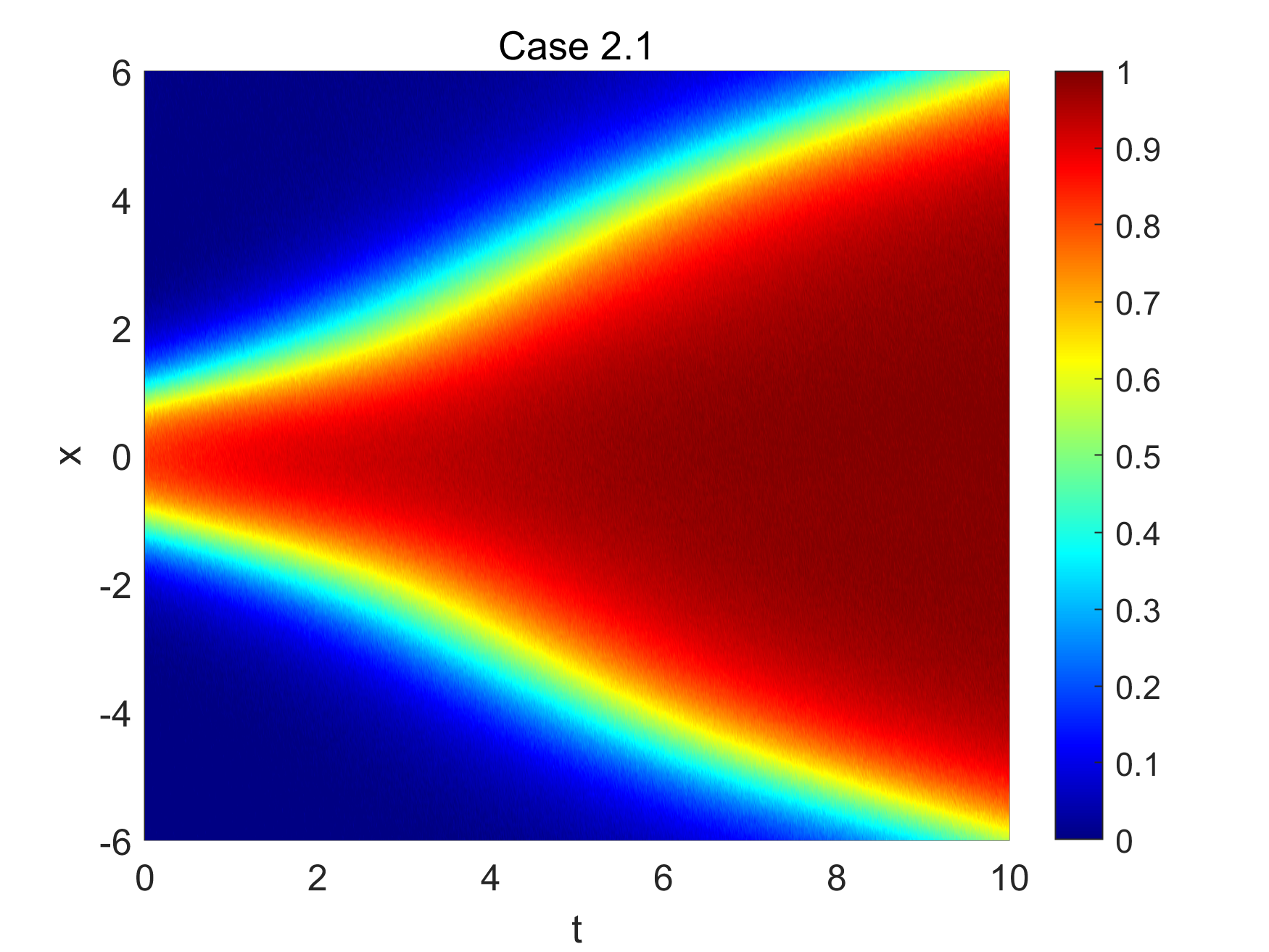}
			\end{minipage}
		}
		\subfloat[Variance field estimation]{
			\begin{minipage}[t]{0.315\linewidth}
				\includegraphics[width=1\linewidth]{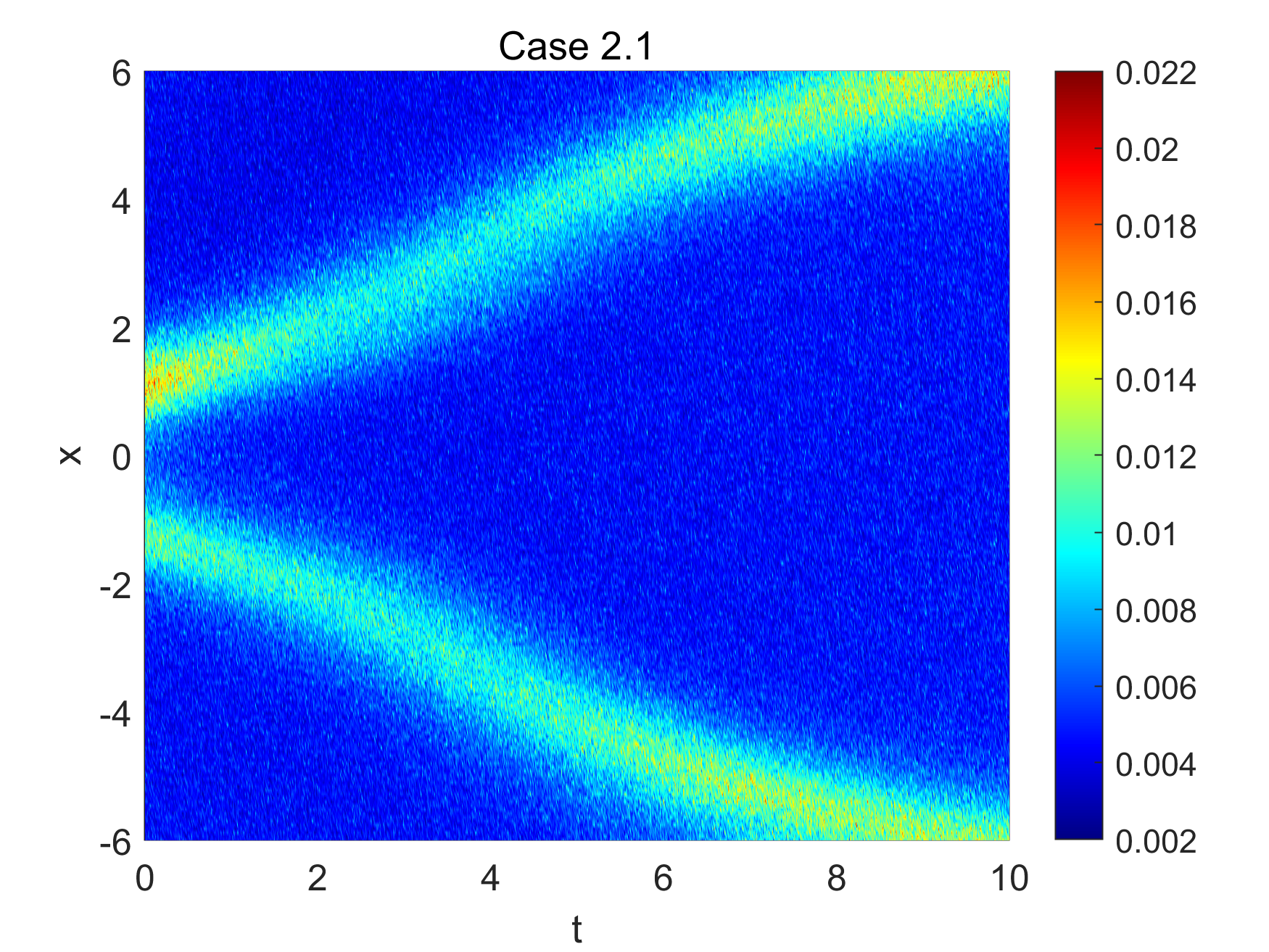}
			\end{minipage}
		}
		\subfloat[Mean field absolute error]{
			\begin{minipage}[t]{0.315\linewidth}
				\includegraphics[width=1\linewidth]{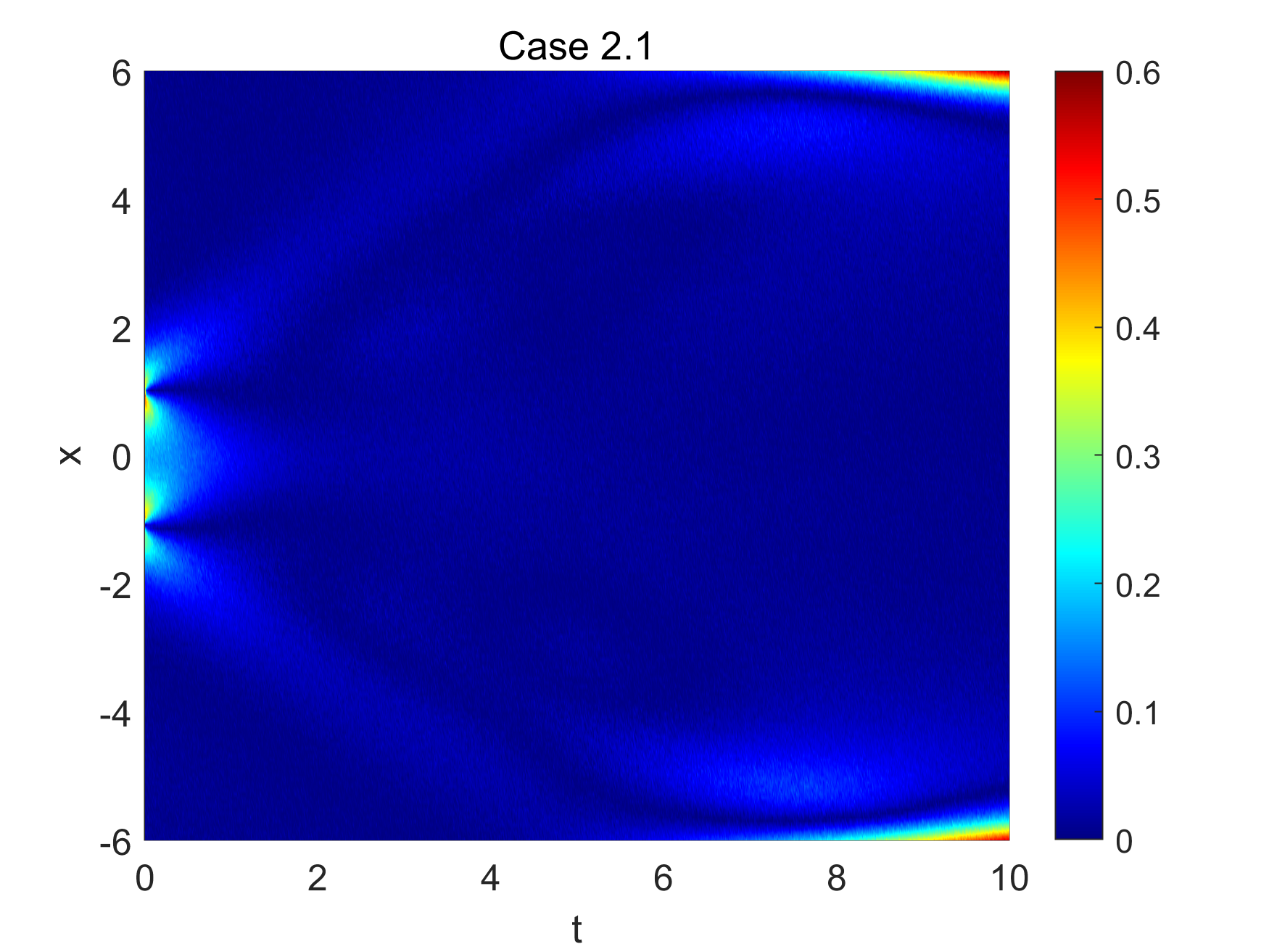}
			\end{minipage}
		}\\
		\subfloat[Mean field estimation]{
			\begin{minipage}[t]{0.315\linewidth}
				\includegraphics[width=1\linewidth]{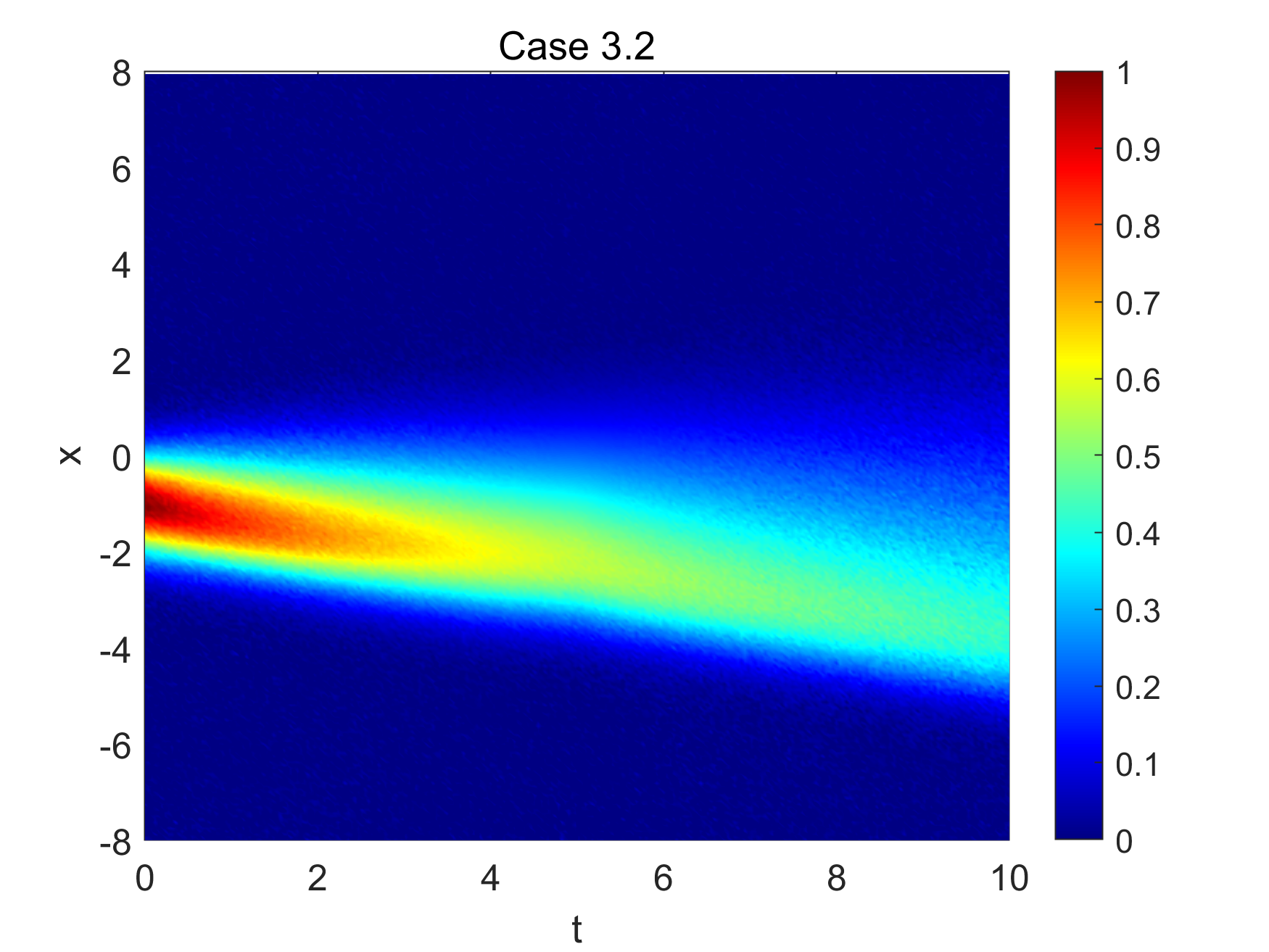}
			\end{minipage}
		}
		\subfloat[Variance field estimation]{
			\begin{minipage}[t]{0.315\linewidth}
				\includegraphics[width=1\linewidth]{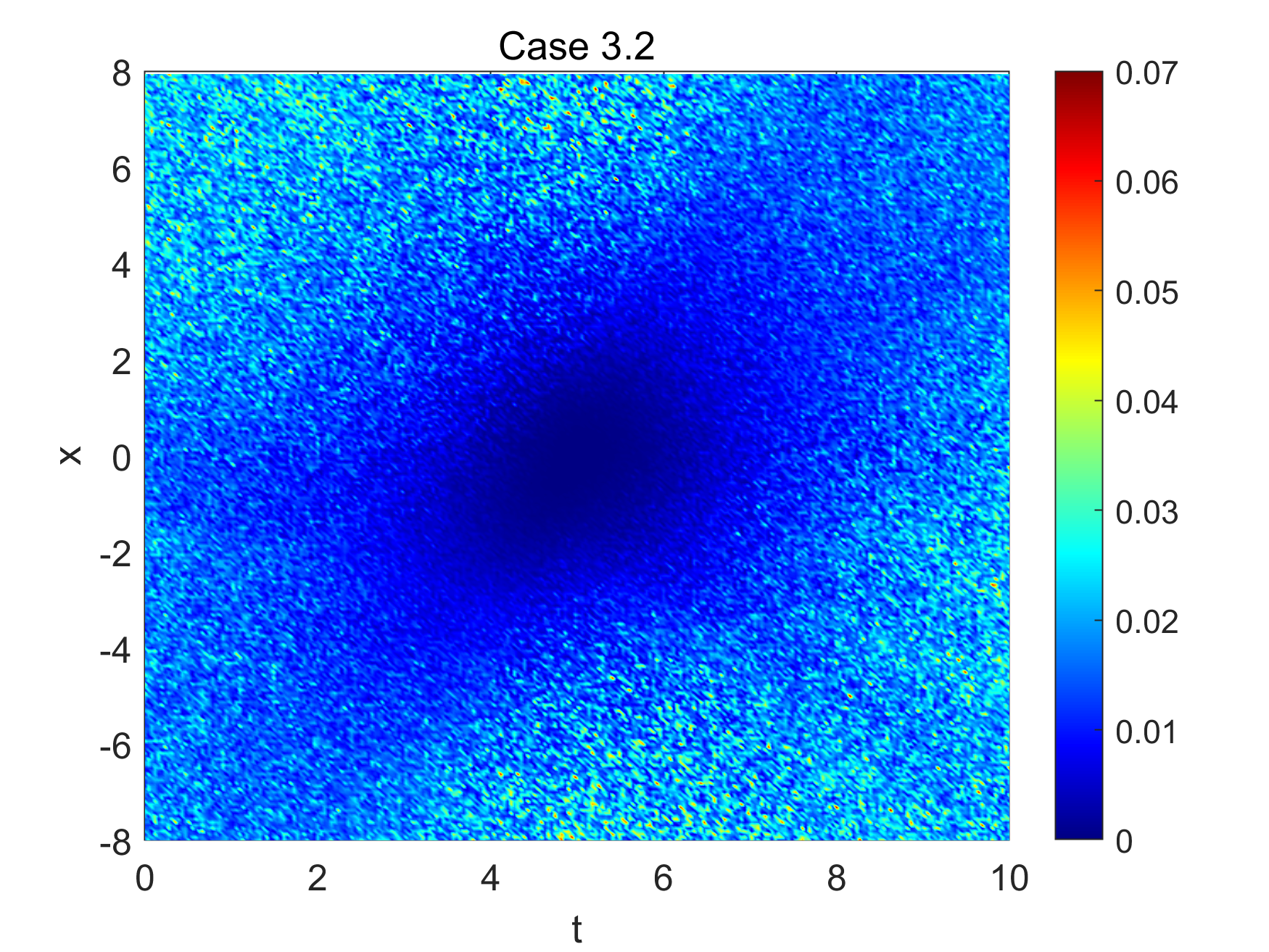}
			\end{minipage}
		}
		\subfloat[Mean field absolute error]{
			\begin{minipage}[t]{0.315\linewidth}
				\includegraphics[width=1\linewidth]{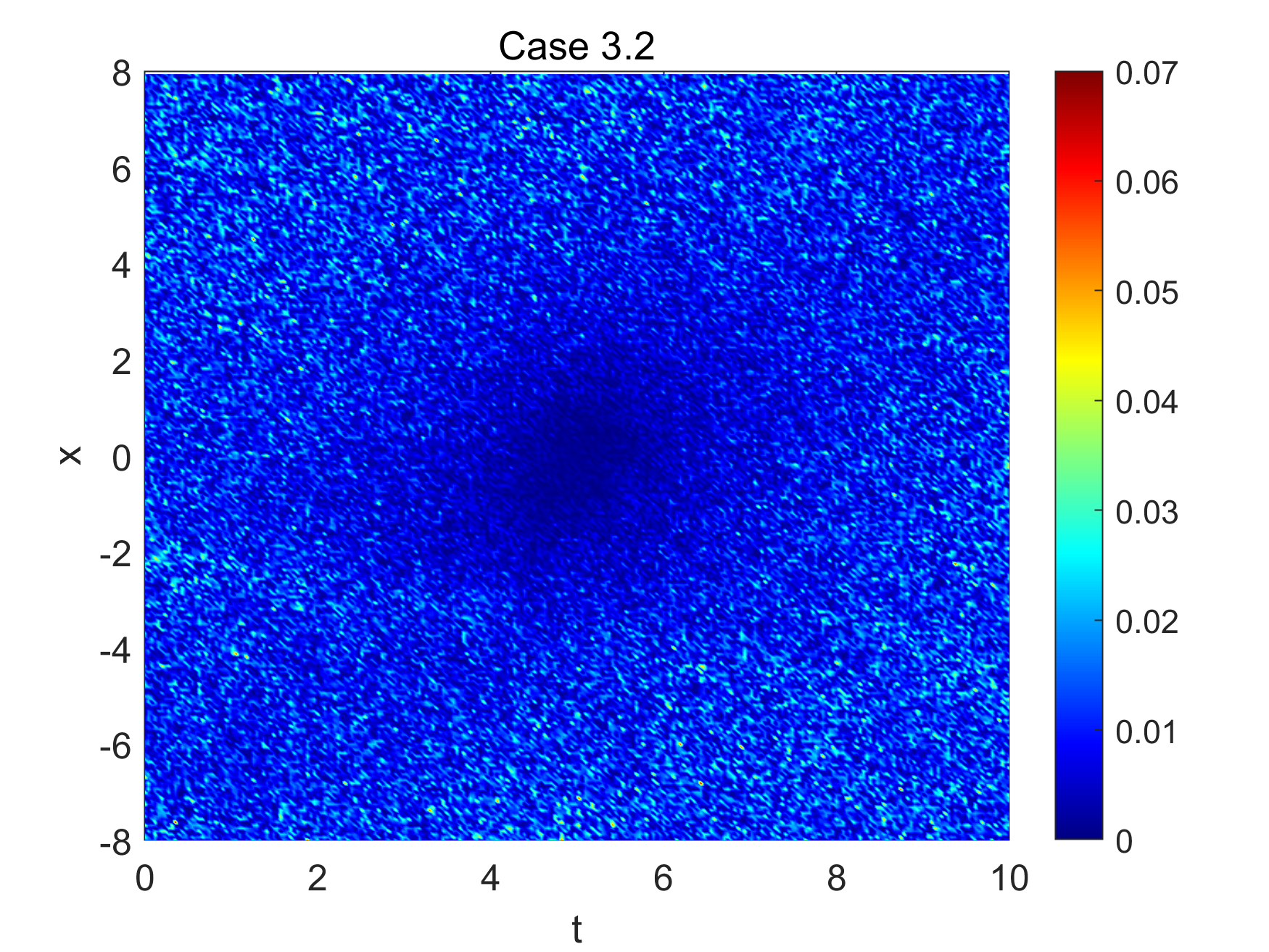}
			\end{minipage}
		}\\
		\subfloat[Mean field estimation]{
			\begin{minipage}[t]{0.315\linewidth}
				\includegraphics[width=1\linewidth]{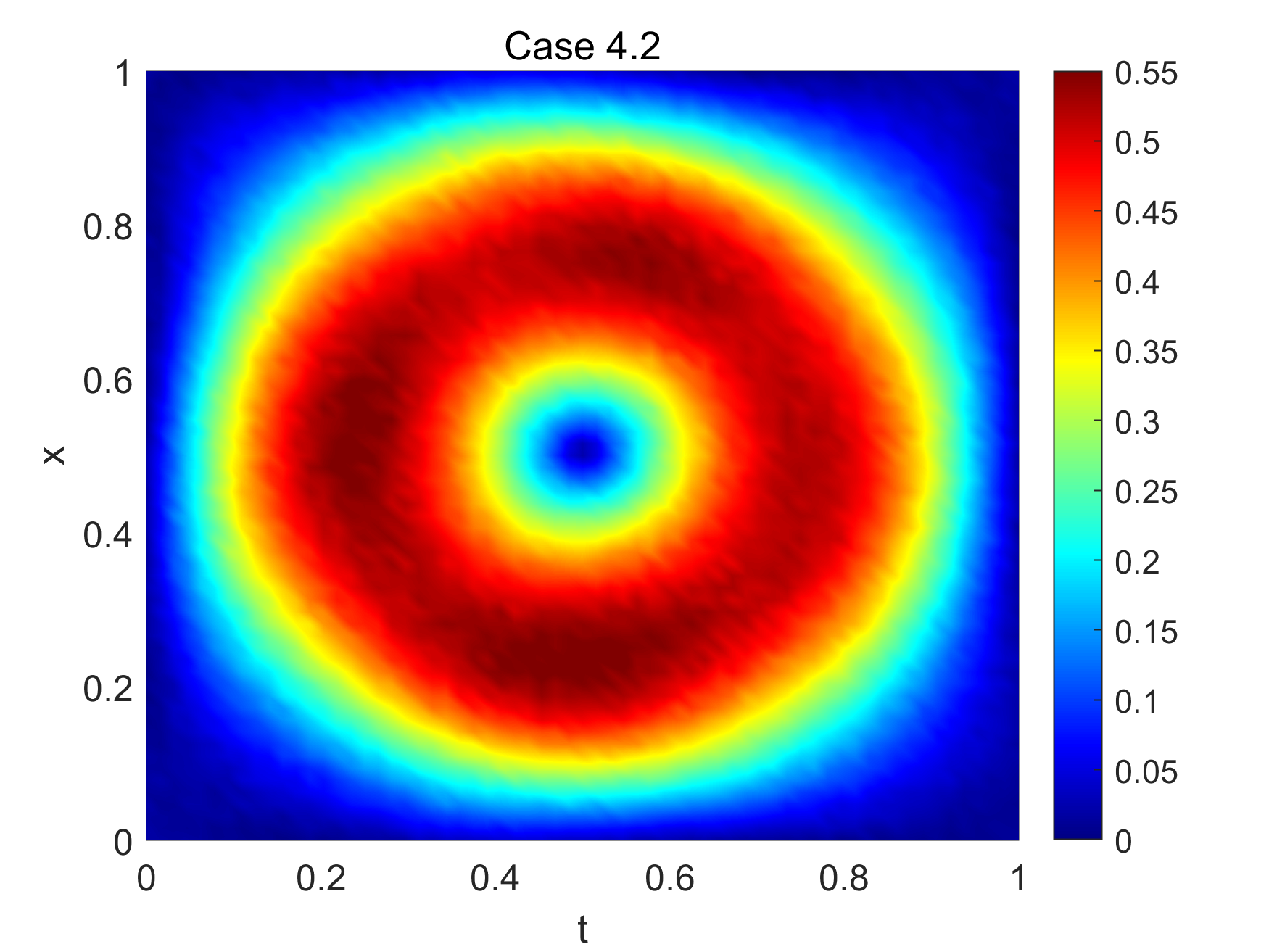}
			\end{minipage}
		}
		\subfloat[Variance field estimation]{
			\begin{minipage}[t]{0.315\linewidth}
				\includegraphics[width=1\linewidth]{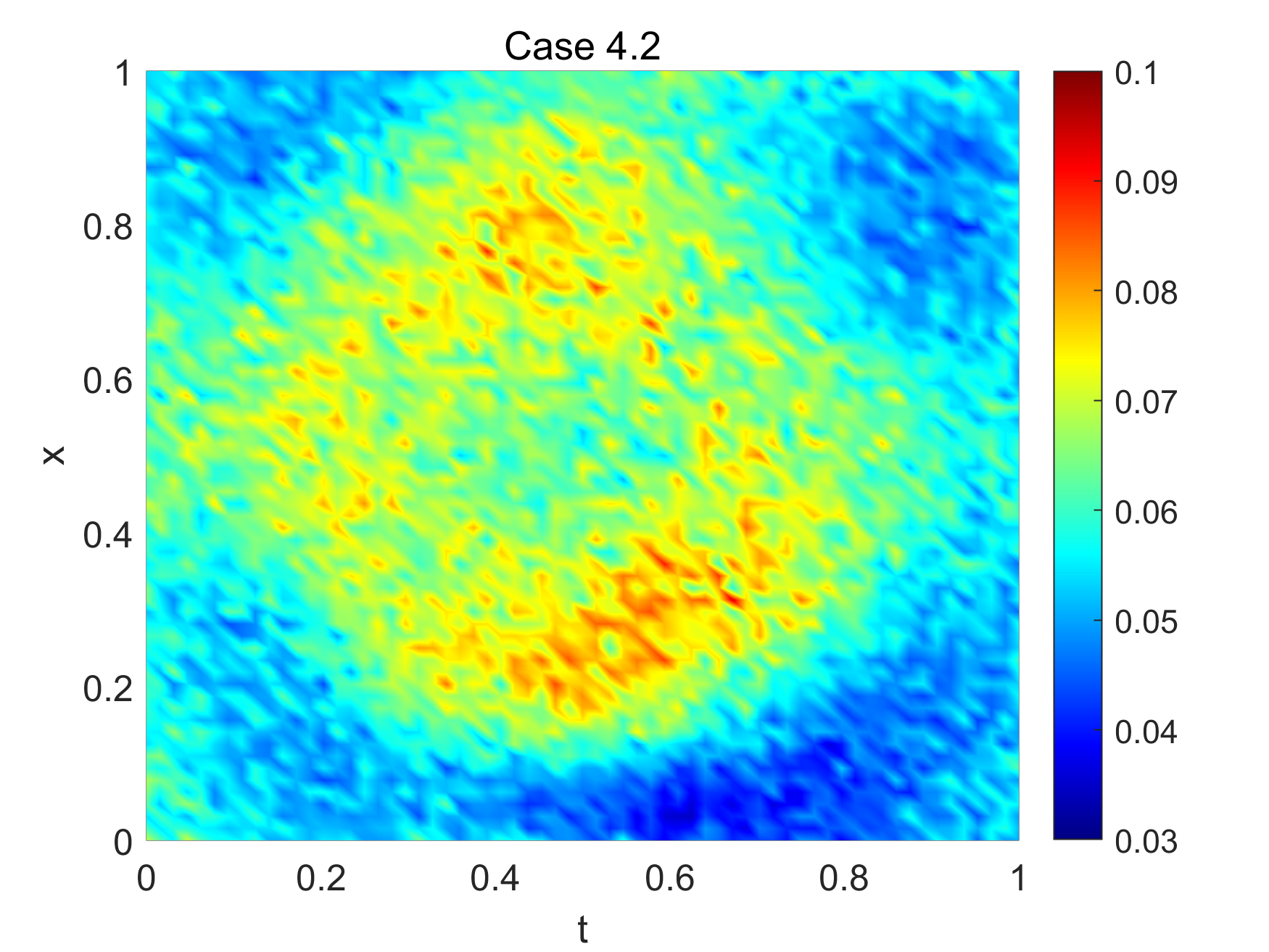}
			\end{minipage}
		}
		\subfloat[Mean field absolute error]{
			\begin{minipage}[t]{0.315\linewidth}
				\includegraphics[width=1\linewidth]{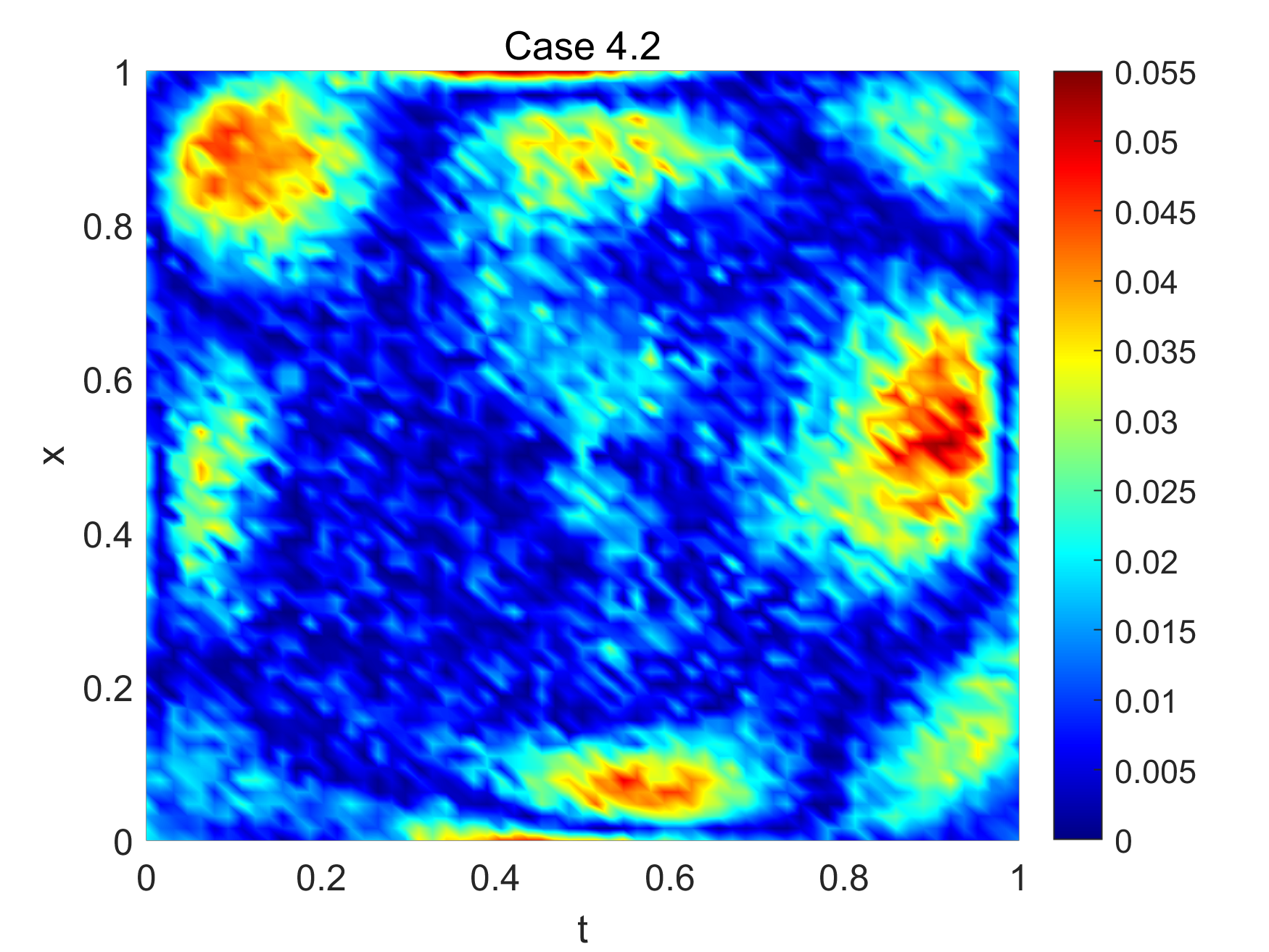}
			\end{minipage}
		}\\
		\caption{Predictive variability assessment under $4\%$ relative Gaussian observation noise using MC dropout in second-stage CCD-PINNs. \label{fig16}}
	\end{figure}

    For each noise level reported in Table~\ref{tab12}, the dropout probability is set to $p=0.05$, and $100$ stochastic forward passes are performed in Stage~2 training. During each pass, a random dropout mask produces a realization of the solution field and a finite-dimensional coefficient parameter vector. The latter is used to construct the corresponding hard piecewise-constant function. The sample solution mean and variance fields, the means and standard deviations of coefficient levels and change-point locations are then computed from these stochastic realizations. For the representative cases with $4\%$ relative additive Gaussian observation noise, the sample mean field, the sample variance field, and the absolute error between the mean field and the reference solution are presented in Figure~\ref{fig16}. Except for the dropout and Monte Carlo sampling settings, all network architectures, training points, optimizers, learning rates, stage divisions, GMM-BDMC detection procedures, and parameter initializations are kept the same as in the previous numerical experiments.

	\begin{table}[p]
		\centering
		\caption{Testing the impact of relative Gaussian observation noise on JVC-PINNs framework. \label{tab12}}
		\scalebox{0.95}{
			\setlength{\tabcolsep}{2.5pt}
			\begin{tabular}{ccccccccccccc}
				\toprule
				\multirow{2}*{Case} & Noise & MSE of & \multicolumn{2}{c}{Coefficient} & Relative & \multicolumn{2}{c}{MC Dropout} & \multicolumn{2}{c}{Change Point} & Relative & \multicolumn{2}{c}{MC Dropout} \\
				~ & Level & Solution & True & Predict & Error & Mean & Sigma & True & Predict & Error & Mean & Sigma \\
				\midrule
				\multirow{12}*{2.1} & \multirow{2}*{0} & \multirow{2}*{6.3578e-6} & 0.1 & 0.0980 & $-$2.0034\% & 0.0976 & 0.0014 & \multirow{2}*{5} & \multirow{2}*{4.9846} & \multirow{2}*{$-$0.3074\%} & \multirow{2}*{4.9845} & \multirow{2}*{0.1354} \\
				~ & ~ & ~ & 0.2 & 0.1981 & $-$0.9534\% & 0.1978 & 0.0021 & ~ & ~ & ~ & ~ & ~ \\
				\cmidrule(lr){2-13}
				~ & \multirow{2}*{0.01} & \multirow{2}*{7.3178e-6} & 0.1 & 0.0954 & $-$4.6441\% & 0.0949 & 0.0043 & \multirow{2}*{5} & \multirow{2}*{4.9152} & \multirow{2}*{$-$1.6967\%} & \multirow{2}*{4.9172} & \multirow{2}*{0.1651} \\
				~ & ~ & ~ & 0.2 & 0.1901 & $-$4.9322\% & 0.1946 & 0.0049 & ~ & ~ & ~ & ~ & ~ \\
				\cmidrule(lr){2-13}
				~ & \multirow{2}*{0.04} & \multirow{2}*{1.2118e-5} & 0.1 & 0.0930 & $-$7.0342\% & 0.0884 & 0.0059 & \multirow{2}*{5} & \multirow{2}*{4.8095} & \multirow{2}*{$-$3.8102\%} & \multirow{2}*{4.8200} & \multirow{2}*{0.2372} \\
				~ & ~ & ~ & 0.2 & 0.1852 & $-$7.4204\% & 0.1831 & 0.0130 & ~ & ~ & ~ & ~ & ~ \\
				\cmidrule(lr){2-13}
				~ & \multirow{2}*{0.09} & \multirow{2}*{2.6518e-5} & 0.1 & 0.0907 & $-$9.2647\% & 0.0931 & 0.0222 & \multirow{2}*{5} & \multirow{2}*{4.5923} & \multirow{2}*{$-$8.1541\%} & \multirow{2}*{4.5752} & \multirow{2}*{0.3226} \\
				~ & ~ & ~ & 0.2 & 0.1806 & $-$9.7205\% & 0.1734 & 0.0269 & ~ & ~ & ~ & ~ & ~ \\
				\cmidrule(lr){2-13}
				~ & \multirow{2}*{0.16} & \multirow{2}*{6.0118e-5} & 0.1 & 0.0774 & $-$22.6142\% & 0.0800 & 0.0252 & \multirow{2}*{5} & \multirow{2}*{3.9380} & \multirow{2}*{$-$21.2409\%} & \multirow{2}*{3.9723} & \multirow{2}*{0.3895} \\
				~ & ~ & ~ & 0.2 & 0.1524 & $-$23.7829\% & 0.1549 & 0.0349 & ~ & ~ & ~ & ~ & ~ \\
				\cmidrule(lr){2-13}
				~ & \multirow{2}*{0.25} & \multirow{2}*{1.2636e-4} & 0.1 & 0.0604 & $-$39.6312\% & 0.0602 & 0.0377 & \multirow{2}*{5} & \multirow{2}*{6.6709} & \multirow{2}*{+33.4175\%} & \multirow{2}*{6.7119} & \multirow{2}*{0.4458} \\
				~ & ~ & ~ & 0.2 & 0.1163 & $-$41.8608\% & 0.1179 & 0.0462 & ~ & ~ & ~ & ~ & ~ \\
				\midrule
				\multirow{18}*{3.2} & \multirow{3}*{0} & \multirow{3}*{7.5590e-10} & 0.5 & 0.4998 & $-$0.0387\% & 0.4996 & 0.0048 & \multirow{3}*{5} & \multirow{3}*{4.9995} & \multirow{3}*{$-$0.0093\%} & \multirow{3}*{4.9957} & \multirow{3}*{0.0508} \\
				~ & ~ & ~ & 1 & 0.9998 & $-$0.0152\% & 0.9995 & 0.0067 & ~ & ~ & ~ & ~ & ~ \\
				\cmidrule(lr){4-6}
				~ & ~ & ~ & 0.1 & 0.1000 & $-$0.0183\% & 0.1000 & 0.0003 & ~ & ~ & ~ & ~ & ~ \\
				\cmidrule(lr){2-13}
				~ & \multirow{3}*{0.01} & \multirow{3}*{1.2208e-6} & 0.5 & 0.4906 & $-$1.8731\% & 0.4927 & 0.0185 & \multirow{3}*{5} & \multirow{3}*{5.0114} & \multirow{3}*{+0.2271\%} & \multirow{3}*{5.0057} & \multirow{3}*{0.0765} \\
				~ & ~ & ~ & 1 & 0.9584 & $-$4.1602\% & 0.9545 & 0.0239 & ~ & ~ & ~ & ~ & ~ \\
				\cmidrule(lr){4-6}
				~ & ~ & ~ & 0.1 & 0.0951 & $-$4.9228\% & 0.0935 & 0.0049 & ~ & ~ & ~ & ~ & ~ \\
				\cmidrule(lr){2-13}
				~ & \multirow{3}*{0.04} & \multirow{3}*{6.3208e-6} & 0.5 & 0.4727 & $-$5.4619\% & 0.4666 & 0.0357 & \multirow{3}*{5} & \multirow{3}*{4.9443} & \multirow{3}*{$-$1.1142\%} & \multirow{3}*{4.9411} & \multirow{3}*{0.1305} \\
				~ & ~ & ~ & 1 & 0.9258 & $-$7.4153\% & 0.9295 & 0.0498 & ~ & ~ & ~ & ~ & ~ \\
				\cmidrule(lr){4-6}
				~ & ~ & ~ & 0.1 & 0.0926 & $-$7.4098\% & 0.0955 & 0.0098 & ~ & ~ & ~ & ~ & ~ \\
				\cmidrule(lr){2-13}
				~ & \multirow{3}*{0.09} & \multirow{3}*{1.9621e-5} & 0.5 & 0.4624 & $-$7.5218\% & 0.4683 & 0.0646 & \multirow{3}*{5} & \multirow{3}*{5.1559} & \multirow{3}*{+3.1172\%} & \multirow{3}*{5.1526} & \multirow{3}*{0.2047} \\
				~ & ~ & ~ & 1 & 0.9028 & $-$9.7157\% & 0.8930 & 0.0855 & ~ & ~ & ~ & ~ & ~ \\
				\cmidrule(lr){4-6}
				~ & ~ & ~ & 0.1 & 0.0903 & $-$9.7104\% & 0.0872 & 0.0171 & ~ & ~ & ~ & ~ & ~ \\
				\cmidrule(lr){2-13}
				~ & \multirow{3}*{0.16} & \multirow{3}*{4.8321e-5} & 0.5 & 0.4267 & $-$14.6534\% & 0.4413 & 0.1074 & \multirow{3}*{5} & \multirow{3}*{4.4893} & \multirow{3}*{$-$10.2144\%} & \multirow{3}*{4.5043} & \multirow{3}*{0.2966} \\
				~ & ~ & ~ & 1 & 0.8321 & $-$16.7872\% & 0.8282 & 0.1305 & ~ & ~ & ~ & ~ & ~ \\
				\cmidrule(lr){4-6}
				~ & ~ & ~ & 0.1 & 0.0762 & $-$23.7578\% & 0.0727 & 0.0314 & ~ & ~ & ~ & ~ & ~ \\
				\cmidrule(lr){2-13}
				~ & \multirow{3}*{0.25} & \multirow{3}*{1.0250e-4} & 0.5 & 0.3226 & $-$35.4733\% & 0.3162 & 0.1672 & \multirow{3}*{5} & \multirow{3}*{6.4227} & \multirow{3}*{+28.4544\%} & \multirow{3}*{6.4113} & \multirow{3}*{0.3994} \\
				~ & ~ & ~ & 1 & 0.6152 & $-$38.4778\% & 0.6403 & 0.1858 & ~ & ~ & ~ & ~ & ~ \\
				\cmidrule(lr){4-6}
				~ & ~ & ~ & 0.1 & 0.0536 & $-$46.4055\% & 0.0531 & 0.0357 & ~ & ~ & ~ & ~ & ~ \\
				\midrule
				\multirow{12}*{4.2} & \multirow{2}*{0} & \multirow{2}*{2.1226e-5} & 0.01 & 0.0101 & +1.4376\% & 0.0103 & 0.0002 & \multirow{2}*{0.5} & \multirow{2}*{0.4998} & \multirow{2}*{$-$0.0489\%} & \multirow{2}*{0.4955} & \multirow{2}*{0.0199} \\
				~ & ~ & ~ & 0.02 & 0.0198 & $-$0.7740\% & 0.0190 & 0.0008 & ~ & ~ & ~ & ~ & ~ \\
				\cmidrule(lr){2-13}
				~ & \multirow{2}*{0.01} & \multirow{2}*{2.3966e-5} & 0.01 & 0.0096 & $-$4.2495\% & 0.0097 & 0.0009 & \multirow{2}*{0.5} & \multirow{2}*{0.4988} & \multirow{2}*{$-$0.2304\%} & \multirow{2}*{0.4828} & \multirow{2}*{0.0259} \\
				~ & ~ & ~ & 0.02 & 0.0190 & $-$4.9231\% & 0.0194 & 0.0010 & ~ & ~ & ~ & ~ & ~ \\
				\cmidrule(lr){2-13}
				~ & \multirow{2}*{0.04} & \multirow{2}*{3.5066e-5} & 0.01 & 0.0093 & $-$7.1653\% & 0.0092 & 0.0011 & \multirow{2}*{0.5} & \multirow{2}*{0.4947} & \multirow{2}*{$-$1.0563\%} & \multirow{2}*{0.5131} & \multirow{2}*{0.0332} \\
				~ & ~ & ~ & 0.02 & 0.0185 & $-$7.4102\% & 0.0183 & 0.0032 & ~ & ~ & ~ & ~ & ~ \\
				\cmidrule(lr){2-13}
				~ & \multirow{2}*{0.09} & \multirow{2}*{6.3166e-5} & 0.01 & 0.0091 & $-$9.4374\% & 0.0090 & 0.0016 & \multirow{2}*{0.5} & \multirow{2}*{0.5411} & \multirow{2}*{+8.2204\%} & \multirow{2}*{0.5457} & \multirow{2}*{0.0608} \\
				~ & ~ & ~ & 0.02 & 0.0181 & $-$9.7108\% & 0.0179 & 0.0051 & ~ & ~ & ~ & ~ & ~ \\
				\cmidrule(lr){2-13}
				~ & \multirow{2}*{0.16} & \multirow{2}*{1.2267e-4} & 0.01 & 0.0077 & $-$23.0368\% & 0.0077 & 0.0024 & \multirow{2}*{0.5} & \multirow{2}*{0.6070} & \multirow{2}*{+21.3957\%} & \multirow{2}*{0.6265} & \multirow{2}*{0.0848} \\
				~ & ~ & ~ & 0.02 & 0.0152 & $-$23.7596\% & 0.0147 & 0.0045 & ~ & ~ & ~ & ~ & ~ \\
				\cmidrule(lr){2-13}
				~ & \multirow{2}*{0.25} & \multirow{2}*{2.3373e-4} & 0.01 & 0.0059 & $-$41.4715\% & 0.0058 & 0.0051 & \multirow{2}*{0.5} & \multirow{2}*{0.3575} & \multirow{2}*{$-$28.4959\%} & \multirow{2}*{0.3422} & \multirow{2}*{0.0937} \\
				~ & ~ & ~ & 0.02 & 0.0116 & $-$42.1305\% & 0.0117 & 0.0071 & ~ & ~ & ~ & ~ & ~ \\
				\bottomrule
		\end{tabular}}
	\end{table}

    The results are reported in Table~\ref{tab12}, which contains the MSE of solution, deterministic Stage~2 estimates, the sample means and standard deviations obtained from stochastic forward passes through both the Stage~2 main network and coefficient sub network. When the noise level is small or moderate, especially at or below $0.04$, JVC-PINNs maintains relatively accurate solution approximation, coefficient estimation, and jump localization in the tested cases. The MSE of the predicted solution increases gradually with the noise level, while the signed relative errors of the recovered coefficients and change points remain comparatively limited in this range. This indicates that the physics constraints and statistical regularization help prevent overfitting to noisy observations in the considered examples.

    As the noise level further increases to $0.16$ and $0.25$, the degradation becomes more evident. The coefficient estimates become more biased, and the change-point errors increase significantly. This behavior illustrates a potential failure mechanism of the framework under severe noise. The first-stage coefficient samples are strongly distorted, causing the GMM-BDMC intervals and candidate jump regions to deviate from the underlying regimes. Consequently, the second-stage constrained refinement is initialized from less reliable statistical information. Nevertheless, the gradual deterioration shown in Table~\ref{tab12} indicates that the framework loses accuracy progressively rather than through abrupt numerical instability.

    As the observation noise increases, the Stage~1 coefficient samples become more dispersed, leading to larger component standard deviations estimated by GMM-BDMC and, consequently, wider coefficient search intervals. Although the Stage~2 coefficient estimates remain constrained within these intervals, the wider feasible ranges provide weaker regularization and allow different dropout masks to produce more variable plateau values and change-point locations. Therefore, the increase in the MC-dropout sample standard deviation at high noise levels reflects the combined effects of less informative statistical constraints and greater sensitivity of the Stage~2 coefficient sub network to network perturbations. This behavior can be interpreted as a diagnostic of reduced reconstruction reliability.

    Under noisy observations, relying only on a single deterministic output of the main network may be unreliable because local perturbations in the data can be amplified during the reconstruction process. Reporting both mean and variance fields helps identify reconstructions that are sensitive to network perturbations. The predictive-variability visualizations in Figure~\ref{fig16} show that the sample variance is spatially nonuniform and tends to increase in regions where the solution has stronger local variation or where the reconstruction is more sensitive to noisy observations. Meanwhile, the sample mean remains close to the reference solution in most regions, and the absolute error is broadly consistent with the high-variance areas. These results indicate that MC dropout can supplement the point estimates of the solution fields with a diagnostic of their sensitivity to network perturbations. The variance field can also provide heuristic guidance for subsequent data acquisition and numerical refinement. In regions with large predictive variability, one may deploy sensors with stronger noise resistance, collect additional observation points, or increase the sampling density of physics residual points during training. Therefore, the MC dropout module can be used as a diagnostic tool for assessing reconstruction stability and as a heuristic indicator for improving experimental design and computational accuracy.

    In summary, the MC-dropout experiment conducted in this study provides a diagnostic of predictive variability in reconstructed solution fields and estimated coefficient parameters. Because the GMM-BDMC statistical learning results obtained from Stage~1 coefficient sub-network output determine the number of finite-dimensional parameters in the target piecewise-constant coefficient function during Stage~2 PINNs training, the numbers of coefficient states and change points in time-varying coefficient scenario inferred by GMM-BDMC are held fixed during MC sampling. Consequently, the reported sample means and standard deviations for parameters, solution mean and variance fields characterize sensitivity to network perturbations conditional on the inferred piecewise-constant structure. However, these results do not constitute full uncertainty quantification for the inverse problem and should not be interpreted as calibrated posterior statistics or credible intervals. Moreover, a more complete uncertainty-quantification framework would also need to propagate the joint uncertainty in the coefficient structure and solution field.
    
	\section{Limitation}\label{sec8}
	The proposed JVC-PINNs framework is developed primarily for PDE inverse problems with finitely many piecewise-constant coefficients. This design is most explicit in the CCD-PINNs stage, where the unknown coefficient is replaced by a hard piecewise-constant representation. For constant coefficients, the framework does not introduce spurious change regions, as demonstrated by the wave equation Case~1.1 and the Navier-Stokes equation Case~4.1. For smoothly varying coefficients or coefficients that are only approximately piecewise constant, the GWS-PINNs stage may already provide a satisfactory continuous approximation of the parameter field, as illustrated by Burgers equation Case~3.5 and Case~3.6. In such cases, imposing a hard discontinuous structure in the second stage may be unnecessary and may introduce additional approximation bias.
	
	The present work does not establish theoretical results on identifiability, convergence, or error bounds. Existing analyses of PINNs have established approximation, optimization, and generalization results for various smooth PDE settings. The problems considered here involve the additional difficulty of discontinuous PDE coefficient fields. Across coefficient interfaces, the solution may fail to possess the classical regularity required by standard smooth-PDE analyses, even though the solution remains continuous over the domains considered in the numerical examples. This structure suggests that some existing PINNs methods' analyses can be extended to the present setting by treating the discontinuity interfaces separately or by using piecewise-regularity arguments. Establishing identifiability conditions, convergence guarantees, and quantitative error estimates for the JVC-PINNs framework remains an important direction for future work.
	
	The accuracy of CCD-PINNs depends strongly on the coefficient samples produced by GWS-PINNs. The statistical mixture learner estimates coefficient regimes, heuristic search intervals, and candidate transition regions from the outputs of the Stage~1 sub network. Errors in Stage~1 sampling may therefore propagate directly to the Stage~2 refinement. Large observational noise, sparse data, or unstable optimization can blur the sampled coefficient levels and distort the transition regions, leading to inaccurate GMM-BDMC inference and unreliable constrained estimation.
	
	The numerical studies in this work are based on synthetic data generated from prescribed PDE models. This setting allows the coefficient values and discontinuity locations to be evaluated in a controlled manner. In contrast, measured heterogeneous physical fields may contain sensor bias, correlated noise, missing observations, unmodeled forcing terms, or more complex material structures. The performance of the proposed framework on real experimental data remains to be investigated.
	
	The present experiments use regular computational domains and standard initial-boundary configurations. The coefficient jumps are assumed to occur within the interior of the spatiotemporal domain, away from the initial slice and the spatial boundary. Irregular geometries, complex boundary conditions, moving interfaces, and coefficient variations that intersect the initial or boundary surfaces are not considered in the current implementation.
    
	The governing PDE, source term, initial and boundary data are assumed to be known and correctly specified. This represents a strong model prior. When the governing PDE is unknown, symbolic-learning or PDE-discovery techniques may be required before applying the proposed framework. Model mismatch, incorrect boundary conditions, and inaccurate initial data are also outside the scope of the present study. Because the physics-informed loss is constructed directly from the prescribed PDE prior, JVC-PINNs may be sensitive to errors in the assumed model. Improving the robustness of the framework under incomplete physical priors is therefore an important direction for future research.
    
    The fully Bayesian uncertainty-quantification framework also remains an important direction for future research. The present study uses MC dropout only as a preliminary diagnostic of conditional predictive variability of the main network for solution reconstruction and the sub network for parameter estimation in Stage~2 training. This approach does not quantify posterior uncertainty in the number of coefficient regimes or in the spatial interface geometry, nor does it propagate such uncertainty through the coupled coefficient-solution system. A more comprehensive treatment should jointly model and propagate uncertainty in the observations, solution field, coefficient states, transition locations, interface geometry, network parameters, and PDE-model discrepancy.
	
	These limitations suggest several directions for future work. Studies based on noisy real-world measurements, irregular domains, boundary-intersecting jumps, unknown governing equations, model mismatch, and fully Bayesian uncertainty quantification would substantially broaden the applicability of the proposed framework. Since this work focuses primarily on methodological development, it serves as an initial study of the JVC-PINNs framework for inverse problems for partial differential equations with jump discontinuities in coefficients and demonstrates its feasibility through numerical experiments on representative PDE models.
	
	\section{Conclusion and Discussion}\label{sec9}
	This study developed a two-stage physics-informed learning framework, termed JVC-PINNs, for inverse problems governed by PDEs with discontinuously varying coefficients. The proposed method addresses the challenge of recovering jump-discontinuous parameter fields from limited and potentially noisy observations when the coefficient values, number of regimes, and transition locations are all unknown. In the first stage, a dual-network GWS-PINNs architecture is employed, in which the main network approximates the PDE solution and an auxiliary coefficient sub network produces a relaxed continuous surrogate of the underlying discontinuous coefficient field. A gradient-adaptive weighting strategy is incorporated into the physics residual to improve training stability and increase the reliability of coefficient sampling in approximately constant regions near potential discontinuities.
    
    The sampled coefficient values are subsequently analyzed using Bayesian statistical mixture modeling combined with birth-death model selection, referred to as GMM-BDMC. This procedure estimates the number of coefficient states and provides heuristic search intervals for the coefficient values together with candidate transition regions. In the second stage, the inverse problem is reformulated using the constrained CCD-PINNs estimator, in which the smooth coefficient surrogate is replaced by an explicit hard piecewise-constant parameterization. The coefficient values and discontinuity locations are then refined within the search intervals and candidate regions supplied by the statistical learning stage, thereby reducing the search space of the inverse problem.
    
    Numerical experiments on representative PDE systems with jump-discontinuous coefficients show that the proposed framework can identify coefficient regimes, localize candidate transition regions, and reconstruct the corresponding solution fields. Across the tested examples, JVC-PINNs provides accurate coefficient and solution estimates, particularly for abrupt temporal or spatial coefficient transitions. The results also show that the physics-informed sampling, statistical regime identification, and constrained piecewise-constant refinement stages complement one another in the identification of discontinuous coefficient fields. These findings suggest that JVC-PINNs provides a flexible computational approach for inverse problems involving discontinuous parameter structures in nonstationary and heterogeneous spatiotemporal systems.
    
    Several directions remain for future research. A deeper theoretical understanding of the proposed framework, including convergence, identifiability, error propagation, and fully Bayesian uncertainty quantification, is needed to further establish its reliability. Improving computational efficiency and scalability is also important for high-dimensional PDEs, multi-parameter systems, and large-scale datasets. Adaptive sampling, parallel computing, reduced-order modeling, and dimensionality-reduction techniques may be useful in this regard. Future studies should also evaluate the framework using measured heterogeneous physical fields, irregular domains, complex boundary conditions, unknown governing equations, model mismatch, and parameter jumps that intersect initial or boundary surfaces. These extensions may require combining JVC-PINNs with PDE discovery, symbolic learning, robust optimization, or uncertainty-aware inference.
    
    Potential application areas include aerospace engineering, transportation, geophysics, biological systems, materials science, and smart manufacturing, particularly in settings that require online parameter estimation or real-time decision support. This work provides an initial methodological study of JVC-PINNs and validates its feasibility using representative synthetic PDE examples. Addressing the above directions would broaden the applicability and practical utility of the proposed framework.
	
	\section*{Acknowledgement}
	This work was supported by the Postdoctoral Fellowship Program of CPSF under Grant Number GZC20261675 and the Fundamental Research Funds for the Central Universities.

	
	
\end{document}